\documentclass[10pt,journal,compsoc]{IEEEtran}
\ifCLASSOPTIONcompsoc
  \usepackage[nocompress]{cite}
\else
  \usepackage{cite}
\fi
\ifCLASSINFOpdf
\else
\fi
\usepackage{amsmath}
\usepackage{amsthm}
\usepackage{graphicx}
\usepackage{amssymb}
\usepackage{multirow}
\usepackage{lettrine}
\usepackage{subfigure}
\usepackage{bm}
\usepackage{mathrsfs}
\usepackage[noend]{algpseudocode}
\usepackage[ruled]{algorithm2e}
\usepackage{ragged2e}
\usepackage{colortbl}
\usepackage{titletoc}
\usepackage{tikz}
\usepackage[framemethod=tikz]{mdframed}
\usepackage{ctable}
\usepackage {enumitem} 
\setlist {nolistsep} 

\usepackage{caption}

\usepackage[colorlinks,
linkcolor=blue,
anchorcolor=orange,
citecolor=orange
]{hyperref}

\usepackage{amsfonts,amssymb}
\DeclareMathOperator*{\Max}{\bm{\mathsf{max}}}

\DeclareMathOperator*{\expe}{\mathbb{E}} 
\DeclareMathOperator*{\mcu}{\mathcal{U}} 
\DeclareMathOperator*{\mcv}{\mathcal{V}}
\DeclareMathOperator*{\mcs}{\mathcal{S}}

\DeclareMathOperator*{\mcq}{\mathcal{Q}}
\DeclareMathOperator*{\mco}{\mathcal{O}}

\DeclareMathOperator*{\mfr}{\mathfrak{R}}
\DeclareMathOperator*{\mch}{\mathcal{H}}

\DeclareMathOperator*{\bss}{\boldsymbol{\sigma}}
\DeclareMathOperator*{\sg}{\sigma}

\DeclareMathOperator*{\ps}{\Phi(\mathcal{S})}

\newcommand{\tj}[1]{\tilde{j}}
\newcommand{\tk}[1]{\tilde{k}}

\newtheorem{prop}{Proposition}
\newtheorem{thm}{Theorem}
\newtheorem{defi}{Definition} 
\newtheorem{assum}{Assumption} 
\newtheorem{coro}{Corollary} 

\newtheorem{lem}{Lemma}
\newtheorem{rem}{Remark}
\newtheorem*{pf}{Proof}

\newtheorem*{rdef2}{Restate of Definition \ref{def1}}
\newtheorem*{rthm1}{Restate of Theorem \ref{cml:lem2}}
\newtheorem*{rthm3}{Restate of Theorem \ref{theo1}}
\newtheorem*{rthm2}{Restate of Theorem \ref{thm3}}
\newtheorem*{rthm4}{Restate of Theorem \ref{theo3}}

\begin{document}
\title{Rethinking Collaborative Metric Learning: \\ Toward an Efficient Alternative \\ without Negative Sampling}
\author{Shilong~Bao,
	Qianqian~Xu*,~\IEEEmembership{Senior~Member,~IEEE},
	Zhiyong Yang, \\
	Xiaochun~Cao,~\IEEEmembership{Senior~Member,~IEEE},
	and~Qingming Huang*,~\IEEEmembership{Fellow,~IEEE}% <-this % stops a space
	\IEEEcompsocitemizethanks{\IEEEcompsocthanksitem * Corresponding authors}
	\IEEEcompsocitemizethanks{\IEEEcompsocthanksitem Shilong Bao is with State Key Laboratory of Information Security (SKLOIS), Institute of Information Engineering, Chinese Academy of Sciences, Beijing 100093, China, and also with School of Cyber Security, University of Chinese Academy of Sciences, Beijing 100049, China (email: baoshilong@iie.ac.cn).}
	\IEEEcompsocitemizethanks{\IEEEcompsocthanksitem Qianqian Xu is with the Key Laboratory of Intelligent Information Processing, Institute of Computing Technology, Chinese Academy of Sciences, Beijing 100190, China, (email: xuqianqian@ict.ac.cn).}% <-this % stops a space
	\IEEEcompsocitemizethanks{\IEEEcompsocthanksitem Zhiyong Yang is with the School of Computer Science and Technology, University of Chinese Academy of Sciences, Beijing 101408, China (email: yangzhiyong21@ucas.ac.cn).}
	\IEEEcompsocitemizethanks{\IEEEcompsocthanksitem Xiaochun Cao is with State Key Laboratory of Information Security (SKLOIS), Institute of Information Engineering, Chinese Academy of Sciences, Beijing 100093, China, also with School of Cyber Security, University of Chinese Academy of Sciences, Beijing 100049, China(email: caoxiaochun@iie.ac.cn).}% <-this % stops a space
	\IEEEcompsocitemizethanks{\IEEEcompsocthanksitem Qingming Huang is with the School of Computer Science and Technology, University of Chinese Academy of Sciences, Beijing 101408, China, also with the Key Laboratory of Big Data Mining and Knowledge Management (BDKM), University of Chinese Academy of Sciences, Beijing 101408, China, also with the Key Laboratory of Intelligent Information Processing, Institute of Computing Technology, Chinese Academy of Sciences, Beijing 100190, China, and also with Peng Cheng Laboratory, Shenzhen 518055, China (e-mail: qmhuang@ucas.ac.cn).}}

\markboth{TO APPEAR IN IEEE TRANSACTIONS ON PATTERN ANALYSIS AND MACHINE INTELLIGENCE}%
{Shell \MakeLowercase{\textit{et al.}}: Bare Demo of IEEEtran.cls for Computer Society Journals}

% Remember, if you use this you must FCMLl \IEEEpubidadjcol in the second


\IEEEtitleabstractindextext{%
\begin{abstract}
\justifying
The recently proposed Collaborative Metric Learning (CML) paradigm has aroused wide interest in the area of recommendation systems (RS) owing to its simplicity and effectiveness. Typically, the existing literature of CML depends largely on the \textit{negative sampling} strategy to alleviate the time-consuming burden of pairwise computation. However, in this work, by taking a theoretical analysis, we find that negative sampling would lead to a biased estimation of the generalization error. Specifically, we show that the sampling-based CML would introduce a bias term in the generalization bound, which is quantified by the per-user \textit{Total Variance} (TV) between the distribution induced by negative sampling and the ground truth distribution. This suggests that  optimizing the sampling-based CML loss function does not ensure a small generalization error even with sufficiently large training data. Moreover, we show that the bias term will vanish without the negative sampling strategy. Motivated by this, we propose an efficient alternative without negative sampling for CML named \textit{Sampling-Free Collaborative Metric Learning} (SFCML), to get rid of the sampling bias in a practical sense. Finally, comprehensive experiments over seven benchmark datasets speak to the superiority of the proposed algorithm.  

%demonstrate , 
\end{abstract}

}

\maketitle

\IEEEdisplaynontitleabstractindextext

% \fi
\IEEEpeerreviewmaketitle

\section{Introduction}

\IEEEPARstart{N}{owadays}, the explosion of Internet data poses an inevitable challenge of how to help users access their desirable information (say which book/news to read, which resturtant to eat and which anime to see, etc.). Consequently, \textit{recommendation system} (short for RS) \cite{cao2019unifying, wang2017item, DBLP:conf/pakdd/WangZ0HC20, DBLP:conf/nips/MaZ0Y019, DBLP:conf/sigir/CuiWLOYS11, DBLP:conf/kdd/MaZYCW020, DBLP:conf/sigir/0001C17,DBLP:journals/aei/LvZWWW20} has recently emerged as a major solution and has broad applications in modern Internet enterprises, such as Amazon, Alibaba and Facebook. 

The main purpose of RS is to leverage user preference prediction based on the historical data produced from user-item interactions. In practice, such interactions often exist as implicit feedback \cite{oard1998implicit,hu2008collaborative} where no explicit ratings and only actions are provided (such as browses, clicks, purchases, etc.). It is well-known that implicit feedback only contains indirect records from user behavior, without explicit knowledge about their negative intentions. This poses a great challenge to RS-targeted machine learning algorithms and brings about a wave of relevant studies \cite{lerche2016using,zhang2019deep,su2009survey}. The vast majority of such work follows a standard paradigm known as One-Class Collaborative Filtering (OCCF) \cite{DBLP:conf/icdm/PanZCLLSY08, DBLP:journals/www/YaoTYXZSL19, DBLP:conf/icml/HeckelR17, DBLP:conf/sigir/ChenZ0NLC17, DBLP:conf/sigir/Xin0ZZJ19,DBLP:journals/corr/abs-2006-04153,DBLP:conf/ijcai/0001DWTTC18}, where the items not being observed are usually assumed to be of less interest for a given user. 

Over the past decade, Matrix Factorization (MF)-based algorithms are one of the most representative techniques among the OCCF community, where the user's preference toward an item is captured by an inner product between their latent factors, such as \cite{DBLP:journals/ijon/ZhangR21, DBLP:conf/aaai/ChenL019, DBLP:journals/kbs/LiZWCP17}. Unfortunately, some literature pointed out that the inner product violates the triangle inequality, which may lead to sub-optimal performance of recommendation \cite{hsieh2017collaborative,zhang2019deep}. To tackle this problem,  \textit{Collaborative Metric Learning} (CML)  \cite{hsieh2017collaborative} presents a novel OCCF framework via incorporating the strengths of \textit{metric learning} \cite{DBLP:conf/mm/LiuCSWNK19, DBLP:journals/access/JanarthanTRLZY20} into the \textit{Collaborative Filtering} (CF) framework, and achieves a reasonable performance on a wide range of RS benchmark datasets. Nonetheless, the full-batch objective function of CML is featured with an $\mco(\sum_{i=1}^{M}n^+_{i} n^-_i)$ complexity, where $M$ is the number of users and $n^+_i(n^-_i)$ is the number of positive (unobserved) items for user $u_i$. Practically, CML adopts the standard negative sampling strategy \cite{DBLP:conf/www/HeLZNHC17, DBLP:conf/sigir/Wang0WFC19, DBLP:conf/kdd/Wang00LC19, DBLP:conf/wsdm/ChenZLM19} to improve the efficiency, where merely a limited amount of unobserved items are selected to optimize the model for a given user. Hereafter, a series of the related studies has been carried out to improve the performance of CML, such as popularity-based \cite{DBLP:conf/kdd/ChenSSH17, DBLP:conf/sigir/WuVSSR19, tran2019improving}, two-stage \cite{tran2019improving}, and hard mining \cite{DBLP:conf/acml/CanevetF14, DBLP:conf/sigir/WangYZGXWZZ17, DBLP:conf/www/ParkC19, DBLP:conf/ijcai/DingQ00J19} negative sampling strategies, translation-based CML \cite{DBLP:conf/icdm/ParkKXY18, DBLP:conf/www/TayTH18} and co-occurrence embedding regularized metric learning (CRML) \cite{DBLP:journals/nn/WuZNC20}. \textit{Without loss of generality, since the existing methods of CML cannot bypass the negative sampling in the training phase, we call this kind of algorithms sampling-based CML in this paper.}

Different from the sampling-based CML, we argue that the negative sampling strategy essentially alters the intrinsic distribution of the training data, leading to a biased estimation for the expected loss function over the unseen data. Therefore, in this paper, we are interested in the following question:
\begin{center}
 \textit{\textbf{Could the sampling-based CML guarantee a good generalization performance?}}
\end{center}

In search of an answer to the question, we provide a systematic analysis of the generalization ability of the CML framework, based on the Rademacher Complexity-based arguments. The major challenge is that the CML framework adopts a pairwise loss function to capture the preference comparison between positive and the unobserved items, which could not be expressed as a sum of independently identically distributed (i.i.d) loss term. As a result, the standard theoretical arguments \cite{shalev2014understanding} are no longer available for our task. Therefore, we first extend the standard symmetrization regime and the definition of Rademacher complexity \cite{DBLP:conf/colt/BartlettM01, DBLP:books/daglib/0034861} according to a specifically designed symmetrization strategy.

On top of the proposed complexity measure, we prove that the sampling-based CML would introduce an extra per-user \textit{Total Variance} (TV) term in its generalization upper bound, which reflects the discrepancy between the sampling-strategy-induced distribution and the ground truth distribution. This implies that minimizing the sampling-based CML over the training data cannot ensure a small generalization error when the induced distribution behaves away from its ground truth. Meanwhile, we also demonstrate that the biased term will vanish in the sampling-free version. Therefore, in order to obtain a reasonable performance, we propose to learn CML in a sampling-free manner to get rid of the bias. However, as we mentioned above, we must face the heavy computational burden with a non-sampling favor.

{{Facing this challenge, we start the {first exploration} to develop an efficient alternative for CML without negative sampling.}} Specifically, by posing a $\ell_2$ hyper-sphere constraint over the embedding space, we figure out the closed connection between CML and another technique called AUC optimization \cite{DBLP:conf/icml/GaoJZZ13,DBLP:conf/ijcai/GaoZ15}. Motivated by this fact, we construct an acceleration method to evaluate the full sample loss and gradient on top of a semi-regular comparison graph.

Finally, a systematic empirical study is conducted on seven real-world RS datasets, including MovieLens-100k, CiteULike, MovieLens-1m, Steam-200k, Anime, MovieLens-20m and Amazon-Book, the results of which consistently speak to the efficacy of our proposed method.

In a nutshell, the main contributions are summarized as follows:
\begin{itemize}
	\item The pairwise formulation of the CML loss function makes it impossible to employ standard Rademacher complexity-based measures to analyze the generalization ability of CML framework. To address this issue, we propose an extended Rademacher complexity with the strength of a novel symmetrization scheme.    
	
    \item According to the proposed complexity measure, we start an early trial to present theoretical analyses of the generalization ability of the CML framework with (without) the help of negative sampling strategy, which reflects the biased issue of the sampling-based process and suggests the benefit of the sampling-free manner.  
    
	\item Motivated by the theoretical findings, we propose  an efficient alternative called \textit{Sampling-Free Collaborative Metric Learning} (SFCML) to deal with the full samples based CML, where the loss and gradient evaluations are accelerated with a graph-based reformulation.
\end{itemize}

The rest of the paper is organized as follows. Sec.\ref{sec2} presents a review of the most related studies. Sec.\ref{CAL: preliminary} presents  a brief introduction of the CML framework. In Sec.\ref{CAL:bound}, we propose the theoretical analysis for sampling-based and sampling-free CML. In Sec.\ref{CAL: method}, the SFCML method is proposed. In Sec.\ref{CAL: exp}, we conduct empirical studies to show the efficacy of our proposed algorithm on seven real-world RS datasets. Finally, Sec.\ref{conclusion} presents a concluding remark about the work.    

\section{Prior Art\label{sec2}} 

In this section, we briefly review the closely related studies along with our main topic, including one-class collaborative filtering and the existing solutions without negative sampling.
 
\subsection{One-Class Collaborative Filtering}

In many real-world applications, the vast majority of interactions are implicitly expressed by users’ behaviors, e.g, downloads of movies, clicks of products and browses of news. In order to develop RS from such implicit feedback, researchers usually formulate the recommendation task as the \textit{One-Class Collaborative Filtering} (OCCF) problem \cite{DBLP:conf/icdm/PanZCLLSY08, DBLP:journals/www/YaoTYXZSL19, DBLP:conf/icml/HeckelR17, DBLP:conf/kdd/PanS09, DBLP:journals/corr/abs-2006-04153, paquet2013one, sindhwani2010one, pappas2013sentiment}. The existing OCCF algorithms can be mainly categorized into two fashions: a) Pointwise fashion: The goal of the pointwise-based algorithms is to recover the missing signals by minimizing the error between the estimated and observed implicit feedback \cite{fang2011matrix, DBLP:conf/ijcai/0001DWTTC18}. b) Pairwise fashion: Specifically, the pairwise-based solutions aim to construct a system that the observed items should be ranked higher than the unobserved items \cite{pan2013gbpr, he2016ups}. 

\noindent \textbf{Matrix Factorization (MF) based Algorithm}. Over the past decades, the Matrix Factorization (MF)-based algorithms are one of the most classical OCCF solutions \cite{DBLP:conf/sigir/HeZKC16, sindhwani2010one}. The key idea of MF is to represent each user and item as a latent factor to recover the missing entries in a unified space. Typically, a user's preference toward an item is represented as an inner product between their latent factors. For example, under the pointwise setting, \cite{DBLP:conf/sigir/HeZKC16} proposes an item-oriented MF method with implicit feedback. Neural Collaborative Filtering (NCF) \cite{DBLP:conf/www/HeLZNHC17} develops a general framework that unifies the MF and the neural networks together, and then regards the recommendation task as a regression problem. In addition, a pairwise-based algorithm is developed for MF-based recommendation \cite{DBLP:conf/uai/RendleFGS09}. However, some literature pointed out that the inner product violates the triangle inequality, which may lead to sub-optimal performance \cite{hsieh2017collaborative,zhang2019deep}.

\noindent \textbf{Collaborative Metric Learning based Algorithm}. Recently, there rises a new trend in the community to alleviate the intrinsic problem of MF-based algorithms \cite{DBLP:conf/www/Turnbull19, DBLP:conf/www/XuZCC20,DBLP:conf/www/HeLZNHC17}. Among them, a popular idea is to borrow the strengths of metric learning \cite{DBLP:conf/cvpr/LiuWYLRS17, weinberger2009distance}, due to its simplicity and effectiveness \cite{DBLP:conf/mm/BaoXMYCH19, li2017social, park2018collaborative, feng2015personalized, DBLP:conf/www/TayTH18}. Noteworthy is the work known as Collaborative Metric Learning (CML) \cite{hsieh2017collaborative}, which is the first successful integration of metric learning and CF. Generally speaking, the idea of CML is to represent users and items in a unified Euclidean space, where the proximity between users and items naturally captures the preference. As a typical trait, CML employs a negative sampling strategy \cite{DBLP:conf/kdd/YangDZYZT20, DBLP:conf/www/DingF0YLJ18, DBLP:conf/nips/MaZ0Y019, DBLP:conf/www/WangX000C20, DBLP:journals/corr/abs-1811-04411, DBLP:conf/wsdm/RendleF14} to generate the contrastive pairs, which could mitigate the high computational burden of pairwise learning. Thereafter, many efforts have been made to improve the recommendation performance of CML. The relevant studies fall into two camps. The first camp employs a model-oriented strategy, where the CML model is enriched with either more complicated structures or side information. Typically, inspired by the knowledge translation mechanism in knowledge graph embedding, \cite{DBLP:conf/icdm/ParkKXY18, DBLP:conf/www/TayTH18} propose to learn an exclusive latent relation vector for each user-item interaction to explore the preference of each user and item more precisely. Co-occurrence embedding Regularized Metric Learning (CRML) \cite{DBLP:journals/nn/WuZNC20} presents an effective approach that optimizes the representations of both users and items by considering the global statistical information of user-user and item-item pairs. Another camp of the related work attempts to improve the effectiveness of the negative sampling process. For example, Popularity-based negative sampling (PopS) \cite{DBLP:conf/kdd/ChenSSH17, tran2019improving} proposes to sample negative items based on their popularity to minimize the pairwise ranking loss. Tran et al. \cite{tran2019improving} propose a two-stage negative sampling (2stS), which first samples an items' candidate according to their popularity and then selects a negative item from the candidate based on their inner product with positive items. In addition, hard sample mining \cite{hsieh2017collaborative, DBLP:conf/acml/CanevetF14, DBLP:conf/www/ParkC19, DBLP:conf/ijcai/DingQ00J19} manners are also adopted to obtain negative items. However, in this paper, we prove that negative sampling would introduce a bias term in the generalization bound (shown in Sec.\ref{CAL:bound}), such that optimizing CML with the negative sampling strategies may fail to obtain a reasonable generalization performance. Consequently, different from both directions, we will study how to perform CML without the help of negative sampling.
\subsection{Learning without negative sampling \label{neg}}
%In practice, the number of observed implicit feedback is extremely limited, while the unobserved data is of such large volumes. Therefore, in order to learn from such large-scale data, there generally exists two kinds of methods: a) \textit{negative sampling} strategy , i.e., only sample a few items (usually set as a constant $U$) from the unlabeled data and regard them as negatives for training; and b) \textit{non-sampling} strategy \cite{DBLP:conf/sigir/ChenZWMLLM19, , }, i.e., treat the remaining unobserved data as negatives for learning. 

Recently, some researchers have pointed out that negative sampling results in insufficient training of model and thus leads to performance degradation \cite{DBLP:conf/uai/YuanX0GZCJ18, DBLP:conf/sigir/ChenZMLM20, DBLP:conf/acl/HeXYJ18, DBLP:conf/sigir/HeZKC16}. This is because it inevitably discards some informative samples during the training process. On the contrary, non-sampling could avoid this problem since taking all training data into account directly induces a better solution. However, the main bottleneck is training efficiency for the sampling-free manner. Therefore, a growing number of researchers attempt to alleviate the efficiency issue of learning with a non-sampling paradigm in different tasks. He et al. \cite{DBLP:conf/sigir/HeZKC16} design an efficient element-wise
alternating least squares (eALS) with non-uniform missing data. Yuan et al. \cite{DBLP:conf/uai/YuanX0GZCJ18} present a generic and fast batch gradient descent optimizer $f_{BGD}$  that can learn embedding from the whole training dataset without a negative sampling strategy. Therefore, it can naturally be applied to the factorization-based CF recommendation. \cite{DBLP:journals/tois/ChenZZLM20} proposes a general matrix factorization algorithm, i.e., Efficient Neural Matrix Factorization (ENMF) without non-sampling, which shows both effective and efficient performance. In order to sufficiently capture collaborative information among users, items and entities in knowledge graph (KG), a novel jointly non-sampling learning model for KG enhanced recommendation (JNSKR) \cite{DBLP:conf/sigir/ChenZMLM20} is proposed. In addition, recently, Chen et al. \cite{DBLP:conf/aaai/ChenZZMLM20} study the flaws in heterogeneous collaborative filtering (HCF) and develop a novel and state-of-the-art non-sampling learning framework for the recommendation, i.e., Efficient Heterogeneous Collaborative Filtering (EHCF) . 

The recent development of recommendations with implicit feedback has revolutionary motivated the advanced studies of learning without negative sampling. Nonetheless, optimizing CML in a sampling-free manner has long been left behind this wave of revolution. Specifically, \textbf{the existing non-sampling algorithms are designed for a factorization-based RS model or a pointwise loss function, which is not applicable for the CML framework based on the pairwise ranking loss}. On stark contrary, different from the above non-sampling recommendation models, in this work, we study how to learn CML in a sampling-free manner efficiently. Most importantly, we also present the issue of negative sampling from the theoretical perspective, which has not been explored in previous work.

\begin{table}
	\centering
	\setlength{\abovecaptionskip}{5pt}    
	\setlength{\belowcaptionskip}{15pt}    
	\setlength{\tabcolsep}{8pt}
	\caption{A summary of key notations and descriptions in ths work.}
	\scalebox{0.9}{
	\begin{tabular}{ll}
		\toprule
		Notations & Descriptions \\
		\midrule
		$M$ & number of users\\
		$N$ & number of items\\
		$\mcu$ & the set of users\\
		$\mcv$ & the set of items \\
		$\mathcal{S}_i^+$ & the interacted items of $u_i$\\
		$v_*^+$ & the item contained in $\mathcal{S}_i^+$ \\
		$v_*^-$ & the item not belonging to $\mathcal{S}_i^+$ \\
		$y_{ij}$ & user $u_i$'s preference toward item $v_j$ \\
		$f$ & learned score function \\
		$d$ & the dimension of space\\
		$\bm{W}_u$ & learned transformation weight of users \\
		$\bm{W}_v$ & learned transformation weight of items\\
		$\boldsymbol{e}_{u_i}$ & embeddings of user $u_i$\\ 
		$\boldsymbol{e}_{v_j}$ & embeddings of item $v_j$ \\
		$\boldsymbol{d}(i, j)$ & Euclidean distance between $u_i$ and $v_j$\\
		$\boldsymbol{u}_i$ & one-hot encoding of user $u_i$\\
		$\boldsymbol{v}_j$ & one-hot encoding of item $v_j$\\ 
		$\ell^{(i)}_{hinge}$ & Hinge loss function \\
		$\ell^{(i)}_{sq}$ & Square loss function\\ 
		$n_i^+$ & number of observed items for user $u_i$\\
		$n_i^-$& number of unobserved items for user $u_i$\\
		$\tilde{\boldsymbol{\mathbb{P}}}^{(i)}$ & the sampling-strategy-induced distribution\\
		$\hat{\boldsymbol{\mathbb{P}}}^{(i)}$ & the ground-truth distribution\\
		$\boldsymbol{L}^{(i)}$ & the Laplacian matrix of user $u_i$\\
		$\mathcal{G}^{(i)}$& the graph of user $u_i$\\
		$\mathcal{D}^{(i)}$ & the adjacent matrix of user $u_i$\\
		$\mathcal{S}^{(i)}$ & the vertex set of $\mathcal{G}^{(i)}$\\
		$\mathcal{E}^{(i)}$& the edge set of $\mathcal{G}^{(i)}$\\
		$\nabla_{\boldsymbol{e}_{u_i}}$ & gradient of variable $\boldsymbol{e}_{u_i}$\\
		$\nabla_{\bm{W}_v}$ & gradient of variable $\bm{W}_v$\\
		$\boldsymbol{y}^{(i)}$ & vector of $u_i$'s preferences \\
		$\boldsymbol{f}^{(i)}$ & user $u_i$'s score vectors toward all items\\
		$\mathcal{H}_R$ & the hypothesis space \\
		$\left\|\bm{W}\right\|_{*}$ & induced 2 norm of matrix $\bm{W}$\\
		$\xi_t(\bm{W}_v)$ & the $t$-th largest singular value of $\bm{W}_v$ \\
		\bottomrule
	\end{tabular}
	\label{notation}
}
\end{table}
\section{Preliminary} \label{CAL: preliminary}
In this section, we first introduce some notations and the task of recommendation learning from implicit feedback. Then, the concepts of CML framework are briefly presented. For clarity, a summary of key notations and their corresponding descriptions throughout this paper are listed in Tab.\ref{notation}.

%Finally, we discuss the flaws of existing CML algorithms: \textbf{a) directly learning with all samples is not desirable as the unaffordable computation complexity}; \textbf{b) adopting negative sampling to improve the efficiency would introduce extra bias (the Total Variance term in Sec.\ref{CAL:bound}), which degrades the performance of recommendation}. These issues are the main driving force for us to develop an efficient alternative without negative sampling in Sec.\ref{CAL: method}.

\subsection{Task Definition}
In this work, our model is developed by the \textit{implicit feedback} signals (e.g., clicks, thump ups, likes, etc.). Given the historical user-item interaction records, our primary aim is to infer the preference of users and then recommend unseen items that he/she is most likely to be interested in. Mathematically, assume that there are $M$ users and $N$ items in practical recommendation system, denoted as $\mathcal{U} = \{u_1, u_2, \dots, u_M\}$ and $\mathcal{V} = \{v_1, v_2, \dots, v_N\}$, respectively. Let $\mathcal{S}_i^+$ be the set of interacted items of user $u_i$. Then, the interaction record of user $u_i$ toward item $v_j$ is defined as:
\begin{equation}
y_{ij} = \left\{
\begin{aligned}
& 1, \ \ if \ \ v_j  \in \mathcal{S}_i^+ \wedge v_j \in \mathcal{V};\\
& 0,  \ \ else.
\end{aligned}
\right.
\end{equation}
where $y_{ij}=1$ indicates the observed/positive actions with item $v_j$, and $y_{ij} = 0$ could imply user that dislikes or is unaware of the existence of item $v_j$ \cite{DBLP:conf/www/TayTH18, DBLP:conf/www/HeLZNHC17}.

Motivated by the preference paradigm of OCCF, here for observed interactions, we assume that the users tend to have higher preferences for these interacted items than other unobserved items.
% As a result, ones usually explore the preference of users toward items $\mathcal{V}$ with a pairwise manner
%\[
%\mathcal{Z} = \{(u_i, v_j, v_k)| u_i \in \mathcal{U}, v_j, v_k \in \mathcal{V}, v_j \in \mathcal{S}^{(i)}_+, v_k \notin \mathcal{S}^{(i)}_+\},
%\]
%where $(u_i, v_j, v_k)$ could present a preference relation that user $u_i$ prefers $v_j$ to $v_k$. 
Consequently, given a target user $u_i \in \mathcal{U}$ and his/her interaction records, the main task of recommendation is to seek a preference score function $f(v_j|u_i)$ out of a predefined hypothesis class $\mathcal{H}_R$ containing the candidate models. Finally, given a proper choice of $f$, the RS system will then recommend the items having the top-K ranking score.

\subsection{Preference Consistency Model} \label{precon}
Now we start to introduce how the preference score functions are formulated in CML. Borrowing the wisdom from metric learning, in order to capture the preferences, a distance metric should be learned from the implicit feedback. On top of the metric space, the users' preferences could be naturally specified by the value of distance through the learned metric.

To this end, we first project each user and item in a joint Euclidean space through the following lookup transformations \cite{DBLP:conf/recsys/QuadranaKHC17, DBLP:conf/wsdm/ChenXZT0QZ18, DBLP:journals/tkde/CuiWPZ19}:
\begin{equation}
\begin{aligned}
\boldsymbol{e}_{u_i} &= \bm{W}_u^\top \boldsymbol{u}_i, \\
\boldsymbol{e}_{v_j} &= \bm{W}_v^\top \boldsymbol{v}_j,
\label{emb}
\end{aligned}
\end{equation}
where $\boldsymbol{e}_{u_i} \in \mathbb{R}^d$ and $\boldsymbol{e}_{v_j} \in \mathbb{R}^d$ are the embeddings of user $u_i$ and item $v_j$, respectively; $d$ is the dimension of space; $\bm{W}_u \in \mathbb{R}^{M \times d}$, $\bm{W}_v \in \mathbb{R}^{N \times d}$ are the learned transformation weight; $\boldsymbol{u}_i$ and $\boldsymbol{v}_j$ are two different one-hot encodings in which the nonzero elements in $\boldsymbol{u}_i$ and $\boldsymbol{v}_j$ correspond to the index of a particular user $u_i$ and item $v_j$, respectively.

If the metric space is Euclidean, the value of metric between user $u_i$ and item $v_j$ can be intuitively measured by their distance:
%\begin{equation}
%\begin{aligned}
%\boldsymbol{d}(i, j) = ||\boldsymbol{e}_{u_i} - \boldsymbol{e}_{v_j}||^2.
%\end{aligned}
%\end{equation}
\[
\boldsymbol{d}(i, j) = ||\boldsymbol{e}_{u_i} - \boldsymbol{e}_{v_j}||^2.
\]

Subsequently, in order to capture the preference of user $u_i$, one should push away the observed items and unobserved items through the lens of  distance constraints. Therefore, if user $u_i$ likes item $v_j$ (i.e., $y_{ij} = 1$), a small value for $\boldsymbol{d}(i, j)$ should be assigned. If the opposite is the case (i.e., $y_{ij} = 0$), we then hope a large $\boldsymbol{d}(i, j)$.  Mathematically, the following inequality should be held to reflect the relative preference of $u_i$ toward different items $v_j$ and $v_k$:
\begin{equation}
\begin{cases}
\ \  \boldsymbol{d}(i, j) < \boldsymbol{d}(i,k), & v_j^+ \in \mathcal{S}_i^+, v_k^- \notin \mathcal{S}_i^+; \\
\ \ \boldsymbol{d}(i, j) > \boldsymbol{d}(i, k), &  v_j^- \notin \mathcal{S}_i^+, v_k^+ \in \mathcal{S}_i^+
\end{cases}\label{Eq3}
\end{equation}  

where, with a slight abuse of notation, we let $v_*^+$ represent the item involved in $\mathcal{S}_i^+$ and that for $\mathcal{S}^-_i$ is denoted by $v_*^-$ ($*$ represents any item here).

Motivated by Eq.(\ref{Eq3}), we only need to control the relative preference rather than their magnitude, since scaling both sides of inequalities does not change the partial order. Hence, the CML framework is to minimize the following  pairwise empirical risk to reflect such preference consistency:
\begin{equation}
\begin{aligned}
	\hat{\mathcal{R}}_{\mcs}^{\text{cml}}(f) &= \frac{1}{M} \sum_{u_i \in \mcu} \frac{1}{n_i^+n_i^-} \sum_{j=1}^{n_i^+} \sum_{k=1}^{n_i^-} \ell^{(i)}(v_j^+, v_k^-),  \\
%	\ell_{jk}^i &= \boldsymbol{d}(i, j) - \boldsymbol{d}(i, k), \\
	s.t. \ \ &||\boldsymbol{e}_{u_i}||^2 \le R, \ \ \ ||\boldsymbol{e}_{v_j}||^2 \le R, \ \ u_i \in \mathcal{U}, v_j \in \mathcal{V}.
\end{aligned}\label{e4}
\end{equation}
In respect of Eq.(\ref{e4}), we have the following explanations. Denote the number of observed (unobserved) items for a given user $u_i$ as $n^+_i$($n^-_i$). Since the overall amount of items in the system is fixed, we naturally come to the fact that $n_i^+ + n_i^- = N$, where $N$ is the total number of the items. Moreover, the user/item embeddings are constrained within a $\ell_2$ ball with radius $R$ to ensure a normalization. Finally, $\ell^{(i)}(v_j^+, v_k^-)$ is a differentiable ranking loss which is often set as the hinge loss
\[
	\ell^{(i)}_{hinge}(v_j^+, v_k^-) = \Max\left(0,  \lambda + \boldsymbol{d}(i,j) - \boldsymbol{d}(i, k)\right),
\] 
where $\lambda>0$ is the safe margin to ensure sufficient partition across different types of items.

  Different from the traditional inner-product-based OCCF framework, CML induces the triangle inequality \cite{DBLP:conf/nips/XingNJR02, DBLP:journals/ftml/Kulis13} by means of the Euclidean metric. As a result, the learned embedding can automatically cluster 1) co-liked items from  the same user and 2) co-liked items of similar users, which suggests a better global consistency of the preference ranking\cite{hsieh2017collaborative, DBLP:conf/mm/BaoXMYCH19}. 
  
  Finally, when the training is completed, one can easily leverage $f(v_j|u_i) = -\boldsymbol{d}(i, j)$ to calculate the rank of each item $v_j$ and generate recommendations for user $u_i$.

\subsection{Learning with Negative Sampling} \label{sec.3.3}
Despite the strength of metric learning, CML leads to a heavy computational burden:
\textit{In Eq.(\ref{e4}), every item that the user interacted with needs to be paired with all remaining items, which induces a $\mathcal{O}(\sum_{i=1}^Mn_i^+n_i^-)$ time and space complexity.} 

At present, to alleviate this situation, most of the existing literature usually resorts to the \textit{negative sampling} strategy, i.e., sampling a few items from unobserved sets as negatives, such as uniform sampling \cite{DBLP:conf/recsys/RendleKZA20, DBLP:conf/www/HeLZNHC17, DBLP:conf/sigir/Wang0WFC19}, popular-based sampling \cite{DBLP:conf/kdd/ChenSSH17, DBLP:conf/sigir/WuVSSR19, tran2019improving}, two-stage sampling technique \cite{tran2019improving} and hard sampling strategy \cite{DBLP:conf/acml/CanevetF14, hsieh2017collaborative, DBLP:conf/sigir/WangYZGXWZZ17, DBLP:conf/www/ParkC19, DBLP:conf/ijcai/DingQ00J19}. 

Specifically, one usually samples $U$ negative items for each positive user-item $(u_i, v_j^+)$ interaction based on a pre-designed sampling distribution. We could formulate such a process as modification toward the objective function in expectation. For user $i$, the negative samples are drawn from a sparse and discrete distribution $\tilde{\boldsymbol{\mathbb{P}}}^{(i)}$, then the empirical expectation of the sampling-based CML could be rewritten as:
\begin{equation}
	\begin{aligned}
		\tilde{\mathcal{R}}_{\mcs}^{\text{cml}}(f) 
		&= \frac{1}{M} \sum_{u_i \in \mathcal{U}} \sum_{j=1}^{n_i^+} \sum_{k=1}^{n_i^-} \tilde{\boldsymbol{\mathbb{P}}}^{(i)}_{jk} \cdot \ell^{(i)}(v_j^+, v_k^-) \\
		s.t. &\left\|\boldsymbol{e}_{u_i}\right\|^2 \le R, \ \ \ \left\|\boldsymbol{e}_{v_j}\right\|^2 \le R, \ \ u_i \in \mathcal{U}, v_j \in \mathcal{V},
	\end{aligned}\label{e5}
\end{equation}
where $\tilde{\boldsymbol{\mathbb{P}}}^{(i)}_{jk} = \mathbb{P}(\bm{v}_j^+, \bm{v}_k^-)$ represents the probability that item $v_k^-$ is sampled as a negative instance for $v_j^+$. The whole distribution for user $i$ is then expressed in a compact form :
\[\tilde{\boldsymbol{\mathbb{P}}}^{(i)}_{j} = [\tilde{\boldsymbol{\mathbb{P}}}^{(i)}_{j1}, \tilde{\boldsymbol{\mathbb{P}}}^{(i)}_{j2}, \dots, \tilde{\boldsymbol{\mathbb{P}}}^{(i)}_{jn_i^-}],\] 
which is a sparse vector with $U$ non-zero terms. Here,  $\tilde{\boldsymbol{\mathbb{P}}}^{(i)}_{jk} \neq 0$ only if item $v_k^-$ is sampled as one of the negatives for $v_j^+$. Moreover, it is interesting to note that, the orginal objective function could be also regarded as a special case of the new formulation via setting $ \mathbb{P}(\bm{v}_j^+, \bm{v}_k^-) \equiv \frac{1}{n_i^+n_i^-}>0$. In this sence, the negative sampling strategy introduces sparisity to the original distribution, which leads to a lighter $\mathcal{O}(\sum_{i=1}^Mn_i^+U)$ complexity than $\mathcal{O}(\sum_{i=1}^Mn_i^+n_i^-)$ (note that $U \ll n_i^-)$. 

Although the negative sampling schemes could reduce the computation burden of Eq.(\ref{e4}), \textbf{such strategies may cause some unexpected issues}, including
\begin{itemize}
	\item The sampling distributions and the size of sampling (constant $U$) to a large extent determine the performance, which is generally difficult to choose. Moreover, this also makes the sampling-based CML unstable.
	\item Comparing Eq.(\ref{e4}) with Eq.(\ref{e5}), the sampling-based CML is a biased estimation of the original loss function.  In other words, from the theoretical analysis in the next section (Sec.\ref{CAL:bound}), one can see that \emph{\textbf{optimizing the sampling-based loss function will not necessarily lead to a small generalization error}}. 
\end{itemize}

%\textbf{(1) convergence slows down considerably due to the gradually negligible gradient magnitude when training}. As we can see from Eq.(\ref{e5}), the magnitude of the gradient at each step depends on how CML equipped with current $\theta$ distinguishes between the positives item and sampled negative items. This gradient is easily close to $0$, meaning that the model can learn nothing and thus $\theta$ would not be updated.  

%\textbf{(2) Sampling bias}. Needless to say, one usually has no prior knowledge of the true distribution of data. Therefore, optimizing CML with an empirically defined sampling distribution is usually biased. \textbf{(3) Performance degradation} The aforementioned issues may make the learning process vulnerable and be stuck in a local minimum, which degrades the performance of recommendation.

\section{Generalization Bounds for CML framework} \label{CAL:bound}
In this section, we provide a systematic theoretical discussion of the generalization ability of the CML framework. To do this, the basic notations and assumptions of theoretical analysis are first introduced. Subsequently, we extend the standard symmetrization regime and the definition of Rademacher Complexity. Finally, based on the proposed complexity arguments, we present the generalization bounds for the CML framework, including sampling-based and sampling-free manners.

\noindent \textbf{Asymptotic Notations}. In order to make our theoretical discussions more clear, we first provide some asymptotic notations that will be adopted throughout the generalization analysis.
\begin{itemize}
	\item $x \lesssim y$ represents that there exists some universal constant $C > 0$ such that $x \le Cy$.
	\item Similarly, $x \gtrsim y$ means that there exists some universal constant $C > 0$ such that $x \ge Cy$.
	\item $x \asymp y$ is equivalent to $y \lesssim x \lesssim y$.
\end{itemize}
Notably, other notations could be found in Tab.\ref{notation}.
\subsection{Basic Assumptions on the Item Embeddings}
Our analysis relies on two fundamental regularities of the item embedding matrix $\bm{W}_v \in \mathbb{R}^{N\times d}$, where we generally assume that $d \ll N$ in the practical RS. 

Here, we need the basic notion of the induced matrix 2-Norm. Given a matrix $\bm{W}_v$,  $\xi_1(\bm{W}_v) \ge \xi_2(\bm{W}_v) \ge, \cdots,  \ge \xi_d(\bm{W}_v)$, are the singular values of $\bm{W}_v$. Then we have the following definition.  
\begin{defi}\label{def3} (\textbf{Induced 2-Norm of a Matrix}). Let $\left\|\bm{W}\right\|_{*}$ be the induced 2-norm of a weight matrix $\bm{W}$. Then, the induced norm of $\bm{W}$ is defined as follows:
	\begin{equation}
	\begin{aligned}
	\left\|\bm{W}\right\|_{*} = \max_{||\bm{x}||_2 =1 } \left\|\bm{W}\bm{x}\right\|_2 = \xi_{1}(\bm{W}).
	\end{aligned}
	\end{equation}
\end{defi}

%In the following formulations, without loss of generality, we just assume there is only one user (taking $u_i$ as an example) in our systems, since it is computed separately for every user and thus can be easily extend to $M$ users. 

\begin{assum}(\textbf{Basic Assumptions}).\label{assu1} 
	We have the following assumptions:
	\begin{enumerate}
		\item[(a)] \textbf{Embedding Diversity Assumption:} Let $d$ be the dimensionality of the embedding space, we assume that: 
		
		\[\frac{\xi_1(\bm{W}_v)}{\xi_d(\bm{W}_v)} \asymp 1\]
		\item[(b)] \textbf{Embedding Capacity Assumption:} We assume that d is sufficiently large such that:
		\begin{equation*}
		(N)^{1/2} \lesssim d \ll N.
		\end{equation*}
	\end{enumerate}
%where the notation $x \lesssim y$ and $x \gtrsim y$ represent that there exists some universal constant $C > 0$ holding that $x \le Cy$ and $x \ge Cy$, respectively. According to this, $x \asymp y$ is introduced to be equivalent to $y \lesssim x \lesssim y$.
\end{assum}

\begin{rem}
	We have the following remarks about the assumptions above.
	\begin{enumerate}
		\item[(a)] Note that we call Assm.\ref{assu1}-(a) the embedding diversity assumption, because when $\frac{\xi_1(\bm{W}_v)}{\xi_d(\bm{W}_v)} \asymp 1$, the embedding matrix is of full column rank with each dimension exhibiting similar importance. 
		\item[(b)] Note that, since $\left\|\bm{e}_{v_j}\right\|_2^2 \lesssim 1$, $\left\|\bm{W}_v\right\|_{F}^2 \lesssim N$. Moreover, 
		\[
		\left\|\bm{W}_v\right\|_{F}^2 = \xi_1^2 + \xi_2^2 + \dots + \xi_d^2.
		\]  
		Then, we have 
		\begin{equation}\label{ineq:wv}
		\left\|\bm{W}_v\right\|_* \lesssim \sqrt{\frac{n^+_i + n^-_i}{d}} = \sqrt{\frac{N}{d}} \lesssim d^{1/2}
		\end{equation}
	\end{enumerate}
\end{rem}

Overall, Assum.\ref{assu1} ensures that the learned embeddings are sufficiently informative to support a well-trained model.

\subsection{The Rademacher Complexity Measure for CML Framework}
According to the model constraints of CML, the user-item embeddings are chosen uniformly from the following embedding hypothesis space:
\begin{equation}
\mathcal{H}_R = \bigg\{ \bm{e}: \bm{e}\in \mathbb{R}^d, ~\left||\bm{e}\right\|^2 \le R \bigg\},
\end{equation}
where $\bm{e}_{u_i} \in \mathcal{H}_R, ~ u_i \in \mcu$ and $\bm{e}_{v_j} \in \mathcal{H}_R, ~ v_j \in \mcv$.

Based on the given hypothesis space, we will present a worse case generalization analysis to show that even the worst choice of the user-item embedding set has a reasonably small generalization error given a small training error. Following the standard learning theory arguments, such a bound relies on the Rademacher complexity measure of the given hypothesis $\mathcal{H}$. Traditionally, the Rademacher complexity is derived from the symmetrization technique as an upper bound for the largest deviation over a given hypothesis $\mathcal{H}$:
\[
\expe_{\mathcal{S}}\left[\sup_{f \in \mathcal{H}} \expe_{\mathcal{S}}(\hat{\mathcal{R}}_{\mathcal{S}}) - \hat{\mathcal{R}}_{{\mathcal{S}}} \right].
\]

Unfortunately, the standard argument of the symmetrization technique requires the empirical risk $\hat{\mathcal{R}}_{\mathcal{S}}$ to be a sum of independent terms. This is not available for the CML framework-based loss, which is essentially a sum of pairwise terms. For example, the terms $\ell^{(i)}(v_j^+, v_k^-)$ and $\ell^{(i)}(\tilde{v}_j^+, \tilde{v}_k^-)$ are interdependent as long as one of them is the same (i.e., $v_j^+=\tilde{v}_j^+$ or $v_k^- = \tilde{v}_k^-$).

To solve this problem, we present a novel symmetrization technique to construct an extended Rademacher complexity defined as follows:

\begin{defi} \label{def1} (CML \textbf{Rademacher Complexity}). Given the sample set $\mcs = \mathop{\cup}\limits_{u_i \in \mcu} \mcs_i$ where $\mcs_i = \{v_j^+\}_{j=1}^{n_i^+} \cup \{v_k^-\}_{k=1}^{n_i^-}, n_i^+ + n_i^- = N$ and the hypothesis space $\mathcal{H}_R$, then the empirical CML Rademacher Complexity with respect to the sample $\mcs$ is defined as:
	\begin{equation}
	\hat{\mfr}_{\ell, \mcs}^{\text{cml}}(\mathcal{H}_R) = \frac{1}{M}\sum_{u_i \in \mathcal{U}} \mathop{\expe}\limits_{\bss_i} \left[\mathop{\sup}\limits_{\mathcal{H}_R}  \frac{1}{n_i^+n_i^-} \sum_{j=1}^{n_i^+} \sum_{k=1}^{n_i^-} 
	\mathcal{Q}_{(i)}^{jk} \right], \label{ded6}
	\end{equation}
	where 
	\[
	\mathcal{Q}_{(i)}^{jk} = \frac{\sigma^+_{ij} + \sigma^-_{ik}}{2} \cdot \ell^{(i)}(v_j^+, v_k^-);
	\]
	$\bss_i =[\sg^+_{i1}, \sg^+_{i2}, \dots, \sg^+_{in_i^+}, \sg^-_{i1},\sg^-_{i2}, \dots,\sg^-_{in_i^-}]$
	is i.i.d Rademacher random variables uniformly chosen from $\{-1, +1\}$, i.e., $\mathbb{P}(\sg=1) = \mathbb{P}(\sg=-1) = 0.5$. Next, the population version of the Rademacher Complexity of CML is expressed as $\mfr_{\ell, \mcs}^{\text{cml}}(\mathcal{H}_R) = \mathop{\expe}\limits_{\mcs}\left[\hat{\mfr}_{\ell,\mcs}^{\text{cml}}(  \mathcal{H}_R)\right]$. 
\end{defi} 
According to this extended form of Rademacher complexity, we establish a symmetrization result expressed in the following theorem. The proof of Thm.\ref{cml:lem2} is involved in Appendix \ref{cml:sysm} in the supplementary materials.

\begin{thm}[\textbf{CML Symmetrization}] \label{cml:lem2} Let $\mcs$ and $ \mathcal{S}'$ be the two independent datasets of interactions that only one sample is different. In terms of any the hypothesis set $\mathcal{H}_R$ and loss function $\ell$, the following holds:
	\begin{equation}
	\begin{aligned}
	&\expe_{\mathcal{S}}\left[\sup_{\mathcal{H}_R} \left[ \expe_{\mathcal{S}}(\hat{\mathcal{R}}^{\text{cml}}_{\mathcal{S}}) - \hat{\mathcal{R}}^{\text{cml}}_{{\mathcal{S}}} \right] \right] \\  
	&\le \expe_{\mcs, \mathcal{S}'} \left[\mathop{\sup}\limits_{\mathcal{H}_R} \left[ \left(\hat{\mathcal{R}}^{\text{cml}}_{\mathcal{S}'}(f) - \hat{\mathcal{R}}^{\text{cml}}_{\mcs}(f)\right) \right] \right] \\
	&\le 2 \mathfrak{R}_{\ell,\mcs}^{\text{cml}}(\mathcal{H}_R).
	\end{aligned}
	\end{equation}
\end{thm}

Moreover, the generalization analysis also relies on an upper bound of the empirical Rademacher complexity $\hat{\mathfrak{R}}_{\ell,\mcs}^{\text{cml}}(\mathcal{H}_R)$. Here, we need the following notion of the effective sample size.

\begin{defi}\label{defi3}(\textbf{Essential Sample Size}) Given the dataset $\mathcal{S}$, and $n^+_i$, $n^-_i$ for each specific user, the effective sample size is defined as:
	\begin{equation*}
	\tilde{N} =\left( \sum_{u_i \in \mathcal{U}} \sqrt{{ \frac{1}{n^+_i} + \frac{1}{n^-_i}  }} \right)^{-2}.
	\end{equation*}	
\end{defi}
Note that we call $\tilde{N}$ Essential Sample Size since it behaves like  an ordinary sample size in the traditional generalization bound. Specifically, the traditional generalization bounds enjoy an order of $O(({1}/{N})^{1/2})$, while our results scale as $O(({1}/{\tilde{N}})^{1/2})$. Moreover, the following remark presents an interesting property of the Essential Sample Size.

\begin{rem}
	Compared with the true sample size $N$, $\tilde{N}$ could better reflect the long-tail nature of the implicit feedback since increasing $n^+_i$ brings way shaper influence to $\tilde{N}$ than $n^-_i$ (Note that the number of unobserved items ($n^-_i$) often dominates the observed ones ($n^+_i$) in practical RS).
\end{rem}

Based on the effective sample size, we reach the following upper bound for the Rademacher complexity. Refer to Appendix \ref{SFCML:rade} for the details of its proof.
\begin{thm} \label{thm3} (\textbf{Upper Bound of empirical Rademacher Complexity}). Given the sample dataset $\mcs = \mathop{\cup}\limits_{u_i \in \mcu} \mcs_i$ where $\mcs_i = \{v_j^+\}_{j=1}^{n_i^+} \cup \{v_k^-\}_{k=1}^{n_i^-}, n_i^+ + n_i^- = N$. If $\ell$ is $\phi$-Lipschitz continuous, then the following inequality holds:
	\begin{equation}
	\begin{aligned}
	\hat{\mfr}_{\ell, \mcs}^{\text{cml}}(\mathcal{H}_R) &\lesssim \phi \cdot \frac{{\Max(\lambda, \sqrt{R \cdot d})}}{M} \cdot \tilde{N}^{-1/2}.
	\end{aligned}\label{eq15}
	\end{equation}	
\end{thm}
Based on the two theorems above, we provide generalization bounds for the sampling-based and sampling-free CML framework respectively in the following two subsections.
% Equipped with Def.\ref{def3} and Assu.\ref{assu1}, in the following, we would show the generalization ability of CML, sampling-based and FCML, to show the strengths and weakness of these methods, based on the widely used Rademacher Complexity-based worst-case analysis. Let $f$ be chosen from the hypothesis space $\mch$. As we all know, when the expectation risk $\mathcal{R}_{\ell}(f)$ is small, we said that the model has good generalization ability to other unseen items. Different from the conventional pointwise problems, these CML-based algorithms are introduced the interdependent pairwise losse}s, which makes the traditional generalization analysis not applicable for our problem. To solve this, we extend the standard symmetrization scheme (refer to Lem.\ref{cml:sysm} and Lem.\ref{cal:sysm} in the appendix for more details) \cite{DBLP:conf/colt/BartlettM01, DBLP:books/daglib/0034861} and give an improved definition of Rademacher Complexity, such that the machine learning theory becomes applicable to our problem.
\subsection{Generalization Bound of Sampling-Free CML}
We start our discussion with the sampling-free CML. Specifically, the main result is summarized in the following theorem.  
\begin{thm}\label{theo1} (\textbf{Generalization Upper Bound of CML} \textbf{with Eq.(\ref{e4})}). Let $\mathcal{H}_R$ be the hypothesis space and $\ell$ be $\phi$-Lipschitz continuous. Given the sample set $\mcs = \mathop{\cup}\limits_{u_i \in \mcu} \mcs_i$ where $\mcs^{(i)} = \{v_j^+\}_{j=1}^{n_i^+} \cup \{v_k^-\}_{k=1}^{n_i^-}, n_i^+ + n_i^- = N$, for any $\delta \in (0,1)$, with probability at least $1 - \delta$, the following inequation holds:
	\begin{equation}
	\begin{aligned}
	\mathcal{R}_{\ell}^{\text{cml}}(f) &\lesssim \hat{\mathcal{R}}^{\text{cml}}_{\mcs}(f) + \phi \cdot \frac{\Max(\lambda, \sqrt{R \cdot d})}{M} \cdot \sqrt{\frac{1}{\tilde{N}}}   \\
	&+ \phi \cdot  \frac{R}{M} \cdot \sqrt{\frac{\log2/\delta}{2}} \cdot \sqrt{\frac{1}{\tilde{N}}},
	\label{cml:final}
	\end{aligned}
	\end{equation}
	where $	\mathcal{R}_{\ell}^{\text{cml}}(f)$ is the expectation risk.
\end{thm}

%\begin{pf} We refer the readers to see more details of the proof in appendix \ref{rthm:eq4}.

%	To prove this, one should first ensure the Bounded Difference Property  (Definition.\ref{def:bdp}) satisfied and then apply the \textit{Mcdiarmid's Inequality} (Lem.\ref{lem:mc}). Then, the key of this proof is to leverage the following \text{CML} symmetrization.
%	\begin{lem}[CML Symmetrization Regime] \label{cml:lem1} Let $\mcs$ and $ \mathcal{S}'$ be the two independent dataset of interactions that only one sample is different. In terms of any the hypothesis set $\mch$ and loss function $\ell^{(i)}$, the following holds:
%		\begin{equation}
%		\begin{aligned}
%		\expe_{\mcs, \mathcal{S}'} \left[\mathop{\sup}\limits_{f \in \mch} \left(\hat{\mathcal{R}}^{\text{cml}}_{\mathcal{S}'}(f) - \hat{\mathcal{R}}^{\text{cml}}_{\mcs}(f)\right)\right] &\le 2 \mathfrak{R}_{\mcs}^{\text{cml}}(\  \mathcal{H}_R)
%		\end{aligned}
%		\end{equation}
%	\end{lem}

%\end{pf}

\begin{pf} Equipped with Assum.\ref{assu1} and Def.\ref{def1}, by applying Talagrand contraction (Lem.\ref{lem: tala}), we could complete the proof. More details are presented in the Appendix \ref{SFCML:rthm3}.
\end{pf}

Based on a proper choice of $\lambda$, the theorem above shows that the generalization gap 

\[\Delta_\mathcal{S} = \mathcal{R}_{\ell}^{\text{cml}}(f) - \hat{\mathcal{R}}^{\text{cml}}_{\mcs}(f)\]
satisfies $\Delta_\mathcal{S} \lesssim  \frac{1}{M}\sqrt{\frac{d}{\tilde{N}}}$. This shows that  $\Delta_\mathcal{S}$ vanishes with a sufficiently large data size and a moderate magnitude of $d$.

\subsection{Generalization Bound of Sampling-based CML}
For the sampling-based CML framework we have the following generalization upper bound.
\begin{thm}\label{theo3} (\textbf{Generalization Upper Bound of sampling-based CML Eq.(\ref{e5})}). Let $\mathcal{H}_R$ be the hypothesis set and $\ell$ be $\phi$-Lipschitz. Given the sample set $\mcs = \mathop{\cup}\limits_{u_i \in \mcu} \mcs_i$ where $\mcs^{(i)} = \{v_j^+\}_{j=1}^{n_i^+} \cup \{v_k^-\}_{k=1}^{n_i^-}, n_i^+ + n_i^- = N$, for any $\delta \in (0,1)$, with probability at least $1 - \delta$, the following holds for all possible embedding $\mathcal{H}_R$:
	\begin{equation}
	\begin{aligned}
	\mathcal{R}_{\ell}^{\text{cml}}(f) &\lesssim \hat{\mathcal{R}}^{\text{cml}}_{\mcs}(f) + \phi\cdot \frac{\Max(\lambda, \sqrt{R \cdot d})}{M} \cdot \sqrt{\frac{1}{\tilde{N}}} \\ 
	&+ \frac{(\lambda + 4R)}{M} \cdot \sum_{u_i \in \mathcal{U}} D_{TV}(\hat{\mathbb{P}}^{(i)}, \tilde{\mathbb{P}}^{(i)}) \\ 
	&+ \phi \cdot  \frac{R}{M} \cdot \sqrt{\frac{\log2/\delta}{2}} \cdot \sqrt{\frac{1}{\tilde{N}}}
	\end{aligned}\label{thm1:eq51}
	\end{equation}
	where $\hat{\mathbb{P}}^{(i)}$ is the original distribution with $\hat{\mathbb{P}}^{(i)}_{ik} \equiv \frac{1}{n^+_in^-_i}$; $D_{TV}(\hat{\mathbb{P}}^{(i)}, \tilde{\mathbb{P}}^{(i)}) = \frac{1}{2} \cdot\left\|\hat{\mathbb{P}}^{(i)} - \tilde{\mathbb{P}}^{(i)} \right\|_1$ is the per-user \textit{Total Variance (TV)}  between two probability distributions $\hat{\mathbb{P}}^{(i)}$ and $\tilde{\mathbb{P}}^{(i)}$ on $\mcs$ for a specific user $u_i$, which characterizes the difference between two probability distributions. 
\end{thm}
\begin{pf} The proof of Thm.\ref{theo3} follows those of Thm.\ref{theo1} and Thm.\ref{thm3}, and we refer the readers to see more details in the Appendix \ref{SFCML:sampling-based}. 
\end{pf}

It is easy to see that the generalization upper bound for sampling-based CML has an extra term 
\[\frac{(\lambda + 4R)}{M} \cdot \sum_{u_i \in \mathcal{U}} D_{TV}\left(\hat{\mathbb{P}}^{(i)}, \tilde{\mathbb{P}}^{(i)}\right),\]
which captures the distance between the sampling-strategy-induced distribution and the ground-truth distribution for per user. This brings about a biased estimation.

\subsection{Summary and Discussion} \label{sec4.5}
Thm.\ref{theo3} intuitively reveals the shortcomings of sampling-based CML, due to the extra bias term $D_{TV}(\hat{\mathbb{P}}^{(i)}, \tilde{\mathbb{P}}^{(i)})$. This term describes the distribution deviation between leveraging whole samples for learning and adopting the negative sampling strategy. In order to eliminate this deviation and improve the performance of recommendation, we must learn from all samples instead of adopting negative sampling. \textit{This motivates us to develop an efficient alternative without negative sampling.}

\section{Sampling-Free Acceleration} \label{CAL: method}
Following our theoretical results, we propose an efficient Sample-Free CML (SFCML) acceleration algorithm in this section.

%Since we focus on developing an efficient algorithm for CML without negative sampling, we could set $p(\bm{v}_j^+, \bm{v}_k^-) \equiv \frac{1}{n_i^+n_i^-}>0$ to obtain our full samples-based algorithm following Eq.(\ref{e5}). 

\subsection{Modifying the Pairwise Loss}  \label{sec.5.1}
Since we focus on developing an efficient algorithm for CML without negative sampling, our acceleration method is based on a modification of the full samples-based CML loss function Eq.(\ref{e4}). First of all, we replace the hinge loss with widely adopted square loss $\ell_{sq}(x) = (\lambda - x)^2$. In addition, without loss of generality, we restrict the bounded norm of all users and items on a $R$-radius hyper-sphere rather than a bounded ball. Putting them into Eq.(\ref{e4}), we arrive at our new sampling-free CML loss function:
\begin{equation}
\begin{aligned}
\hat{\mathcal{R}}_{\mathcal{S}}^{\text{sfcml}}(f) &= \frac{1}{M} \sum_{u_i \in \mcu} \frac{1}{n_i^+n_i^-} \sum_{j=1}^{n_i^+} \sum_{k=1}^{n_i^-} \ell_{sq}^{(i)}(v_j^+, v_k^-),  \\
s.t. \left\|\boldsymbol{e}_{u_i}\right\|^2 &= R, \ \ \ \left\|\boldsymbol{e}_{v_j}\right\|^2 = R, \ \ u_i \in \mathcal{U}, v_j \in \mathcal{V},
\end{aligned}\label{eq6}
\end{equation}
where 
\begin{equation}
\begin{aligned}
\ell_{sq}^{(i)}(v_j^+, v_k^-) &= \left(\lambda + \boldsymbol{d}(i,j) - \boldsymbol{d}(i, k)\right)^2.
\end{aligned}\label{square_margin}
\end{equation}
and $\lambda > 0$ could also be regarded as the safe margin.
Since we let $\boldsymbol{e}_{u_i}$ and $\boldsymbol{e}_{v_j}$ distribute on a hyper-sphere with $R$ radius ($R=1.0$ in the experiment), i.e., $\left\|\boldsymbol{e}_{u_i}\right\|^2 = R$ and $\left\|\boldsymbol{e}_{v_j}\right\|^2 = R$, $\ell_{sq}^{(i)}(v_j^+, v_k^-)$ could be further simplified as
\[
\ell_{sq}^{(i)}(v_j^+, v_k^-) = \left(\lambda - 2\boldsymbol{e}_{u_i}^\top(\boldsymbol{e}_{v_j^+} - \boldsymbol{e}_{v_k^-})\right)^2.
\]
Therefore, the final CML loss function is modified as follows:
\begin{equation}
	\begin{aligned}
		\hat{\mathcal{R}}_{\mathcal{S}}^{\text{sfcml}}(f) &= \frac{1}{M} \sum_{u_i \in \mcu} \frac{1}{n_i^+n_i^-} \sum_{j=1}^{n_i^+} \sum_{k=1}^{n_i^-} \left(\lambda - 2\boldsymbol{e}_{u_i}^\top(\boldsymbol{e}_{v_j^+} - \boldsymbol{e}_{v_k^-})\right)^2,  \\
		\ \ \ s.t. & \left\|\boldsymbol{e}_{u_i}\right\|^2 = R, \ \ \ \left\|\boldsymbol{e}_{v_j}\right\|^2 = R, \ \ u_i \in \mathcal{U}, v_j \in \mathcal{V}.
	\end{aligned}\label{eq17}
\end{equation}

% 
%Intuitively, optimizing Eq.(\ref{eq17}) could still leverage the distance restrictions in Eq.(\ref{Eq3}). The only difference here is that the Euclidean distance could be reformulated as the inner product ($f(v_j|u_i) = 2\boldsymbol{e}_{u_i}^\top \boldsymbol{e}_{v_j}$) for items/users embeddings located in a fixed-radius hyper-sphere. 

%Different from the vanilla CML,  we employ the squared loss instead of the hinge loss. 
%In the subsection, we will see that this leads to an efficient acceleration algorihtm.
%  

Motivated by the modified pairwise loss, the following corollary demonstrates that the bias term caused by the per-user TV term will vanish in the generalization upper bound of SFCML. This proves the effectiveness of our proposed SFCML method.
\begin{coro}\label{coro1}
	 (\textbf{Generalization Upper Bound of SFCML} \textbf{with Eq.(\ref{eq17})}). Given the sample set $\mcs = \mathop{\cup}\limits_{u_i \in \mcu} \mcs_i$ where $\mcs_i = \{v_j^+\}_{j=1}^{n_i^+} \cup \{v_k^-\}_{k=1}^{n_i^-}, n_i^+ + n_i^- = N$, for any $\delta \in (0,1)$, with probability at least $1 - \delta$, the following inequation holds:
	\begin{equation}
	\begin{aligned}
	\mathcal{R}_{\ell}^{\text{cml}}(f) &\lesssim \hat{\mathcal{R}}^{\text{sfcml}}_{\mcs}(f) \\
	&+ (\lambda + 4R) \cdot \frac{\Max(\lambda, \sqrt{R \cdot d})}{M} \cdot \sqrt{\frac{1}{\tilde{N}}}   \\
	&+ \frac{(\lambda + 4R) \cdot R}{M} \cdot \sqrt{\frac{\log2/\delta}{2}} \cdot \sqrt{\frac{1}{\tilde{N}}}.
	\label{sfcml:final}
	\end{aligned}
	\end{equation}
\end{coro}

\begin{pf} 
	 Note that, changing the constraint of embeddings from $\left\|\boldsymbol{e}_{u_i}\right\|^2 \le R, \left\|\boldsymbol{e}_{v_j}\right\|^2 \le R$ to $\left\|\boldsymbol{e}_{u_i}\right\|^2 = R, \left\|\boldsymbol{e}_{v_j}\right\|^2 = R$ will not change the result of Thm.\ref{theo1}, since the constraints are only employed to bound the supremum. It is easy to show that $\ell_{sq}^{(i)}$ is $(\lambda + 4R)$-Lipschitz continuous. 
	Then the proof is completed via setting $\phi = \lambda + 4R$ in Thm.\ref{theo1}.
	\qed
\end{pf}

%Meanwhile, from another perspective, when completing the optimization, the score of observed items will be larger than the unobserved items, and thus one can adopt $f(v_j|u_i) = 2\boldsymbol{e}_{u_i}^\top \boldsymbol{e}_{v_j}$ to carry out the recommendation for a specific user $u_i$. In addition, by enforcing all embedding in this space to have the same $\ell_2$-norm, the model only focuses on learning angular information, which can improve the robustness and discriminate power of the metric. It is also adopted in \textit{Face Recognition} (FR) problems \cite{DBLP:conf/cvpr/WangWZJGZL018, DBLP:conf/cvpr/LiuWYLRS17}.

%It is also an extensively adopted technique, such as \textit{Network Embedding} (NE) \cite{DBLP:journals/pami/WilsonHPD14, DBLP:conf/nips/LiuZLLDZS17, DBLP:conf/kdd/MengZH0Z020}, \textit{Face Recognition} (FR) \cite{DBLP:conf/cvpr/WangWZJGZL018, DBLP:conf/cvpr/LiuWYLRS17} and \textit{Collaborative Filtering} (CF) \cite{tran2019improving}. We refer the readers to Sec.\ref{sec3.6} for more interpretations. 

%Needless to say, in order to ensure the strictness and fairness of the experiments, in Sec.\ref{sec4s}, in terms of CML with sampling strategies, we also normalize all the users and items embeddings to the $R$ sphere instead of being bounded within the $R$-ball.

%that the coordinate origin at the centre of the hypersphere 

\subsection{Efficient Alternative without Negative Sampling} \label{sec.5.2}
With the help of the squared loss, it is interesting to note that Eq.(\ref{eq17}) could be regarded as an AUC optimization problem \cite{DBLP:conf/icml/GaoJZZ13, DBLP:conf/icdm/ZhouYS20}. Recall that area under the ROC curve (AUC) measures the probability of the score of a positive sample higher than a negative sample \cite{yang2020stochastic, hanley1982meaning, DBLP:conf/nips/CortesM03}. Then, assuming there are no ties in the scores of samples, AUC could be optimized by the following empirical minimization problem \cite{DBLP:conf/icml/GaoJZZ13, DBLP:conf/icml/ZhaoHJY11}: 
$$
\min_{f} \frac{1}{n_+n_-}\sum_{x_+} \sum_{x_-} \ell_{0-1}\left(f(x_+) - f(x_-) \right).
$$
 Generally speaking, $x_+$ ($x_-$) denotes the positive (negative) samples, and $n_+$ ($n_-$) represents the number of positive (negative) instances. $f$ is the score/decision function representing the probability of an instance to be predicted as a positive sample. $\ell_{0-1}(z)$ is the 0-1 loss, which returns 1 if $z < 0$, otherwise $0$ is returned. Since $\ell_{0-1}$ is not continuous, it is often replaced by a continuous surrogate loss $\ell_{sur}$ \cite{DBLP:conf/ijcai/GaoZ15, ying2016stochastic, DBLP:conf/uai/LyuY18}, which induces a surrogate AUC optimization problem:
\begin{equation}
	\min_{f} \frac{1}{n_+n_-}\sum_{x_+} \sum_{x_-} \ell_{sur}\left(f(x_+) -  f(x_-)\right). \label{auc_rs}
\end{equation}
Therefore, if we regard $f$ and $\ell_{sur}$ in Eq.(\ref{auc_rs}) as $f(v_j|u_i) = 2\boldsymbol{e}_{u_i}^\top\boldsymbol{e}_{v_j}$ and $\ell_{sq}$ respectively, it then recovers our objective function.

Next, we will elaborate on how to develop an efficient algorithm without negative sampling strategies. Note that, since the loss function is calculated separately for different users, we only consider one specific user (taking $u_i \in \mathcal{U}$ as an example), while the overall objective function could simply be obtained by taking an average. For the convenience of the subsequent description, we let $\hat{\mathcal{R}}_{\mcs_i}^{\text{sfcml}}(f)$ be the empirical risk of the specific user $u_i$. 

At first, we can find that, for every single item $v_j^+ \in \mathcal{S}_i^+$, it is only paired with the remaining negative items $v_k^- \notin \mathcal{S}_i^+$. This observation helps us to decouple the time-consuming pairwise computation and develop our efficient algorithm SFCML. Let us construct a graph defined as $\mathcal{G}^{(i)} = (\mathcal{S}^{(i)}, \mathcal{E}^{(i)}, \mathcal{D}^{(i)})$, where the vertex set $\mathcal{S}^{(i)} = \{(v_j, y_{ij})|v_j \in \mathcal{V}\}$ is the set of preferences for user $u_i$, $\mathcal{E}^{(i)} = \{(j, k) | y_{ij} \ne y_{ik}\}$ is the edge of the graph and $\mathcal{D}^{(i)}$ is the adjacent matrix  defined as follows:

\begin{equation}
\mathcal{D}^{(i)}_{jk}=
\begin{cases}
\ \ \ \frac{1}{n_i^+n_i^-}, & (j, k) \in \mathcal{E}^{(i)}, \\
\ \ \ \ \ \ \ \ 0, & (j, k) \notin \mathcal{E}^{(i)}.
\end{cases}\label{Eq6}
\end{equation}

Subsequently, the graph Laplacian matrix could be defined as:
\begin{equation}
\boldsymbol{L}^{(i)} = diag(\mathcal{D}^{(i)}\boldsymbol{1}) - \mathcal{D}^{(i)}, \label{Eq7}
\end{equation} 
where $\boldsymbol{1} \in \mathbb{R}^{N}$ is an all-one vector; $diag(\mathcal{D}^{(i)}\boldsymbol{1})\in \mathbb{R}^{N \times N}$ is a diagonal matrix, with the $diag(\mathcal{D}^{(i)}\boldsymbol{1})_{i,i}$ representing the degree of vertex $i$ in the graph.

Based the graph-theoretic machinery, Eq.(\ref{eq17}) can be reformulated as follows:
\begin{equation}
\begin{aligned}
& \hat{\mathcal{R}}_{\mcs_i}^{\text{sfcml}}(f) = (\boldsymbol{f}^{(i)} - \lambda \cdot \boldsymbol{y}^{(i)} )^\top\boldsymbol{L}^{(i)}(\boldsymbol{f}^{(i)} - \lambda \cdot \boldsymbol{y}^{(i)} ), \\
& s.t. \ \ \left\|\boldsymbol{e}_{u_i}\right\|^2 = R, \ \ \ \left\|\boldsymbol{e}_{v_j}\right\|^2 = R, \ \  v_j \in \mathcal{V},
\end{aligned}\label{eq8}
\end{equation}
where $\boldsymbol{y}^{(i)} = [y_{i1}, y_{i2}, \dots, y_{iN}]^\top$ is a vector representing the preferences of user $u_i$ toward all items, $\boldsymbol{f}^{(i)} = [2\boldsymbol{e}_{u_i}^\top\boldsymbol{e}_{v_1}, 2\boldsymbol{e}_{u_i}^\top\boldsymbol{e}_{v_2}, \dots, 2\boldsymbol{e}_{u_i}^\top\boldsymbol{e}_{v_N}]^\top$ could be seen as a score vector in terms of all items.

\textit{However, Eq.(\ref{eq8}) still brings a rather heavy computation burden due to the inefficiency of naive matrix multiplication of Eq.(\ref{eq8}), almost $\mco(N^2 + N)$ for each user. }

This drives us to adopt the following proposition to further accelerate the calculation.
\begin{prop}\label{prop1}
	Let both $p$ and $q$ be the positive integers. Then, for any matrix $\boldsymbol{P} \in \mathbb{R}^{N \times p}$ and $\boldsymbol{Q} \in \mathbb{R}^{N \times q}$, the calculation of $\boldsymbol{P}^\top\boldsymbol{L}^{(i)} \in \mathbb{R}^{p \times N}$ could be nearly finished within $\mco(pN)$ and $\boldsymbol{P}^\top\boldsymbol{L}^{(i)}\boldsymbol{Q} \in \mathbb{R}$ could be almost completed within $\mco(pqN)$.
\end{prop}

\begin{pf}
	Firstly, according to the above definition, we can show that, $\mathcal{D}^{(i)}$ can be reformulated as
	\begin{equation}
	\mathcal{D}^{(i)} = \frac{1}{n_{i}^+n_{i}^-}[\boldsymbol{y}^{(i)}(\boldsymbol{1} - \boldsymbol{y}^{(i)})^\top + (\boldsymbol{1} - \boldsymbol{y}^{(i)})(\boldsymbol{y}^{(i)})^\top], \label{eq9}
	\end{equation}
	where $\boldsymbol{1} \in \mathbb{R}^N$ is a vector where all values are $1$.
	
	Meanwhile, given $\mathcal{D}^{(i)}$, $\boldsymbol{L}^{(i)} \in \mathbb{R}^{N \times N}$ could be rewritten as 
	\begin{equation}
	\begin{aligned}
	\boldsymbol{L}^{(i)} &= diag(\mathcal{D}^{(i)}\boldsymbol{1}) - \mathcal{D}^{(i)} \\
	&= diag\left(\frac{\boldsymbol{y}^{(i)}}{n_{i}^+} + \frac{(\boldsymbol{1} -\boldsymbol{y}^{(i)})}{n_{i}^-}\right) - \mathcal{D}^{(i)}
	\end{aligned}\label{eq10}
	\end{equation}
	Correspondingly, according to Eq.(\ref{eq10}), the following equation holds for $\boldsymbol{P}^\top\boldsymbol{L}^{(i)} \in \mathbb{R}^{p \times N}$
	\begin{equation}
		\begin{aligned}
			\boldsymbol{P}^\top\boldsymbol{L}^{(i)} =& \boldsymbol{P}^\top \bigg(diag\left(\frac{\boldsymbol{y}^{(i)}}{n_{i}^+} + \frac{(\boldsymbol{1} -\boldsymbol{y}^{(i)})}{n_{i}^-}\right)\bigg) \\
			&- \frac{\boldsymbol{P}^\top\boldsymbol{y}^{(i)}(\boldsymbol{1} - \boldsymbol{y}^{(i)})^\top}{n_{i}^+n_{i}^-} \\
			&- \frac{\boldsymbol{P}^\top(\boldsymbol{1} -\boldsymbol{y}^{(i)}) (\boldsymbol{y}^{(i)})^\top}{n_{i}^+n_{i}^-}
		\end{aligned} \label{new24}
	\end{equation}
	According to Eq.(\ref{new24}), it is easy to conclude that $\boldsymbol{P}^\top\boldsymbol{L}^{(i)}$ could be completed within almost $\mco(pN)$ while it should be computed within almost $\mco(pN^2)$ with the naive matrix multiplication.
	
	In addition, with the acceleration of $\boldsymbol{P}^\top\boldsymbol{L}^{(i)} \in \mathbb{R}^{p \times N}$, it is obvious to show that the calculation of $\boldsymbol{P}^\top\boldsymbol{L}^{(i)}\boldsymbol{Q} \in \mathbb{R}$ is reduced from $\mco(pqN^2)$ to almost $\mco(pqN)$. 
	
	This proved the proposition.\qed
\end{pf}
\begin{rem}
	Equipped with Prop.\ref{prop1}, Eq.(\ref{eq8}) could be almost finished within $\mco(N) = \mco(n_i^+ + n_i^-)$ by replacing either of the two parts in Eq.(\ref{eq8}), i.e., let $\boldsymbol{P} =  (\boldsymbol{f}^{(i)} - \lambda \cdot \boldsymbol{y}^{(i)} ) \in \mathbb{R}^N$ or $\boldsymbol{Q} = (\boldsymbol{f}^{(i)} - \lambda \cdot \boldsymbol{y}^{(i)} ) \in \mathbb{R}^N$. This is a significant efficiency improvement against naive CML with $\mco(n_i^+n_i^-)$ time complexity for each user, due to $n_i^+ + n_i^- \ll n_i^+n_i^-$ in the practical RS. At the same time, Prop.\ref{prop1} also guarantees that our algorithm SFCML enjoys practically $\mco(N)$ space complexity per user, while the naive CML (refer to Eq.(\ref{e4})) has almost $\mco(n_i^+n_i^-) = \mco(N^2)$ space complexity per user.
\label{sec5.2_remark_3}
\end{rem}

\subsection{Optimization and Algorithm} \label{CAL:opt}
\subsubsection{The Overall Objective Function}
Now we come to the objective function of our efficient alternative SFCML with all users by taking an average
\begin{equation}
\begin{aligned}
& \hat{\mathcal{R}}_{\mcs}^{\text{\text{sfcml}}}(f) = \frac{1}{M}\sum_{u_i \in \mathcal{U}} \hat{\mathcal{R}}^{\text{\text{sfcml}}}_{\mcs_i}(f), \\
s.t. \ \ ||\boldsymbol{e}_{u_i}||^2 &= R, \ \ \ ||\boldsymbol{e}_{v_j}||^2 = R, \ \ u_i \in \mathcal{U}, v_j \in \mathcal{V}.
\end{aligned}\label{eq14}
\end{equation}

%Thanks to the property of Prop.\ref{prop1}, the well-designed algorithm FCML could be completed within $\mco(M \cdot (n_i^+ + n_i^-))$ for all samples without any negative sampling strategies, while the conventional CML to do this is within $\mco(M\cdot n_i^+n_i^-)$. Note that, $(n_i^+ + n_i^-)$ is usually much smaller than $n_i^+n_i^-$, i.e., $(n_i^+ + n_i^-) \ll n_i^+n_i^-$, which brings several orders of magnitude efficiency improvements under the context of practical recommendation systems. 

\subsubsection{Optimization}
We employ the gradient descent method as the optimizer to learn our proposed algorithm SFCML. The optimization of user $u_i$ and items' weight $\bm{W}_v$ could be summarized as follows.

\noindent\textbf{Optimization of user $u_i$}. In order to minimize Eq.(\ref{eq14}), we first rewrite the score function of $u_i$ as $\boldsymbol{f}^{(i)} = 2\bm{W}_v\boldsymbol{e}_{u_i}$ where $\bm{W}_v \in \mathbb{R}^{N \times d}$ is the learned transformation weight (review Eq.(\ref{emb})). Then, the gradient descent method updates the variable $\boldsymbol{e}_{u_i}$ according to:
\begin{equation}
\begin{aligned}
\boldsymbol{e}_{u_i} = \boldsymbol{e}_{u_i} - \eta \cdot \nabla_{\boldsymbol{e}_{u_i}}\hat{\mathcal{R}}^{\text{sfcml}}_{\mcs_i}(f)
\end{aligned} \label{CAL:e_u_i}
\end{equation}
where $\eta$ is the learning rate, and
\begin{equation*}
\nabla_{\boldsymbol{e}_{u_i}}\hat{\mathcal{R}}^{\text{sfcml}}_{\mcs_i}(f) = 2\bm{W}_v^\top \boldsymbol{L}^{(i)} \left(\boldsymbol{f}^{(i)} - \lambda \cdot \boldsymbol{y}^{(i)} \right).
\end{equation*}
Note that, here we just present the derivation of $\boldsymbol{e}_{u_i}$. One can easily match the $\boldsymbol{e}_{u_i}$ and $\bm{W}_u$ based on Eq.(\ref{emb}), i.e., $\boldsymbol{e}_{u_i}$ corresponding the $i$-th row in the matrix $\bm{W}_u$. 

\begin{rem} Owing to the strengths of Prop.\ref{prop1}, the computation complexity of Eq.(\ref{CAL:e_u_i}) is still reasonable, which could be almost completed within $\mco(dN) = \mco(d(n_i^+ + n_i^-))$ for a specific user $u_i$.
\end{rem}

\noindent\textbf{Optimization of items' weight $\bm{W}_v$}. In the same way, the gradient descent method updates the variable $\bm{W}_v$ according to:
\begin{equation}
\begin{aligned}
\bm{W}_v = \bm{W}_v - \eta \cdot \nabla_{\bm{W}_v}\hat{\mathcal{R}}^{\text{sfcml}}_{\mcs_i}(f)
\end{aligned}\label{CAL:v_j}
\end{equation}
where 
\[
\nabla_{\bm{W}_v}\hat{\mathcal{R}}^{\text{sfcml}}_{\mcs_i}(f) = 2\boldsymbol{L}^{(i)} \left(\boldsymbol{f}^{(i)} - \lambda \cdot \boldsymbol{y}^{(i)} \right)\boldsymbol{e}_{u_i}^\top
\]

\begin{rem}
	Similarly, following the Prop.\ref{prop1}, we could demonstrate that, for any  $\boldsymbol{Q} \in \mathbb{R}^{N \times q}$ where $q$ is a positive integer, $\boldsymbol{L}^{(i)}\boldsymbol{Q} \in \mathbb{R}^{N \times q}$ could be finished within $\mco(qN)$. According to this, by setting $\boldsymbol{Q} = (\boldsymbol{f}^{(i)} - \lambda \cdot \boldsymbol{y}^{(i)}) \in \mathbb{R}^N$, Eq.(\ref{CAL:v_j}) could also be updated within $\mco(dN + N)= \mco\left((d+1)(n_i^+ + n_i^-)\right)$ for all items' embeddings.
\end{rem}

Finally, we summarize all the details of SFCML in  Alg.\ref{algorithm1}.

\begin{algorithm}[!t]
	\caption{Sampling-Free Collaborative Metric Learning}
	\label{algorithm1}
	\LinesNumbered
	\KwIn{User set $\mathcal{U} = \{u_1, u_2, \dots, u_M\}$}
	\KwIn{Item set $\mathcal{V} = \{v_1, v_2, \dots, v_N\}$} 
	\KwIn{Preference set: $\{\boldsymbol{y}^{(i)}| u_i \in \mcu\}$} 
	\KwIn{Hypersphere radius: $R$}
	\KwIn{Safe margin $\lambda$} 
	\KwIn{Learning rate $\eta$} 
	\KwOut{User transformation matrix: $\bm{W}_u$}
	\KwOut{Item transformation matrix: $\bm{W}_v$}
	Initialize $\bm{W}_u$\;
	Initialize $\bm{W}_v$\;
	\While{Not Converged}{
		Restrict the norm of all users' and items' embeddings on a $R$-radius hypersphere\;
		\For{$u_i$ in user set $\mathcal{U}$}{
			Project user $u_i$ and items into the metric space by Eq.(\ref{emb})\;
			Index the corresponding preference set $\boldsymbol{y}^{(i)}$\;
			Optimize the weights via Eq.(\ref{CAL:e_u_i}) and Eq.(\ref{CAL:v_j})\;
			
			Restrict the norm of all items' embeddings on a $R$-radius hypersphere\;
		}
	}
	Restrict the norm of all users' embeddings on a $R$-radius hypersphere\;
	\Return $\bm{W}_u$ and $\bm{W}_v$
\end{algorithm}

\begin{table*}[]
	\centering
	\setlength{\abovecaptionskip}{6pt}    
	\setlength{\belowcaptionskip}{15pt}    
	\setlength{\tabcolsep}{9pt}
	\caption{Basic Information of the Datasets. \%Density is defined as $\frac{\#Ratings}{\#Users \times \#Items} \times 100\% $.}	
	\label{table1}
	\scalebox{0.95}{
	\begin{tabular}{c|ccccccc}
		\toprule
		Datasets & MovieLens-100K & CiteULike-T & MovieLens-1M & Steam-200k & Anime & MovieLens-20M & Amazon-Book\\
		\midrule
		Domain & Movie & Paper & Movie & Game & Anime & Movie & Book\\
		\#Users & 938 & 7,947 & 11,209  & 3,757 & 54,190 & 136,677 & 64,937 \\
		\#Items & 1,447 & 25,975 & 7,491 & 5,113 & 6,967 & 17,679 & 181,152\\
		\#Ratings & 55,361 & 125,580 & 85,341  & 115,139 & 7,634,542 & 9,986,829 & 2,880,930\\
		\%Density & 4.0788  & 0.0608 & 0.1016 & 0.5994 & 2.0221 & 0.4133 & 0.0245\\
		\bottomrule
	\end{tabular}
}
\end{table*}

\section{Experiments \label{CAL: exp}}

In this section, we conduct comprehensive experiments on a wide range of benchmark datasets to show the superiority of our proposed method. 
\subsection{Dataset Descriptions} \label{data_descr}

We perform the empirical studies over seven widely adopted benchmark datasets to evaluate the performance, including:
\begin{itemize}

	\item \textbf{MovieLens} \footnote{\url{https://grouplens.org/datasets/movielens/}} - A series of benchmark datasets are popularly and widely used in RS. There are many versions of MovieLens, and we adopt  \textbf{MovieLens-100k}\footnote{\url{https://grouplens.org/datasets/movielens/100k/}}, \textbf{MovieLens-1m}\footnote{\url{https://grouplens.org/datasets/movielens/1m/}} and \textbf{MovieLens-20m}\footnote{\url{https://grouplens.org/datasets/movielens/20m/}} here to test the performance. Specifically, it includes ratings ranging from 1 to 5 on various movies. Following the previous work \cite{DBLP:conf/kdd/WangWY15, hsieh2017collaborative}, if the score of item $v_j$ rated by user $u_i$ is no less than 4, we regard item $v_j$ as a positive item for user $u_i$. \\

	\item \textbf{CiteULike}\footnote{\url{http://www.citeulike.org/faq/data.adp}} \cite{DBLP:conf/ijcai/WangCL13} - An implicit feedback dataset that allows users to create their own collections of articles. There are two configurations of CiteULike collected from CiteULike and Google Scholar. Following \cite{hsieh2017collaborative}, we adopt \textbf{CiteULike-T} here to evaluate the performance.
	\item \textbf{Steam-200k\footnote{\url{https://www.kaggle.com/tamber/steam-video-games}}} - This dataset is collected from the Steam which is the world's most popular PC gaming hub. The observed behaviors of users include 'purchase' and 'play' signals. In order to obtain the implicit feedback, if user has purchased a game as well as the playing hours $play > 0$, we treat this game as a positive item. 
	\item\textbf{Anime\footnote{\url{https://www.kaggle.com/CooperUnion/anime-recommendations-database}}} - This dataset is collected by the myanimelist.net API which records the preferences of users toward several animes. Its ratings range from $-1$ to $10$, where $-1$ represents the user watched an anime but didn't make an assessment for it. Moreover, the higher the ratings for an item is, and the more the user likes it. Similarly, if the item $v_j$'s ratings produced by user $u_i$ is no less than 5, we regard item $v_j$ as a positive item in terms of user $u_i$. 
	\item\textbf{Amazon-Book \footnotemark} \cite{DBLP:conf/www/HeM16} - This dataset includes ratings and various metadata collected from Amazon. The ratings therein range from 1 to 5. We conduct the same pre-process as MoiveLens to obtain the implicit signals. 
\footnotetext{\url{https://jmcauley.ucsd.edu/data/amazon/}}
\end{itemize}
More detailed statistics with respect to these datasets are summarized in Tab.\ref{table1}.

\subsection{Competitors}
Note that, the starting point of this work is to develop an efficient CML-based algorithm without negative sampling to get rid of the bias caused by the sampling-based CML. Therefore, to show the superiority of the proposed algorithm, we evaluate SFCML against the following 12 competitors:
\begin{itemize}
	\item \textbf{itemKNN} \cite{linden2003amazon, DBLP:conf/www/SarwarKKR01} is a simple but classical item-based collaborative filtering method. It recommends new items to the target user based on the similarities with his/her interacted items. Generally speaking, one usually adopts the cosine function to measure the similarities between different items.
%	\item \textbf{BPR-MF} \cite{DBLP:conf/uai/RendleFGS09} is one ofthe classical matrix factorization methods, which optimizes the pairwise ranking between the positive and negative items.
	\item \textbf{Generalized Matrix Factorization} (GMF) can be regarded as a generalized and extended MF method, which is more expressive than the traditional  MF algorithm. It is one of the instantiates in \cite{DBLP:conf/www/HeLZNHC17}, which applies a linear kernel to model the latent user-item interactions.  
	\item \textbf{Multi-Layer Perceptron} (MLP) is a deep learning-based framework \cite{DBLP:conf/www/HeLZNHC17}, which adopts a non-linearity multilayer perceptron to learn the interaction between users and items. In this way, the model could be endowed with reasonable flexibility and non-linearity to capture the preference of users. 
	\item \textbf{Neural network-based Collaborative Filtering} (NCF)\footnotemark \cite{DBLP:conf/www/HeLZNHC17} is a popular and competitive deep learning-based framework bridging the gap of GMF and MLP. NCF concatenates the output of GMF and MLP, and regards the recommendation task as a regression problem. Notably, it computes the ranking scores with a neural network instead of the inner product. 
	\item \textbf{Uniform Negative Sampling} (UniS) \cite{DBLP:conf/icdm/PanZCLLSY08,hsieh2017collaborative} leverages a uniform negative Sampling strategy to alleviate the heavy burden of computations for CML. Specifically, for every user, uniformly sample $U$ items from unobserved interactions as negatives to minimize Eq.(\ref{e5}).
	\item \textbf{Popularity-based Negative Sampling} (PopS) \cite{DBLP:conf/kdd/ChenSSH17, DBLP:conf/sigir/WuVSSR19, tran2019improving} optimizes the pairwise ranking loss Eq.(\ref{e5}) with a popularity-based negative sampling strategy, i.e., sampling $U$ negative candidates from unobserved interactions based on their frequencies.
	\item  \textbf{Two-Stage Negative Sampling} (2stS) \footnotemark \cite{tran2019improving} is an effective and competitive method. To increase the number of informative items to optimize Eq.(\ref{e5}), 2stS adopts a two-stage sampling strategy. Firstly, a candidate set of items are sampled based on their popularity. Secondly, according to their inner product values with anchors (positive items), the most informative samples are selected from this candidate.
	\item \textbf{Hard Negative Sampling} (HarS) \footnotemark \cite{hsieh2017collaborative} is similar to the negative sample mining process widely adopted in the object detection \cite{DBLP:conf/acml/CanevetF14, DBLP:conf/sigir/WangYZGXWZZ17, DBLP:conf/www/ParkC19, DBLP:conf/ijcai/DingQ00J19}. Specifically, it includes two stages: 1) uniformly sample $U$ candidates from unobserved items; 2) select the hard item from the candidates as negative item to train based on the distance between targeted user and items. Note that, when the number of $U$ is set as $1$, the HarS is the same as the UniS strategy.
		\item \textbf{Collaborative Translational Metric Learning (TransCF)} \cite{DBLP:conf/icdm/ParkKXY18} is a translation-based method. Specifically, such translation-based algorithms employ $\bm{d}(i, j) = ||\boldsymbol{e}_{u_i} + \boldsymbol{e}_{r_{ij}} - \boldsymbol{e}_{v_j}||^2$ as the distance/score between user $u_i$ and item $v_j$ instead of $||\boldsymbol{e}_{u_i} - \boldsymbol{e}_{v_j}||^2$, where $\boldsymbol{e}_{r_{ij}}$ is a specific translation vector for $u_i$ and $v_j$. In light of this, TransCF discovers such user–item translation vectors via the users' relationships with their neighbor items.
	\item \textbf{Latent Relational Metric Learning (LRML)} \cite{DBLP:conf/www/TayTH18} is also a translation-based CML method. As a whole, the key idea of LRML is similar to TransCF. The main difference is how to access the translation vectors effectively. Concretely, TransCF leverages the neighborhood information of users and items to acquire the translation vectors while LRML introduces an attention-based memory-augmented neural architecture to learn the exclusive and optimal translation vectors. 
	\item \textbf{Co-occurrence embedding Regularized Metric Learning (CRML)} \cite{DBLP:journals/nn/WuZNC20} considers the global statistical information of user-user and item-item pairs by involving a co-occurrence embedding to regularize the metric learning model. Then, CRML regards the optimization problem as a multi-task learning problem to boost the performance of CML, including the primary CML recommendation task and two auxiliary representation learning tasks. 
	\item \textbf{Efficient Heterogeneous Collaborative Filtering (EHCF)}  \cite{DBLP:conf/aaai/ChenZZMLM20} is a state-of-the-art non-sampling-based neural framework, which presents a sampling-free strategy to optimize an NCF-like model using the whole heterogeneous data without negative sampling.
%	\item \textbf{Naive Collaborative Metric Learning (NaiveCML)} is the initial method that does not adopt any negative sampling strategies to acceleration its computations, which can be treated as the upper bound performance here. Unfortunately, due to the limitation of the memory and heavy computation burden on larger dataset, we only report the experimental results on two tiny datasets, i.e., MovieLens-100k and Steam-200k. 
\end{itemize}

\footnotetext[9]{\url{ https://github.com/guoyang9/NCF}}
\footnotetext[10]{\url{https://github.com/deezer/sigir2019-2stagesampling}}
\footnotetext[11]{\url{https://github.com/changun/CollMetric}}

\noindent \textbf{Discussions of the competitors}. The most related competitors to our proposed SFCML roughly fall into two groups: \textbf{a)} \textbf{Sampling-based CML methods}, including UniS, PopS, 2stS, HarS, TransCF, LRML and CRML. \textbf{b)} \textbf{The state-of-the-art sampling-free algorithms}, i.e., EHCF. Our work differs from both of them. In terms of a), some of them (i.e., UniS, PopS, 2stS and HarS) try to improve the performance of CML by directly developing more effective negative sampling strategies. The others (including TransCF, LRML and CRML) introduce more complicated structures or auxiliary statistical information to improve the CML. \textit{Yet they still need to adopt the negative sampling (usually employing one of the above-mentioned sampling strategies) in the training phase to alleviate the heavy burden of pairwise computations. Such circumstance implies they would still encounter the generalization problem more or less as discussed in Sec.\ref{sec4.5}.} Different from the existing sampling-based CML algorithms, SFCML attempts to boost the recommendation performance from a sampling-free aspect, i.e., directly optimize CML leveraging the whole data under a relatively acceptable efficiency. In terms of b), however, it still differs significantly from our work. Practically, EHCF presents an effective heterogeneous CF-based framework learned from the whole data instead of negative sampling. \textit{However, such a method is specifically tailored for the pointwise-based recommendation (such as NCF-like algorithms and some other factorization-based models), which is not suitable for the CML framework based on the pairwise ranking loss.}

\subsection{Evaluation Metrics}
In some typical recommendation systems, users often care about the top-$K$ items in recommendation lists, so the most relevant items should be ranked first as much as possible. In light of this, we evaluate the performance of competitors and our algorithm with the following extensively adopted six metrics: \textbf{Precision} (P@$K$), \textbf{Recall} (R@$K$), \textbf{Normalized Discounted Cumulative Gain} (NDCG@$K$), \textbf{Mean Average Precision} (MAP), \textbf{Mean Reciprocal Rank} (MRR) and \textbf{Area Under ROC Curve} (AUC). Note that, for all the above metrics, the higher the metric is, the better the performance the algorithm achieves. See Appendix.\ref{app_metrics} for more details.

\begin{figure}[!t]
	\centering
		\includegraphics[width=0.5\textwidth]{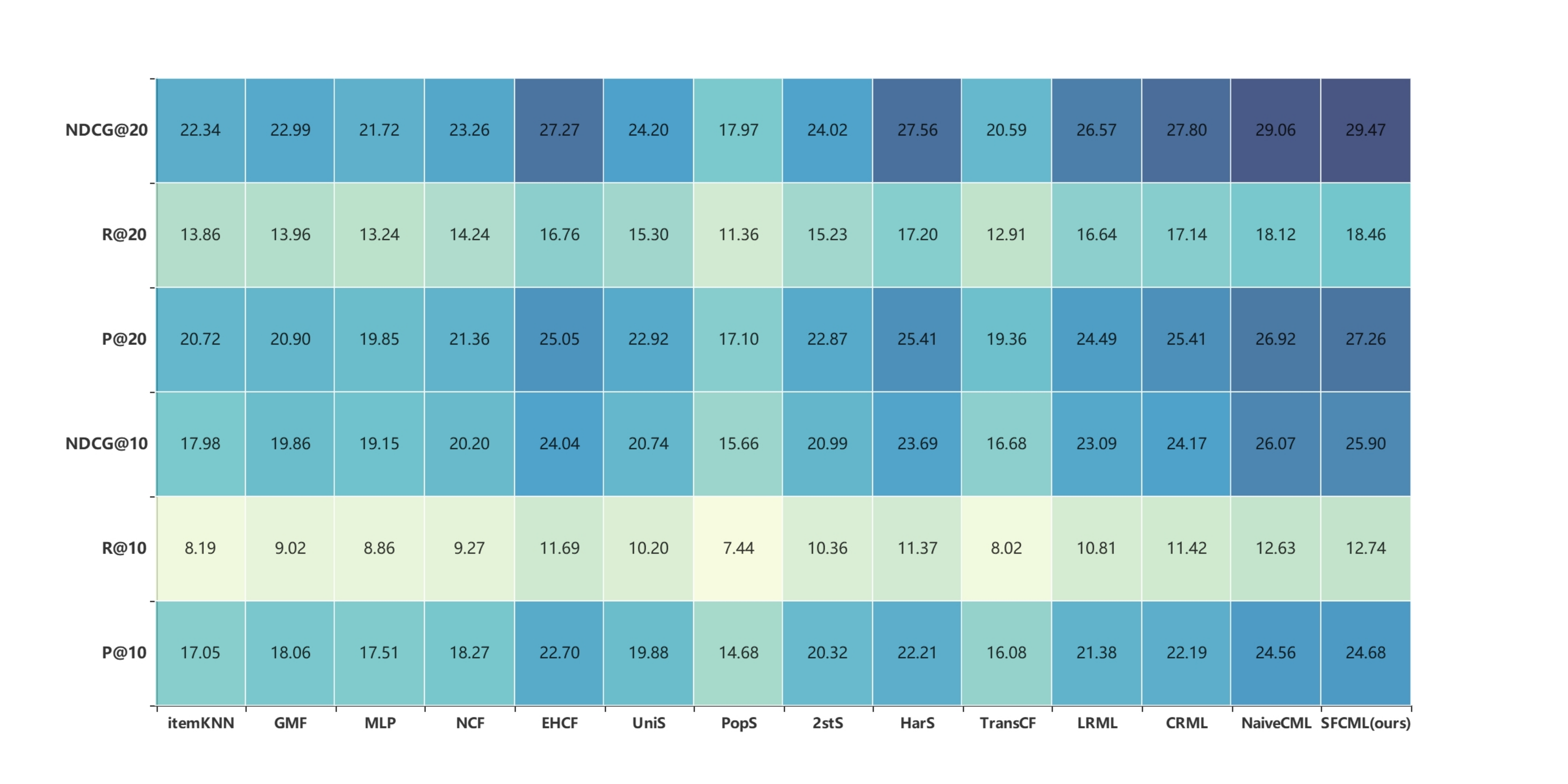}
		\caption{Heat map of performance results on MovieLens-100k in terms of $K=\{10, 20\}$. Please see Appendix.\ref{more_evaluation_results} for more results. }
		\label{ab}
\end{figure}

\begin{table*}[!ht]
	\centering
	\caption{Performance comparisons on MovieLens-100k, CiteULike and MovieLens-1m datasets, where '-' means that we cannot complete the experiments due to the out-of-memory issue. The best and second-best are highlighted in bold and underlined, respectively.}
	\scalebox{1.0}{
		\begin{tabular}{c|c|ccccccccc}
			\toprule
			& Method & \cellcolor[rgb]{ 1.0,  1.0,  1.0} P@3 & R@3 & NDCG@3 & P@5 & R@5 & NDCG@5 & MAP & MRR & AUC \\
			\midrule
			\multirow{15}[5]{*}{MovieLens-100k} & itemKNN & \cellcolor[rgb]{ 1.0,  1.0,  1.0} 11.35 & 2.41 & 11.57 & 12.96 & 4.11 & 13.45 & 8.49 & 24.63 & 85.68 \\
			& GMF & \cellcolor[rgb]{ 1.0,  1.0,  1.0} 14.35 & 3.37 & 15.20 & 16.43 & 5.79 & 17.21 & 9.82 & 31.00 & 86.12 \\
			& MLP & \cellcolor[rgb]{ 1.0,  1.0,  1.0} 14.98 & 3.93 & 15.57 & 15.51 & 5.70 & 16.54 & 10.09 & 31.99 & \cellcolor[rgb]{ .996,  .973,  .961}87.09  \\
			& NCF & \cellcolor[rgb]{ .996,  .973,  .961} 15.94  & \cellcolor[rgb]{ .996,  .973,  .961}4.11  & \cellcolor[rgb]{ .996,  .973,  .961}16.75  & \cellcolor[rgb]{ 1,  .961,  .937}17.26  & \cellcolor[rgb]{ 1,  .961,  .937}6.45  & \cellcolor[rgb]{ 1,  .961,  .937}18.25  & \cellcolor[rgb]{ .996,  .973,  .961}11.35  & \cellcolor[rgb]{ .996,  .973,  .961}34.34  & \cellcolor[rgb]{ .996,  .973,  .961}88.03  \\
			& EHCF & \cellcolor[rgb]{ .996,  .89,  .839} 21.13  & \cellcolor[rgb]{ .996,  .89,  .839}6.99  & \cellcolor[rgb]{ .996,  .89,  .839}21.80 & \cellcolor[rgb]{ .996,  .949,  .925}20.89 & \cellcolor[rgb]{ .988,  .925,  .882}8.82  & \cellcolor[rgb]{ .988,  .925,  .882}22.08  & \cellcolor[rgb]{ .996,  .89,  .839}16.51  & \cellcolor[rgb]{ .996,  .89,  .839}41.77  & \cellcolor[rgb]{ .996,  .89,  .839}92.18  \\
			\cmidrule{2-11}      & UniS & \cellcolor[rgb]{ 1,  .961,  .937} 15.94 & \cellcolor[rgb]{ 1,  .961,  .937}4.43  & \cellcolor[rgb]{ 1,  .961,  .937}16.06  & \cellcolor[rgb]{ 1,  .961,  .937}17.04  & \cellcolor[rgb]{ 1,  .961,  .937}6.23 & \cellcolor[rgb]{ 1,  .961,  .937}17.40  & \cellcolor[rgb]{ 1,  .961,  .937}13.21  & \cellcolor[rgb]{ 1,  .961,  .937}33.07  & \cellcolor[rgb]{ 1,  .961,  .937}92.27  \\
			& PopS & \cellcolor[rgb]{ 1.0,  1.0,  1.0} 13.05 & 3.99 & 13.36 & 13.38 & 5.10 & 13.93 & 9.49 & 29.13 & 80.51 \\
			& 2stS & \cellcolor[rgb]{ 1.0,  1.0,  1.0} 15.50 & \cellcolor[rgb]{ .996,  .973,  .961}4.42  & 15.77 & \cellcolor[rgb]{ .996,  .973,  .961}16.76  & \cellcolor[rgb]{ .996,  .973,  .961}6.21  & \cellcolor[rgb]{ .996,  .973,  .961}17.18  & \cellcolor[rgb]{ 1,  .961,  .937}13.35  & \cellcolor[rgb]{ .996,  .973,  .961}32.95  & \cellcolor[rgb]{ .996,  .949,  .925}92.01 \\
			& HarS & \cellcolor[rgb]{ .996,  .949,  .925} 20.76 & \cellcolor[rgb]{ .996,  .949,  .925}6.51 & \cellcolor[rgb]{ .992,  .953,  .929}21.05 & \cellcolor[rgb]{ .996,  .89,  .839}21.36  & \cellcolor[rgb]{ .996,  .89,  .839}8.86  & \cellcolor[rgb]{ .988,  .925,  .882}22.10 & \cellcolor[rgb]{ .996,  .949,  .925}15.94 & \cellcolor[rgb]{ 1,  .961,  .937}40.02  & \cellcolor[rgb]{ .996,  .949,  .925}91.66  \\
			& TransCF & \cellcolor[rgb]{ 1.0,  1.0,  1.0} 12.90 & 3.72 & 13.32 & 14.35 & 5.70 & 14.76 & \cellcolor[rgb]{ .996,  .973,  .961}11.19  & 29.88 & \cellcolor[rgb]{ .996,  .973,  .961}87.53  \\
			& LRML & \cellcolor[rgb]{ .996,  .949,  .925} 20.65  & \cellcolor[rgb]{ .988,  .925,  .882}6.65 & \cellcolor[rgb]{ .996,  .949,  .925}21.44 & \cellcolor[rgb]{ .996,  .949,  .925}20.36  & \cellcolor[rgb]{ .996,  .949,  .925}8.24  & \cellcolor[rgb]{ .996,  .949,  .925}21.75 & \cellcolor[rgb]{ 1,  .961,  .937}13.48 & \cellcolor[rgb]{ 1,  .961,  .937}37.93 & \cellcolor[rgb]{ 1,  .961,  .937}90.38  \\
			& CRML & \cellcolor[rgb]{ .988,  .925,  .882} 20.94  & \cellcolor[rgb]{ .988,  .925,  .882}6.43  & \cellcolor[rgb]{ .996,  .89,  .839}21.80 & \cellcolor[rgb]{ .988,  .925,  .882}21.14  & \cellcolor[rgb]{ .996,  .949,  .925}8.53 & \cellcolor[rgb]{ .996,  .89,  .839}22.44  & \cellcolor[rgb]{ .988,  .925,  .882}16.33 & \cellcolor[rgb]{ .988,  .925,  .882}41.14  & \cellcolor[rgb]{ .988,  .925,  .882}92.07  \\
			%		& NaiveCML & \cellcolor[rgb]{ .98,  .863,  .788}22.28 & \cellcolor[rgb]{ .98,  .863,  .788}7.02 & \cellcolor[rgb]{ .98,  .863,  .788}22.57 & \cellcolor[rgb]{ .973,  .796,  .678}\textbf{23.82 } & \cellcolor[rgb]{ .98,  .863,  .788}9.74 & \cellcolor[rgb]{ .973,  .796,  .678}\textbf{24.36 } & \cellcolor[rgb]{ .98,  .863,  .788}17.44 & \cellcolor[rgb]{ .98,  .863,  .788}42.40 & \cellcolor[rgb]{ .98,  .863,  .788}93.00 \\
			%		\cmidrule{2-11}      & SFCML(ours) & \cellcolor[rgb]{ .973,  .796,  .678}\textbf{23.17} & \cellcolor[rgb]{ .973,  .796,  .678}\textbf{7.38} & \cellcolor[rgb]{ .973,  .796,  .678}\textbf{23.11} & \cellcolor[rgb]{ .98,  .863,  .788}23.71 & \cellcolor[rgb]{ .973,  .796,  .678}\textbf{9.82} & \cellcolor[rgb]{ .98,  .863,  .788}24.30 & \cellcolor[rgb]{ .973,  .796,  .678}\textbf{17.53} & \cellcolor[rgb]{ .973,  .796,  .678}\textbf{42.51} & \cellcolor[rgb]{ .973,  .796,  .678}\textbf{93.01} \\
			
%			&NaiveCML & \cellcolor[rgb]{ .98,  .863,  .788}\underline{22.02} & \cellcolor[rgb]{ .98,  .863,  .788}6.95 & \cellcolor[rgb]{ .98,  .863,  .788}\underline{22.53} & \cellcolor[rgb]{ .973,  .796,  .678}\textbf{23.93} & \cellcolor[rgb]{ .98,  .863,  .788}\underline{9.80} & \cellcolor[rgb]{ .973,  .796,  .678}\textbf{24.50} & \cellcolor[rgb]{ .973,  .796,  .678}\textbf{17.61} & \cellcolor[rgb]{ .973,  .796,  .678}\textbf{42.57} & \cellcolor[rgb]{ .973,  .796,  .678}\textbf{93.11} \\
			&NaiveCML & \cellcolor[rgb]{ .98,  .863,  .788} \underline{22.51} & \cellcolor[rgb]{ .98,  .863,  .788}\underline{7.26} & \cellcolor[rgb]{ .98,  .863,  .788}\underline{22.79} & \cellcolor[rgb]{ .973,  .796,  .678}\textbf{23.85} & \cellcolor[rgb]{ .98,  .863,  .788}\underline{9.81} & \cellcolor[rgb]{ .98,  .863,  .788}\underline{24.42} & \cellcolor[rgb]{ .98,  .863,  .788}\underline{17.62} & \cellcolor[rgb]{ .98,  .863,  .788}\underline{42.35} & \cellcolor[rgb]{ .973,  .796,  .678}\textbf{93.24} \\
			
%			\cmidrule{2-11} &	SFCML(ours) & \cellcolor[rgb]{ .973,  .796,  .678}\textbf{23.17} & \cellcolor[rgb]{ .973,  .796,  .678}\textbf{7.38} & \cellcolor[rgb]{ .973,  .796,  .678}\textbf{23.11} & \cellcolor[rgb]{ .98,  .863,  .788}\underline{23.71} & \cellcolor[rgb]{ .973,  .796,  .678}\textbf{9.82} & \cellcolor[rgb]{ .98,  .863,  .788}\underline{24.30} & \cellcolor[rgb]{ .98,  .863,  .788}\underline{17.53} & \cellcolor[rgb]{ .98,  .863,  .788}\underline{42.51} & \cellcolor[rgb]{ .98,  .863,  .788}\underline{93.01} \\

			\cmidrule{2-11} &	SFCML(ours) & \cellcolor[rgb]{ .973,  .796,  .678} \textbf{23.40} & \cellcolor[rgb]{ .973,  .796,  .678}\textbf{7.62} & \cellcolor[rgb]{ .973,  .796,  .678}\textbf{23.63} & \cellcolor[rgb]{ .98,  .863,  .788}\underline{23.74} & \cellcolor[rgb]{ .973,  .796,  .678}\textbf{9.95} & \cellcolor[rgb]{ .973,  .796,  .678}\textbf{24.65} & \cellcolor[rgb]{ .973,  .796,  .678}\textbf{18.00} & \cellcolor[rgb]{ .973,  .796,  .678}\textbf{43.13} & \cellcolor[rgb]{ .98,  .863,  .788}\underline{93.11} \\
			\midrule
			\multirow{15}[5]{*}{CiteULike} & itemKNN & \cellcolor[rgb]{ 1.0,  1.0,  1.0} 1.20 & 0.83 & 1.23 & 1.15 & 0.77 & 1.16 & 1.44 & 3.78 & 69.94 \\
			& GMF & \cellcolor[rgb]{ 1.0,  1.0,  1.0} 1.86 & 0.96 & 2.05 & 2.15 & 0.97 & 2.40 & 1.34 & 5.53 & 65.38 \\
			& MLP &\cellcolor[rgb]{ 1.0,  1.0,  1.0} 1.76 & 0.77 & 1.94 & 2.42 & 0.98 & \cellcolor[rgb]{ .984,  .992,  .976}2.67  & \cellcolor[rgb]{ .984,  .992,  .976}1.52  & 5.70 & \cellcolor[rgb]{ .961,  .98,  .941}78.14 \\
			& NCF & \cellcolor[rgb]{ .984,  .992,  .976} 2.06  & \cellcolor[rgb]{ .984,  .992,  .976}1.04  & 2.21 & 2.36 & \cellcolor[rgb]{ .984,  .992,  .976}1.16  & \cellcolor[rgb]{ .984,  .992,  .976}2.64  & \cellcolor[rgb]{ .984,  .992,  .976}1.66  & 6.20 & \cellcolor[rgb]{ .961,  .98,  .941}77.88 \\
			& EHCF & \cellcolor[rgb]{ .961,  .98,  .941} 4.91  & \cellcolor[rgb]{ .961,  .98,  .941}2.69  & \cellcolor[rgb]{ .961,  .98,  .941}5.21  & \cellcolor[rgb]{ .961,  .98,  .941}5.88  & \cellcolor[rgb]{ .961,  .98,  .941}3.07  & \cellcolor[rgb]{ .984,  .992,  .976}6.29  & \cellcolor[rgb]{ .984,  .992,  .976}3.78  & \cellcolor[rgb]{ .984,  .992,  .976}12.60 & \cellcolor[rgb]{ .961,  .98,  .941}76.40 \\
			\cmidrule{2-11}      & UniS & \cellcolor[rgb]{ .945,  .969,  .929} 5.84  & \cellcolor[rgb]{ .945,  .969,  .929}3.04  & \cellcolor[rgb]{ .945,  .969,  .929}6.07  & \cellcolor[rgb]{ .945,  .969,  .929}7.58  & \cellcolor[rgb]{ .945,  .969,  .929}3.96  & \cellcolor[rgb]{ .961,  .98,  .941}7.91  & \cellcolor[rgb]{ .961,  .98,  .941}4.67  & \cellcolor[rgb]{ .961,  .98,  .941}14.51  & \cellcolor[rgb]{ .776,  .878,  .706}\textbf{86.62} \\
			& PopS & \cellcolor[rgb]{ .835,  .91,  .784} \underline{7.25}  & \cellcolor[rgb]{ .835,  .91,  .784}\underline{3.82}  & \cellcolor[rgb]{ .835,  .91,  .784}\underline{7.55}  & \cellcolor[rgb]{ .835,  .91,  .784}\underline{9.14}  & \cellcolor[rgb]{ .835,  .91,  .784}\underline{4.96}  & \cellcolor[rgb]{ .835,  .91,  .784}\underline{9.72}  & \cellcolor[rgb]{ .835,  .91,  .784}\underline{5.47}  & \cellcolor[rgb]{ .835,  .91,  .784}\underline{17.06}  & \cellcolor[rgb]{ .835,  .91,  .784}\underline{84.79}  \\
			& 2stS & \cellcolor[rgb]{ .886,  .937,  .851} 7.16  & \cellcolor[rgb]{ .886,  .937,  .851}3.69  & \cellcolor[rgb]{ .886,  .937,  .851}7.44  & \cellcolor[rgb]{ .886,  .937,  .851}8.95  & \cellcolor[rgb]{ .886,  .937,  .851}4.71  & \cellcolor[rgb]{ .886,  .937,  .851}9.66  & \cellcolor[rgb]{ .914,  .953,  .89}5.34  & \cellcolor[rgb]{ .886,  .937,  .851}16.96  & \cellcolor[rgb]{ .835,  .91,  .784}\underline{84.79}  \\
			& HarS & \cellcolor[rgb]{ .945,  .969,  .929} 6.05  & \cellcolor[rgb]{ .945,  .969,  .929}3.11  & \cellcolor[rgb]{ .945,  .969,  .929}6.35  & \cellcolor[rgb]{ .945,  .969,  .929}7.84  & \cellcolor[rgb]{ .945,  .969,  .929}4.16  & \cellcolor[rgb]{ .945,  .969,  .929}8.29  & \cellcolor[rgb]{ .945,  .969,  .929}5.02  & \cellcolor[rgb]{ .945,  .969,  .929}15.46  & \cellcolor[rgb]{ .886,  .937,  .851}83.80 \\
			& TransCF & \cellcolor[rgb]{ .945,  .969,  .929} 5.84  & \cellcolor[rgb]{ .945,  .969,  .929}3.12  & \cellcolor[rgb]{ .945,  .969,  .929}6.21  & \cellcolor[rgb]{ .945,  .969,  .929}7.29  & \cellcolor[rgb]{ .945,  .969,  .929}3.92  & \cellcolor[rgb]{ .961,  .98,  .941}7.76  & \cellcolor[rgb]{ .961,  .98,  .941}4.40 & \cellcolor[rgb]{ .961,  .98,  .941}14.43  & \cellcolor[rgb]{ .886,  .937,  .851}83.62  \\
			& LRML & \cellcolor[rgb]{ .984,  .992,  .976} 2.93  & \cellcolor[rgb]{ .984,  .992,  .976}1.38 & \cellcolor[rgb]{ .984,  .992,  .976}3.05  & \cellcolor[rgb]{ .984,  .992,  .976}3.84  & \cellcolor[rgb]{ .984,  .992,  .976}1.88 & \cellcolor[rgb]{ .984,  .992,  .976}4.13  & \cellcolor[rgb]{ .984,  .992,  .976}2.05  & \cellcolor[rgb]{ .984,  .992,  .976}7.96  & \cellcolor[rgb]{ .984,  .992,  .976}74.89  \\
			& CRML & \cellcolor[rgb]{ .914,  .953,  .89} 6.71  & \cellcolor[rgb]{ .914,  .953,  .89}3.47  & \cellcolor[rgb]{ .914,  .953,  .89}7.08  & \cellcolor[rgb]{ .914,  .953,  .89}8.61  & \cellcolor[rgb]{ .914,  .953,  .89}4.65  & \cellcolor[rgb]{ .914,  .953,  .89}9.21  & \cellcolor[rgb]{ .886,  .937,  .851}5.40 & \cellcolor[rgb]{ .914,  .953,  .89}16.85  & \cellcolor[rgb]{ .886,  .937,  .851}84.09 \\
			& NaiveCML & \cellcolor[rgb]{ 1.0,  1.0,  1.0} - & - & - & - & - & - & - & - & - \\
			\cmidrule{2-11}      & SFCML(ours) & \cellcolor[rgb]{ .776,  .878,  .706} \textbf{8.28 } & \cellcolor[rgb]{ .776,  .878,  .706}\textbf{4.65 } & \cellcolor[rgb]{ .776,  .878,  .706}\textbf{8.57 } & \cellcolor[rgb]{ .776,  .878,  .706}\textbf{9.69 } & \cellcolor[rgb]{ .776,  .878,  .706}\textbf{5.38 } & \cellcolor[rgb]{ .776,  .878,  .706}\textbf{10.29 } & \cellcolor[rgb]{ .776,  .878,  .706}\textbf{6.70} & \cellcolor[rgb]{ .776,  .878,  .706}\textbf{19.43 } & \cellcolor[rgb]{ .914,  .953,  .89}83.44  \\
			\midrule
			\multirow{15}[5]{*}{MovieLens-1m} & itemKNN & \cellcolor[rgb]{ 1.0,  1.0,  1.0} 12.24 & \cellcolor[rgb]{ .961,  .976,  .992}2.90 & 12.41 & \cellcolor[rgb]{ .961,  .976,  .992}12.43  & 4.29 & 12.79 & \cellcolor[rgb]{ .965,  .976,  .988}8.34  & \cellcolor[rgb]{ .961,  .976,  .992}26.16  & \cellcolor[rgb]{ .965,  .976,  .988}88.70 \\
			& GMF & \cellcolor[rgb]{ .965,  .976,  .988} 14.03  & \cellcolor[rgb]{ .965,  .976,  .988}2.79  & \cellcolor[rgb]{ .965,  .976,  .988}14.35  & \cellcolor[rgb]{ .965,  .976,  .988}14.28  & \cellcolor[rgb]{ .965,  .976,  .988}4.08  & \cellcolor[rgb]{ .965,  .976,  .988}14.80 & \cellcolor[rgb]{ .965,  .976,  .988}8.24  & \cellcolor[rgb]{ .965,  .976,  .988}29.51  & \cellcolor[rgb]{ .965,  .976,  .988}88.56  \\
			& MLP & \cellcolor[rgb]{ .965,  .976,  .988} 13.95  & \cellcolor[rgb]{ .965,  .976,  .988}2.78  & \cellcolor[rgb]{ .965,  .976,  .988}14.22  & \cellcolor[rgb]{ .965,  .976,  .988}14.06  & \cellcolor[rgb]{ .965,  .976,  .988}3.98  & \cellcolor[rgb]{ .965,  .976,  .988}14.56  & \cellcolor[rgb]{ .965,  .976,  .988}8.30  & \cellcolor[rgb]{ .965,  .976,  .988}29.33  & \cellcolor[rgb]{ .965,  .976,  .988}88.88  \\
			& NCF & \cellcolor[rgb]{ .906,  .941,  .973} 16.43  & \cellcolor[rgb]{ .965,  .976,  .988}3.20 & \cellcolor[rgb]{ .906,  .941,  .973}16.87  & \cellcolor[rgb]{ .906,  .941,  .973}16.73  & \cellcolor[rgb]{ .863,  .918,  .965}4.68  & \cellcolor[rgb]{ .863,  .918,  .965}17.40 & \cellcolor[rgb]{ .906,  .941,  .973}9.69  & \cellcolor[rgb]{ .906,  .941,  .973}33.23  & \cellcolor[rgb]{ .906,  .941,  .973}90.07  \\
			& EHCF & \cellcolor[rgb]{ .863,  .918,  .965} 17.82  & \cellcolor[rgb]{ .808,  .882,  .949}\underline{4.21}  & \cellcolor[rgb]{ .863,  .918,  .965}18.18  & \cellcolor[rgb]{ .863,  .918,  .965}18.08 & \cellcolor[rgb]{ .808,  .882,  .949}\underline{6.06}  & \cellcolor[rgb]{ .863,  .918,  .965}18.67  & \cellcolor[rgb]{ .863,  .918,  .965}12.12  & \cellcolor[rgb]{ .863,  .918,  .965}36.23  & \cellcolor[rgb]{ .906,  .941,  .973}90.42 \\
			\cmidrule{2-11} & UniS & \cellcolor[rgb]{ .965,  .976,  .988} 12.46  & \cellcolor[rgb]{ .961,  .976,  .992}2.40 & 12.60 & \cellcolor[rgb]{ .961,  .976,  .992}12.98  & \cellcolor[rgb]{ .961,  .976,  .992}3.72  & \cellcolor[rgb]{ .961,  .976,  .992}13.31  & \cellcolor[rgb]{ .965,  .976,  .988}8.47  & \cellcolor[rgb]{ .965,  .976,  .988}27.10 & \cellcolor[rgb]{ .863,  .918,  .965}91.84  \\
			& PopS & \cellcolor[rgb]{ 1.0,  1.0,  1.0} 9.07 & 1.98 & 9.32 & 9.07 & 2.94 & 9.52 & 5.39 & 21.74 & 81.02 \\
			& 2stS & \cellcolor[rgb]{ .965,  .976,  .988}12.42  & \cellcolor[rgb]{ .961,  .976,  .992}2.27  & 12.80 & \cellcolor[rgb]{ .961,  .976,  .992}12.64  & \cellcolor[rgb]{ .961,  .976,  .992}3.46  & \cellcolor[rgb]{ .961,  .976,  .992}13.27  & \cellcolor[rgb]{ .965,  .976,  .988}8.43  & \cellcolor[rgb]{ .965,  .976,  .988}27.48  & \cellcolor[rgb]{ .906,  .941,  .973}89.96  \\
			& HarS & \cellcolor[rgb]{ .843,  .906,  .961} 18.75  & \cellcolor[rgb]{ .906,  .941,  .973}4.04  & \cellcolor[rgb]{ .843,  .906,  .961}19.23  & \cellcolor[rgb]{ .843,  .906,  .961}19.09  & \cellcolor[rgb]{ .843,  .906,  .961}5.93  & \cellcolor[rgb]{ .843,  .906,  .961}19.88  & \cellcolor[rgb]{ .843,  .906,  .961}12.84  & \cellcolor[rgb]{ .843,  .906,  .961}37.48  & \cellcolor[rgb]{ .843,  .906,  .961}92.78  \\
			& TransCF & \cellcolor[rgb]{ 1.0,  1.0,  1.0} 10.55 & 2.25 & 10.77 & \cellcolor[rgb]{ .961,  .976,  .992}10.16  & \cellcolor[rgb]{ .961,  .976,  .992}3.17  & 10.75 & 6.44 & \cellcolor[rgb]{ .961,  .976,  .992}23.75  & \cellcolor[rgb]{ .965,  .976,  .988}86.94  \\
			& LRML & \cellcolor[rgb]{ .906,  .941,  .973} 15.37  & \cellcolor[rgb]{ .965,  .976,  .988}2.91  & \cellcolor[rgb]{ .965,  .976,  .988}15.78  & \cellcolor[rgb]{ .965,  .976,  .988}15.84  & \cellcolor[rgb]{ .906,  .941,  .973}4.37  & \cellcolor[rgb]{ .906,  .941,  .973}16.43  & \cellcolor[rgb]{ .906,  .941,  .973}9.25  & \cellcolor[rgb]{ .906,  .941,  .973}31.67  & \cellcolor[rgb]{ .906,  .941,  .973}90.21  \\
			& CRML & \cellcolor[rgb]{ .808,  .882,  .949} \underline{19.13}  & \cellcolor[rgb]{ .843,  .906,  .961}4.10 & \cellcolor[rgb]{ .808,  .882,  .949}\underline{19.64} & \cellcolor[rgb]{ .808,  .882,  .949}\underline{19.28} & \cellcolor[rgb]{ .843,  .906,  .961}5.99  & \cellcolor[rgb]{ .808,  .882,  .949}\underline{20.13} & \cellcolor[rgb]{ .808,  .882,  .949}\underline{13.13}  & \cellcolor[rgb]{ .808,  .882,  .949}\underline{37.80} & \cellcolor[rgb]{ .808,  .882,  .949}\underline{93.90} \\
			& NaiveCML & \cellcolor[rgb]{ 1.0,  1.0,  1.0} - & - & - & - & - & - & - & - & - \\
			\cmidrule{2-11}      & SFCML(ours) & \cellcolor[rgb]{ .741,  .843,  .933} \textbf{22.71} & \cellcolor[rgb]{ .741,  .843,  .933}\textbf{5.39} & \cellcolor[rgb]{ .741,  .843,  .933}\textbf{23.18} & \cellcolor[rgb]{ .741,  .843,  .933}\textbf{22.66} & \cellcolor[rgb]{ .741,  .843,  .933}\textbf{7.56} & \cellcolor[rgb]{ .741,  .843,  .933}\textbf{23.49} & \cellcolor[rgb]{ .741,  .843,  .933}\textbf{15.30} & \cellcolor[rgb]{ .741,  .843,  .933}\textbf{42.88} & \cellcolor[rgb]{ .741,  .843,  .933}\textbf{94.02} \\
			\bottomrule    
		\end{tabular}%
	}
	\label{results1}%
\end{table*}%

\subsection{Implementation Details} 
All the experiments are conducted on a Ubuntu 16.04.6 server equipped with 256GB RAM, Intel(R) Xeon(R) Gold-5218 CPU, and an RTX 3090 GPU. We implement our model with PyTorch \footnotemark \cite{paszke2017automatic} and adopt \textit{Adagrad} \cite{duchi2011adaptive} as the optimizer to minimize the objective loss function. For all datasets, each user's interactions are divided into training/validation/test sets with a $60\%/20\%/20\%$ split. Based on this split ratio, to ensure that each user has at least one positive interaction in training/validation/test, we thus filter out users that have less than five interactions. We conduct the grid search to find the best parameters based on the validation set and report the corresponding performance on the test set.  Specifically, for all methods, the batch size is set to $256$ and the learning rate is tuned amongst $\{0.001, 0.003, 0.005, 0.01, 0.03, 0.05\}$. The number of epochs is set as $200$ on all datasets. In addition, to further avoid the over-fitting problem, if the performance according to the AUC metric (with error range $\epsilon=10^{-5}$) on the validation set does not improve after $15$ epochs, the early-stopping is executed. With respect to the CML-based algorithms, the dimension of embedding $d$ is fixed as $256$, and the margin $\lambda$ is tuned amongst $\{1.0, 1.5, 2.0\}$. Moreover, we test the sampling-based CML with different sampling constant $U=\{1, 3, 5, 8, 10\}$ and then report the best performance according to AUC metric. For the other parameters of baseline models, we follow their tuning strategies in the original papers. Finally, in terms of the top-$K$ recommendation, we evaluate the performance at $K \in \{3, 5, 10, 20\}$, respectively.
\footnotetext{\url{https://pytorch.org/}}

\begin{table*}[!ht]
	\centering
	\caption{Performance comparisons on Steam-200k and Anime datasets, where '-' means that we cannot complete the experiments due to the out-of-memory issue. The best and second-best are highlighted in bold and underlined, respectively.}
	\scalebox{1.0}{
		\begin{tabular}{c|c|ccccccccc}
			\toprule
			& Method & \cellcolor[rgb]{ 1.0,  1.0,  1.0} P@3 & R@3 & NDCG@3 & P@5 & R@5 & NDCG@5 & MAP & MRR & AUC \\
			\midrule
			\multirow{15}[5]{*}{Steam-200k} & itemKNN & \multicolumn{1}{c}{\cellcolor[rgb]{ 1.0,  1.0,  1.0} 12.58} & \multicolumn{1}{c}{9.47} & \multicolumn{1}{c}{13.23} & \multicolumn{1}{c}{6.47} & 3.9 & \multicolumn{1}{c}{7.23} & \multicolumn{1}{c}{11.74} & \multicolumn{1}{c}{23.33} & \cellcolor[rgb]{ .996,  .98,  .973}86.81 \\
			& GMF & \multicolumn{1}{c}{\cellcolor[rgb]{ 1.0,  1.0,  1.0} 9.28} & \multicolumn{1}{c}{3.85} & \multicolumn{1}{c}{9.52} & \multicolumn{1}{c}{12.94} & 5.73 & \multicolumn{1}{c}{13.41} & \multicolumn{1}{c}{7.32} & \multicolumn{1}{c}{22.35} & \cellcolor[rgb]{ .996,  .98,  .973}87.25 \\
			& MLP & \multicolumn{1}{c}{\cellcolor[rgb]{ .996,  .98,  .973} 13.09} & \multicolumn{1}{c}{6.64} & \multicolumn{1}{c}{13.66} & \multicolumn{1}{c}{\cellcolor[rgb]{ .996,  .98,  .973}14.34} & \cellcolor[rgb]{ .996,  .973,  .961}6.92  & \multicolumn{1}{c}{\cellcolor[rgb]{ .996,  .98,  .973}15.48} & \multicolumn{1}{c}{10.02} & \multicolumn{1}{c}{\cellcolor[rgb]{ .996,  .973,  .961}28.64 } & \cellcolor[rgb]{ .996,  .973,  .961}91.45  \\
			& NCF & \multicolumn{1}{c}{\cellcolor[rgb]{ 1.0,  1.0,  1.0} 12.97} & \multicolumn{1}{c}{6.58} & \multicolumn{1}{c}{13.58} & \multicolumn{1}{c}{\cellcolor[rgb]{ .996,  .98,  .973}14.26} & \cellcolor[rgb]{ .996,  .973,  .961}6.90 & \multicolumn{1}{c}{\cellcolor[rgb]{ .996,  .98,  .973}15.42} & \multicolumn{1}{c}{10.10} & \multicolumn{1}{c}{\cellcolor[rgb]{ .996,  .973,  .961}28.71 } & \cellcolor[rgb]{ .996,  .973,  .961}91.51  \\
			& EHCF & \multicolumn{1}{c}{\cellcolor[rgb]{ .996,  .89,  .839} 22.76} & \multicolumn{1}{c}{\cellcolor[rgb]{ .996,  .89,  .839}13.33}  & \multicolumn{1}{c}{\cellcolor[rgb]{ .996,  .89,  .839}24.07} & \multicolumn{1}{c}{\cellcolor[rgb]{ .98,  .863,  .788}21.13} & \cellcolor[rgb]{ .98,  .863,  .788}10.87 & \multicolumn{1}{c}{\cellcolor[rgb]{ .996,  .89,  .839}22.57} & \multicolumn{1}{c}{\cellcolor[rgb]{ .98,  .863,  .788}19.98} & \multicolumn{1}{c}{\cellcolor[rgb]{ .98,  .863,  .788}43.82} & \cellcolor[rgb]{ .996,  .949,  .925}93.17 \\
			\cmidrule{2-11}      & UniS & \multicolumn{1}{c}{\cellcolor[rgb]{ .996,  .98,  .973} 13.33} & \multicolumn{1}{c}{\cellcolor[rgb]{ .996,  .98,  .973}8.14} & \multicolumn{1}{c}{13.73} & \multicolumn{1}{c}{12.09} & \cellcolor[rgb]{ .996,  .98,  .973}5.73 & \multicolumn{1}{c}{12.65} & \multicolumn{1}{c}{\cellcolor[rgb]{ .996,  .98,  .973}13.00} & \multicolumn{1}{c}{\cellcolor[rgb]{ .996,  .973,  .961}27.25 } & \cellcolor[rgb]{ .996,  .949,  .925}93.59 \\
			& PopS & \multicolumn{1}{c}{\cellcolor[rgb]{ .996,  .949,  .925} 18.84} & \multicolumn{1}{c}{\cellcolor[rgb]{ .996,  .949,  .925}11.54} & \multicolumn{1}{c}{\cellcolor[rgb]{ .996,  .949,  .925}19.51} & \multicolumn{1}{c}{\cellcolor[rgb]{ .996,  .973,  .961}16.23 } & \cellcolor[rgb]{ .996,  .973,  .961}8.26  & \multicolumn{1}{c}{\cellcolor[rgb]{ .996,  .973,  .961}17.09 } & \multicolumn{1}{c}{\cellcolor[rgb]{ .996,  .973,  .961}15.33 } & \multicolumn{1}{c}{\cellcolor[rgb]{ .996,  .973,  .961}35.38 } & 84.46 \\
			& 2stS & \multicolumn{1}{c}{\cellcolor[rgb]{ .996,  .98,  .973} 13.65} & \multicolumn{1}{c}{\cellcolor[rgb]{ .996,  .98,  .973}8.33} & \multicolumn{1}{c}{\cellcolor[rgb]{ .996,  .98,  .973}14.22} & \multicolumn{1}{c}{12.44} & \cellcolor[rgb]{ .996,  .98,  .973}5.84 & \multicolumn{1}{c}{\cellcolor[rgb]{ .996,  .98,  .973}13.02} & \multicolumn{1}{c}{\cellcolor[rgb]{ .996,  .98,  .973}13.44} & \multicolumn{1}{c}{\cellcolor[rgb]{ .996,  .98,  .973}27.91} & \cellcolor[rgb]{ .996,  .973,  .961}92.22  \\
			& HarS & \multicolumn{1}{c}{\cellcolor[rgb]{ .996,  .949,  .925} 20.27} & \multicolumn{1}{c}{\cellcolor[rgb]{ .996,  .89,  .839}11.50} & \multicolumn{1}{c}{\cellcolor[rgb]{ .996,  .949,  .925}20.87} & \multicolumn{1}{c}{\cellcolor[rgb]{ .996,  .949,  .925}20.80} & \cellcolor[rgb]{ .996,  .949,  .925}10.04 & \multicolumn{1}{c}{\cellcolor[rgb]{ .996,  .949,  .925}21.39} & \multicolumn{1}{c}{\cellcolor[rgb]{ .996,  .949,  .925}18.16} & \multicolumn{1}{c}{\cellcolor[rgb]{ .996,  .949,  .925}37.97} & \cellcolor[rgb]{ .996,  .89,  .839}94.17  \\
			& TransCF & \multicolumn{1}{c}{\cellcolor[rgb]{ .996,  .98,  .973} 15.23} & \multicolumn{1}{c}{\cellcolor[rgb]{ .996,  .973,  .961}9.48 } & \multicolumn{1}{c}{\cellcolor[rgb]{ .996,  .973,  .961}15.93 } & \multicolumn{1}{c}{12.88} & 6.75 & \multicolumn{1}{c}{\cellcolor[rgb]{ .996,  .98,  .973}13.74} & \multicolumn{1}{c}{\cellcolor[rgb]{ .996,  .98,  .973}13.64} & \multicolumn{1}{c}{\cellcolor[rgb]{ .996,  .973,  .961}31.04 } & \cellcolor[rgb]{ .996,  .973,  .961}91.91  \\
			& LRML & \multicolumn{1}{c}{\cellcolor[rgb]{ .996,  .98,  .973} 14.54} & \multicolumn{1}{c}{\cellcolor[rgb]{ .996,  .98,  .973}7.14} & \multicolumn{1}{c}{\cellcolor[rgb]{ .996,  .98,  .973}14.86} & \multicolumn{1}{c}{\cellcolor[rgb]{ .996,  .973,  .961}16.15 } & \cellcolor[rgb]{ .996,  .973,  .961}7.72  & \multicolumn{1}{c}{\cellcolor[rgb]{ .996,  .973,  .961}17.28 } & \multicolumn{1}{c}{11.45} & \multicolumn{1}{c}{\cellcolor[rgb]{ .996,  .973,  .961}30.70} & \cellcolor[rgb]{ .996,  .973,  .961}92.06  \\
			& CRML & \multicolumn{1}{c}{\cellcolor[rgb]{ .996,  .89,  .839} 20.51 } & \multicolumn{1}{c}{\cellcolor[rgb]{ .996,  .949,  .925}11.40} & \multicolumn{1}{c}{\cellcolor[rgb]{ .996,  .89,  .839}21.47 } & \multicolumn{1}{c}{\cellcolor[rgb]{ .996,  .89,  .839}21.10} & \cellcolor[rgb]{ .996,  .89,  .839}10.17  & \multicolumn{1}{c}{\cellcolor[rgb]{ .996,  .89,  .839}22.27 } & \multicolumn{1}{c}{\cellcolor[rgb]{ .996,  .89,  .839}18.42 } & \multicolumn{1}{c}{\cellcolor[rgb]{ .996,  .89,  .839}39.46} & \cellcolor[rgb]{ .98,  .863,  .788}94.19 \\
%			& NaiveCML & \multicolumn{1}{c}{\cellcolor[rgb]{ .98,  .863,  .788}\underline{23.45}} & \multicolumn{1}{c}{\cellcolor[rgb]{ .98,  .863,  .788}\underline{13.79}} & \multicolumn{1}{c}{\cellcolor[rgb]{ .98,  .863,  .788}\underline{24.70}} & \multicolumn{1}{c}{\cellcolor[rgb]{ .98,  .863,  .788}\underline{21.63}} & \cellcolor[rgb]{ .98,  .863,  .788}10.80 & \multicolumn{1}{c}{\cellcolor[rgb]{ .973,  .796,  .678}\textbf{23.19}} & \multicolumn{1}{c}{\cellcolor[rgb]{ .98,  .863,  .788}19.91} & \multicolumn{1}{c}{\cellcolor[rgb]{ .996,  .89,  .839}43.54 } & \cellcolor[rgb]{ .996,  .89,  .839}94.00 \\
			& NaiveCML & \cellcolor[rgb]{ .98,  .863,  .788} \underline{25.78} & \cellcolor[rgb]{ .98,  .863,  .788}\underline{15.22} & \cellcolor[rgb]{ .98,  .863,  .788}\underline{27.01} & \cellcolor[rgb]{ .973,  .796,  .678}\textbf{24.17} & \cellcolor[rgb]{ .973,  .796,  .678}\textbf{12.29} & \cellcolor[rgb]{ .973,  .796,  .678}\textbf{25.63} & \cellcolor[rgb]{ .98,  .863,  .788}\underline{21.67} & \cellcolor[rgb]{ .98,  .863,  .788}\underline{46.74} & \cellcolor[rgb]{ .973,  .796,  .678}\textbf{94.94} \\
			\cmidrule{2-11}      & SFCML(ours)& \cellcolor[rgb]{ .973,  .796,  .678} \textbf{26.34} & \cellcolor[rgb]{ .973,  .796,  .678}\textbf{16.00} & \cellcolor[rgb]{ .973,  .796,  .678}\textbf{27.23} & \cellcolor[rgb]{ .98,  .863,  .788}\underline{23.76} & \cellcolor[rgb]{ .98,  .863,  .788}\underline{12.25} & \cellcolor[rgb]{ .98,  .863,  .788}\underline{24.84} & \cellcolor[rgb]{ .973,  .796,  .678}\textbf{23.17} & \cellcolor[rgb]{ .973,  .796,  .678}\textbf{47.03} & \cellcolor[rgb]{ .98,  .863,  .788}\underline{94.58} \\
%			
%			\cmidrule{2-11}      & SFCML(ours) & \cellcolor[rgb]{ .973,  .796,  .678}\textbf{23.91} & \cellcolor[rgb]{ .973,  .796,  .678}\textbf{14.43} & \cellcolor[rgb]{ .973,  .796,  .678}\textbf{24.74} & \cellcolor[rgb]{ .973,  .796,  .678}\textbf{21.72} & \cellcolor[rgb]{ .973,  .796,  .678}\textbf{11.07} & \cellcolor[rgb]{ .98,  .863,  .788}\underline{22.70} & \cellcolor[rgb]{ .973,  .796,  .678}\textbf{21.19} & \cellcolor[rgb]{ .973,  .796,  .678}\textbf{43.89} & \cellcolor[rgb]{ .973,  .796,  .678}\textbf{94.57} \\
			\midrule    
			\multirow{15}[5]{*}{Anime} & itemKNN & \cellcolor[rgb]{ .984,  .992,  .976} 16.93  & \cellcolor[rgb]{ .945,  .969,  .929} 3.51  & \cellcolor[rgb]{ .984,  .992,  .976} 17.15  & \cellcolor[rgb]{ .945,  .969,  .929} 16.21  & \cellcolor[rgb]{ .945,  .969,  .929} 4.97  & \cellcolor[rgb]{ .984,  .992,  .976} 16.67  & \cellcolor[rgb]{ .984,  .992,  .976} 9.79  & \cellcolor[rgb]{ .984,  .992,  .976} 33.76  & \cellcolor[rgb]{ .945,  .969,  .929} 93.52  \\
			& GMF & \cellcolor[rgb]{ .945,  .969,  .929} 18.69  & \cellcolor[rgb]{ .945,  .969,  .929}3.31  & \cellcolor[rgb]{ .945,  .969,  .929}19.40  & \cellcolor[rgb]{ .945,  .969,  .929}17.26  & \cellcolor[rgb]{ .945,  .969,  .929}4.76  & \cellcolor[rgb]{ .945,  .969,  .929}18.40  & \cellcolor[rgb]{ .984,  .992,  .976}9.57  & \cellcolor[rgb]{ .945,  .969,  .929}37.26  & \cellcolor[rgb]{ .945,  .969,  .929}92.32  \\
			& MLP & \cellcolor[rgb]{ .945,  .969,  .929} 20.45  & \cellcolor[rgb]{ .945,  .969,  .929}3.71  & \cellcolor[rgb]{ .945,  .969,  .929}21.24  & \cellcolor[rgb]{ .945,  .969,  .929}19.13  & \cellcolor[rgb]{ .945,  .969,  .929}5.35  & \cellcolor[rgb]{ .945,  .969,  .929}20.32  & 10.95 & \cellcolor[rgb]{ .945,  .969,  .929}39.96  & \cellcolor[rgb]{ .945,  .969,  .929}93.94  \\
			& NCF & \cellcolor[rgb]{ .914,  .953,  .89} 24.09  & \cellcolor[rgb]{ .914,  .953,  .89}4.29  & \cellcolor[rgb]{ .914,  .953,  .89}24.84  & \cellcolor[rgb]{ .914,  .953,  .89}22.90 & \cellcolor[rgb]{ .914,  .953,  .89}6.28  & \cellcolor[rgb]{ .914,  .953,  .89}24.06  & \cellcolor[rgb]{ .914,  .953,  .89}13.41  & \cellcolor[rgb]{ .914,  .953,  .89}44.47  & \cellcolor[rgb]{ .914,  .953,  .89}94.93  \\
			& EHCF & \cellcolor[rgb]{ .835,  .91,  .784} \underline{28.72}  & \cellcolor[rgb]{ .835,  .91,  .784}\underline{5.88}  & \cellcolor[rgb]{ .835,  .91,  .784}\underline{29.45}  & \cellcolor[rgb]{ .835,  .91,  .784}\underline{27.55}  & \cellcolor[rgb]{ .835,  .91,  .784}\underline{8.56}  & \cellcolor[rgb]{ .835,  .91,  .784}\underline{28.65}  & \cellcolor[rgb]{ .835,  .91,  .784}\underline{18.06}  & \cellcolor[rgb]{ .835,  .91,  .784}\underline{50.55}  & \cellcolor[rgb]{ .835,  .91,  .784}96.03  \\
			\cmidrule{2-11}           & UniS & \cellcolor[rgb]{ 1.0,  1.0,  1.0} 14.81 & 2.33 & 15.16 & 14.24 & 3.48 & 14.83 & \cellcolor[rgb]{ .984,  .992,  .976}9.30 & 30.76 & \cellcolor[rgb]{ .914,  .953,  .89} 94.99  \\
			& PopS & \cellcolor[rgb]{ 1.0,  1.0,  1.0} 15.20 & \cellcolor[rgb]{ .984,  .992,  .976} 2.92  & 15.66 & 14.40 & \cellcolor[rgb]{ .984,  .992,  .976}4.22  & 15.10 & 7.80 & 31.93 & 87.66 \\
			& 2stS & \cellcolor[rgb]{ .984,  .992,  .976} 16.86  & \cellcolor[rgb]{ .984,  .992,  .976}2.84  & \cellcolor[rgb]{ .984,  .992,  .976}17.28  & \cellcolor[rgb]{ .945,  .969,  .929}16.31  & \cellcolor[rgb]{ .984,  .992,  .976}4.27  & \cellcolor[rgb]{ .984,  .992,  .976}16.96  & \cellcolor[rgb]{ .945,  .969,  .929}10.34  & \cellcolor[rgb]{ .984,  .992,  .976}34.39  & \cellcolor[rgb]{ .945,  .969,  .929}93.72  \\
			& HarS & \cellcolor[rgb]{ .945,  .969,  .929} 20.57  & \cellcolor[rgb]{ .945,  .969,  .929}3.54  &  \cellcolor[rgb]{ .945,  .969,  .929}21.07  & \cellcolor[rgb]{ .945,  .969,  .929}19.87  & \cellcolor[rgb]{ .945,  .969,  .929}5.30 & \cellcolor[rgb]{ .945,  .969,  .929}20.65  & \cellcolor[rgb]{ .945,  .969,  .929}12.77  & \cellcolor[rgb]{ .945,  .969,  .929}39.57  & \cellcolor[rgb]{ .914,  .953,  .89}94.97  \\
			& TransCF & \cellcolor[rgb]{ 1.0,  1.0,  1.0} 13.78 & 2.73 & 14.50 & 12.33 & 3.69 & 13.42 & 7.44 & 29.82 & \cellcolor[rgb]{ .984,  .992,  .976} 91.21  \\
			& LRML & \cellcolor[rgb]{ .945,  .969,  .929} 17.69  & \cellcolor[rgb]{ .984,  .992,  .976}3.06  & \cellcolor[rgb]{ .945,  .969,  .929}18.42  & \cellcolor[rgb]{ .945,  .969,  .929}16.49  & \cellcolor[rgb]{ .945,  .969,  .929}4.41  & \cellcolor[rgb]{ .945,  .969,  .929}17.56  & \cellcolor[rgb]{ .984,  .992,  .976}9.17  & \cellcolor[rgb]{ .945,  .969,  .929}36.03  & \cellcolor[rgb]{ .945,  .969,  .929}92.25  \\
			& CRML & \cellcolor[rgb]{ .886,  .937,  .851} 27.05  & \cellcolor[rgb]{ .886,  .937,  .851}5.05  & \cellcolor[rgb]{ .886,  .937,  .851}27.73  & \cellcolor[rgb]{ .886,  .937,  .851}25.93  & \cellcolor[rgb]{ .886,  .937,  .851}7.41  & \cellcolor[rgb]{ .886,  .937,  .851}27.00 & \cellcolor[rgb]{ .886,  .937,  .851}17.23  & \cellcolor[rgb]{ .886,  .937,  .851}48.16  & \cellcolor[rgb]{ .886,  .937,  .851}\underline{96.74}  \\
			& NaiveCML & \cellcolor[rgb]{ 1.0,  1.0,  1.0} - & - & - & - & - & - & - & - & - \\
			\cmidrule{2-11}      & SFCML(ours) & \cellcolor[rgb]{ .776,  .878,  .706} \textbf{30.31} & \cellcolor[rgb]{ .776,  .878,  .706}\textbf{6.36} & \cellcolor[rgb]{ .776,  .878,  .706}\textbf{30.82 } & \cellcolor[rgb]{ .776,  .878,  .706}\textbf{29.50} & \cellcolor[rgb]{ .776,  .878,  .706}\textbf{9.37} & \cellcolor[rgb]{ .776,  .878,  .706}\textbf{30.30} & \cellcolor[rgb]{ .776,  .878,  .706}\textbf{20.48} & \cellcolor[rgb]{ .776,  .878,  .706}\textbf{52.12} & \cellcolor[rgb]{ .776,  .878,  .706} \textbf{97.04} \\
			\bottomrule   
		\end{tabular}%
	}
	\label{results2}%
\end{table*}%

\subsection{Experiments Results}

\subsubsection{Overall Performance}
Some experimental results are presented in Tab.\ref{results1} and Tab.\ref{results2}. The others are shown in Appendix.\ref{more_evaluation_results} due to the limitation of space. From these results, we can draw the following interesting observations:
\begin{itemize}
	\item Our proposed SFCML shows competitive performance on all benchmark datasets, and, in most cases, its performance surpasses all the involved competitors. For example, the significant improvement of performance between SFCML and the best competitor (achieved by EHCF) on the MovieLens-100k dataset are $2.27\%$, $1.49\%$ and $1.36\%$ with respect to P@$3$, MAP and MRR, respectively. This validates the superiority of our proposed SFCML algorithm. 

%	\item Except for Anime and MovieLens-20m datasets, the deep learning-based algorithms show inferior performance than the sampling-based CML competitors in most cases. A possible reason is that the Anime and MovieLens-20m datasets provide sufficient feedback to get rid of over-fitting complicated models like NCF and EHCF. However, the other datasets, especially CiteULike and Amazon-Book, only have access to sparse feedback information, which makes deep-learning-based models generalize poorly to the test dataset. By contrast, most of the CML framework algorithms demonstrate reasonable performance on these datasets, except LRML on Amazon-Book. The reason might be the same as the deep-learning-based models, the insufficient interactions hurt the performance of memory-augmented neural architecture involved in LRML.

	\item The deep learning-based algorithms (such NCF) show inferior performance than the sampling-based CML competitors on CiteULike and Amazon-Book datasets. A possible reason is that deep-learning-based models can only have access to sparse feedback information on CiteULike and Amazon-Book datasets, which makes them generalize poorly to the test set. The same reason also causes the degraded performance of LRML, where a memory-augmented neural architecture is applied to CML. 
	
	\item With respect to the CML-based frameworks, the performance improvement between SFCML and other sampling-based CML is significant. The reason might be that the sampling-based CML essentially alters the intrinsic distribution of the training data. This results in a biased estimation for the expected risk over the unseen data and thus degrades the recommendation performance. On the contrary, SFCML, in a sampling-free manner, could boost the generalization performance. This confirms the arguments of this work and the effectiveness of SFCML.
	\item For the non-sampling algorithms, we see that, SFCML consistently outperforms the state-of-the-art sampling-free algorithm EHCF in most cases, which shows the superiority of our proposed SFCML method. 
\end{itemize}
In summary, the above empirical discussions could support the aforementioned theoretical arguments, i.e., the sampling-based CML methods could not guarantee a small generalization error and sometimes result in sub-optimal generalization performance. By contrast, SFCML could boost the performance since it can get rid of this bias via the sampling-free algorithm. 

%\subsubsection{Comparison against Other Competitors}
%
%\subsubsection{Comparison against metric-based learning Competitors}

\begin{figure*}[!ht]
	\centering
		\subfigure[R@5]{
			\includegraphics[width=0.65\columnwidth]{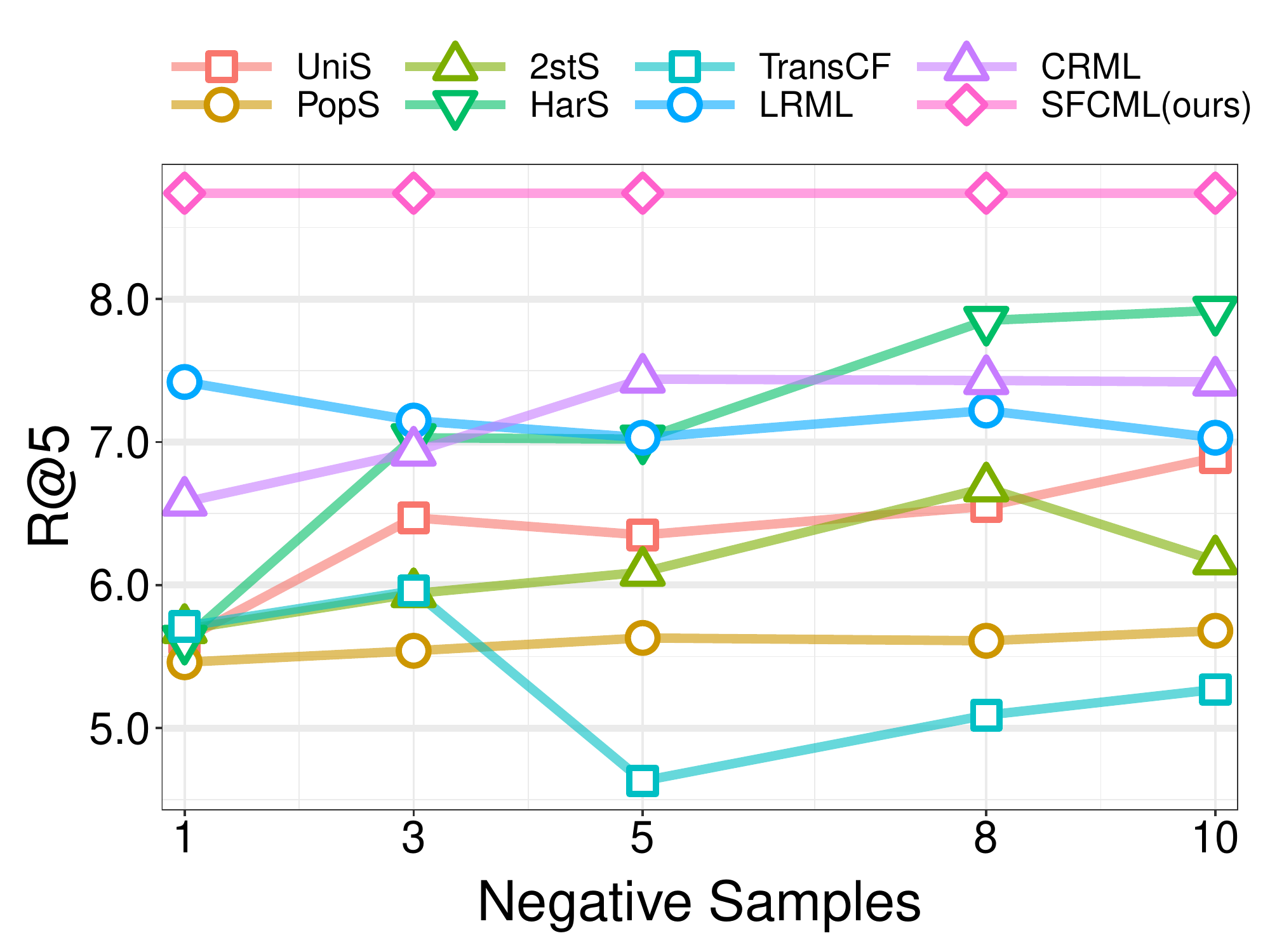}
			\label{ml-1m/R@5}
		}
		\subfigure[MAP]{
			\includegraphics[width=0.65\columnwidth]{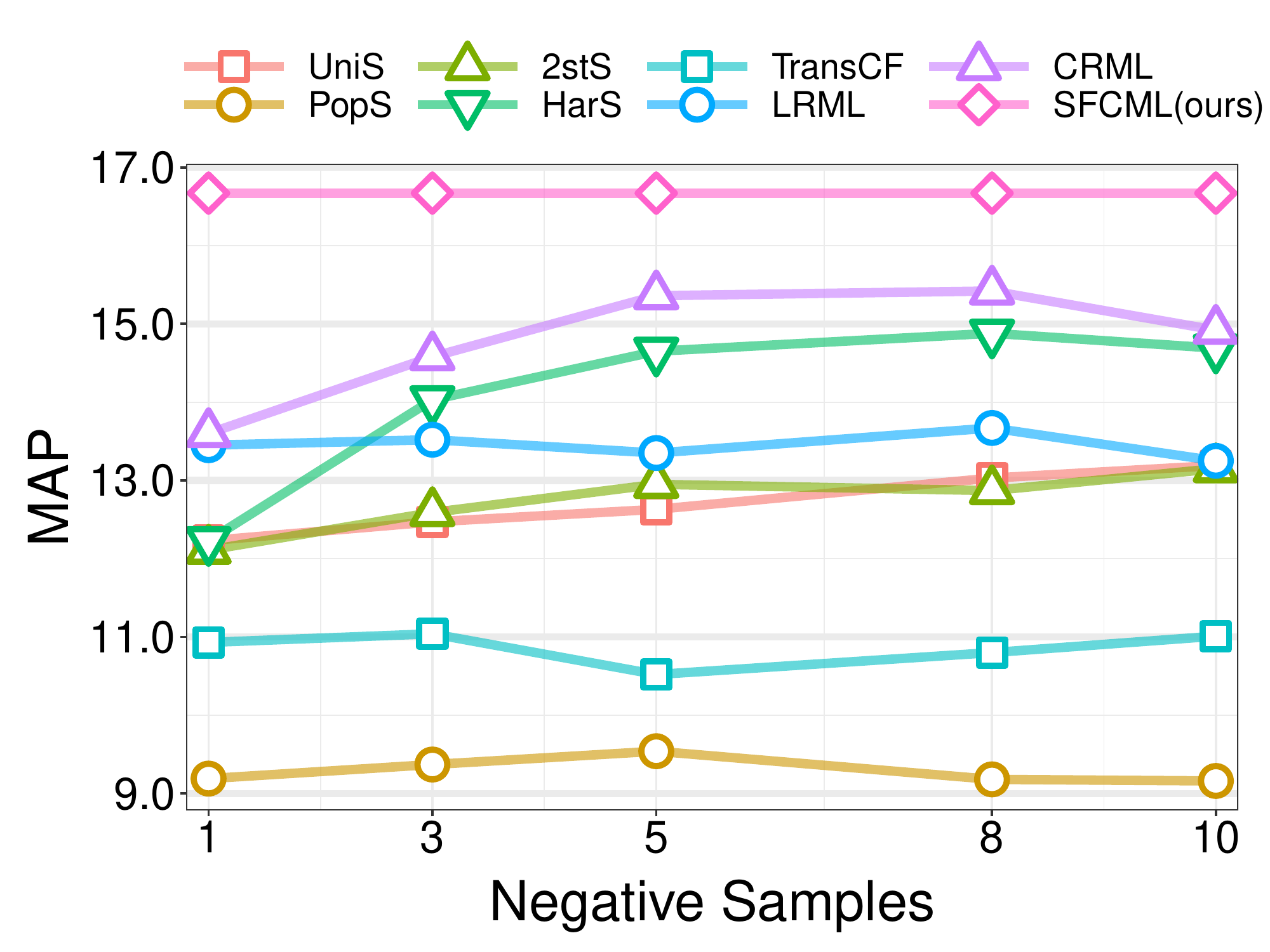}
			\label{ml-1m/MAP}
		}
		\subfigure[MRR]{
			\includegraphics[width=0.65\columnwidth]{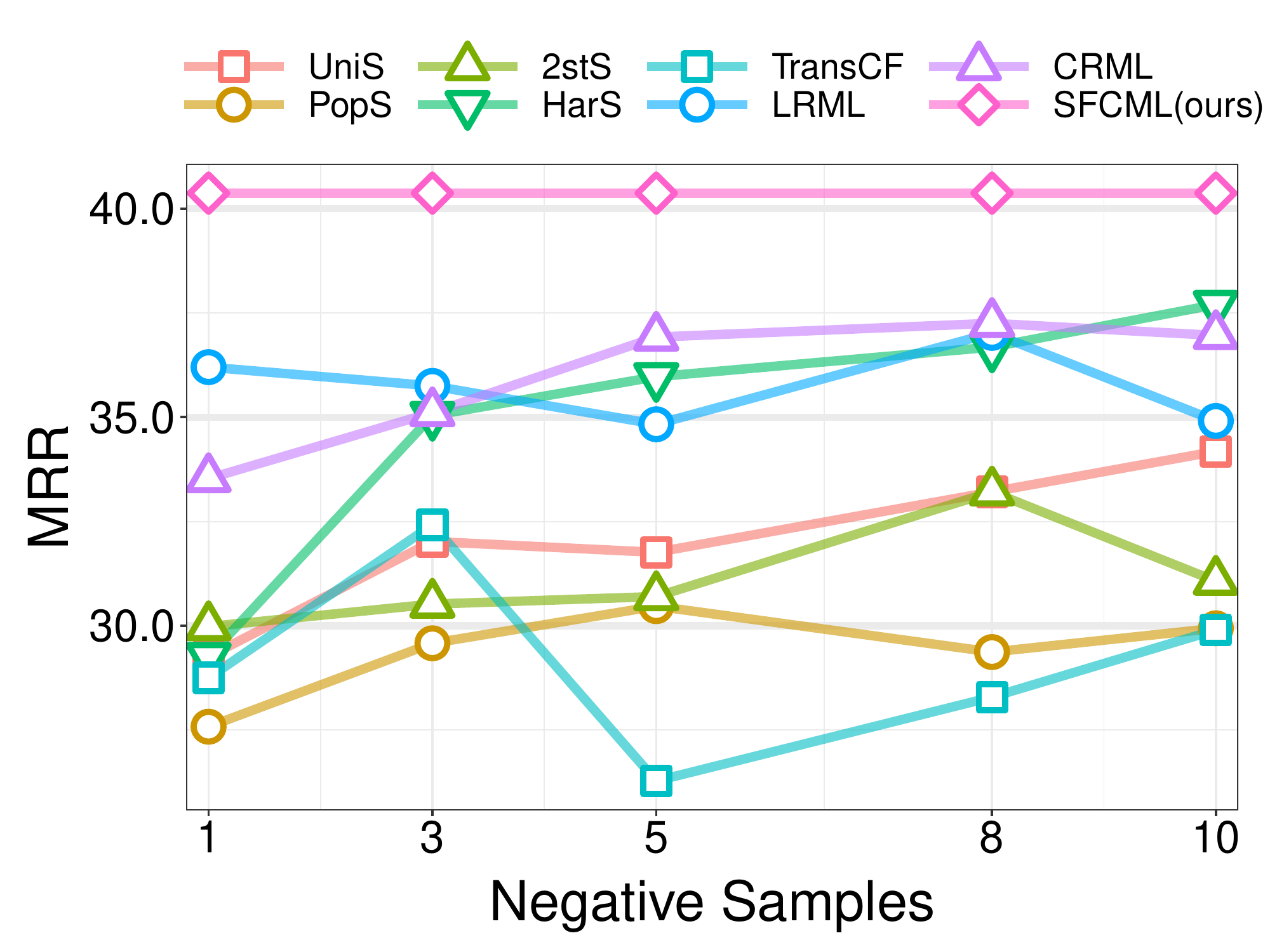}
			\label{ml-1m/MRR}
		}
		\caption{Performance comparisons on validation set of MovieLens-100k with respect to different negative sampling strategies and different sampling numbers $U \in \{1, 3, 5, 8, 10\}$. Please refer to Appendix.\ref{add_neg_nums} for more results.} 
		\label{neg_num_100k}
\end{figure*}

\begin{figure*}[!ht]
	\centering
	\subfigure[MovieLens-1m-1]{
		\includegraphics[width=0.315\textwidth]{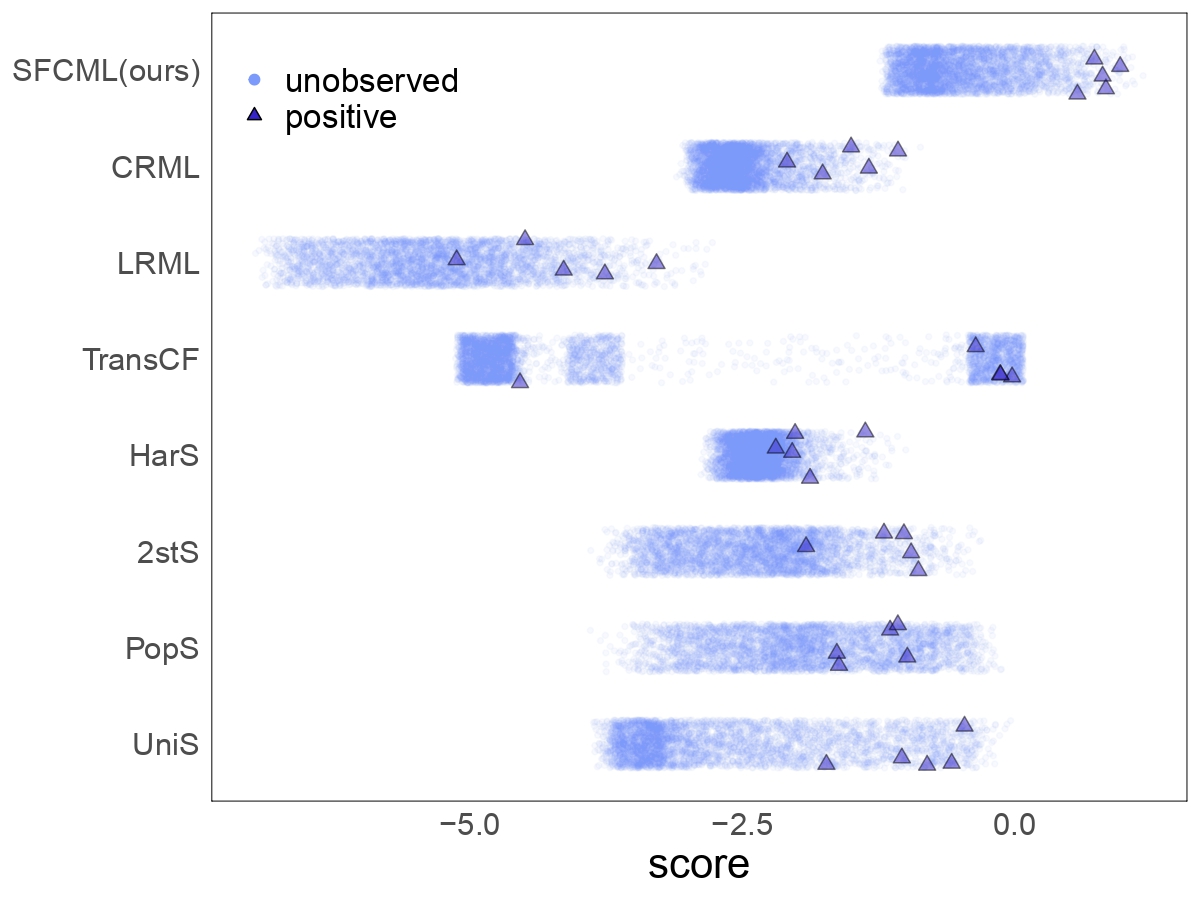}
		\label{ml_1m_1}
	}
	\subfigure[MovieLens-1m-2]{
		\includegraphics[width=0.315\textwidth]{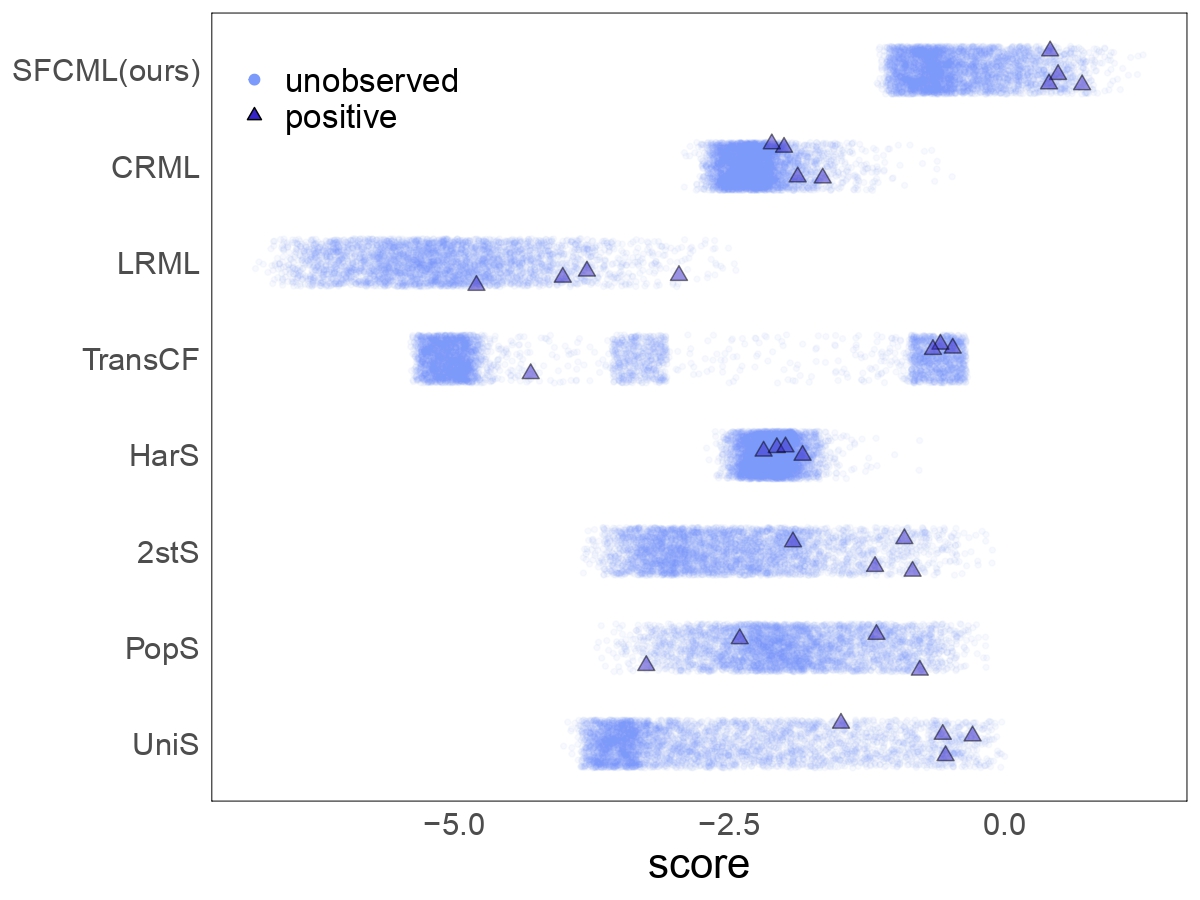}
		\label{ml_1m_2}
	}
	\subfigure[MovieLens-1m-3]{
		\includegraphics[width=0.315\textwidth]{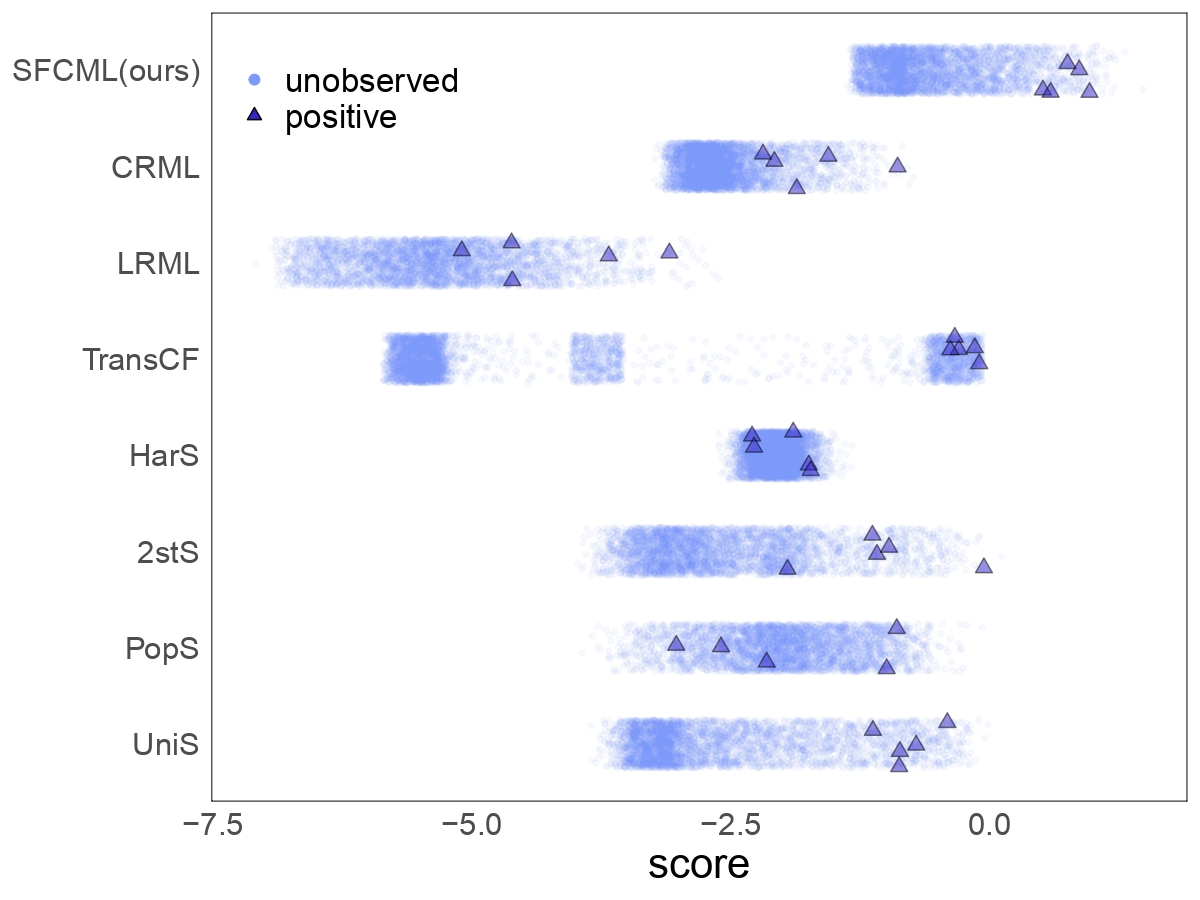}
		\label{ml_1m_3}
	}
	\caption{The graphical visualization of score distribution of positive and unobserved items on MovieLens-1m.}
	\label{jetter_ml_1m}
\end{figure*}

\begin{figure}[!ht]
	\centering
	\subfigure[MovieLens-1m-1]{
		\includegraphics[width=0.23\textwidth]{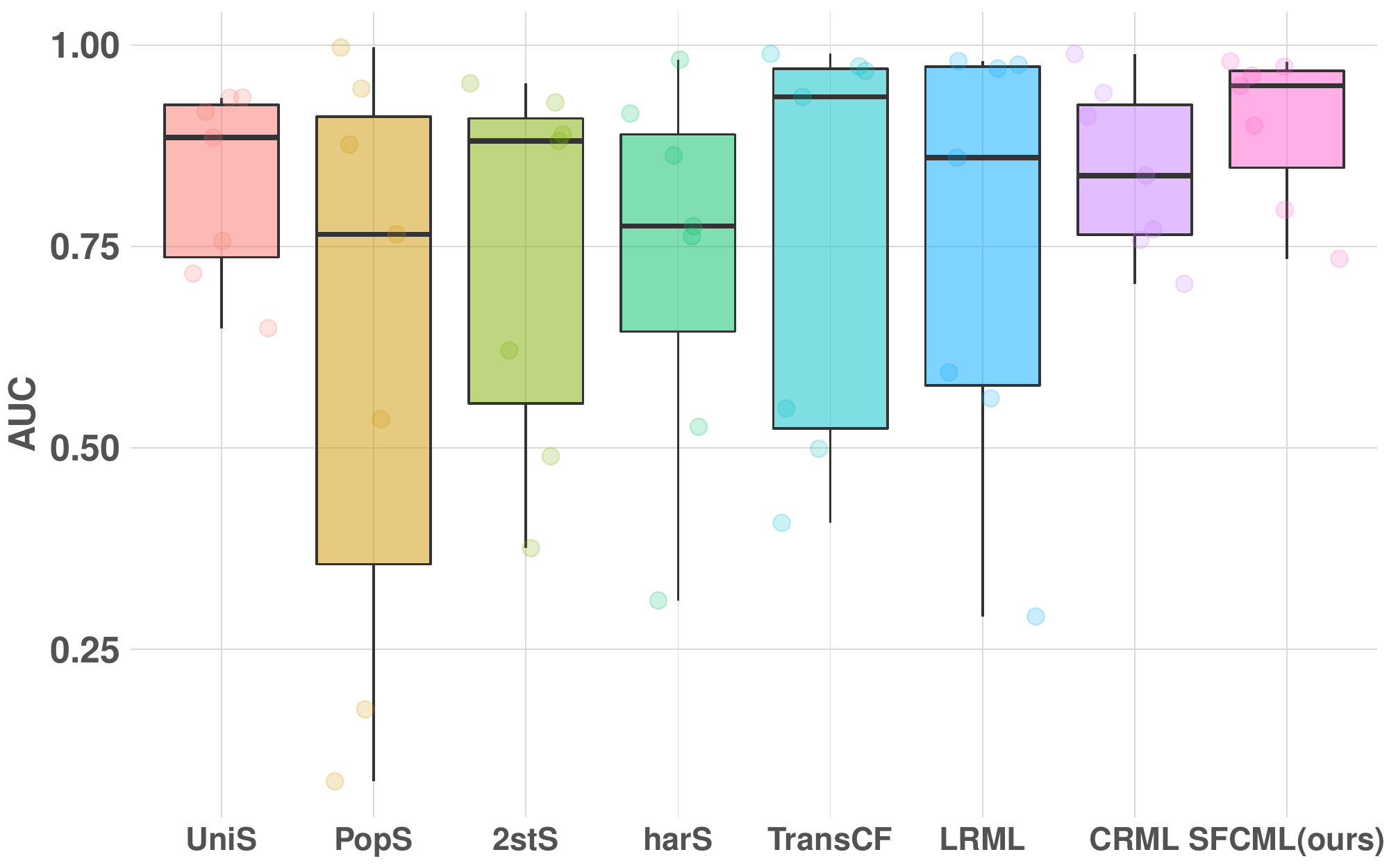}
		\label{main_ml-1}
	}
	\subfigure[MovieLens-1m-2]{
		\includegraphics[width=0.23\textwidth]{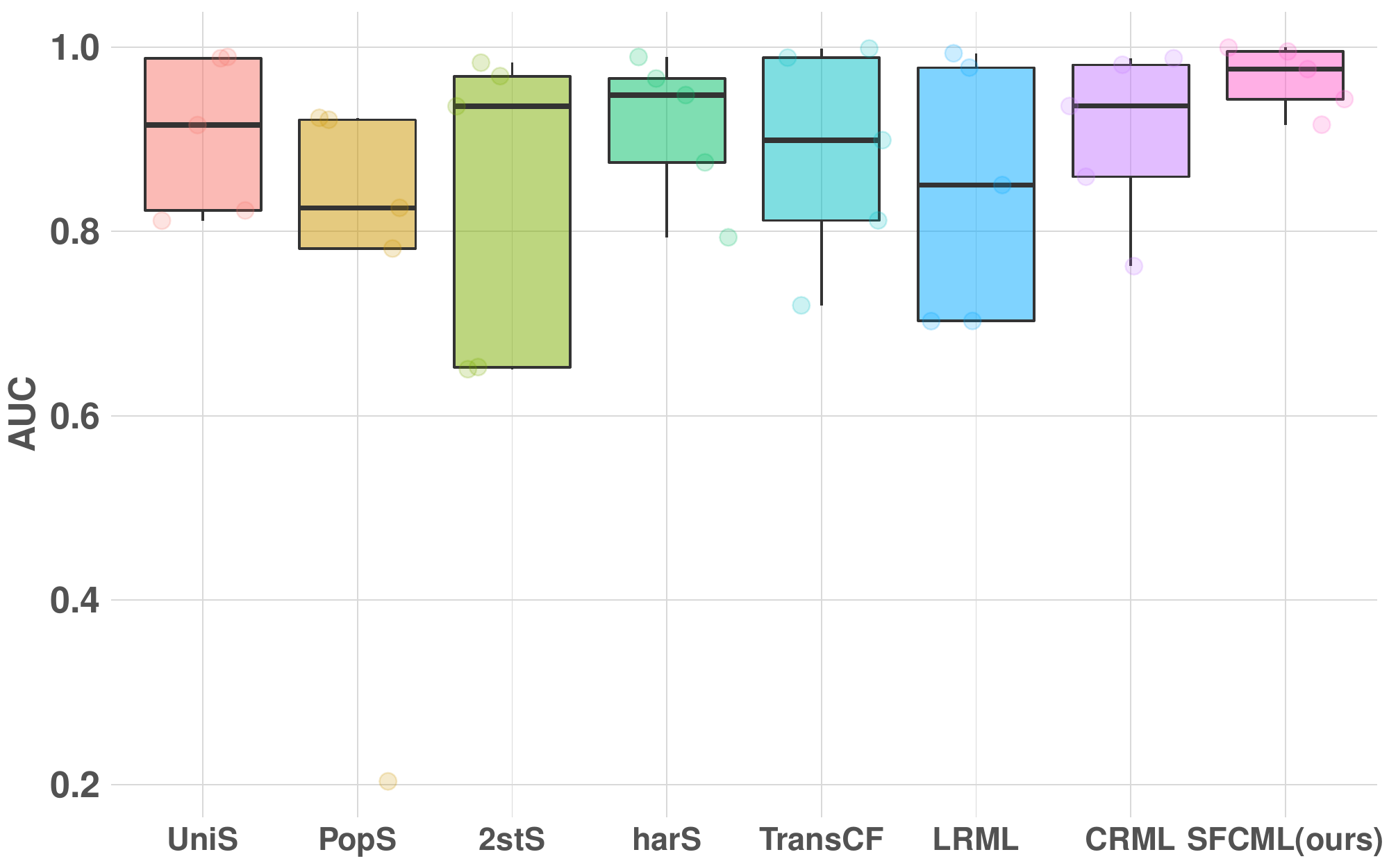}
		\label{main_ml-2}
	}
	
	\subfigure[Anime-1]{
		\includegraphics[width=0.23\textwidth]{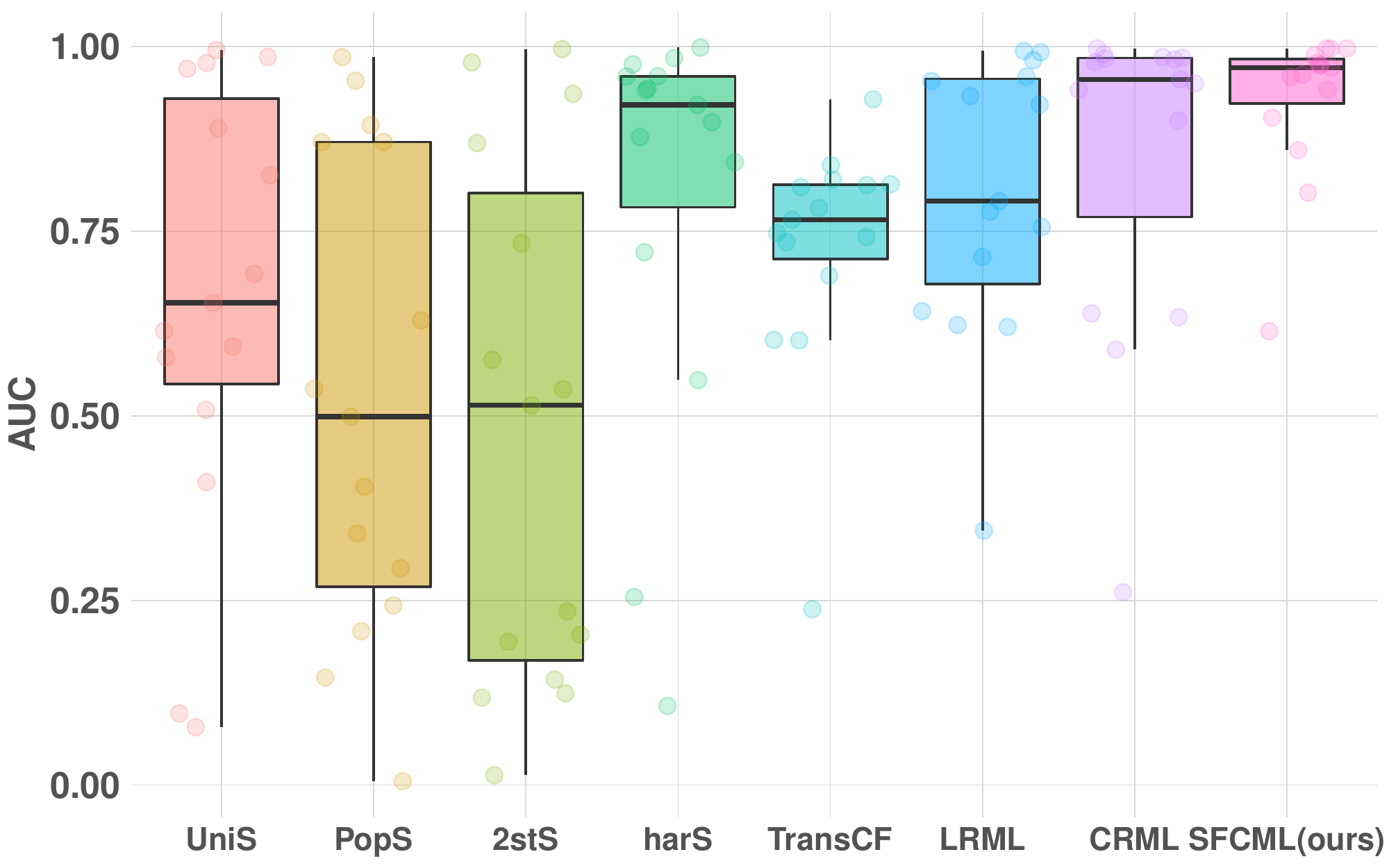}
		\label{Anime-1}
	}
	\subfigure[Anime-2]{
		\includegraphics[width=0.23\textwidth]{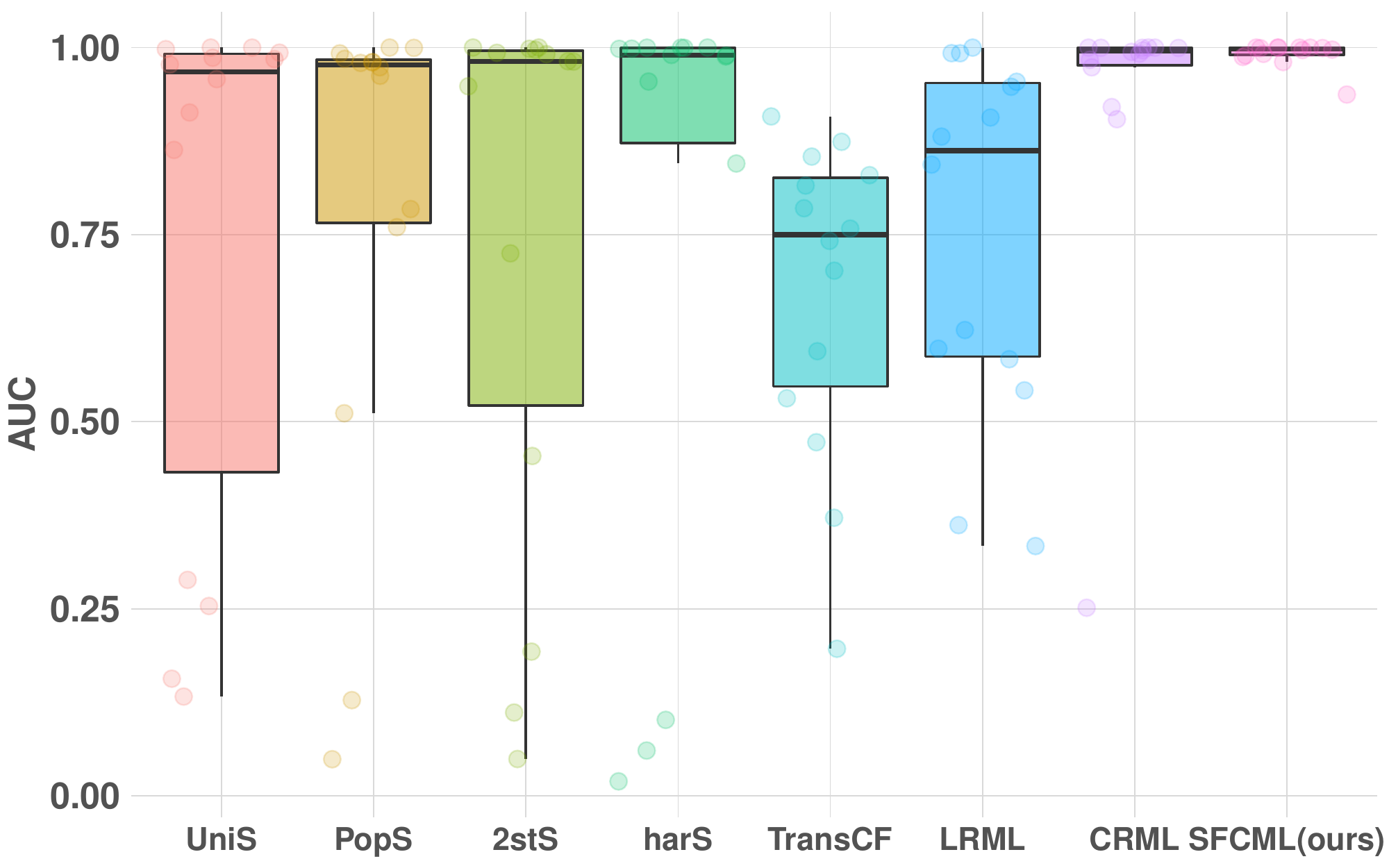}
		\label{Anime-2}
	}
	
	\caption{Fine-grained AUC performance with respect to two users on MovieLens-1m and Anime, respectively. Please refer to Appendix.\ref{auc: app} for more visualizations.}
	\label{auc:citeulike}
\end{figure}

\subsubsection{Comparison against NaiveCML} 
In order to further demonstrate the effectiveness of the acceleration of SFCML, we also report the performance comparison between SFCML and NaiveCML. Unfortunately, as we discussed in Rem.\ref{sec5.2_remark_3}, NaiveCML leads to almost $\mco(\sum_{i=1}^{M} n_i^+n_i^-) = \mco(MN^2)$ space complexity to store the contrastive triplets, while SFCML achieves almost $\mco(MN)$ time and space complexity. Therefore, due to the out-of-memory exception, we only conduct the experiments of NaiveCML on MovieLens-100k and Steam-200k datasets. The performance results are reported in Tab.\ref{results1}, Tab.\ref{results2}, Fig.\ref{ab} and Fig.\ref{ab.sub.200k} (in the Appendix.\ref{more_evaluation_results}). From the empirical results, we can observe that, SFCML can achieve comparable performance against NaiveCML. Since different loss functions usually lead to different optimization goals, it is reasonable that there exists a slight performance gap between SFCML and NaiveCML. This warrants the rationality and effectiveness of our proposed algorithms.

%
%\item \textbf{Naive Collaborative Metric Learning (NaiveCML)} is the initial method that does not adopt any negative sampling strategies to acceleration its computations, which can be treated as the upper bound performance here. Unfortunately, due to the limitation of the memory and heavy computation burden on larger dataset, we only report the experimental results on two tiny datasets, i.e., MovieLens-100k and Steam-200k. 
\subsubsection{Adverse evidence of sampling-based CML}
As we argued in Sec.\ref{sec.3.3}, the recommendation performance of sampling-based CML is largely determined by the negative sampling strategy and the number of sampled items. Since it is difficult to find the best sampling strategy and the number of sampled items, this makes the sampling-based CML methods perform unstably. To validate this claim, we report the performance of different negative sampling strategies and the sampling number $U$ on validation set of the MovieLens-100k dataset. Fig.\ref{neg_num_100k} shows the empirical results in terms of CML framework competitors. The sampling number $U$ is chosen from $\{1, 3, 5, 8, 10\}$. As depicted in Fig.\ref{neg_num_100k}, we can conclude that, given a fixed sampling number, choosing different negative sampling strategies end up with quite different performances. This is due to the fact that different sampling strategies utilize different items to optimize the model. Moreover, given a fixed sampling strategy, the sampling-based CML algorithms also exhibit quite different performances across different $U$. Theoretically, altering different negative sampling strategies and the number of sampled items both induce a different sampling distribution, resulting in a nonzero $D_{TV}$ term in its generalization upper bound (see Thm.\ref{theo3}). This may lead to different generalization performance for sampling-based CML. By contrast, SFCML consistently outperforms all its sampling-based counterparts through the lens of a sampling-free fashion.

%Different from sampling-based CML, since our SFCML is developed in a sampling-free manner without negative sampling, its performance is certain and superior on all metrics. 

%In a nutshell, the above discussions show the issue of sampling-based CML algorithms and validates the superiority of SFCML.

\subsubsection{Fine-grained Performance Visualization} 
Recall that, in Sec.\ref{precon}, with respect to user $u_i$, we say that a contrastive pair $(v_j^+, v_k^-)$ meets the preference consistency if the score $f(v_j^+|u_i) > f(v_k^-|u_i)$. This could be exactly measured by AUC, as discussed in Sec.\ref{sec.5.2}. According to this, at first, we separately report the fine-grained AUC comparison with respect to six users on the MovieLens-1m and Anime datasets, respectively. Specifically, in terms of each positive item, we evaluate its AUC score against all remaining unobserved items. The results are shown in Fig.\ref{auc:citeulike}. We can notice that the AUC values of SFCML are consistently higher than the sampling-based competitors. This indicates that SFCML tends to provide a better preference consistency. By contrast, in most cases, the sampling-based CML results in a heavy tail positive score distribution (i.e., lower AUC score of positive items). This is because they merely leverage a small part of unobserved items to train CML while ignoring other informative samples. In this way, the relative preference consistency could not be well preserved, leading to the degradation of performance. Secondly, we separately select three users on these two datasets. Then, for each user, we plot all of their scores in a scatted plot. The visualization results are presented in Fig.\ref{jetter_ml_1m} and Fig.\ref{jetter_Anime} in Appendix.\ref{auc: app}. The dots here are unobserved instances, and the triangles are positive ones. In the plots, the unobserved examples form a long band, and the dark part of the band corresponds to a higher density of the unobserved distribution. For our algorithm, we see that the triangles tend to form clear cliques and they only overlap with the top and light part of the unobserved band. While for other algorithms, the triangles either span the band or overlap with a dark part of the band. This shows that our proposed method could better separate the positive and unobserved examples apart.

%demonstrate the score density distributions of positive and negative items.  According to the visualizations, we can observe that, SFCML owns a better ranking ability making the scores of positive items higher than negative ones. However, the sampling-based algorithms either have a large number of negative items distributed at top with high scores (such as TransCF), or lead to a lower score of positive samples (e.g., LRML). Moreover, the density of negative samples near the overlaps of positive and negative instances is lower than other sampling-based CML methods, implying that there are fewer negative items violating the preference consistency. This again demonstrates the superiority of SFCML.
% 
%Note that, due to a large number of samples, with respect to each positive/negative item, we vertically add some random jitters (do not change the score value) to make a better and clearer comparison. 
 
\begin{figure}[!ht]
	\centering
		\subfigure[MovieLens-100k]{
			\includegraphics[width=0.23\textwidth]{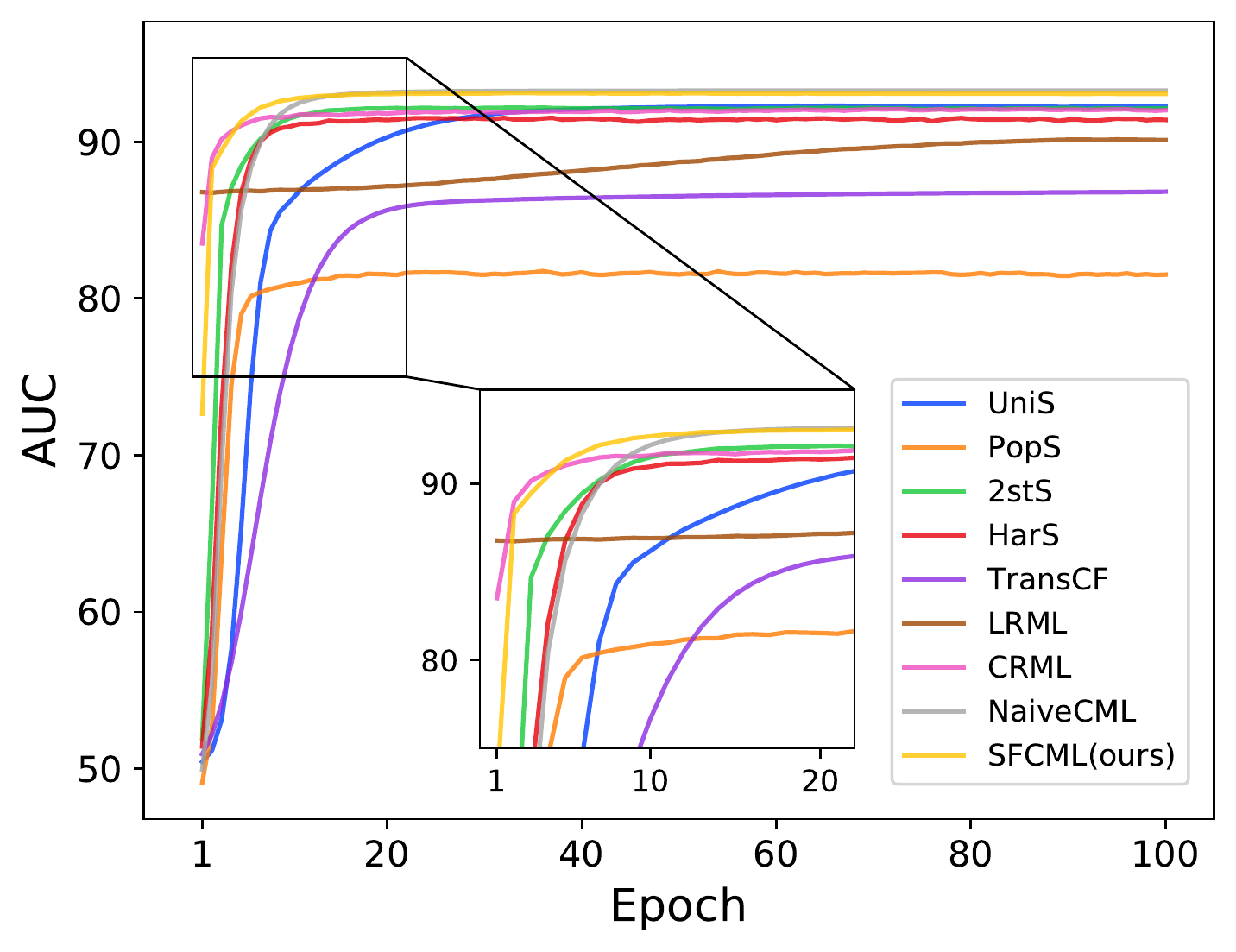}
			\label{line_ml-100k}
		}
		\subfigure[Steam-200k]{
			\includegraphics[width=0.23\textwidth]{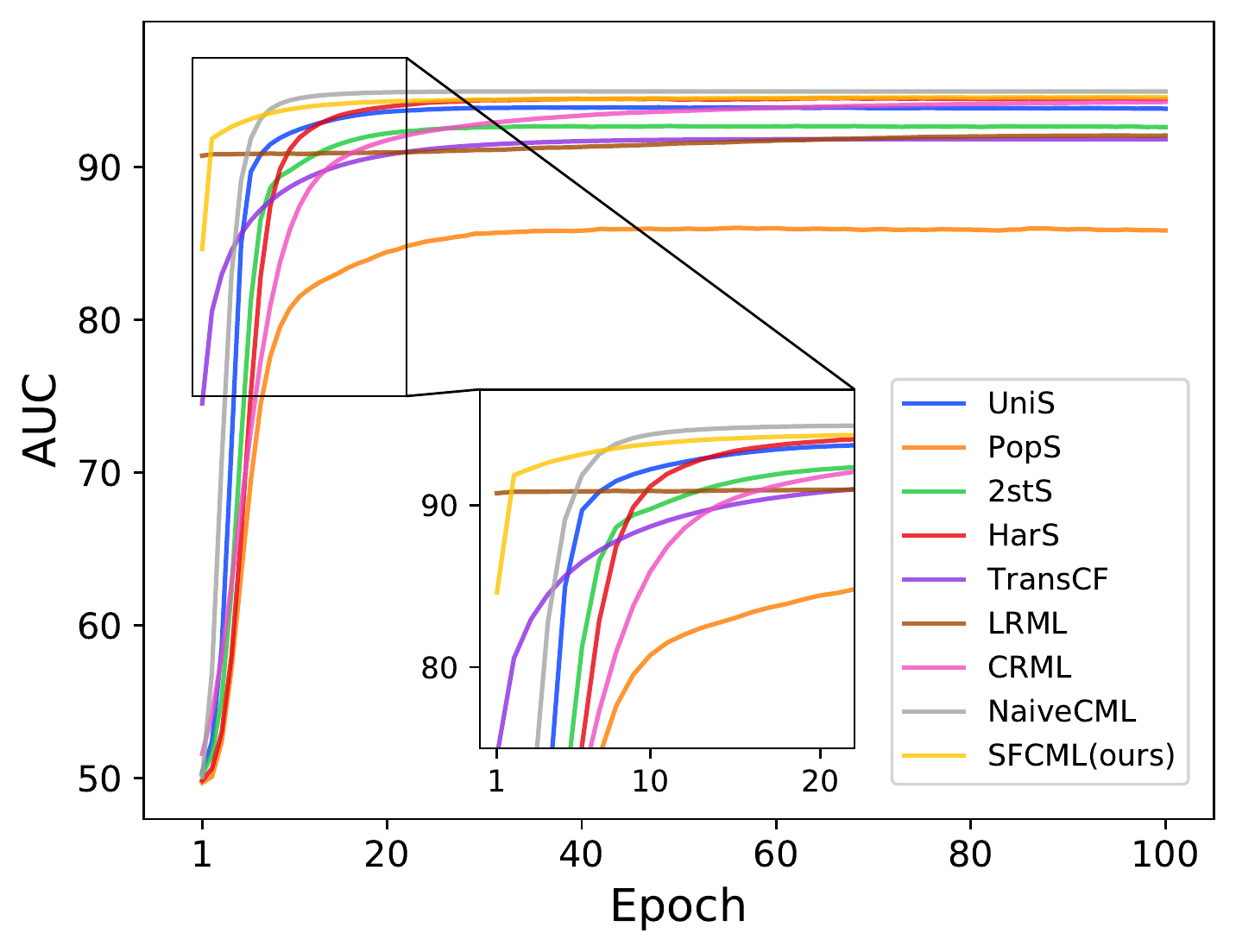}
			\label{line_Steam_200k}
		}
		\caption{Empirical converge analysis of testing AUC of all CML-based algorithms, reporting  $100$ epochs here.}
		\label{empirical_analysis}
\end{figure}

\subsubsection{Empirical analysis} 
Fig.\ref{empirical_analysis} reports the convergence of AUC on MovieLens-100k and Steam-200k datasets. Grounded on the results, we can observe that, with regard to most algorithms, the AUC metric increases to a high value rapidly over only several epochs, and then tends to be stable. Especially, LRML can achieve a competitive performance over fewer iterations owing to the strong expression of its attention-based neural network. Moreover, different sampling-based CML algorithms require different epochs to converge, since they usually leverage different negative items to train the model. Finally, SFCML performs consistently better than other competitors and shows a close performance to the NaiveCML algorithm. 
%Results on other datasets show the same trend and thus they are omitted here. Especially the LRML method, it can achieve 

\begin{figure}[!t]
	\centering
		\subfigure[MovieLens-100k]{
			\includegraphics[width=0.305\columnwidth, height=0.265\columnwidth]{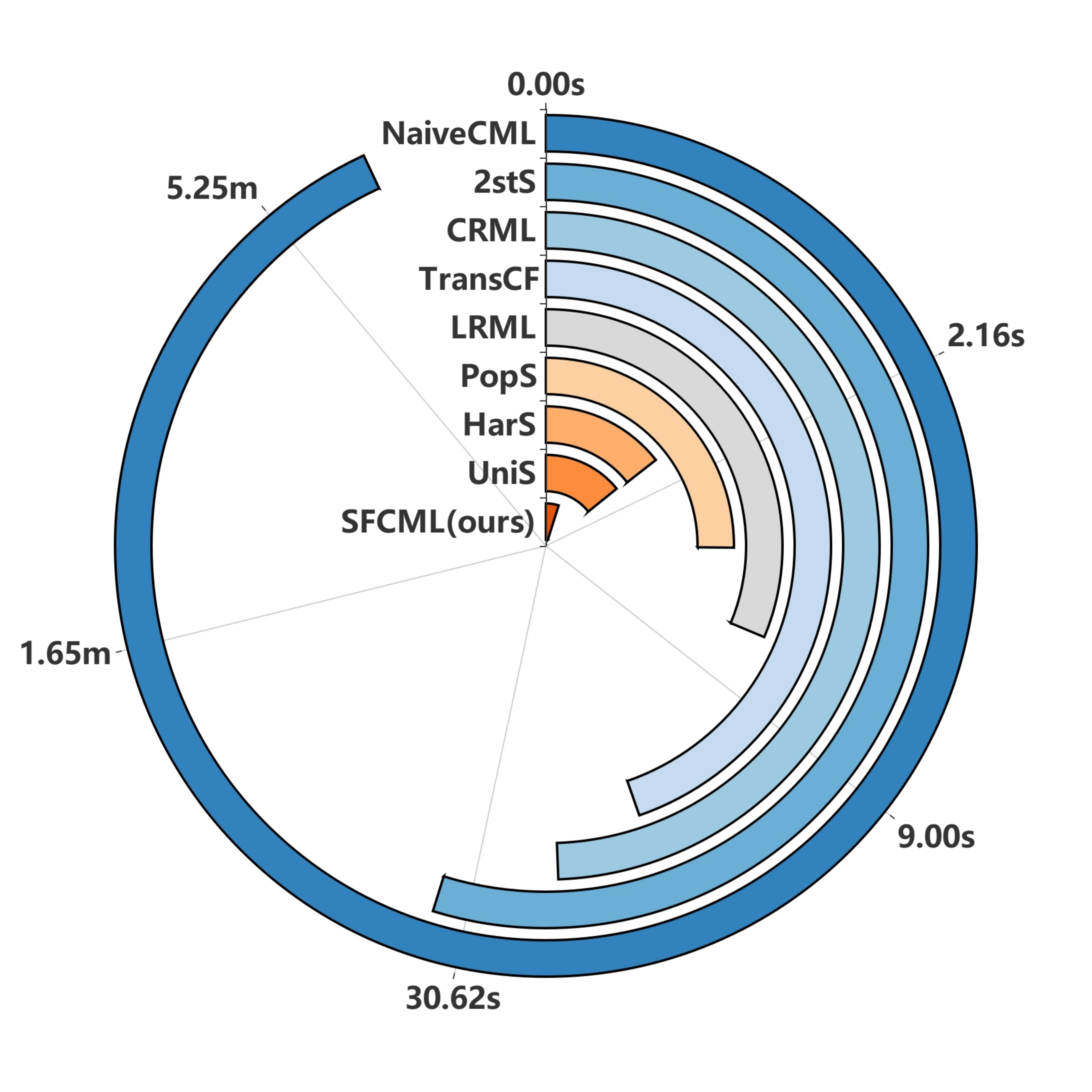}
			\label{ab.sub.ml-100k}
		}
		\subfigure[MovieLens-1m]{
			\includegraphics[width=0.305\columnwidth, height=0.265\columnwidth]{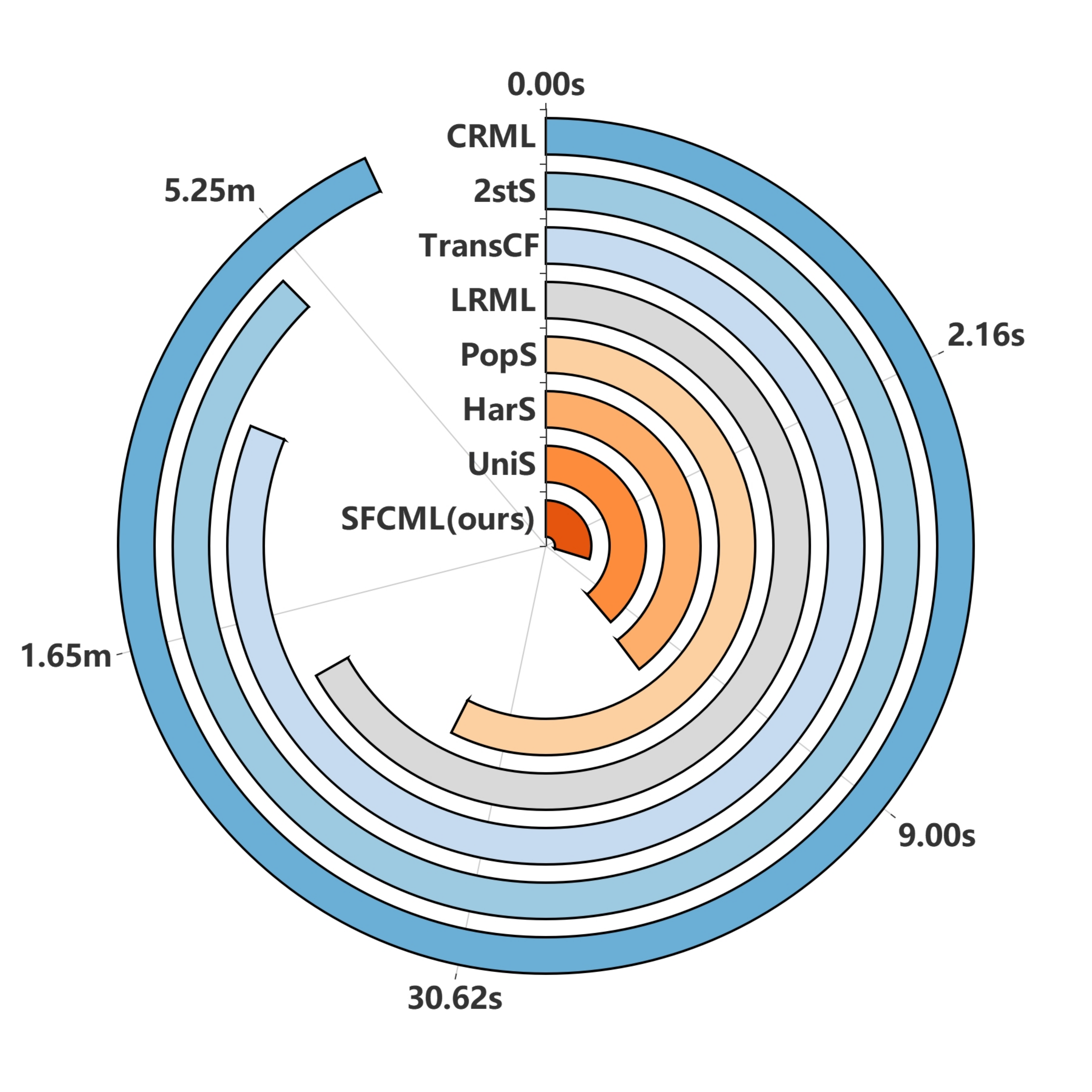}
			\label{ab.sub.ml-1m}
		}
		\subfigure[MovieLens-20m]{
			\includegraphics[width=0.305\columnwidth, height=0.265\columnwidth]{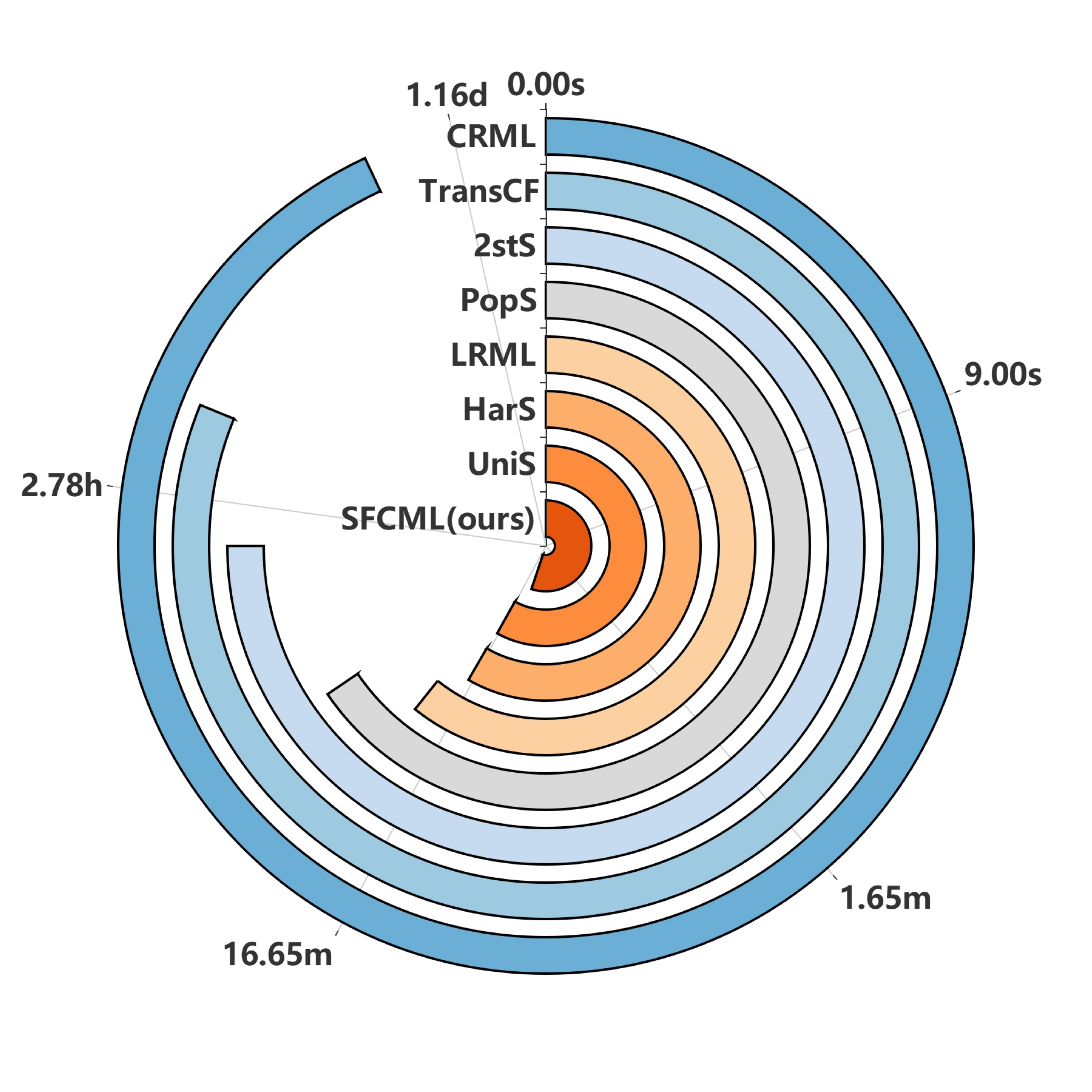}
			\label{ab.sub.ml-20m}
		}
		\caption{Comparisons against average running time with respect to CML framework algorithms and SFCML(ours). The method closer to the center of the circle enjoys better efficiency. Note that, here the 's', 'm', 'h' and 'd' represent the second, minute, hour and day respectively. Please see Appendix.\ref{detail_eff} for more results.}
		\label{runtime}
\end{figure}

\begin{table}[!tbp]
		\centering
		\caption{Statistics information of ml-100k dataset with different preference thresholds $t$.}
		\scalebox{1.0}{
			\begin{tabular}{c|cccc}
				\toprule
				Thresholds  & \multicolumn{1}{c}{\#Users} & \multicolumn{1}{c}{\#Items} & \#Ratings & \%Density \\
				\midrule
				t=1   & 943   & 1,682  & 100,000 & 6.3047 \\
				t=2   & 943   & 1,612  & 93,890 & 5.9264 \\
				t=3   & 943   & 1,574  & 82,520 & 5.5596 \\
				t=4   & 938   & 1,447  & 55,361 & 4.0788 \\
				t=5   & 779   & 1,172  & 20,805 & 2.2788 \\
				\bottomrule
			\end{tabular}%
		}
		\label{tab:addlabel}%
\end{table}%

\subsubsection{Comparison against the efficiency} 
In order to show the efficiency improvements of SFCML against other competitors, we also report the running time of CML competitors, including: a) \textbf{UniS} b) \textbf{PopS} c) \textbf{2stS} d) \textbf{HarS} e) \textbf{TransCF} f) \textbf{LRML} g) \textbf{CRML} h) \textbf{NaiveCML} and i) ours \textbf{SFCML}. Notably, for all sampling-based CML methods, we set the sampling number $U=10$.

Fig.\ref{runtime} shows the average running time over $10$ epochs on all benchmark datasets, and the concrete training time per epoch could be found in the Appendix \ref{detail_eff}. Let us first define the following acceleration ratio (\text{acc.r}):
\begin{equation*}
	\begin{aligned}
		\text{acc.r}
		& = \dfrac{\text{Running time of the slower algorithm}}{\text{Running time of the faster algorithm}}
	\end{aligned}
\end{equation*}
According to Fig.\ref{runtime}, we can draw the following conclusions. At first, SFCML demonstrates even higher efficiency than the sampling-based CML on MovieLens-100k ($4.1$x speed-up against the second-best algorithm), MovieLens-1m ($1.9$x speed-up), Anime ($4.7$x speed-up) and MovieLens-20m ($1.4$x speed-up) datasets. The possible reason lies in that, since the sampling-based algorithms usually need to traverse all observed user-item interactions and then sample unobserved items to generate contrastive pairs for each interaction, this hurts their efficiency for datasets with dense interactions. Accordingly, for the medium/large datasets, such as Anime and MovieLens-20m, TransCF and CRML are much more inefficient than SFCML. Last but not least, compared with the NaiveCML, SFCML significantly reduces the running time without any negative sampling strategies. Although the experiments of NaiveCML on CiteULike, MovieLens-1m, Anime, MovieLens-20m and Amazon-Book could not be finished due to the out-of-memory issue, the efficiency gaps between SFCML and NaiveCML are already sharp on MovieLens-100k and Steam-200k datasets, where the improvements are up to $1114.1$x speed-up and $743.0$x speed-up on MovieLens-100k and Steam-200k, respectively. This validates the effectiveness of proposed accelerations in Sec.\ref{sec.5.2}, making it possible to learn from the whole data under a relatively acceptable efficiency.

% Table generated by Excel2LaTeX from sheet 'statistics_t'

%\begin{table}[]
%	 
%	\centering
%	\caption{Statistics information of ml-100k dataset with different preference threshold $t$.}
%	\begin{tabular}{c|cccc}
%		\toprule
%		Threshold  & \#Ratings & \%Density \\
%		\midrule
%		t=1   & 10.0M   & 6.3047\% \\
%		t=2   & 9.4M  & 5.9264\% \\
%		t=3   & 8.3M  & 5.2329\% \\
%		t=4   & 5.5M  & 3.4676\% \\
%		t=5   & 2.1M  & 1.3240\% \\
%		\bottomrule
%	\end{tabular}%
%	\label{tab:addlabel}%
%  
%\end{table}%

% Table generated by Excel2LaTeX from sheet 'threshold'
\begin{table}[!t]
	\centering
	 
	\caption{The empirical performance of AUC with respect to different preference thresholds $t \in \{1, 2, 3 ,4, 5\}$ on MovieLens-100k. The best and second-best performances are highlighted in bold and underlined, respectively. Please refer to Appendix.\ref{app_pre} for more results. }
	\label{thres_AUC}%
	\scalebox{0.9}{
	\begin{tabular}{c|c|ccccc}
		\toprule
		Method & Trend & \cellcolor[rgb]{ 1.0,  1.0,  1.0} t=1 & t=2 & t=3 & t=4 & t=5 \\
		\midrule
		itemKNN & \begin{minipage}[!t]{0.1\columnwidth}
			\centering
			\raisebox{-.5\height}{\includegraphics[width=\textwidth, height=0.3\textwidth]{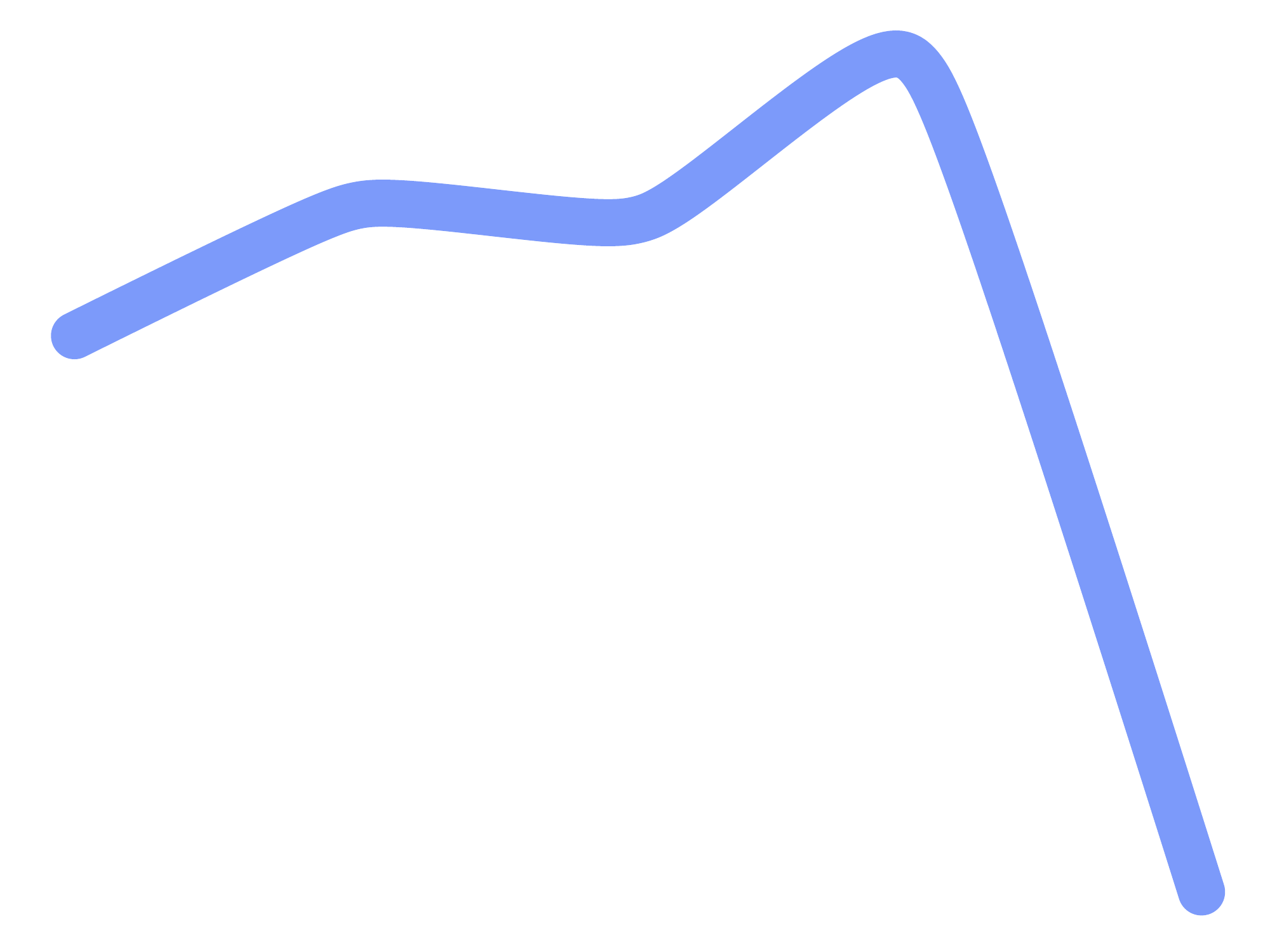}}
		\end{minipage} & \cellcolor[rgb]{ 1.0,  1.0,  1.0}82.60 & \cellcolor[rgb]{ .996,  .984,  .98}84.09 & 83.93 & \cellcolor[rgb]{ .996,  .984,  .98}85.68 & 76.23 \\
		GMF & \begin{minipage}[!t]{0.1\columnwidth}
			\centering
			\raisebox{-.5\height}{\includegraphics[width=\textwidth, height=0.3\textwidth]{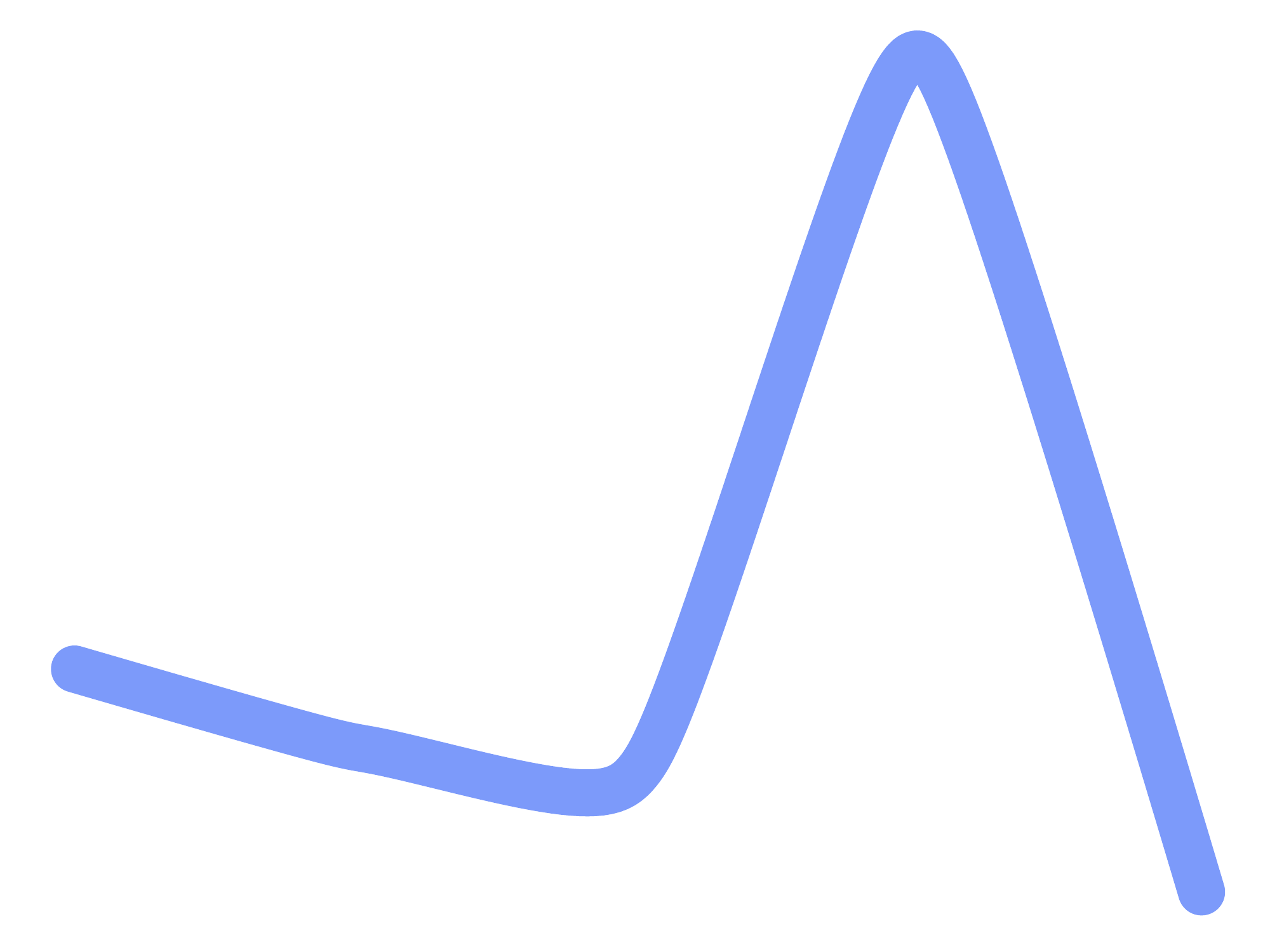}}
		\end{minipage} & \cellcolor[rgb]{ .996,  .984,  .98}84.08 & \cellcolor[rgb]{ .996,  .984,  .98}83.82 & 83.74 & \cellcolor[rgb]{ .996,  .984,  .98}86.12 & \cellcolor[rgb]{ .996,  .984,  .98}83.34 \\
		MLP & \begin{minipage}[!t]{0.1\columnwidth}
			\centering
			\raisebox{-.5\height}{\includegraphics[width=\textwidth, height=0.3\textwidth]{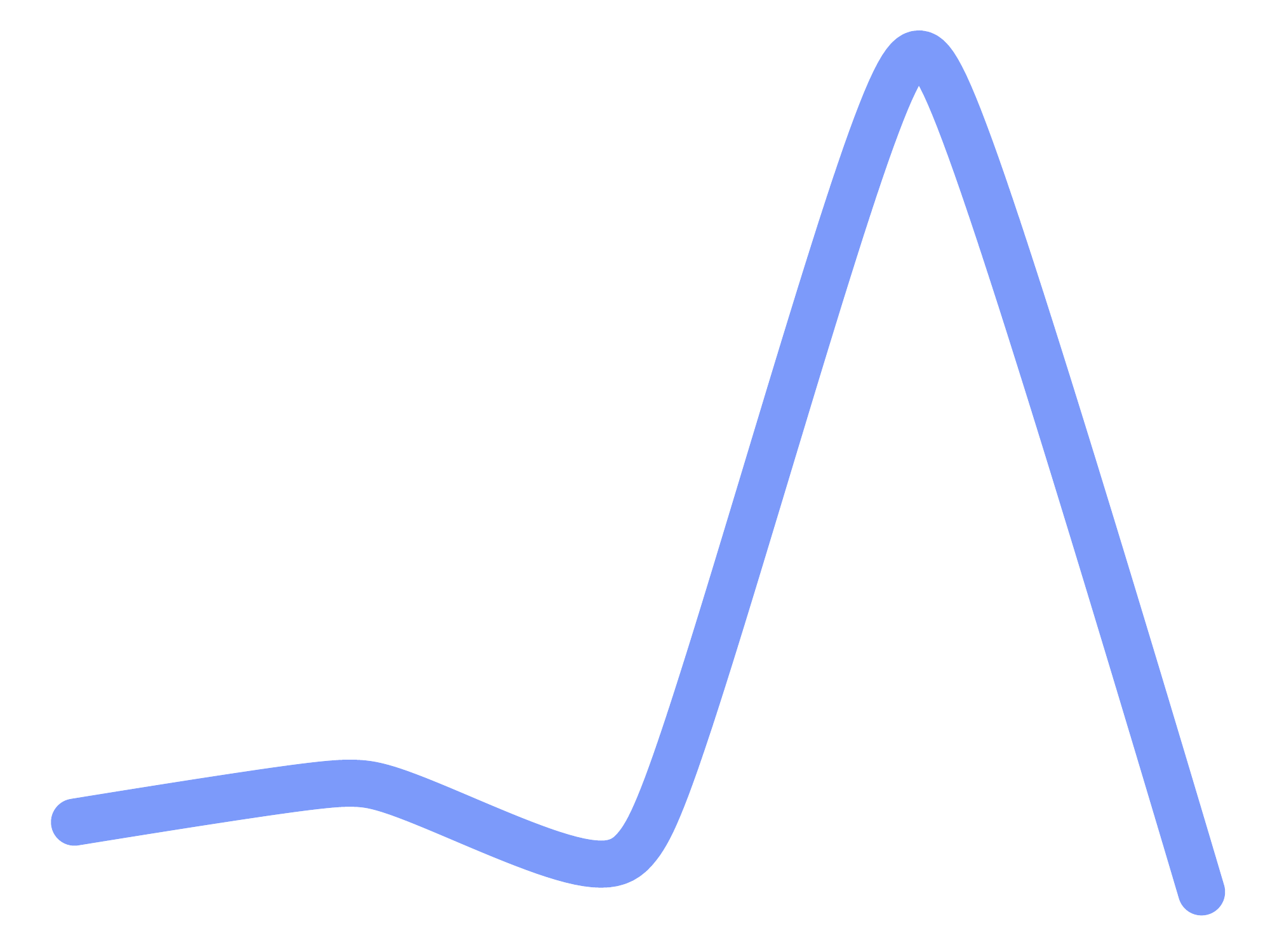}}
		\end{minipage}       & \cellcolor[rgb]{ .996,  .984,  .98}85.11 & \cellcolor[rgb]{ .996,  .984,  .98}85.21 & \cellcolor[rgb]{ .973,  .918,  .918}85.05 & \cellcolor[rgb]{ .996,  .984,  .98}87.09 & \cellcolor[rgb]{ .996,  .984,  .98}84.93 \\
		NCF & \begin{minipage}[!t]{0.1\columnwidth}
			\centering
			\raisebox{-.5\height}{\includegraphics[width=\textwidth, height=0.3\textwidth]{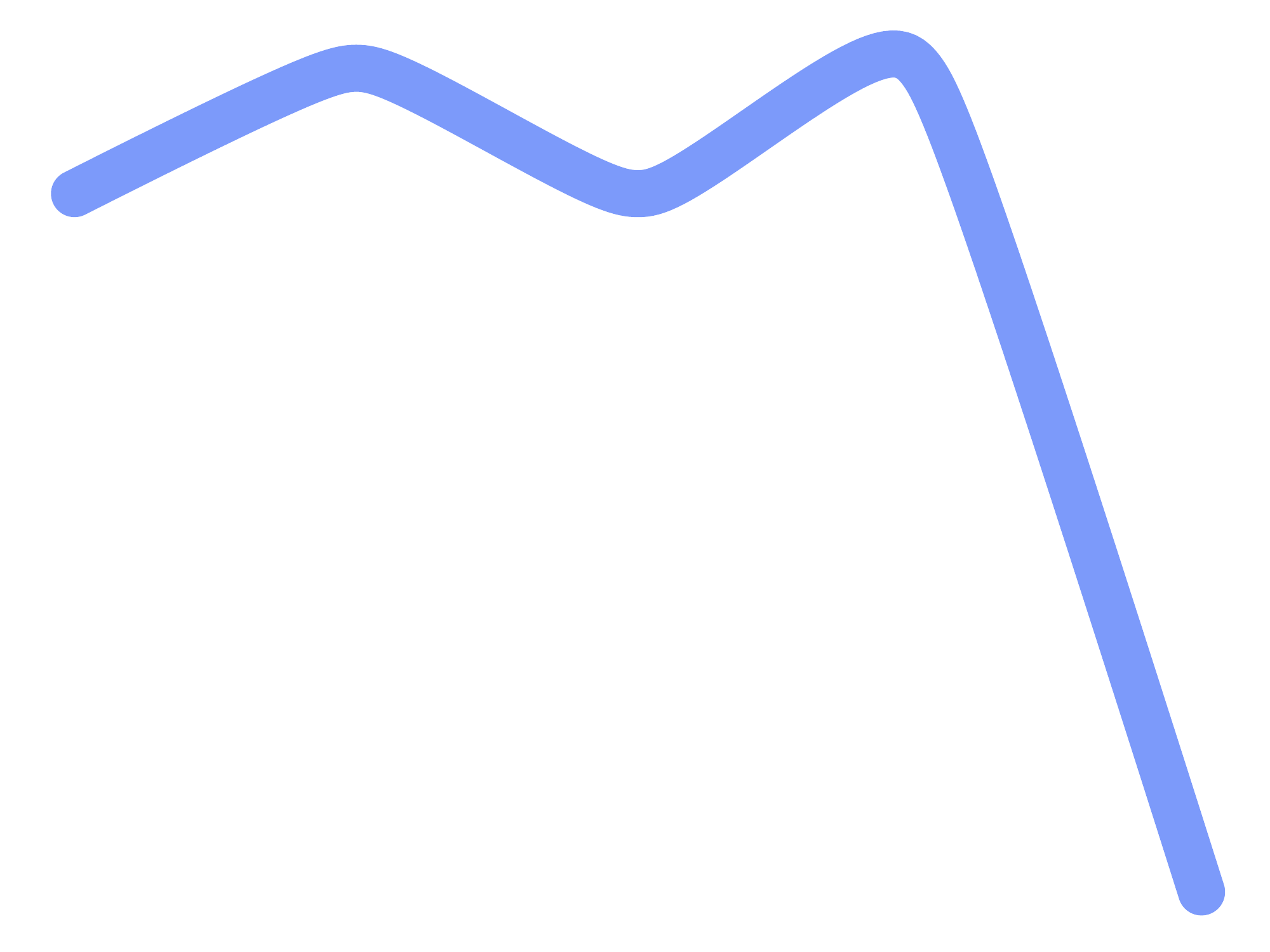}}
		\end{minipage}    & \cellcolor[rgb]{ .996,  .984,  .98}87.52 & \cellcolor[rgb]{ .996,  .984,  .98}88.03 & \cellcolor[rgb]{ .973,  .918,  .918}87.52 & \cellcolor[rgb]{ .996,  .984,  .98}88.03 & \cellcolor[rgb]{ .996,  .984,  .98}84.68 \\
		EHCF & \begin{minipage}[!t]{0.1\columnwidth}
			\centering
			\raisebox{-.5\height}{\includegraphics[width=\textwidth, height=0.3\textwidth]{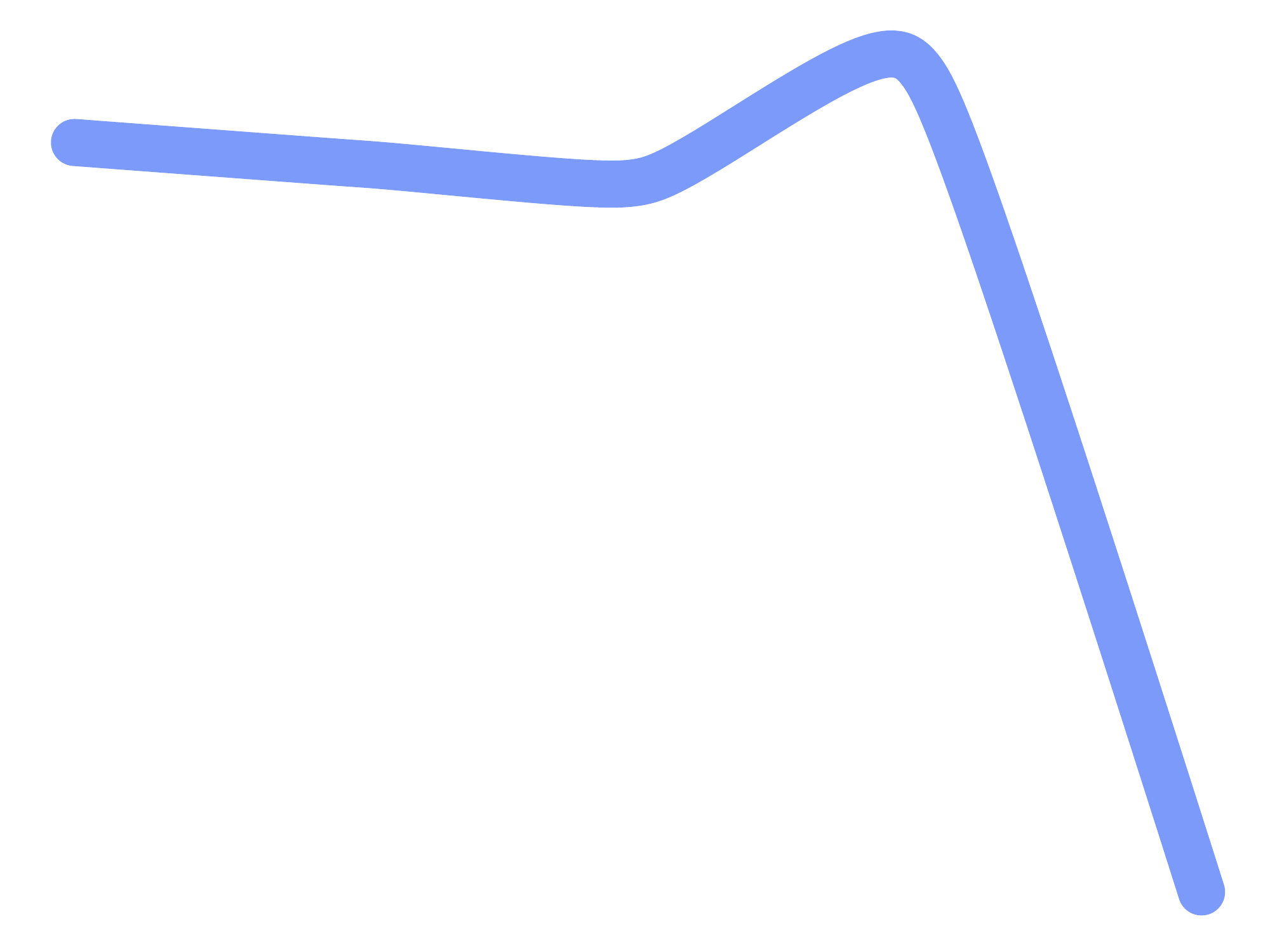}}
		\end{minipage}   & \cellcolor[rgb]{ .945,  .835,  .835}\underline{91.59}  & \cellcolor[rgb]{ .945,  .835,  .835}\underline{91.42}  & \cellcolor[rgb]{ .945,  .835,  .835}\underline{91.27}  & \cellcolor[rgb]{ .961,  .882,  .882}92.18 & \cellcolor[rgb]{ .973,  .918,  .918}85.55 \\
		\cmidrule{1-7}    UniS & \begin{minipage}[b]{0.1\columnwidth}
			\centering
			\raisebox{-.5\height}{\includegraphics[width=\textwidth, height=0.3\textwidth]{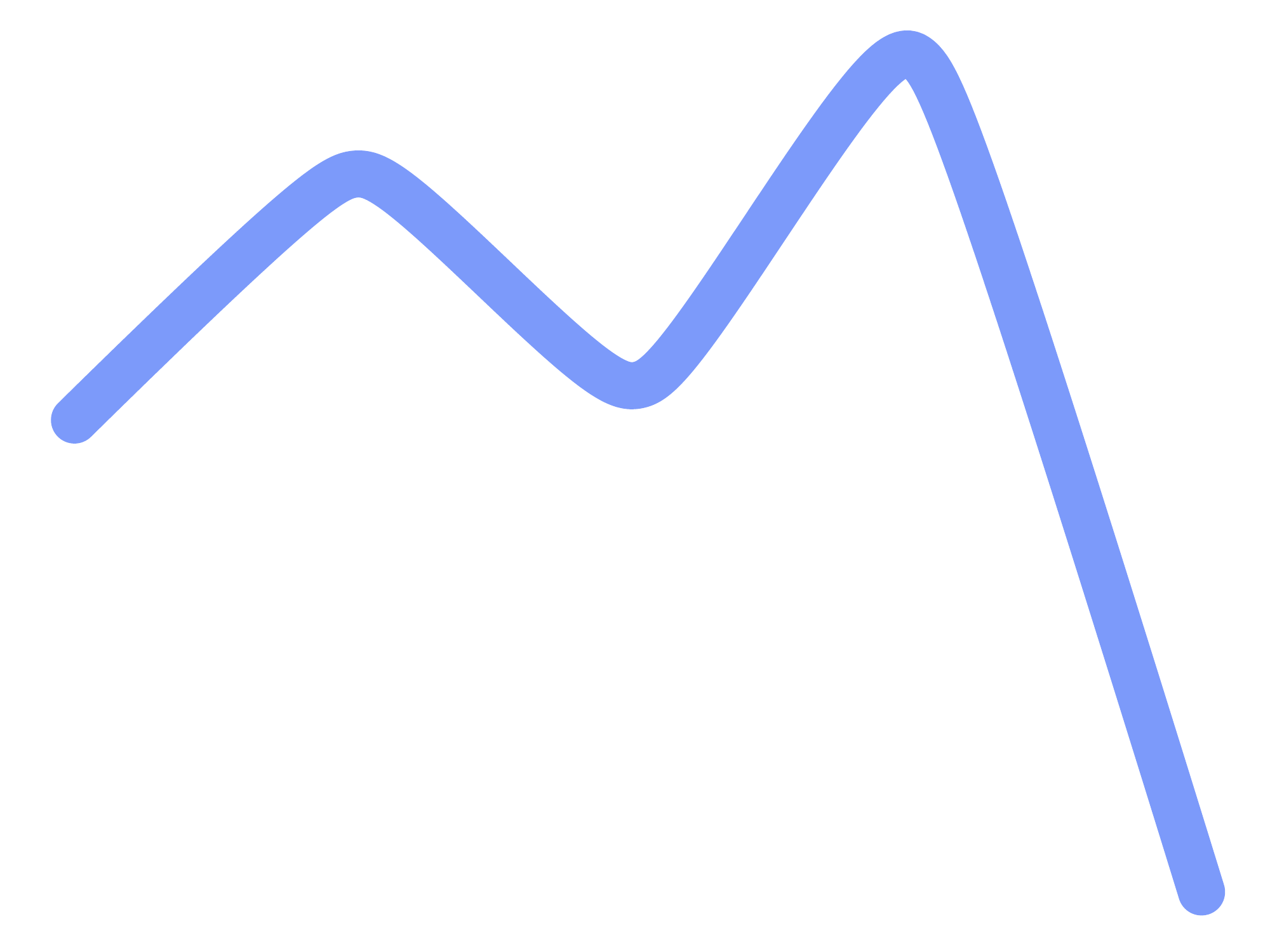}}
		\end{minipage}   & \cellcolor[rgb]{ .973,  .918,  .918}89.59 & \cellcolor[rgb]{ .945,  .835,  .835}\underline{91.42}  & \cellcolor[rgb]{ .961,  .882,  .882}89.85 & \cellcolor[rgb]{ .945,  .835,  .835}\underline{92.27}  & \cellcolor[rgb]{ .961,  .882,  .882}86.08 \\
		PopS & \begin{minipage}[b]{0.1\columnwidth}
			\centering
			\raisebox{-.5\height}{\includegraphics[width=\textwidth, height=0.3\textwidth]{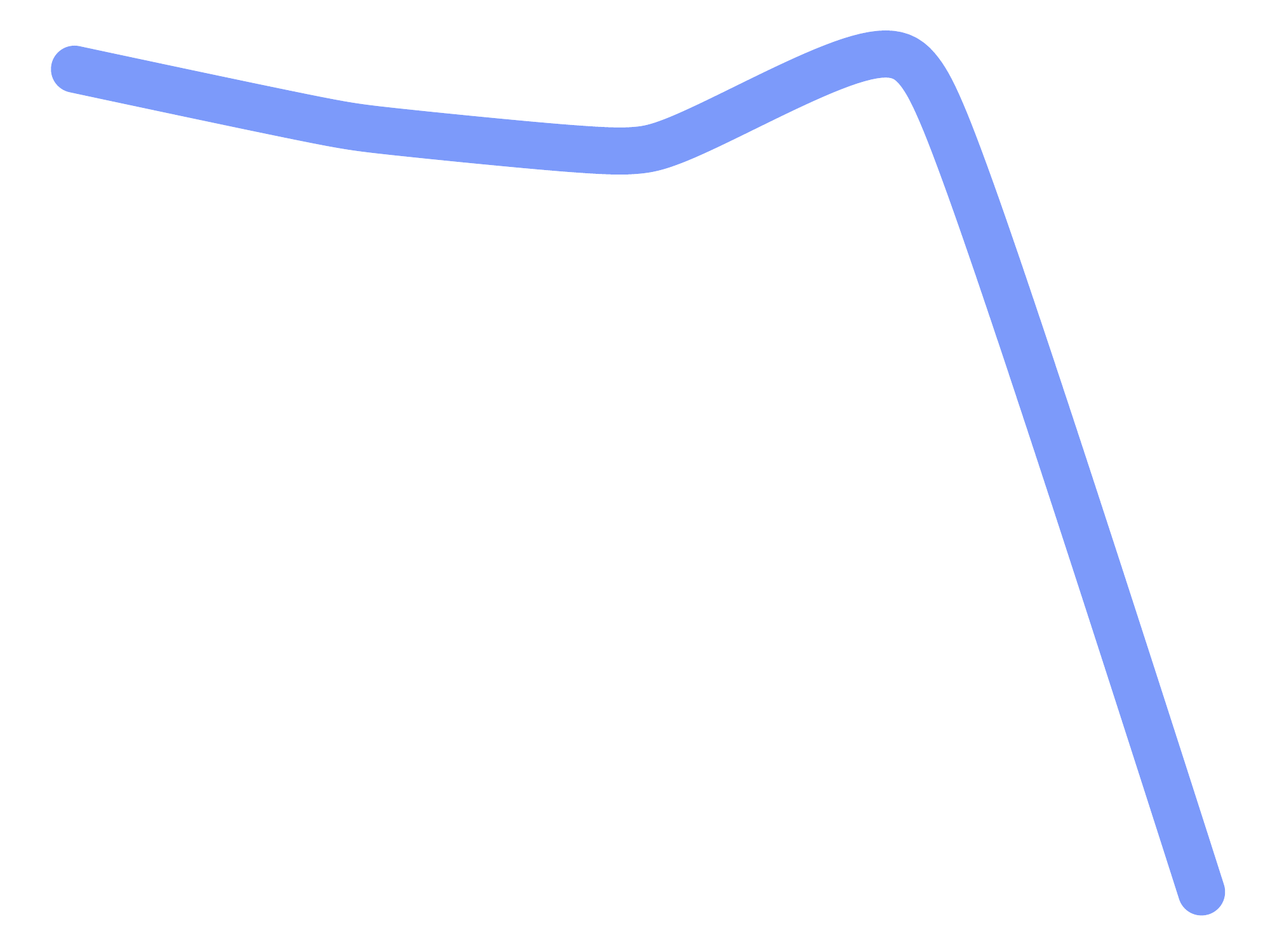}}
		\end{minipage}      & \cellcolor[rgb]{ 1.0,  1.0,  1.0} 80.53 & 80.13 & 79.97 & 80.51 & 74.84 \\
		2stS & \begin{minipage}[b]{0.1\columnwidth}
			\centering
			\raisebox{-.5\height}{\includegraphics[width=\textwidth, height=0.3\textwidth]{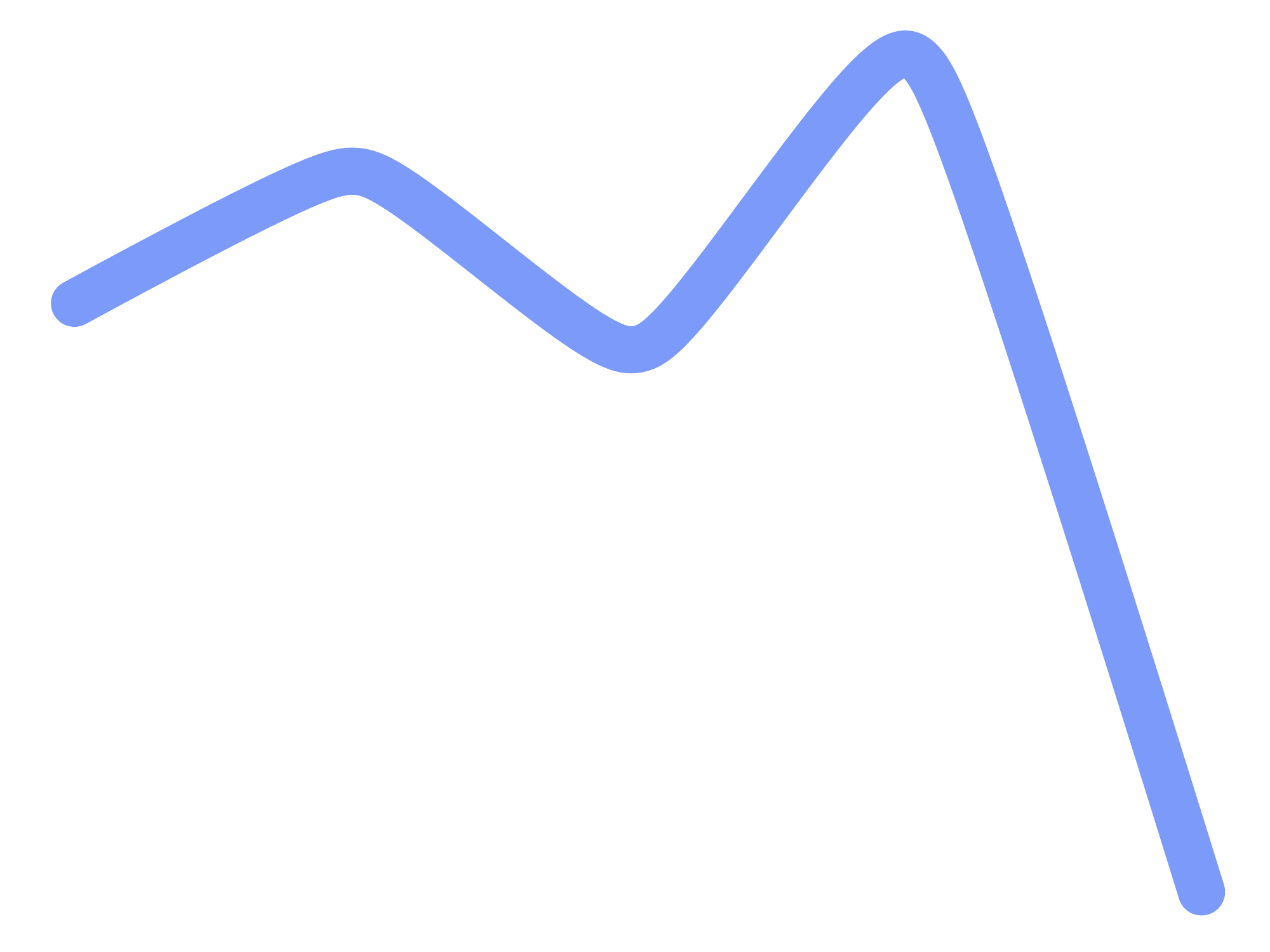}}
		\end{minipage}   & \cellcolor[rgb]{ .973,  .918,  .918}90.21 & \cellcolor[rgb]{ .961,  .882,  .882}91.19 & \cellcolor[rgb]{ .961,  .882,  .882}89.87 & \cellcolor[rgb]{ .961,  .882,  .882}92.01 & \cellcolor[rgb]{ .973,  .918,  .918}85.84 \\
		HarS & \begin{minipage}[b]{0.1\columnwidth}
			\centering
			\raisebox{-.5\height}{\includegraphics[width=\textwidth, height=0.3\textwidth]{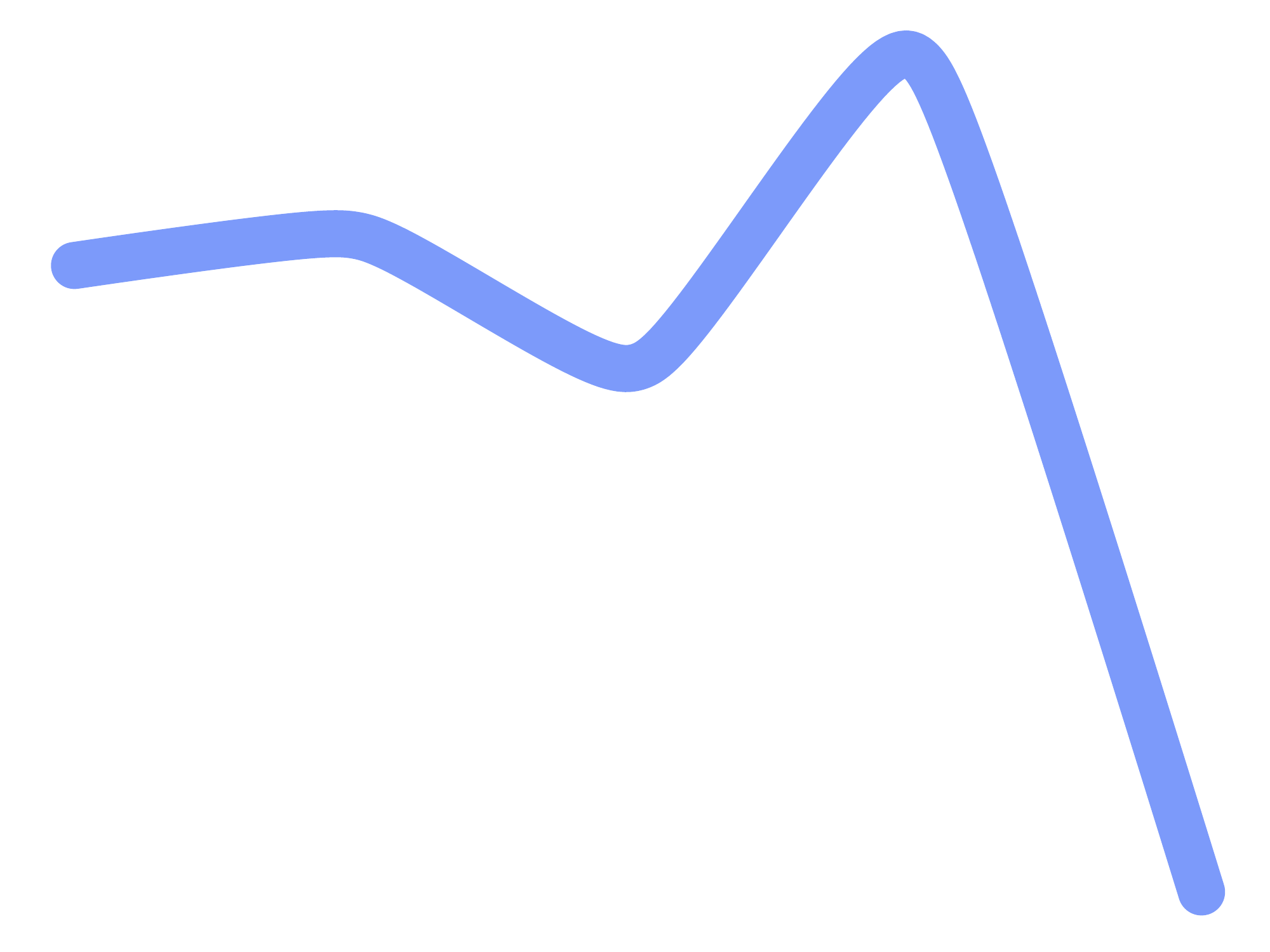}}
		\end{minipage}    & \cellcolor[rgb]{ .961,  .882,  .882}90.91 & \cellcolor[rgb]{ .961,  .882,  .882}91.02 & \cellcolor[rgb]{ .961,  .882,  .882}90.54 & \cellcolor[rgb]{ .961,  .882,  .882}91.66 & \cellcolor[rgb]{ .945,  .835,  .835}\underline{88.62}  \\
		TransCF & \begin{minipage}[b]{0.1\columnwidth}
			\centering
			\raisebox{-.5\height}{\includegraphics[width=\textwidth, height=0.3\textwidth]{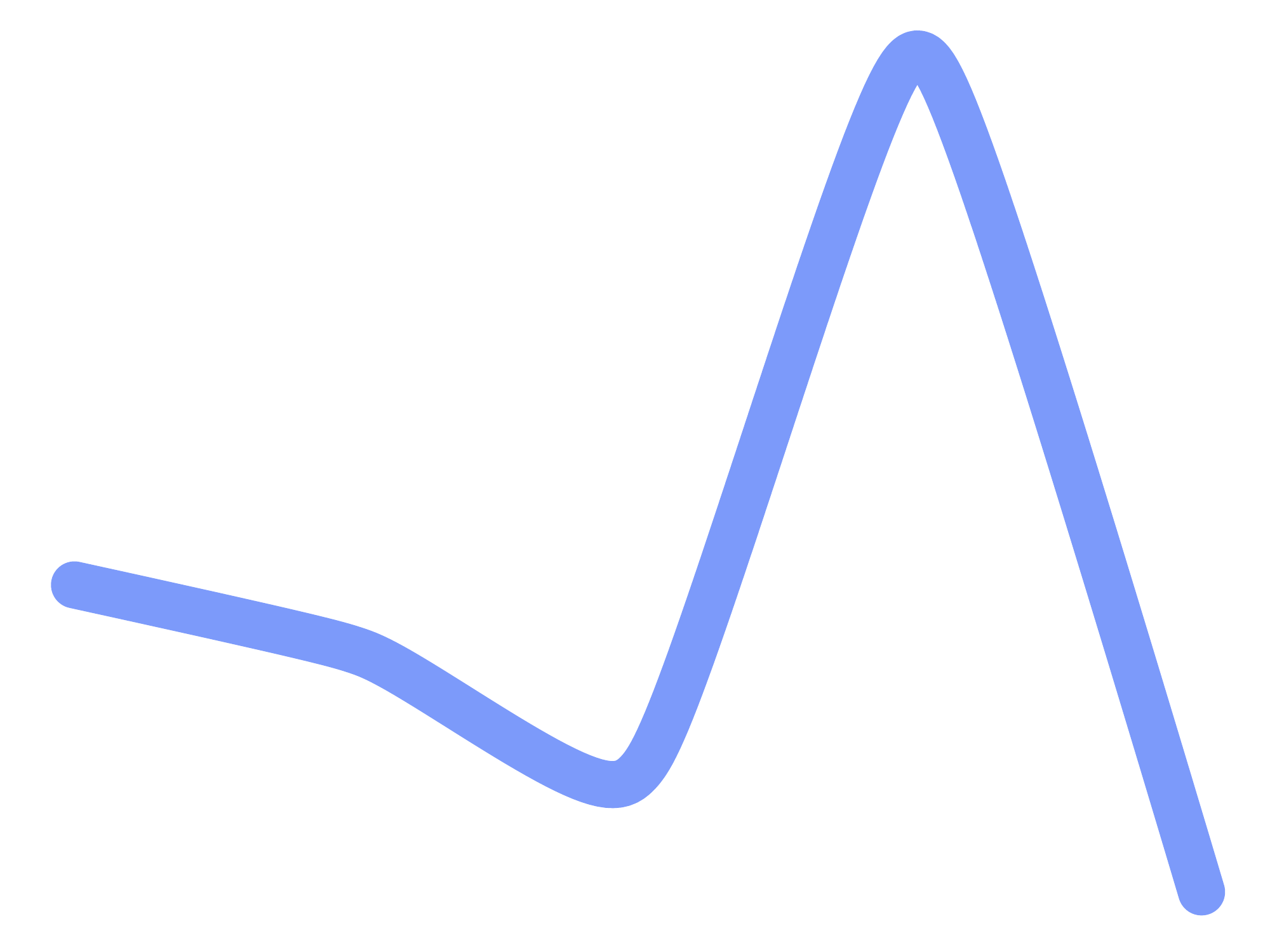}}
		\end{minipage}    & \cellcolor[rgb]{ .996,  .984,  .98}85.75 & \cellcolor[rgb]{ .996,  .984,  .98}85.53 & \cellcolor[rgb]{ .973,  .918,  .918}85.12 & \cellcolor[rgb]{ .996,  .984,  .98}87.53 & \cellcolor[rgb]{ .996,  .984,  .98}84.72 \\
		LRML & \begin{minipage}[b]{0.1\columnwidth}
			\centering
			\raisebox{-.5\height}{\includegraphics[width=\textwidth, height=0.3\textwidth]{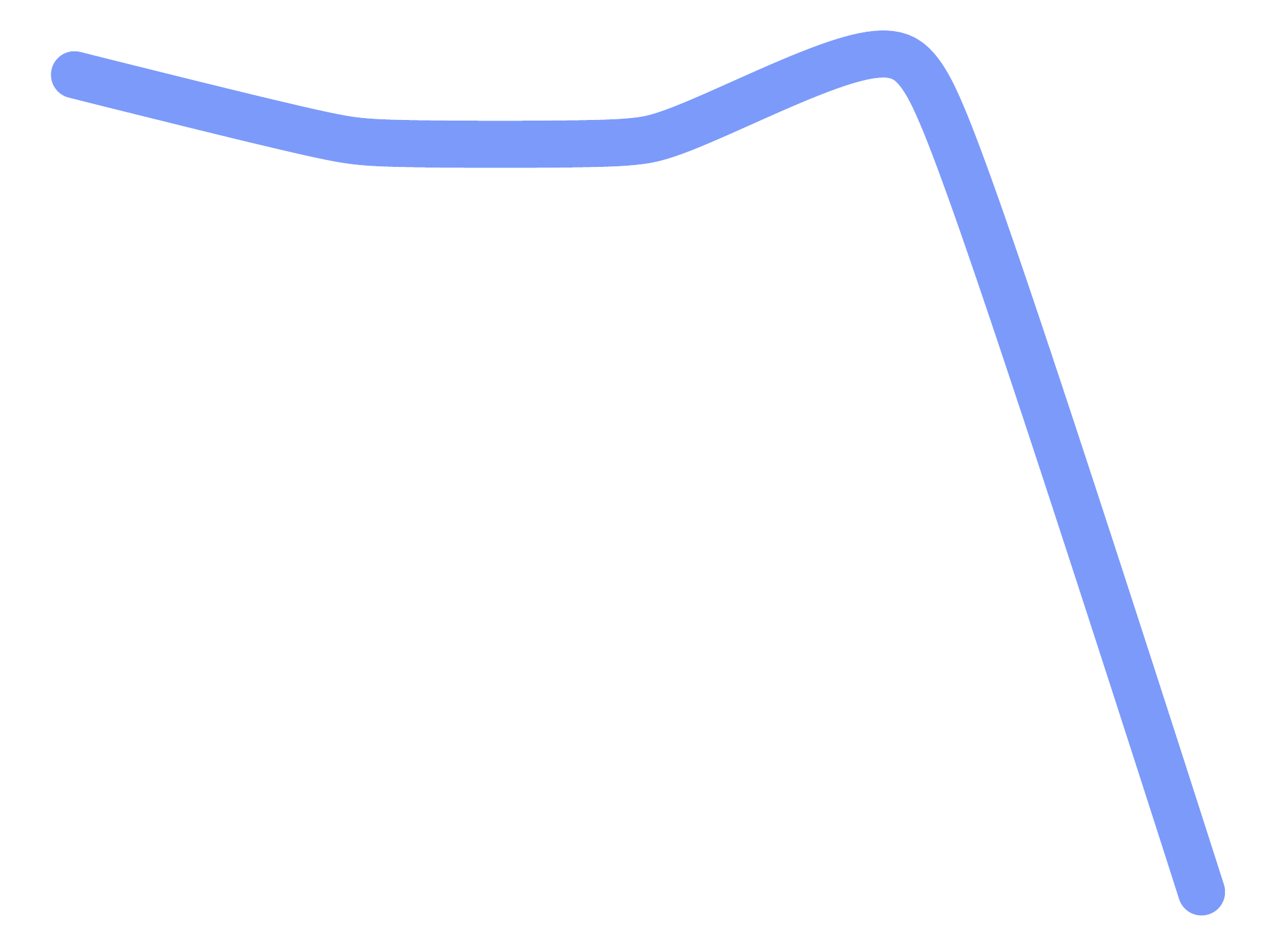}}
		\end{minipage}    & \cellcolor[rgb]{ .973,  .918,  .918}90.37 & \cellcolor[rgb]{ .996,  .984,  .98}89.97 & \cellcolor[rgb]{ .961,  .882,  .882}89.97 & \cellcolor[rgb]{ .973,  .918,  .918}90.38 & \cellcolor[rgb]{ .973,  .918,  .918}85.43 \\
		CRML & \begin{minipage}[b]{0.1\columnwidth}
			\centering
			\raisebox{-.5\height}{\includegraphics[width=\textwidth, height=0.3\textwidth]{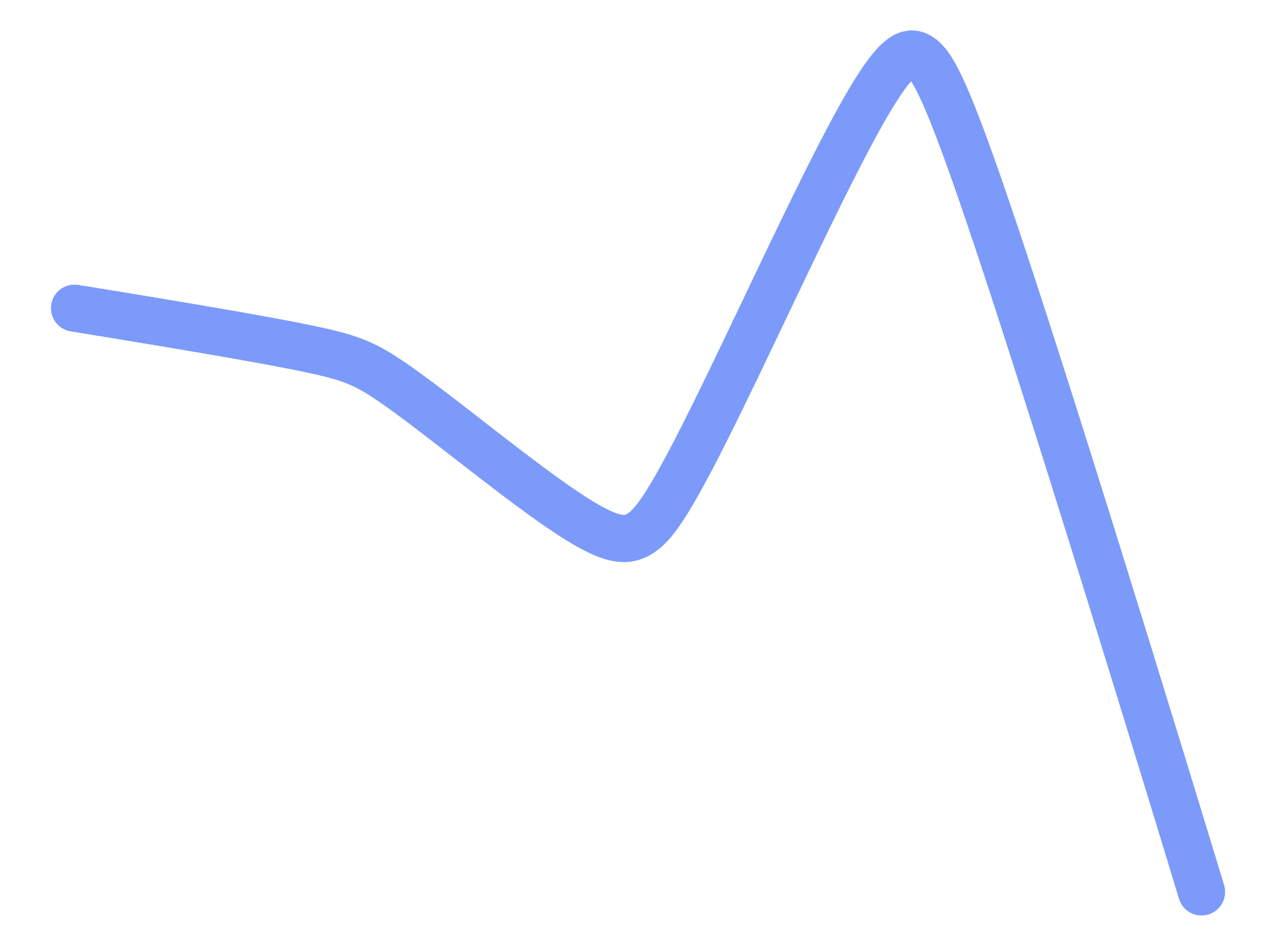}}
		\end{minipage}   & \cellcolor[rgb]{ .961,  .882,  .882}90.76 & \cellcolor[rgb]{ .973,  .918,  .918}90.48 & \cellcolor[rgb]{ .961,  .882,  .882}89.58 & \cellcolor[rgb]{ .961,  .882,  .882}92.07 & \cellcolor[rgb]{ .961,  .882,  .882}87.72 \\
		\cmidrule{1-7}    SFCML(ours) & \begin{minipage}[b]{0.1\columnwidth}
			\centering
			\raisebox{-.5\height}{\includegraphics[width=\textwidth, height=0.3\textwidth]{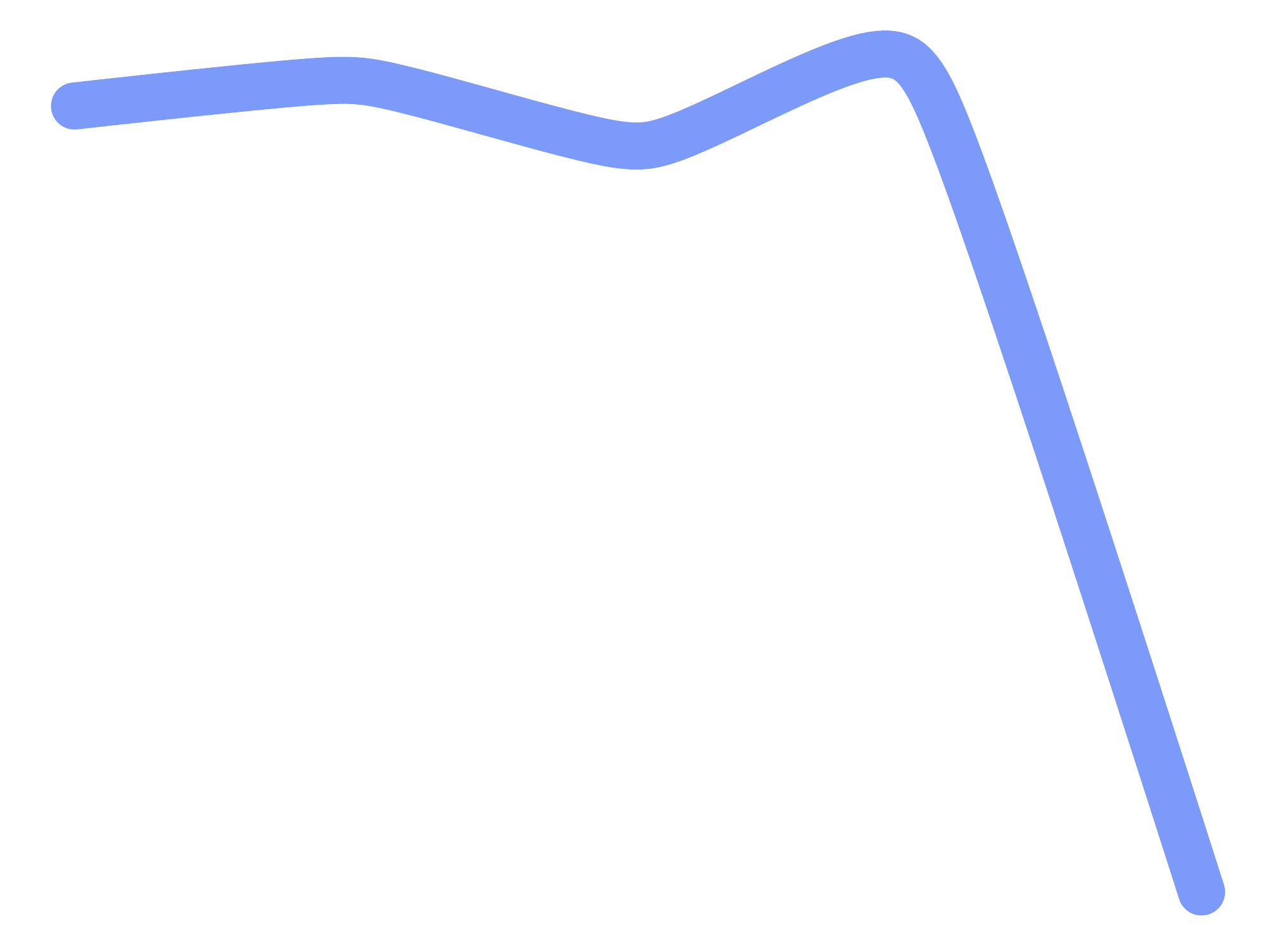}}
		\end{minipage}  
				  & \cellcolor[rgb]{ .929,  .788,  .788}\textbf{92.85} & \cellcolor[rgb]{ .929,  .788,  .788}\textbf{92.97} & \cellcolor[rgb]{ .929,  .788,  .788}\textbf{92.66} & \cellcolor[rgb]{ .929,  .788,  .788}\textbf{93.11} & \cellcolor[rgb]{ .929,  .788,  .788}\textbf{89.11} \\
		\bottomrule
	\end{tabular}%
}
\end{table}%

\subsubsection{Sensitivity Analysis of Preference thresholds}
  
	As we mentioned in Sec.\ref{data_descr}, to exert the explicit feedback to develop the implicit-feedback-driven recommendation, we follow a widely adopted setting in RS \cite{DBLP:conf/kdd/WangWY15, hsieh2017collaborative, tran2019improving, DBLP:conf/mm/BaoXMYCH19}. Concretely, if the score of item $v_j$ rated by user $u_i$ is no less than a preference thresholds $t$, then $v_j$ is treated as a positive item for $u_i$. To figure out the influence of $t$ to the performance, we conduct the sensitivity analysis on MovieLens-100k data with different preference thresholds $t \in \{1, 2, 3, 4, 5\}$. The corresponding statistics of data used in this experiment are listed in Tab.\ref{tab:addlabel} and the empirical results of all algorithms are shown in Tab.\ref{thres_AUC}. Grounded on these results, we have the following conclusions. Firstly, we see that, almost all the algorithms achieve the best AUC performance at $t=4$, which means that the models enjoy the best preference consistency at this threshold. Secondly, for the rest of the metrics, most algorithms tend to demonstrate a similar trend except for the R@$K$ metric with respect to all thresholds. According to the calculations of metrics in Appendix.\ref{app_metrics}, we know that both numerator and denominator of R@$K$ are related to the number of positive interactions $|I_{u_i}|$ in the dataset, while the other metrics are only influenced by the numerator. Since the numerator for all the involved metrics is inversely proportional to $t$, we see that, in most cases, all the metrics except R@$K$ are inversely proportional to $t$ correspondingly. Most importantly, even under different preference thresholds, SFCML still achieves the best performance consistently on all metrics, and the performance improvements are significant compared with other competitors. This further supports the superiority of SFCML. 
	
%	This means that most of the methods are insensitive to the preference thresholds $t$.
	
%	Generally speaking, it is apparent that a higher score given by the user usually indicates a higher level of his/her satisfaction with this product. In light of this intuitiveness, the preference threshold $t$ is usually prone for a relatively large value, since the lower value would introduce false positive interaction to the model, degrading the performance more or less. 

%	# disuse
%	1) When changing $t$ from 1 to 5, the data available to the model is gradually insufficient so that the recommendation task for all methods is become more challenging. This phenomenon is supported by the fact that the performance of all algorithms is not always degrade in most cases with the decreasing of threshold $t$. 

\section{Conclusion}\label{conclusion}
In this paper, we study the issue of sampling-based CML framework and then start an early trial to develop an efficient alternative for CML without negative sampling. Specifically, based on the extended Rademacher Complexity and the specifically designed symmetrization regime, we provide a systematic analysis of the generalization ability of the CML framework. The theoretical analysis shows that the sampling-based CML may fail to obtain a reasonable generalization performance, due to the per-user TV bias term in its generalization upper bound. Meanwhile, we also prove that the biased term would be eliminated in a sampling-free manner. Motivated by this, we propose to learn CML without negative sampling to get rid of the bias and construct an acceleration method to overcome the heavy computational burden. Finally, empirical studies conducted on seven benchmark datasets demonstrate the superiority of our proposed SFCML algorithm. 

\section{acknowledgment}
This work was supported in part by the National Key R\&D Program of China under Grant 2018AAA0102003, in part by National Natural Science Foundation of China: U21B2038, U2001202, U1936208, 61620106009, 62025604, 61931008, 6212200758 and 61976202, in part by the Fundamental Research Funds for the Central Universities, in part by the National Postdoctoral Program for Innovative Talents under Grant BX2021298, in part by the Youth Innovation Promotion Association CAS, in part by the Strategic Priority Research Program of Chinese Academy of Sciences, Grant No. XDB28000000.

\bibliographystyle{IEEEtran}
\bibliography{sample-base-v7}



\begin{IEEEbiography}[{\includegraphics[width=1in,height=1.25in,clip,keepaspectratio]{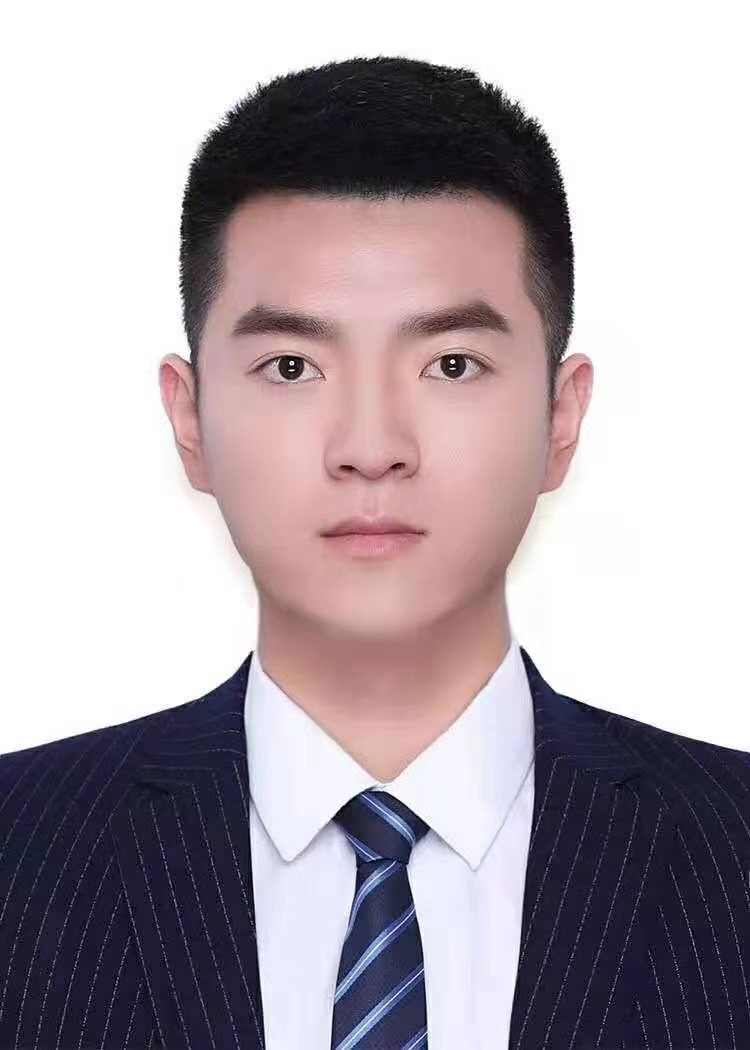}}]{\textbf{Shilong Bao}} 
	received the B.S. degree in College of Computer Science and Technology from Qingdao University in 2019. He is currently pursuing the Ph.D. degree with University of Chinese Academy of Sciences. His research interest is machine learning and data mining. He has authored or coauthored several academic papers in top-tier international conferences and journals including T-PAMI, ICML, and ACM Multimedia. 
\end{IEEEbiography}

\begin{IEEEbiography}
	[{\includegraphics[width=1in,height=1.25in,clip,keepaspectratio]{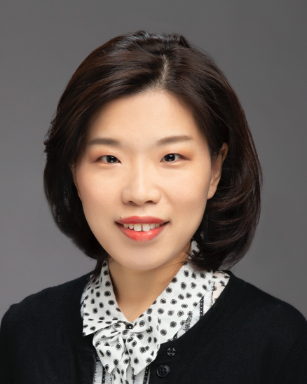}}]{Qianqian Xu} received the B.S. degree in computer science from China University of Mining and Technology in 2007 and the Ph.D. degree in computer science from University of Chinese Academy of Sciences in 2013. She is currently an Associate Professor with the Institute of Computing Technology, Chinese Academy of Sciences, Beijing, China. Her research interests include statistical machine learning, with applications in multimedia and computer vision. She has authored or coauthored 50+ academic papers in prestigious international journals and conferences (including T-PAMI, IJCV, T-IP, NeurIPS, ICML, CVPR, AAAI, etc).	Moreover, she has served as the Senior Program Committee (SPC) of AAAI and IJCAI, Area Chair of ACM MM and ICME, and Reviewer of many leading journals and conferences (including TPAMI, TNNLS, TMM, TCSVT, ICML, NeurIPS, CVPR, ICCV, AAAI, IJCAI, ACM Multimedia, and ICLR).
\end{IEEEbiography}

\begin{IEEEbiography}[{\includegraphics[width=1in,height=1.25in,clip,keepaspectratio]{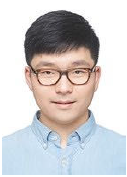}}]{Zhiyong Yang} received the M.Sc. degree in computer science and technology from University of Science and Technology Beijing (USTB) in 2017, and the Ph.D. degree from University of Chinese Academy of Sciences (UCAS) in 2021. He is currently a postdoctoral research fellow with the University of Chinese Academy of Sciences. His research interests lie in machine learning and learning theory, with special focus on AUC optimization, meta-learning/multi-task learning, and learning theory for recommender systems. He has authored or coauthored several academic papers in top-tier international conferences and journals including T-PAMI/ICML/NeurIPS/CVPR. He served as a reviewer for several top-tier journals and conferences such as T-PAMI, ICML, NeurIPS and ICLR.

\end{IEEEbiography}

\begin{IEEEbiography}
	[{\includegraphics[width=1in,height=1.25in,clip,keepaspectratio]{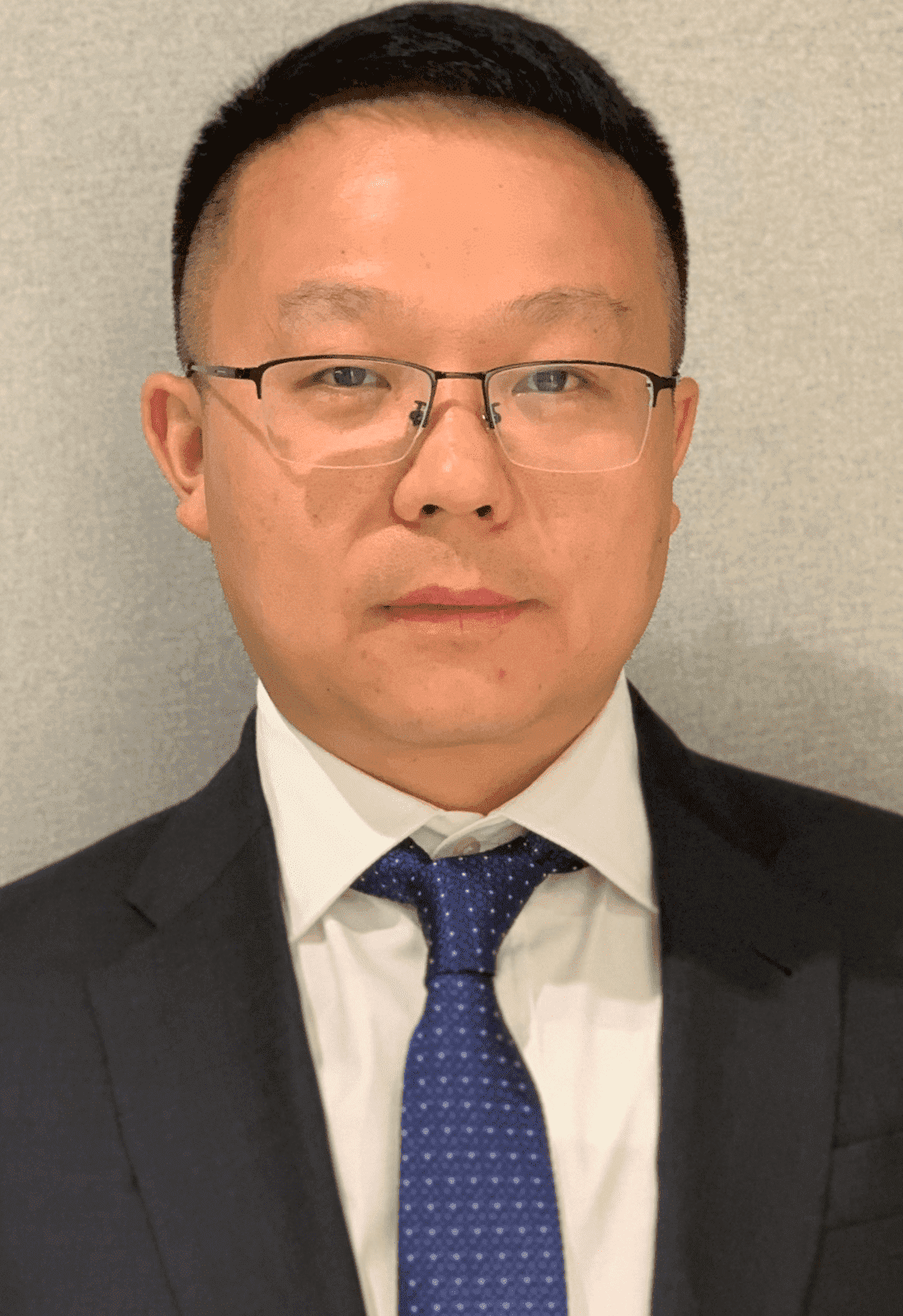}}]{Xiaochun Cao}, Professor of the Institute of Information Engineering, Chinese Academy of Sciences. He received the B.E. and M.E. degrees both in computer science from Beihang University (BUAA), China, and the Ph.D. degree in computer science from the University of Central Florida, USA, with his dissertation nominated for the university level Outstanding Dissertation Award. After graduation, he spent about three years at ObjectVideo Inc. as a Research Scientist. From 2008 to 2012, he was a professor at Tianjin University. He has authored and coauthored over 100 journal and conference papers. In 2004 and 2010, he was the recipients of the Piero Zamperoni best student paper award at the International Conference on Pattern Recognition. He is a fellow of IET and a Senior Member of IEEE. He is an associate editor of IEEE Transactions on Image Processing, IEEE Transactions on Circuits and Systems for Video Technology and IEEE Transactions on Multimedia.
\end{IEEEbiography}

\begin{IEEEbiography}
	[{\includegraphics[width=1in,height=1.25in,clip,keepaspectratio]{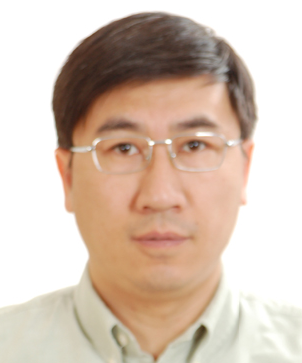}}] {Qingming Huang} is a chair professor in the University of Chinese Academy of Sciences and an adjunct research professor in the Institute of Computing Technology, Chinese Academy of Sciences. He graduated with a Bachelor degree in Computer Science in 1988 and Ph.D. degree in Computer Engineering in 1994, both from Harbin Institute of Technology, China. His research areas include multimedia computing, image processing, computer vision and pattern recognition. He has authored or coauthored more than 400 academic papers in prestigious international journals and top-level international conferences. He was the associate editor of IEEE Trans. on CSVT and Acta Automatica Sinica, and the reviewer of various international journals including IEEE Trans. on PAMI, IEEE Trans. on Image Processing, IEEE Trans. on Multimedia, etc. He is a Fellow of IEEE and has served as general chair, program chair, area chair and TPC member for various conferences, including ACM Multimedia, CVPR, ICCV, ICME, ICMR, PCM, BigMM, PSIVT, etc.
\end{IEEEbiography}

\clearpage
\newpage
\onecolumn
\appendices

\section*{\textcolor{blue}{\Large{Contents}}}
\startcontents[sections]
\printcontents[sections]{l}{1}{\setcounter{tocdepth}{2}}
\newpage

\section{Preliminary for Generalization Analysis}\label{app:gen_pre}
\subsection{Preliminary Lemmas}
In this section, we first briefly review some preparatory knowledge for the proof.
\begin{defi} [Bounded Difference Property]\label{def:bdp}
	Given a group of independent random variables $X_1, X_2, \cdots, X_n$ where $X_t \in \mathbb{X}, \forall t$, $f(X_1,X_2,\cdots, X_n)$ is satisfied with the bounded difference property, if there exists some non-negative constants $c_1, c_2,\cdots, c_n$, such that: 
	\begin{equation}
		\sup_{x_1,x_2,\cdots,x_n, x'_t} \left|f(x_1,\cdots,x_n) - f(x_1,\cdots, x_{t-1},x'_t,\cdots,x_n)\right| \le c_t, ~ \forall t,  1 \le t \le n.
	\end{equation}
\end{defi}

Hereafter, if any functions $f$ holds the Bounded Difference Property, the following Mcdiarmid's inequality is always satisfied.

\begin{lem}[Mcdiarmid's Inequality \cite{mc}] \label{lem:mc} Assume we have $n$ independent random variables $X_1, X_2, \dots, X_n$ that all of them are chosen from the set $\mathcal{X}$. For a function $f: \mathcal{X} \rightarrow \mathbb{R}$, $\forall t, 1 \le t \le n$, if the following inequality holds:
\[
		\sup_{x_1,x_2,\cdots,x_n, x'_t} \left|f(x_1,\cdots,x_n) - f(x_1,\cdots, x_{t-1},x'_t,\cdots,x_n)\right| \le c_t, ~ \forall t,  1 \le t \le n.
\]with $\boldsymbol{x} \neq \boldsymbol{x}'$, then for all $\epsilon >0$, we have
\[\mathbb{P}[ \expe(f) - f \ge \epsilon ] \le \exp\left(\dfrac{-2\epsilon^2}{\sum_{t=1}^nc_t^2} \right), \]
\[
\mathbb{P}[ f - \expe(f) \ge \epsilon ] \le \exp\left(\dfrac{-2\epsilon^2}{\sum_{t=1}^nc_t^2} \right).
\]
\end{lem}

\begin{lem} [$\phi$-Lipschitz Continuous] \label{lem2} Given a set $\mathcal{X}$ and a function $f: \mathcal{X} \rightarrow \mathbb{R}$, if $f$ is continuously differentiable on $\mathcal{X}$ and the derivative of $f$ is Lipschitz continuous on $\mathcal{X}$ with constant $\mu$:
	\[
	\left\|f(x) - f(y)\right\| \le \phi \left\|x - y\right\|.
	\]
	Thereafter, $f$ is said to a $\phi$-Lipschitz continuous function.
\end{lem}

\begin{lem}[Talagrand Contraction Lemma] \label{lem: tala} Let $h_1, h_2, \dots, h_n$ be a series of $\phi$-Lipschitz continuous function and $\sg_1, \dots, \sg_n$ be the independent Rademacher random variables, the following holds
	\[
	\frac{1}{n} \expe_{\sg} \left[ \mathop{\sup}\limits_{f \in \mch} \sum_{i = 1}^n \sigma_i \cdot (h_i \circ f)(z)\right] \le \frac{\phi}{n} \expe_{\sg} \left[ \mathop{\sup}\limits_{f \in \mch} \sum_{i = 1}^n \sigma_i \cdot f(z)\right]
	\]
\end{lem}

\section{Generalization Bounds of CML framework and its proofs} \label{SFCML}

\begin{rdef2} \label{rdef2} (CML \textbf{Rademacher Complexity}). Given the sample set $\mcs = \mathop{\cup}\limits_{u_i \in \mcu} \mcs_i$ and $\mcs_i = \{v_j^+\}_{j=1}^{n_i^+} \cup \{v_k^-\}_{k=1}^{n_i^-}$ with $n_i^+ + n_i^- = N$ and the hypothesis space $\mathcal{H}_R$, then the empirical CML Rademacher Complexity with respect to the sample $\mcs$ is defined as:
\begin{equation}
	\hat{\mfr}_{\ell,\mcs}^{\text{cml}}(\mathcal{H}_R) = \frac{1}{M} \sum_{u_i \in \mathcal{U}} \mathop{\expe}\limits_{\bss_i} \left[\mathop{\sup}\limits_{\mathcal{H}_R} \frac{1}{n_i^+n_i^-}\sum_{j=1}^{n_i^+} \sum_{k=1}^{n_i^-} 
	\mathcal{Q}_{(i)}^{jk} \right], \label{eq65}
\end{equation}
where 
\[
\mathcal{Q}_{(i)}^{jk} = \frac{\sigma^+_{ij} + \sigma^-_{ik}}{2} \cdot \ell^{(i)}(v_j^+, v_k^-);
\]
$\bss_i =[\sg^+_{i1}, \sg^+_{i2}, \dots, \sg^+_{in_i^+}, \sg^-_{i1},\sg^-_{i2}, \dots,\sg^-_{in_i^-}]^\top$ 
is i.i.d Rademacher random variables uniformly chosen from $\{-1, +1\}$ and we have $\mathbb{P}(\sg=1) = \mathbb{P}(\sg=-1) = 0.5$. Next, the population version of the Rademacher Complexity of CML is represented as $\mfr_{\ell, \mcs}^{\text{cml}}(\mathcal{H}_R) = \mathop{\expe}\limits_{\mcs}\left[\hat{\mfr}_{\ell,\mcs}^{\text{cml}}(\mathcal{H}_R)\right]$. 
\end{rdef2}

%To make the following proofs more clearly, we first introduces some operation notions
%\begin{itemize}
%	\item $\textcolor{blue}{(*)}$ means applying the inequality: $\sup (x + y) \le \sup(x) + \sup(y)$.
%	\item $\textcolor{blue}{(**)}$ is the fact $\left\|\boldsymbol{x}\right\|_1 \le \sqrt{b} \left\|\boldsymbol{x}\right\|_2, \boldsymbol{x} \in \mathbb{R}^b$.
%	\item $\textcolor{blue}{(***)}$ is achieved by the Cauchy-Buniakowsky-Schwarz Inequality $\left\|\boldsymbol{x}^\top \boldsymbol{y}\right\| \le \left\|\boldsymbol{x}\right\| \cdot \left\|\boldsymbol{y}\right\|$.
%	\item  $\textcolor{blue}{(a)}$ means following the Lem.\ref{lem: tala}.
%	\item $\textcolor{blue}{(b)}$ leverages the fact that $\boldsymbol{v}_k$ and $\boldsymbol{v}_j$ are two one-hot vectors that only one nonzero term there.
%	\item $\textcolor{blue}{(Jen.)}$ follows the \textit{Jensen's Inequality}.
%\end{itemize}
\subsection{CML Symmetrization scheme}
In this section, we provide a brief proof of our extended symmetrization scheme for CML framework. The basic idea is that exchanging item instead of term does not change the value of 
\[
\expe_{\mcs, \mathcal{S}'} \left[\mathop{\sup}\limits_{\mathcal{H}_R} \left(\hat{\mathcal{R}}^{\text{cml}}_{\mathcal{S}'}(f) - \hat{\mathcal{R}}^{\text{cml}}_{\mcs}(f)\right)\right]
\]
\begin{rthm1}[\textbf{CML Symmetrization}] \label{cml:sysm} Let $\mcs$ and $ \mathcal{S}'$ be the two independent datasets of interactions that only one sample is different. In terms of any the hypothesis set $\mathcal{H}_R$ and loss function $\ell$, the following holds:
	\begin{equation}
		\begin{aligned}
			\expe_{\mcs, \mathcal{S}'} \left[\mathop{\sup}\limits_{\mathcal{H}_R} \left(\hat{\mathcal{R}}^{\text{cml}}_{\mathcal{S}'}(f) - \hat{\mathcal{R}}^{\text{cml}}_{\mcs}(f)\right)\right] &\le 2 \mathfrak{R}_{\ell,\mcs}^{\text{cml}}(\mathcal{H}_R)
		\end{aligned} \label{eql}
	\end{equation}
\end{rthm1}

\begin{pf} \label{lem6:proof} 
	%	To prove this lemma, we first assume that there is only one different item in user $u_i$, and the remaining interactions between two independent datasets $\mcs$ and $\mathcal{S}'$ are totally identical. Then, the difference of $\hat{\mathcal{R}}^{\text{cml}}_{\mathcal{S}'}(f) - \hat{\mathcal{R}}^{\text{cml}}_{\mcs}(f)$ with the remaining interactions except for $u_i$'s sample is apparently $0$. Thereafter, we only need to consider the specific user $u_i$ to prove the CML symmetrization in the following. 
	
	Define 
	\begin{equation}
		\begin{aligned}
			\mathcal{Q}^{jk}_{i,\sg} = \frac{\sg^+_{ij}+\sg^-_{ik}}{2} \ell^{(i)}(\tilde{v}_j^+, \tilde{v}_k^-) + \frac{\sg^+_{ij}-\sg^-_{ik}}{2} \ell^{(i)}(\tilde{v}_j^+, v_k^-) - \frac{\sg^+_{ij} - \sg^-_{ik}}{2} \ell^{(i)}(v_j^+, \tilde{v}_k^-) - \frac{\sg^+_{ij} + \sg^-_{ik}}{2} \ell^{(i)}(v_j^+, v_k^-)
		\end{aligned}\label{cml:Q}
	\end{equation}
	where $\sg^+_{ij}, \forall j = 1,2,\dots, n_i^+$ and $\sg^-_{ik}, \forall k = 1,2,\dots, n_i^-$ are i.i.d Rademacher random variables uniformly chosen from $\{-1, +1\}$ with $\mathbb{P}(\sg=1) = \mathbb{P}(\sg=-1) = 0.5$. 
	
	To complete the proof, first of all, we need to prove that 
	\begin{equation}
		\begin{aligned}
			\expe_{\mcs, \mathcal{S}'} \left[\mathop{\sup}\limits_{ \mathcal{H}_R} \left(\hat{\mathcal{R}}^{\text{cml}}_{\mathcal{S}'}(f) - \hat{\mathcal{R}}^{\text{cml}}_{\mcs}(f)\right)\right] &\le 
			\frac{1}{M} \sum_{u_i \in \mcu} \expe_{\mcs_i, \mathcal{S}'_i} \expe_{\sg_i} \left[\mathop{\sup}\limits_{\mathcal{H}_R} \frac{1}{n_i^+n_i^-} \sum_{j=1}^{n_i^+} \sum_{k=1}^{n_i^-} \mathcal{Q}^{jk}_{i, \sg} \right],
		\end{aligned} \label{equ62}
	\end{equation}
	
	To prove this, we have
	
	\begin{equation}
	\begin{aligned}
		& \expe_{\mcs, \mathcal{S}'} \left[\mathop{\sup}\limits_{ \mathcal{H}_R} \left(\hat{\mathcal{R}}^{\text{cml}}_{\mathcal{S}'}(f) - \hat{\mathcal{R}}^{\text{cml}}_{\mcs}(f)\right)\right] \\
		=& \expe_{\mcs, \mathcal{S}'} \left[\mathop{\sup}\limits_{ \mathcal{H}_R} \left(\left(\hat{\mathcal{R}}^{\text{cml}}_{\mathcal{S}_1'}(f) - \hat{\mathcal{R}}^{\text{cml}}_{\mcs_1}(f)\right)\right) \dots \left(\hat{\mathcal{R}}^{\text{cml}}_{\mathcal{S}_i'}(f) - \hat{\mathcal{R}}^{\text{cml}}_{\mcs_i}(f)\right) + \dots + \left(\hat{\mathcal{R}}^{\text{cml}}_{\mathcal{S}_M'}(f) - \hat{\mathcal{R}}^{\text{cml}}_{\mcs_M}(f)\right)\right]\\
		\overset{\textcolor{blue}{(*)}}{\le}& \sum_{u_i \in \mathcal{U}} \expe_{\mathcal{S}_i, \mathcal{S}_i'} \left[\mathop{\sup}\limits_{ \mathcal{H}_R} \left(\hat{\mathcal{R}}^{\text{cml}}_{\mathcal{S}_i'}(f) - \hat{\mathcal{R}}^{\text{cml}}_{\mathcal{S}_i}(f)\right)\right]
	\end{aligned}\label{e2}
	\end{equation}
		where
	\[
	\hat{\mathcal{R}}_{\mathcal{S}_i}^{\text{cml}}(f) = \frac{1}{M} \cdot \frac{1}{n_i^+n_i^-} \sum_{j=1}^{n_i^+} \sum_{k=1}^{n_i^-} \ell^{(i)}(v_j^+, v_k^-).
	\]
	and $\textcolor{blue}{(*)}$ achieves by the inequality: $\sup (x + y) \le \sup(x) + \sup(y)$.
	
	Equipped with Eq.(\ref{e2}), we could hold Eq.(\ref{equ62}) by proving the following equation satisfied:
	\begin{equation}
	\begin{aligned}
	\sum_{u_i \in \mathcal{U}} \expe_{\mathcal{S}_i, \mathcal{S}_i'} \left[\mathop{\sup}\limits_{ \mathcal{H}_R} \left(\hat{\mathcal{R}}^{\text{cml}}_{\mathcal{S}_i'}(f) - \hat{\mathcal{R}}^{\text{cml}}_{\mathcal{S}_i}(f)\right)\right] &= 
	\frac{1}{M} \sum_{u_i \in \mathcal{U}} \expe_{\mathcal{S}_i, \mathcal{S}_i'} \expe_{\sg_i} \left[\mathop{\sup}\limits_{\mathcal{H}_R} \sum_{j=1}^{n_i^+} \sum_{k=1}^{n_i^-} \frac{1}{n_i^+n_i^-} \mathcal{Q}^{jk}_{i, \sg} \right],
	\end{aligned}\label{e3}
	\end{equation}
	
	It is interesting to note that, Eq.(\ref{e3}) is calculated separately for each user. 
	Therefore, we only need to consider each user to prove the CML symmetrization such that the overall symmetrization of CML could be simply proved by taking a sum. Taking $u_i$ as an example, we need to prove the following equation
	\begin{equation}
		\begin{aligned}
			\expe_{\mathcal{S}_i, \mathcal{S}_i'} \left[\mathop{\sup}\limits_{ \mathcal{H}_R} \left(\hat{\mathcal{R}}^{\text{cml}}_{\mathcal{S}_i'}(f) - \hat{\mathcal{R}}^{\text{cml}}_{\mathcal{S}_i}(f)\right)\right] &= 
			\frac{1}{M} \expe_{\mathcal{S}_i, \mathcal{S}_i'} \expe_{\sg_i} \left[\mathop{\sup}\limits_{\mathcal{H}_R} \sum_{j=1}^{n_i^+} \sum_{k=1}^{n_i^-} \frac{1}{n_i^+n_i^-} \mathcal{Q}^{jk}_{i, \sg} \right].
		\end{aligned}\label{e34}
	\end{equation}
	
	Let $\mathcal{S}_i = \{v_j^+\}_{j=1}^{n_i^+} \cup \{v_k^-\}_{k=1}^{n_i^-}$ and $ \mathcal{S}_i' = \{\tilde{v}_j^+ \}_{j=1}^{n_i^+} \cup \{\tilde{v}_k^- \}_{k=1}^{n_i^-}$ be the two independent interaction datasets that only one item is different. To this end, considering the process of exchanging instances, since the items are drawn independently, the following holds
	
	\begin{equation}
		\expe_{\mathcal{S}_i, \mathcal{S}_i'} \left[\mathop{\sup}\limits_{\mathcal{H}_R} \left(\hat{\mathcal{R}}^{\text{cml}}_{\mathcal{S}_i'}(f) - \hat{\mathcal{R}}^{\text{cml}}_{\mathcal{S}_i}(f)\right)\right] = \expe_{\mathcal{S}_i, \mathcal{S}_i'} \left[\mathop{\sup}\limits_{\mathcal{H}_R} \left(\hat{\mathcal{R}}^{\text{cml}}_{\tilde{\mathcal{S}_i}'}(f) - \hat{\mathcal{R}}^{\text{cml}}_{\tilde{\mathcal{S}_i}}(f)\right)\right] \label{eq33}
	\end{equation}
	where $\tilde{\mathcal{S}_i}'$ and $\tilde{\mathcal{S}_i}$ are two different datasets determined by exchanging the corresponding indexed sample between $\mathcal{S}_i'$ and $\mathcal{S}_i$. 
	
	\textbf{Next, we will employ the induction method to prove this conclusion Eq.(\ref{eq33}).}
	
	\noindent \textcolor{blue}{\textbf{Trivial Case}}. Let us first consider that there are only one positive and negative item for the specific user $u_i$, i.e., $n_i^+ = 1$, $n_i^- = 1$, $\mathcal{S}_i = \{v_1^+, v_1^-\}$, $\mathcal{S}_i' = \{\tilde{v}_1^+, \tilde{v}_1^-\}$ and $\sg_i = \{\sg^+_{i1}, \sg^-_{i1}\}$. In this configuration, the value of $\sg_i$ can be permuted as follows:
	\begin{enumerate}
		\item  $\sg^+_{i1} = 1$ and $\sg^-_{i1}=1$. We have
		\begin{equation}
			\mathcal{Q}^{11}_{i,\sg} =  \ell^{(i)}(\tilde{v}_1^+, \tilde{v}_1^-)  -  \ell_i(v_1^+, v_1^-) \label{eq34}
		\end{equation}
		This motivates us to set $\tilde{\mathcal{S}_i}' = \mathcal{S}_i', \tilde{\mathcal{S}_i} = \mathcal{S}_i$ to satisfy Eq.(\ref{eq33}).
		\item $\sg^+_{i1} = -1$ and $\sg^-_{i1}=-1$. Here, we have
		\begin{equation}
			\mathcal{Q}^{11}_{i,\sg} = \ell^{(i)}(v_1^+, v_1^-) - \ell^{(i)}(\tilde{v}_1^+, \tilde{v}_1^-) \label{eq35}
		\end{equation}
		This motivates us to set $\tilde{\mathcal{S}_i}' = \mathcal{S}_i, \tilde{\mathcal{S}_i} = \mathcal{S}_i'$ to satisfy Eq.(\ref{eq33}).
		\item $\sg^+_{i1} = 1$ and $\sg^-_{i1}=-1$. We have
		\begin{equation}
			\mathcal{Q}^{11}_{i, \sg} = \ell^{(i)}(\tilde{v}_1^+, v_1^-) - \ell^{(i)}(v_1^+, \tilde{v}_1^-) \label{eq36}
		\end{equation}
		To hold Eq.(\ref{eq33}), let $\mathcal{S}_i$ and $\mathcal{S}_i'$ exchange sample $v_1^-$ and $\tilde{v}^-_1$, i.e., $\mathcal{S}_i' = \{\tilde{v}^+_1, v_1^-\}$ and $ \mathcal{S}_i = \{v_1^+, \tilde{v}_1^-\}$, and let $\tilde{\mathcal{S}_i}' = \mathcal{S}_i', \tilde{\mathcal{S}_i} = \mathcal{S}_i$ complete the proof.
		\item $\sg^+_{i1} = -1$ and $\sg^-_{i1}=1$. 
		\begin{equation}
			\mathcal{Q}^{11}_{i,\sg} = \ell^{(i)}(v_1^+, \tilde{v}_1^-) - \ell^{(i)}(\tilde{v}_1^+, v_1^-). \label{eq37}
		\end{equation}
		Needless to say, one can let $\tilde{\mathcal{S}_i} = \mathcal{S}_i, \tilde{\mathcal{S}_i}' = \mathcal{S}_i'$ by exchanging $v_1^+$ and $\tilde{v}^+_1$ between $\mathcal{S}_i$ and $\mathcal{S}_i'$ to hold the conclusion.
	\end{enumerate}
	\textbf{In summary, the above discussions support the proof of Eq.(\ref{eq33}) in terms of trivial case.}
	
	\noindent \textcolor{blue}{\textbf{Recursion}}. Given $1 < n_+ < n_i^+, 1 < n_- < n_i^-$ samples, and
	\[
	\mathcal{S}_{i, 0} = \{v_j^+\}_{j=1}^{n_+} \cup \{v_k^-\}_{k=1}^{n_-}, \ \ \mathcal{S}_{i,0}' = \{\tilde{v}_j^+\}_{j=1}^{n_+} \cup \{\tilde{v}_k^-\}_{k=1}^{n_-},  \sigma_{i, 0} = \{\sigma^+_{ij}\}_{j=1}^{n_+} \cup \{\sigma^-_{ik}\}_{k=1}^{n_-},
	\]
	assume $\mathcal{S}_{i, 0}$, $\mathcal{S}_{i, 0}'$ and $\sg_{i,0}$ could hold Eq.(\ref{eq33}) by $\tilde{\mathcal{S}}_{i,0}$ and $\tilde{\mathcal{S}}_{i,0}'$. Next, we will show that for any fresh samples $(v_t^+, \tilde{v}_t^+, \sg_{it}^+)$ or $(v_t^-, \tilde{v}_t^-, \sg_{it}^-)$, our conclusion Eq.(\ref{eq33}) is still satisfied. 
	
	Let us first consider $\sg_{it}^+ = 1$. Obviously, merely $\mcq_{i,\sg}^{tk}, k = 1, 2, \dots, n_-$ has the contribution to this new sample.
	\begin{enumerate}
		\item if $\sg_{ik}^-=1$ is the case, we have
		\[
		\mathcal{Q}^{tk}_{i,\sg} = \ell^{(i)}(\tilde{v}_t^+, \tilde{v}_k^-) - \ell^{(i)}(v_t^+, v_k^-).
		\]
		One can let 
		\[
		\tilde{\mathcal{S}_i}' = \mathcal{S}_{i,0}' \cup \{\tilde{v}_t^+\}, \tilde{\mathcal{S}_i} = \mathcal{S}_{i,0} \cup \{v_t^+\} 
		\]
		to complete the proof.
		\item if $\sg_{ik}^-=-1$ is the case, we have 
		\[
		\mathcal{Q}_{i,\sg}^{tk} = \ell^{(i)}(\tilde{v}_t^+, v_k^-) - \ell^{(i)}(v_t^+, \tilde{v}_k^-),
		\]
		and one can let
		\[
		\tilde{\mathcal{S}_i}' = \mathcal{S}_{i,0}  \cup \{\tilde{v}_t^+\}, \tilde{\mathcal{S}_i} =  \mathcal{S}_{i,0}' \cup \{v_t^+\} 
		\]
		to complete the proof.
	\end{enumerate}
	In addition, with respect to $\sg_{it}^+ = -1$, we have
	\begin{enumerate}
		\item $\sg_{ik}^-=1$.
		\[
		\mathcal{Q}_{i,\sg}^{tk} =  \ell^{(i)}(v_t^+, \tilde{v}_k^-) - \ell^{(i)}(\tilde{v}_t^+, v_k^-),
		\]
		and one can let 
		\[
		\tilde{\mathcal{S}_i}' = \mathcal{S}_{i,0}'  \cup \{v_t^+\}, \tilde{\mathcal{S}_i} =  \mathcal{S}_{i,0} \cup \{\tilde{v}_t^+\}
		\]
		to complete the proof.
		\item $\sg_{ik}^-=-1$. 
		\[
		\mathcal{Q}^{tk}_{i,\sg} = \ell^{(i)}(v_t^+, v_k^-) - \ell^{(i)}(\tilde{v}_t^+, \tilde{v}_k^-), 
		\]
		we have
		\[
		\tilde{\mathcal{S}_i}' = \mathcal{S}_{i,0}  \cup \{v_t^+\} , \tilde{\mathcal{S}_i} =  \mathcal{S}_{i,0}' \cup \{\tilde{v}_t^+\}
		\]
		to statisfy the conclusion.
	\end{enumerate}
	
	In the same way, in terms of $\sg_{it}^-$, we have 
	\begin{enumerate}
		\item $\sg_{it}^- = 1, \sg_{ij}^+ = 1$. This suggests that 
		\[
		\tilde{\mathcal{S}_i}' = \mathcal{S}_{i,0}' \cup \{\tilde{v}_t^-\}, \tilde{\mathcal{S}_i} = \mathcal{S}_{i,0} \cup \{v_t^-\} 
		\]
		\item $\sg_{it}^- = 1, \sg_{ij}^+ = -1$.
		\[
		\tilde{\mathcal{S}_i}' = \mathcal{S}_{i,0}  \cup \{\tilde{v}_t^-\}, \tilde{\mathcal{S}_i} =  \mathcal{S}_{i,0}' \cup \{v_t^-\} 
		\]
		\item $\sg_{it}^- = -1, \sg_{ij}^+ = 1$.
		\[
		\tilde{\mathcal{S}_i}' = \mathcal{S}_{i,0}'  \cup \{v_t^-\}, \tilde{\mathcal{S}_i} =  \mathcal{S}_{i,0} \cup \{\tilde{v}_t^-\}
		\]
		\item $\sg_{it}^- = -1, \sg_{ij}^+ = -1$.
		\[
		\tilde{\mathcal{S}_i}' = \mathcal{S}_{i,0}  \cup \{v_t^-\} , \tilde{\mathcal{S}_i} =  \mathcal{S}_{i,0}' \cup \{\tilde{v}_t^-\}
		\]
	\end{enumerate}
	
	\textbf{It is interesting to note that, in either case, $\tilde{\mathcal{S}_i}$ and $\tilde{\mathcal{S}_i}'$ are both obtained from $\mathcal{S}_i$ and $\mathcal{S}_i'$, which only swaps the single corresponding instance rather than a pair.} Taking the trivial case and recursion into account together, we have proved that 
	
	\begin{equation}
		\begin{aligned}
			\expe_{\mathcal{S}_i, \mathcal{S}_i'} \left[\mathop{\sup}\limits_{\mathcal{H}_R} \left(\hat{\mathcal{R}}^{\text{cml}}_{\mathcal{S}_i'}(f) - \hat{\mathcal{R}}^{\text{cml}}_{\mathcal{S}_i}(f)\right)\right] &= 
			\frac{1}{M} \expe_{\mathcal{S}_i, \mathcal{S}_i'} \expe_{\sg_i} \left[\mathop{\sup}\limits_{\mathcal{H}_R} \sum_{j=1}^{n_i^+} \sum_{k=1}^{n_i^-} \frac{1}{n_i^+n_i^-} \mathcal{Q}^{jk}_{i, \sg} \right].
		\end{aligned}
	\end{equation}
	since we have
	\begin{equation}
	\begin{aligned}
	& \frac{1}{M} \expe_{\mcs_i, \mathcal{S}'_i} \expe_{\sg_i} \left[\mathop{\sup}\limits_{\mathcal{H}_R} \sum_{j=1}^{n_i^+} \sum_{k=1}^{n_i^-} \frac{1}{n_i^+n_i^-} \mathcal{Q}^{jk}_{i, \sg} \right] \\ 
	=& \frac{1}{M} \cdot\frac{1}{2^{N}} \sum_{\sg_i}  \expe_{\mathcal{S}_i, \mathcal{S}_i'} \left[\mathop{\sup}\limits_{\mathcal{H}_R} \frac{1}{n_i^+n_i^-}\sum_{j=1}^{n_i^+} \sum_{k=1}^{n_i^-} \mathcal{Q}^{jk}_{i,\sg} \right] \\
	=& \frac{1}{2^{N}} \sum_{\sg_i} \expe_{\mathcal{S}_i, \mathcal{S}_i'} \left[\mathop{\sup}\limits_{\mathcal{H}_R} \left(\hat{\mathcal{R}}^{\text{cml}}_{\tilde{\mathcal{S}_i}'}(f) - \hat{\mathcal{R}}^{\text{cml}}_{\tilde{\mcs}_i}(f)\right)\right] \\
	=& \frac{1}{2^{N}} \sum_{\sg_i} \expe_{\mathcal{S}_i, \mathcal{S}_i'} \left[\mathop{\sup}\limits_{\mathcal{H}_R} \left(\hat{\mathcal{R}}^{\text{cml}}_{\mathcal{S}_i'}(f) - \hat{\mathcal{R}}^{\text{cml}}_{\mathcal{S}_i}(f)\right)\right]	\\
	=& \frac{2^{N}}{2^{N}} \expe_{\mathcal{S}_i, \mathcal{S}_i'} \left[\mathop{\sup}\limits_{\mathcal{H}_R} \left(\hat{\mathcal{R}}^{\text{cml}}_{\mathcal{S}_i'}(f) - \hat{\mathcal{R}}^{\text{cml}}_{\mathcal{S}_i}(f)\right)\right]  \\
	=& \expe_{\mathcal{S}_i, \mathcal{S}_i'} \left[\mathop{\sup}\limits_{\mathcal{H}_R} \left(\hat{\mathcal{R}}^{\text{cml}}_{\mathcal{S}_i'}(f) - \hat{\mathcal{R}}^{\text{cml}}_{\mathcal{S}_i}(f)\right)\right], \\
	\end{aligned} \label{eq40}
	\end{equation}
	where again $\mcs_i$ and $ \mathcal{S}'_i$ are the two independent datasets of interactions that only one sample is different. 
	
	Equipped with Eq.(\ref{eq40}), by taking a sum over all users, it is easy to obtain Eq.(\ref{e3}) and thus Eq.(\ref{equ62}) holds.
%	
%	\begin{equation}
%		\begin{aligned}
%			\expe_{\mcs, \mathcal{S}'} \left[\mathop{\sup}\limits_{ \mathcal{H}_R} \left(\hat{\mathcal{R}}^{\text{cml}}_{\mathcal{S}'}(f) - \hat{\mathcal{R}}^{\text{cml}}_{\mcs}(f)\right)\right] &= 
%			\frac{1}{M} \sum_{u_i \in \mcu} \expe_{\mcs_i, \mathcal{S}'_i} \expe_{\sg_i} \left[\mathop{\sup}\limits_{\mathcal{H}_R} \sum_{j=1}^{n_i^+} \sum_{k=1}^{n_i^-} \frac{1}{n_i^+n_i^-} \mathcal{Q}^{jk}_{i, \sg} \right].
%		\end{aligned} \label{eq39}
%	\end{equation}
	Finally, according to the property that the sign of the Rademacher random variables do not affect its expectation, we have
	
	\begin{equation}
		\begin{aligned}
			\expe_{\mcs, \mathcal{S}'} \left[\mathop{\sup}\limits_{ \mathcal{H}_R} \left(\hat{\mathcal{R}}^{\text{cml}}_{\mathcal{S}'}(f) - \hat{\mathcal{R}}^{\text{cml}}_{\mcs}(f)\right)\right] &\le 
			\frac{1}{M} \sum_{u_i \in \mcu} \expe_{\mcs_i, \mathcal{S}'_i} \expe_{\sg_i} \left[\mathop{\sup}\limits_{\mathcal{H}_R} \sum_{j=1}^{n_i^+} \sum_{k=1}^{n_i^-} \frac{1}{n_i^+n_i^-} \mathcal{Q}^{jk}_{i, \sg} \right] \\
			& \le \frac{2}{M} \sum_{u_i \in \mcu}  \expe_{\mcs_i,\sg_i} \left[\mathop{\sup}\limits_{\mathcal{H}_R} \frac{1}{n_i^+n_i^-} \sum_{j=1}^{n_i^+} \sum_{k=1}^{n_i^-} \mathcal{Q}_{(i)}^{jk} \right] \\
			&= 2 \mathfrak{R}_{\ell, \mcs}^{\text{cml}}(\mathcal{H}_R)
		\end{aligned} \label{lem6: final}
	\end{equation}
	where again
	\[
	\mathcal{Q}_{(i)}^{jk} = \frac{\sigma^+_{ij} + \sigma^-_{ik}}{2} \cdot \ell^{(i)}(v_j^+, v_k^-)
	\]
	This completed the proof.
\end{pf}

\subsection{Upper Bound of empirical Rademacher Complexity} \label{SFCML:rade}
\begin{rthm2} (\textbf{Upper Bound of empirical Rademacher Complexity}). \label{bounds:rade} Given the user set $\mcu$ and corresponding sample set $\mcs = \mathop{\cup}\limits_{u_i \in \mcu} \mcs_i$ where $\mcs_i = \{v_j^+\}_{j=1}^{n_i^+} \cup \{v_k^-\}_{k=1}^{n_i^-}, n_i^+ + n_i^- = N$. If $\ell$ is a $\phi$ Lipschitz continuous, then the following inequality holds: 
	\begin{equation}
	\begin{aligned}
	\hat{\mfr}_{\ell, \mcs}^{\text{cml}}(\mathcal{H}_R) &\lesssim \frac{\phi}{M} \cdot {\Max(\lambda, \sqrt{R \cdot d})} \cdot \tilde{N}^{-1/2}.
	\end{aligned}\label{eq53}
	\end{equation}	
\end{rthm2}
\begin{pf} At first, according to Definition.\ref{def1}, we have 
	\begin{equation}
	\begin{aligned}
	& \hat{\mfr}_{\ell, \mcs}^{\text{cml}}(\mathcal{H}_R) =
	\frac{1}{M} \sum_{u_i \in \mathcal{U}} \mathop{\expe}\limits_{\bss_i} \left[\mathop{\sup}\limits_{\mathcal{H}_R}  \frac{1}{n_i^+n_i^-} \sum_{j=1}^{n_i^+} \sum_{k=1}^{n_i^-}
	\mathcal{Q}_{(i)}^{jk} \right] \\ 
	=& \frac{1}{M} \left(\mathop{\expe}\limits_{\bss_1} \left[ \mathop{\sup}\limits_{ \mathcal{H}_R} \frac{1}{n_1^+n_1^-} \sum_{j=1}^{n_1^+} \sum_{k=1}^{n_1^-} 
	\mathcal{Q}_{(1)}^{jk}\right] + \dots +  \mathop{\expe}\limits_{\bss_i} \left[ \mathop{\sup}\limits_{\mathcal{H}_R} \frac{1}{n_i^+n_i^-} \sum_{j=1}^{n_i^+} \sum_{k=1}^{n_i^-} 
	\mathcal{Q}_{(i)}^{jk}\right] + \dots + \mathop{\expe}\limits_{\bss_M} \left[ \mathop{\sup}\limits_{\mathcal{H}_R} \frac{1}{n_M^+n_M^-} \sum_{j=1}^{n_M^+} \sum_{k=1}^{n_M^-}
	\mathcal{Q}_{(M)}^{jk}\right]\right),\\
	\end{aligned}\label{rthm3}
	\end{equation}
	where 
	\[
	\mathcal{Q}_{(i)}^{jk} = \frac{\sigma^+_{ij} + \sigma^-_{ik}}{2} \cdot \ell^{(i)}(v_j^+, v_k^-),
	\]
	and $\ell^{(i)}(v_j^+, v_k^-)$ is a differentiable ranking loss, such as hinge-loss and square loss:
	\[
	\ell^{(i)}(v_j^+, v_k^-) = \ell \circ (\lambda + \boldsymbol{d}(i,j) - \boldsymbol{d}(i,k)).
	\]
	Next, according to Eq.(\ref{rthm3}), we first derive the following bound of a specific user $u_i$:
	
	\begin{equation}
	\begin{aligned}
	& \mathop{\expe}\limits_{\bss_i} \left[ \mathop{\sup}\limits_{\mathcal{H}_R} \sum_{j=1}^{n_i^+} \sum_{k=1}^{n_i^-} \frac{1}{n_i^+n_i^-}
	\mathcal{Q}_{(i)}^{jk}\right] = \mathop{\expe}\limits_{\bss_i} \left[ \mathop{\sup}\limits_{\mathcal{H}_R} \frac{1}{n_i^+n_i^-} \sum_{j=1}^{n_i^+} \sum_{k=1}^{n_i^-}
	\frac{\sigma_{ij}^+ + \sigma_{ij}^k}{2} \cdot \ell^{(i)}(v_j^+, v_k^-)\right] \\
	=& \frac{1}{n_i^+n_i^-}\mathop{\expe}\limits_{\bss_i} \left[\mathop{\sup}\limits_{\mathcal{H}_R} \left(\sum_{j=1}^{n_i^+} \sum_{k=1}^{n_i^-} 
	\frac{\sigma^+_{ij}}{2} \cdot \ell^{(i)}(v_j^+, v_k^-) + \sum_{j=1}^{n_i^+} \sum_{k=1}^{n_i^-} 
	\frac{\sigma^-_{ik}}{2} \cdot \ell^{(i)}(v_j^+, v_k^-) \right) \right] \\
	\overset{\textcolor{blue}{(*)}}{\le}& \frac{1}{n_i^+n_i^-}\left(\mathop{\expe}\limits_{\bss_i} \left[\mathop{\sup}\limits_{\mathcal{H}_R} \sum_{j=1}^{n_i^+} \sum_{k=1}^{n_i^-} 
	\frac{\sigma^+_{ij}}{2} \cdot \ell^{(i)}(v_j^+, v_k^-)\right] + \mathop{\expe}\limits_{\bss_i} \left[\mathop{\sup}\limits_{\mathcal{H}_R} \sum_{j=1}^{n_i^+} \sum_{k=1}^{n_i^-} 
	\frac{\sigma^-_{ik}}{2} \cdot \ell^{(i)}(v_j^+, v_k^-)\right]\right) \\
	\overset{\textcolor{blue}{(a)
	}}{\le}& \frac{1}{n_i^+n_i^-}\left(\phi\cdot \mathop{\expe}\limits_{\bss_i} \left[\mathop{\sup}\limits_{ \mathcal{H}_R} \sum_{j=1}^{n_i^+} \sum_{k=1}^{n_i^-} 
	\frac{\sigma^+_{ij}}{2} \cdot (\lambda + \boldsymbol{d}(i,j) - \boldsymbol{d}(i, k)) \right] + \phi\cdot \mathop{\expe}\limits_{\bss_i} \left[\mathop{\sup}\limits_{\mathcal{H}_R} \sum_{j=1}^{n_i^+} \sum_{k=1}^{n_i^-} 
	\frac{\sigma^-_{ik}}{2} \cdot (\lambda + \boldsymbol{d}(i,j) - \boldsymbol{d}(i, k)) \right]\right) \\
	\overset{\textcolor{blue}{(*)}}{\le}& \underbrace{\frac{\phi}{n_i^+n_i^-} \left(\mathop{\expe}\limits_{\bss_i} \left[\mathop{\sup}\limits_{\mathcal{H}_R} \sum_{j=1}^{n_i^+} \sum_{k=1}^{n_i^-} 
		\frac{\sigma^+_{ij}}{2} \cdot \lambda\right] + \mathop{\expe}\limits_{\bss_i} \left[\mathop{\sup}\limits_{\mathcal{H}_R} \sum_{j=1}^{n_i^+} \sum_{k=1}^{n_i^-} 
		\frac{\sigma^-_{ik}}{2} \cdot \lambda\right]\right)}_{\textcolor{orange}{\textcircled{1}}} \\
	\ \ &+ \underbrace{\frac{\phi}{n_i^+n_i^-} \left(\mathop{\expe}\limits_{\bss_i} \left[\mathop{\sup}\limits_{\mathcal{H}_R} \sum_{j=1}^{n_i^+} \sum_{k=1}^{n_i^-} 
		\frac{\sigma^+_{ij}}{2} \cdot (\boldsymbol{d}(i,j) - \boldsymbol{d}(i, k))\right] + \mathop{\expe}\limits_{\bss_i} \left[\mathop{\sup}\limits_{\mathcal{H}_R} \sum_{j=1}^{n_i^+} \sum_{k=1}^{n_i^-} 
		\frac{\sigma^-_{ik}}{2} \cdot (\boldsymbol{d}(i,j) - \boldsymbol{d}(i, k))\right] \right)}_{\textcolor{orange}{\textcircled{2}}}
	\end{aligned}
	\end{equation}
	where \textcolor{blue}{(a)} follows the Lem.\ref{lem: tala} and $\textcolor{blue}{(*)}$ achieves by the inequality: $\sup (x + y) \le \sup(x) + \sup(y)$.
	
	Now, it is easy to show that, for $\textcolor{orange}{\textcircled{1}}$, the following holds:
	\begin{equation}
	\begin{aligned}
	& \frac{\phi}{n_i^+n_i^-} \left(\mathop{\expe}\limits_{\bss_i} \left[\mathop{\sup}\limits_{ \mch_R} \sum_{j=1}^{n_i^+} \sum_{k=1}^{n_i^-} 
	\frac{\sigma^+_{ij}}{2} \cdot \lambda\right] + \mathop{\expe}\limits_{\bss_i} \left[\mathop{\sup}\limits_{\mch_R} \sum_{j=1}^{n_i^+} \sum_{k=1}^{n_i^-} 
	\frac{\sigma^-_{ik}}{2} \cdot \lambda\right]\right) \\
	& \le  \frac{\lambda\phi}{n_i^+n_i^-} \left(\mathop{\expe}\limits_{\bss_i}\left[ \left| \sum_{j=1}^{n_i^+} \sum_{k=1}^{n_i^-} 
	\frac{\sigma^+_{ij}}{2} \right|\right] + \mathop{\expe}\limits_{\bss_i} \left[\left|\sum_{j=1}^{n_i^+} \sum_{k=1}^{n_i^-} 
	\frac{\sigma^-_{ik}}{2}\right|\right]\right) \\
	& \le \frac{\lambda\phi}{n_i^+n_i^-} \left(\mathop{\expe}\limits_{\bss_i}\left[\sum_{k=1}^{n_i^-} \left| \sum_{j=1}^{n_i^+} 
	\frac{\sigma^+_{ij}}{2} \right|\right] + \mathop{\expe}\limits_{\bss_i}\left[ \sum_{j=1}^{n_i^+}\left| \sum_{k=1}^{n_i^-} 
	\frac{\sigma^-_{ik}}{2} \right|\right] \right) \\
	& = \frac{\lambda\phi}{n_i^+n_i^-} \left(\mathop{\expe}\limits_{\bss_i}\left[\sum_{k=1}^{n_i^-} \left| \sum_{j=1}^{n_i^+} 
	\frac{\sigma^+_{ij}}{2} \right| +  \sum_{j=1}^{n_i^+}\left| \sum_{k=1}^{n_i^-} 
	\frac{\sigma^-_{ik}}{2} \right|\right] \right) \\
	& \overset{\textcolor{blue}{(**)}}{\le} \frac{\lambda\phi \cdot \sqrt{n_i^+ + n_i^-}}{n_i^+n_i^-} \left(\mathop{\expe}\limits_{\bss_i}\left[\sqrt{n_i^-\left(\sum_{j=1}^{n_i^+} 
		\frac{\sigma^+_{ij}}{2}\right)^2 + n_i^+\left(\sum_{k=1}^{n_i^-} 
		\frac{\sigma^-_{ik}}{2} \right)^2}\right] \right) \\
	& \overset{\textcolor{blue}{(Jen.)}}{\le} \frac{\lambda \phi \cdot \sqrt{n_i^+ + n_i^-}}{n_i^+n_i^-} \sqrt{n_i^-\mathop{\expe}\limits_{\bss_i}\left[\left(\sum_{j=1}^{n_i^+} 
		\frac{\sigma^+_{ij}}{2}\right)^2\right] + n_i^+\mathop{\expe}\limits_{\bss_i}\left[\left(\sum_{k=1}^{n_i^-} 
		\frac{\sigma^-_{ik}}{2} \right)^2\right]} \\
	& \le \lambda\phi \cdot \sqrt{\frac{n_i^+ + n_i^-}{2n_i^+n_i^-}} \\
	& \lesssim \lambda\phi \cdot \sqrt{\frac{1}{n_i^+} + \frac{1}{n_i^-}} \\
	\end{aligned} \label{eq48}
	\end{equation}
	where $\textcolor{blue}{(**)}$ is because of the fact $\left\|\boldsymbol{x}\right\|_1 \le \sqrt{b} \left\|\boldsymbol{x}\right\|_2, \boldsymbol{x} \in \mathbb{R}^b$, and $\textcolor{blue}{(Jen.)}$ follows the \textit{Jensen's Inequality}.

	Subsequently, for the last term $\textcolor{orange}{\textcircled{2}}$, we have
	\begin{equation}
	\begin{aligned}
	& \frac{\phi}{n_i^+n_i^-} \left(\mathop{\expe}\limits_{\bss_i} \left[\mathop{\sup}\limits_{\mathcal{H}_R} \sum_{j=1}^{n_i^+} \sum_{k=1}^{n_i^-} 
	\frac{\sigma^+_{ij}}{2} \cdot (\boldsymbol{d}(i,j) - \boldsymbol{d}(i, k))\right] + \mathop{\expe}\limits_{\bss_i} \left[\mathop{\sup}\limits_{\mathcal{H}_R} \sum_{j=1}^{n_i^+} \sum_{k=1}^{n_i^-} 
	\frac{\sigma^-_{ik}}{2} \cdot (\boldsymbol{d}(i,j) - \boldsymbol{d}(i, k))\right] \right) \\
	&= \frac{\phi}{n_i^+n_i^-} \left(\mathop{\expe}\limits_{\bss_i} \left[\mathop{\sup}\limits_{\mathcal{H}_R} \sum_{j=1}^{n_i^+} \sum_{k=1}^{n_i^-} 
	\sigma^+_{ij} \cdot \boldsymbol{e}_{u_i}^\top (\boldsymbol{e}_{v_k} - \boldsymbol{e}_{v_j})\right] + \mathop{\expe}\limits_{\bss_i} \left[\mathop{\sup}\limits_{\mathcal{H}_R} \sum_{j=1}^{n_i^+} \sum_{k=1}^{n_i^-} 
	\sigma^-_{ik} \cdot \boldsymbol{e}_{u_i}^\top (\boldsymbol{e}_{v_k} - \boldsymbol{e}_{v_j})\right] \right) \\
	& \le \frac{\phi}{n_i^+n_i^-} \left(\mathop{\expe}\limits_{\bss_i} \left[\mathop{\sup}\limits_{\mathcal{H}_R} \left|\sum_{j=1}^{n_i^+} \sum_{k=1}^{n_i^-} 
	\sigma^+_{ij} \cdot \boldsymbol{e}_{u_i}^\top (\boldsymbol{e}_{v_k} - \boldsymbol{e}_{v_j})\right|\right] + \mathop{\expe}\limits_{\bss_i} \left[\mathop{\sup}\limits_{\mathcal{H}_R} \left|\sum_{j=1}^{n_i^+} \sum_{k=1}^{n_i^-} 
	\sigma^-_{ik} \cdot \boldsymbol{e}_{u_i}^\top (\boldsymbol{e}_{v_k} - \boldsymbol{e}_{v_j})\right|\right] \right) \\
	&\overset{\textcolor{blue}{(***)}}{\le} \frac{\phi}{n_i^+n_i^-} \left(\mathop{\expe}\limits_{\bss_i} \left[\mathop{\sup}\limits_{\mathcal{H}_R} \left\|\boldsymbol{e}_{u_i}\right\| \cdot \left\|\sum_{j=1}^{n_i^+} \sum_{k=1}^{n_i^-} 
	\sigma^+_{ij} \cdot (\boldsymbol{e}_{v_k} - \boldsymbol{e}_{v_j})\right]\right\| + \mathop{\expe}\limits_{\bss_i} \left[\mathop{\sup}\limits_{\mathcal{H}_R} \left\|\boldsymbol{e}_{u_i}\right\| \cdot \left\|\sum_{j=1}^{n_i^+} \sum_{k=1}^{n_i^-} 
	\sigma^-_{ik} \cdot (\boldsymbol{e}_{v_k} - \boldsymbol{e}_{v_j})\right\| \right]\right) \\
	&= \frac{\sqrt{R}\cdot\phi}{n_i^+n_i^-} \left(\mathop{\expe}\limits_{\bss_i} \left[\mathop{\sup}\limits_{\mathcal{H}_R} \left\|\sum_{j=1}^{n_i^+} \sum_{k=1}^{n_i^-} 
	\sigma^+_{ij} \cdot \bm{W}_v^\top(\boldsymbol{v}_k - \boldsymbol{v}_j)\right\|\right] + \mathop{\expe}\limits_{\bss_i} \left[\mathop{\sup}\limits_{\mathcal{H}_R} \left\|\sum_{j=1}^{n_i^+} \sum_{k=1}^{n_i^-} 
	\sigma^-_{ik} \cdot \bm{W}_v^\top(\boldsymbol{v}_k - \boldsymbol{v}_j)\right\|\right]\right) \\
	& \overset{\textcolor{blue}{(***)}}{\le} \frac{\sqrt{R}\cdot \phi}{n_i^+n_i^-} \cdot \left\|\bm{W}_v\right\|_{*} \cdot  \left(\mathop{\expe}\limits_{\bss_i} \left[ \left\|\sum_{j=1}^{n_i^+} \sum_{k=1}^{n_i^-} 
	\sigma^+_{ij} \cdot (\boldsymbol{v}_k - \boldsymbol{v}_j)\right\|\right] + \mathop{\expe}\limits_{\bss_i} \left[ \left\|\sum_{j=1}^{n_i^+} \sum_{k=1}^{n_i^-} 
	\sigma^-_{ik} \cdot (\boldsymbol{v}_k - \boldsymbol{v}_j)\right\|\right] \right) \\
	\end{aligned}
	\end{equation}
	where $\textcolor{blue}{(***)}$ is achieved by the Cauchy-Buniakowsky-Schwarz Inequality $\left\|\boldsymbol{x}^\top \boldsymbol{y}\right\| \le \left\|\boldsymbol{x}\right\| \cdot \left\|\boldsymbol{y}\right\|$ here.
	
	Next, according to Assumption.\ref{assu1}, we have $\left\|\bm{W}_v\right\|_{2} \lesssim \sqrt{\frac{n_i^+ + n_i^-}{d}}$ and $\sqrt{n_i^+ + n_i^-}\lesssim d$, holding that
	\begin{equation}
	\begin{aligned}
	& \frac{\phi}{n_i^+n_i^-} \left(\mathop{\expe}\limits_{\bss_i} \left[\mathop{\sup}\limits_{ \mathcal{H}_R} \sum_{j=1}^{n_i^+} \sum_{k=1}^{n_i^-} 
	\frac{\sigma^+_{ij}}{2} \cdot (\boldsymbol{d}(i,j) - \boldsymbol{d}(i, k))\right] + \mathop{\expe}\limits_{\bss_i} \left[\mathop{\sup}\limits_{\mathcal{H}_R} \sum_{j=1}^{n_i^+} \sum_{k=1}^{n_i^-} 
	\frac{\sigma^-_{ik}}{2} \cdot (\boldsymbol{d}(i,j) - \boldsymbol{d}(i, k))\right] \right) \\
	& \lesssim \frac{\sqrt{Rd}\cdot\phi}{n_i^+n_i^-} \left(\mathop{\expe}\limits_{\bss_i} \left[ \sum_{k=1}^{n_i^-} \left\|\sum_{j=1}^{n_i^+}  
	\sigma^+_{ij} \cdot (\boldsymbol{v}_k - \boldsymbol{v}_j)\right\|\right] + \mathop{\expe}\limits_{\bss_i} \left[ \sum_{j=1}^{n_i^+}\left\| \sum_{k=1}^{n_i^-} 
	\sigma^-_{ik} \cdot (\boldsymbol{v}_k - \boldsymbol{v}_j)\right\|\right] \right) \\
	& \overset{\textcolor{blue}{(**)}}{\lesssim} \phi \cdot \frac{\sqrt{R d\cdot (n_i^+ + n_i^-)}}{n_i^+n_i^-} \left(\mathop{\expe}\limits_{\bss_i}\sqrt{n_i^- \cdot \left\|\sum_{j=1}^{n_i^+}  
		\sigma^+_{ij} \cdot (\boldsymbol{v}_k - \boldsymbol{v}_j)\right\|_{2}^2 + n_i^+ \cdot \left\|\sum_{j=1}^{n_i^+}  
		\sigma^-_{ik} \cdot (\boldsymbol{v}_k - \boldsymbol{v}_j)\right\|_{2}^2 }\right) \\ 
	& \overset{\textcolor{blue}{(Jen.)}}{\lesssim} \phi \cdot \frac{\sqrt{Rd \cdot (n_i^+ + n_i^-)}}{n_i^+n_i^-} \left(\sqrt{n_i^- \cdot \mathop{\expe}\limits_{\bss_i}\left(\left\|\sum_{j=1}^{n_i^+}  
		\sigma^+_{ij} \cdot (\boldsymbol{v}_k - \boldsymbol{v}_j)\right\|_{2}^2 \right) + n_i^+ \cdot \mathop{\expe}\limits_{\bss_i}\left(\left\|\sum_{j=1}^{n_i^+}  
		\sigma^-_{ik} \cdot (\boldsymbol{v}_k - \boldsymbol{v}_j)\right\|_{2}^2\right) }\right) \\
	& \overset{\textcolor{blue}{(b)}}{\lesssim}  \phi \cdot \sqrt{\frac{Rd \cdot (n_i^+ + n_i^-)}{n_i^+n_i^-}}\\
	& \lesssim \phi \cdot \sqrt{R\cdot d} \cdot \sqrt{\frac{1}{n_i^+} + \frac{1}{n_i^-}} 
	\end{aligned}\label{Eq45}
	\end{equation}
	where $\textcolor{blue}{(b)}$ according to the fact that $\boldsymbol{v}_k$ and $\boldsymbol{v}_j$ are two one-hot vectors that only one nonzero term there.
	
	Therefore, taking Eq.(\ref{rthm3}), Eq.(\ref{eq48}), Eq.(\ref{Eq45}) and Definition.\ref{defi3} into account, we hold the following bounds for empirical Rademacher complexity by taking a sum of all users: 
	\begin{equation}
	\begin{aligned}
	\hat{\mfr}_{\ell, \mcs}^{\text{cml}}(\mathcal{H}_R) &\lesssim \phi \cdot \frac{\Max(\lambda, \sqrt{R \cdot d})}{M} \cdot \left(\sum_{u_i \in \mathcal{U}} \sqrt{\frac{1}{n_i^+} + \frac{1}{n_i^-}}\right) \\
	& \lesssim  \frac{\phi}{M} \cdot  \Max(\lambda, \sqrt{R \cdot d}) \cdot \tilde{N}^{-1/2}
	\end{aligned}\label{eq51}
	\end{equation}
	
	This completed the proof.
\end{pf}

\subsection{Generalization Bound of Sampling-Free CML}
\begin{rthm3} \label{SFCML:rthm3}
	(\textbf{Generalization Upper Bound of CML} \textbf{with Eq.(\ref{e4})}). Let $\mathcal{H}_R$ be the hypothesis space and $\ell$ be $\phi$-Lipschitz continuous. Given the sample set $\mcs = \mathop{\cup}\limits_{u_i \in \mcu} \mcs_i$ where $\mcs^{(i)} = \{v_j^+\}_{j=1}^{n_i^+} \cup \{v_k^-\}_{k=1}^{n_i^-}, n_i^+ + n_i^- = N$, for any $\delta \in (0,1)$, with probability at least $1 - \delta$, the following inequation holds:
	\begin{equation}
	\begin{aligned}
	\mathcal{R}_{\ell}^{\text{cml}}(f) &\lesssim \hat{\mathcal{R}}^{\text{cml}}_{\mcs}(f) + \phi \cdot \frac{\Max(\lambda, \sqrt{R \cdot d})}{M} \cdot \sqrt{\frac{1}{\tilde{N}}}   \\
	&+ \phi \cdot \frac{R}{M} \cdot \sqrt{\frac{\log2/\delta}{2}} \cdot \sqrt{\frac{1}{\tilde{N}}},
	\label{cml:final}
	\end{aligned}
	\end{equation}
	where $	\mathcal{R}_{\ell}^{\text{cml}}(f)$ is the expectation risk.
\end{rthm3}

\begin{pf} 
	\noindent \textcolor{blue}{\textbf{Step 1}}. Let
	\[
	\Phi(\mcs) = \mathop{\sup\limits_{\mathcal{H}_R}}  \left(\mathcal{R}_{\ell}^{\text{cml}}(f) - \hat{\mathcal{R}}^{\text{cml}}_{\mcs}(f)\right)
	\]
	 The first aim is to prove that $\Phi(\mcs)$ satisfies the condition of Lem.\ref{lem:mc}. To this end, let $\mcs$ and $\mcs'$ be two independent datasets where exactly one item is different with respect to the specific user $u_i$. Subsequently, we have the following two possible cases for $u_i$:
	%	In order to show the proof more clearly, we separately denote the related preferences of $u_i$ as $\mathcal{S}_i$ and $\mathcal{S}'_i$ respectively, and the remaining users are denoted as $\mathcal{S}_{\neg i}$ and $\mathcal{S}'_{\neg i}$, respectively.
	
	\begin{itemize}
		\item \textbf{Case 1:} \label{SFCML:case1} Only one positive item is different, i.e., 
		\begin{equation}
		\mathcal{S}_i = \{v_j^+\}_{j=1}^{n_i^+} \cup \{v_k^-\}_{k=1}^{n_i^-}, \ \ \ \ \mathcal{S}_i' = (\mathcal{S}_i \backslash	\{v_{t_1}^+\}) \cup \{\tilde{v}_{t_1}^{+}\}, \label{SFCML:eq26}
		\end{equation}
		where $\forall t_1, t_1 = 1, 2, \dots, n_i^+$ and $n_i^+ + n_i^- = N$.
		
		According to this, the upper bound on $\Phi(\mathcal{S}') - \Phi(\mathcal{S})$ could be bounded as follows:
		\begin{equation}
		\begin{aligned}
		\left|\Phi(\mathcal{S}') - \Phi(\mathcal{S})\right| &\le \left|\sup\limits_{\mathcal{H}_R} \left(\hat{\mathcal{R}}^{\text{cml}}_{\mcs}(f) - \hat{\mathcal{R}}^{\text{cml}}_{\mcs'}(f)\right)\right|\\ 
		&\le \sup\limits_{\mathcal{H}_R} \left|\hat{\mathcal{R}}^{\text{cml}}_{\mcs}(f) - \hat{\mathcal{R}}^{\text{cml}}_{\mcs'}(f)\right|	\\
		&= \sup\limits_{\mathcal{H}_R} \left|\hat{\mathcal{R}}^{\text{cml}}_{\mathcal{S}_i}(f) - \hat{\mathcal{R}}^{\text{cml}}_{\mathcal{S}_i'}(f)\right|\\
		\label{SFCML:eq28}
		\end{aligned}
		\end{equation}
		where again
		\[
		\hat{\mathcal{R}}_{\mathcal{S}_i}^{\text{cml}}(f) = \frac{1}{M} \cdot \sum_{j=1}^{n_i^+} \sum_{k=1}^{n_i^-} \ell^{(i)}(v_j^+, v_k^-)
		\]
		and $\ell^{(i)}(v_j^+, v_k^-)$ is a differentiable ranking loss, such as hinge-loss and square loss:
		\[
		\ell^{(i)}(v_j^+, v_k^-) = \ell \circ (\lambda + \boldsymbol{d}(i,j) - \boldsymbol{d}(i,k)).
		\]
		
		\noindent Since $v_j^+$ and $\tilde{v}_j^{+}$ are different in this case, we have
		\begin{equation}
		\begin{aligned}
		\left|\Phi(\mathcal{S}') - \Phi(\mathcal{S})\right| 
		& \le \frac{1}{M}\sup\limits_{\mathcal{H}_R} \left| \frac{1}{n_i^+n_i^-} \sum_{v_j^+} \sum_{k=1}^{n_i^-} \ell^{(i)}(v_j^+, v_k^-) -  \frac{1}{n_i^+n_i^-} \sum_{\tilde{v}_j^{+}} \sum_{k=1}^{n_i^-} \ell^{(i)}(\tilde{v}_j^{+}, v_k^-)\right|\\ 
		& \overset{\textcolor{blue}{(a)
		}}{\le} \frac{1}{M} \cdot \frac{\phi}{n_i^+n_i^-}  \sum_{k=1}^{n_i^-}\left(\boldsymbol{d}(i, j)-\boldsymbol{d}(i, \tilde{j}) \right) \\
		& \le \phi \cdot \frac{4R}{Mn_i^+}
		\label{SFCML:thoe3:1}
		\end{aligned}
		\end{equation}
		\item \textbf{Case 2:} \label{SFCML:case2} Only one negative item is different, i.e.,
		\begin{equation}
		\mathcal{S}_i = \{v_j^+\}_{j=1}^{n_i^+} \cup \{v_k^-\}_{k=1}^{n_i^-}, \ \ \ \ \mathcal{S}_i' = (\mathcal{S}_i \backslash \{v_{t_2}^-\}) \cup \{\tilde{v}_{t_2}^{-}\}. \label{SFCML:eq27}
		\end{equation}
		where $\forall t_2, t_2 = 1, 2, \dots, n_i^-$.
		Similarly, if $v_k^-$ and $\tilde{v}_k^{-}$ are different, we can also hold
		\begin{equation}
		\left|\Phi(\mathcal{S}') - \Phi(\mathcal{S})\right| \le \phi \cdot \frac{4R}{Mn_i^-} \label{SFCML:thoe3:2}
		\end{equation}
	\end{itemize}
	Therefore, according to Eq.(\ref{SFCML:thoe3:1}) and Eq.(\ref{SFCML:thoe3:2}) show that $\Phi(\mathcal{S})$ satisfies the Bounded Difference Property (Lem.\ref{def:bdp}). Subsequently, for any $\delta \in (0,1)$, with probability at least $1 - \delta$, the following holds:
	\begin{equation}
	\Phi(\mathcal{S}) \lesssim \expe_{\mcs}\left[\Phi(\mathcal{S})\right] + \phi \cdot \frac{R}{M} \cdot \sqrt{\frac{\log1/\delta}{2} \cdot \left(\frac{1}{n_i^+} + \frac{1}{n_i^-}\right)} \label{SFCML:thoe3:3}
	\end{equation}
	\noindent \textcolor{blue}{\textbf{Step 2}}. Next, we need to bound the expectation of the right-hand side $\expe\limits_{\mcs}\left[\Phi(\mathcal{S})\right]$ in Eq.(\ref{SFCML:thoe3:3}). We have
	\begin{equation}
	\begin{aligned}
	\expe\limits_{\mcs}\left[\Phi(\mathcal{S})\right] &= \expe_{\mcs}\left[\sup\limits_{ \mathcal{H}_R} \left(\mathcal{R}^{\text{cml}}_{\ell}(f) - \hat{\mathcal{R}}^{\text{cml}}_{\mathcal{S}}(f) \right)\right] \\
	&= \expe_{\mcs}\left[\sup\limits_{\mathcal{H}_R} \expe_{\mathcal{S}'}\left[\hat{\mathcal{R}}^{\text{cml}}_{\mathcal{S}'}(f) - \hat{\mathcal{R}}^{\text{cml}}_{\mathcal{S}}(f)\right]\right] \\
	& \le \expe_{\mcs, \mathcal{S}'} \left[\sup\limits_{\mathcal{H}_R} \left(\hat{\mathcal{R}}^{\text{cml}}_{\mathcal{S}'}(f) - \hat{\mathcal{R}}^{\text{cml}}_{\mathcal{S}}(f)\right)\right] 
	\label{SFCML:eq52}
	\end{aligned}
	\end{equation}
	Now, the most crucial step of the proof is to conduct CML symmetrization, which represents that introducing Rademacher variable $\bss$ (i.e., exchange single instance instead of pairs) does not change the expectation. By applying Thm.\ref{cml:sysm} to Eq.(\ref{SFCML:eq52}), we have
	\begin{equation}
	\begin{aligned}
	\expe_{\mcs}\left[\Phi(\mathcal{S})\right]
	& \le \expe_{\mcs, \mathcal{S}'} \left[\sup\limits_{\mathcal{H}_R} \left(\hat{\mathcal{R}}^{\text{cml}}_{\mathcal{S}'}(f) - \hat{\mathcal{R}}^{\text{cml}}_{\mathcal{S}}(f)\right)\right] \\
	&= \frac{1}{M} \sum_{u_i \in \mcu} \expe_{\mcs_i, \mathcal{S}'_i} \expe_{\sg_i} \left[\mathop{\sup}\limits_{ \mathcal{H}_R} \frac{1}{n_i^+n_i^-} \sum_{j=1}^{n_i^+} \sum_{k=1}^{n_i^-} \mathcal{Q}^{jk}_{i, \sg} \right]\\
	& \le \frac{2}{M} \sum_{u_i \in \mcu} \expe_{\mcs_i,\sg_i} \left[\mathop{\sup}\limits_{\mathcal{H}_R} \frac{1}{n_i^+n_i^-} \sum_{j=1}^{n_i^+} \sum_{k=1}^{n_i^-} \mathcal{Q}_{(i)}^{jk} \right] \\
	&= 2 \mathfrak{R}_{\ell,\mcs}^{\text{cml}}(\mathcal{H}_R)
	\end{aligned} \label{SFCML:eq41}
	\end{equation}
	where 
	\[
	\mathcal{Q}_{(i)}^{jk} = \frac{\sigma^+_{ij} + \sigma^-_{ik}}{2} \cdot \ell^{(i)}(v_j^+, v_k^-).
	\]
	
	\noindent \textcolor{blue}{\textbf{Step 3}}. Similarly, we can follow the proof of $\ps$ to derive a bound with respect of $\mathfrak{R}_{\ell,\mcs}^{\text{cml}}(\mathcal{H}_R)$ by applying Mcdiarmid's Inequality again, and hence, for any $\delta \in (0,1)$, with probability at least $1 - \delta / 2$, the following holds
	\begin{equation}
	\begin{aligned}
	\mathfrak{R}_{\ell,\mcs}^{\text{cml}}(\mathcal{H}_R) \lesssim \hat{\mathfrak{R}_{\ell,\mcs}^{\text{cml}}(\mathcal{H}_R)} + \phi \cdot \frac{R}{M} \cdot \sqrt{\frac{\log2/\delta}{2} \cdot \left(\frac{1}{n_i^+} + \frac{1}{n_i^-}\right)}. 
	\end{aligned} \label{SFCML:eq42}
	\end{equation}
	This immediatly suggests that, for any $\delta \in (0,1)$, with probability at least $1 - \delta$, we have
	\begin{equation}
	\begin{aligned}
	\Phi(\mathcal{S}) \lesssim 2\hat{\mathfrak{R}_{\ell,\mcs}^{\text{cml}}(\mathcal{H}_R)}
	+ \phi \cdot \frac{R}{M} \cdot \sqrt{\frac{\log2/\delta}{2} \cdot \left(\frac{1}{n_i^+} + \frac{1}{n_i^-}\right)}.
	\end{aligned}\label{SFCML:eq43}
	\end{equation}
	
	Finally, based on Thm.\ref{thm3} and the union bound, for any $\delta \in (0,1)$, with probability at least $1 - \delta$, we have proved
	\begin{equation}
	\begin{aligned}
	\mathcal{R}_{\ell}^{\text{cml}}(f) &\lesssim \hat{\mathcal{R}}^{\text{cml}}_{\mcs}(f) + \phi \cdot \frac{\Max(\lambda, \sqrt{R \cdot d})}{M} \cdot \sqrt{\frac{1}{\tilde{N}}}   \\
	&+ \phi \cdot \frac{R}{M} \cdot \sqrt{\frac{\log2/\delta}{2}} \cdot \sqrt{\frac{1}{\tilde{N}}}.
	\end{aligned}
	\end{equation}
\end{pf}

\subsection{Generalization Bound of sampling-based CML} \label{SFCML:sampling-based}
\begin{rthm4} (\textbf{Generalization Upper Bound of sampling-based CML Eq.(\ref{e5})}). Let $\mathcal{H}_R$ be the hypothesis set and $\ell$ be $\phi$-Lipschitz. Given the sample set $\mcs = \mathop{\cup}\limits_{u_i \in \mcu} \mcs_i$ where $\mcs^{(i)} = \{v_j^+\}_{j=1}^{n_i^+} \cup \{v_k^-\}_{k=1}^{n_i^-}, n_i^+ + n_i^- = N$, for any $\delta \in (0,1)$, with probability at least $1 - \delta$, the following equation holds for all possible embedding $\mathcal{H}_R$:
	\begin{equation}
	\begin{aligned}
	\mathcal{R}_{\ell}^{\text{cml}}(f) &\lesssim \hat{\mathcal{R}}^{cml}_{\mcs}(f) + \phi\cdot \frac{\Max(\lambda, \sqrt{R \cdot d})}{M} \cdot \sqrt{\frac{1}{\tilde{N}}} \\ 
	&+ \frac{(\lambda + 4R)}{M} \cdot \sum_{u_i \in \mathcal{U}} D_{TV}(\hat{\mathbb{P}}^{(i)}, \tilde{\mathbb{P}}^{(i)}) \\ 
	&+ \phi \cdot \frac{R}{M} \cdot \sqrt{\frac{\log2/\delta}{2}} \cdot \sqrt{\frac{1}{\tilde{N}}}
	\end{aligned}\label{thm1:eq5}
	\end{equation}
	where $\hat{\mathbb{P}}^{(i)}$ is the original distribution with $\hat{\mathbb{P}}^{(i)}_{ik} \equiv \frac{1}{n^+_in^-_i}$; $D_{TV}(\hat{\mathbb{P}}^{(i)}, \tilde{\mathbb{P}}^{(i)}) = \frac{1}{2} \cdot\left\|\hat{\mathbb{P}}^{(i)} - \tilde{\mathbb{P}}^{(i)} \right\|_1$ is the \textit{Total Variance (TV)} between two probability distributions $\hat{\mathbb{P}}^{(i)}$ and $\tilde{\mathbb{P}}^{(i)}$ on $\mcs$, which characterizes the difference between two probability distributions. 
\end{rthm4}

\begin{pf} Similarly, define 
	\begin{equation}
	\begin{aligned}
	\tilde{\Phi}(\mcs) &= \mathop{\sup\limits_{ \mathcal{H}_R}}  \left(\mathcal{R}_{\ell}^{\text{cml}}(f) - \tilde{\mathcal{R}}^{\text{cml}}_{\mcs}(f) \right) \\
	 &= \mathop{\sup\limits_{\mathcal{H}_R}}  \left(\mathcal{R}_{\ell}^{\text{cml}}(f) - \hat{\mathcal{R}}^{\text{cml}}_{\mcs}(f) + \hat{\mathcal{R}}^{\text{cml}}_{\mcs}(f) - \tilde{\mathcal{R}}^{\text{cml}}_{\mcs}(f) \right)\\
	 & \le \underbrace{\mathop{\sup\limits_{ \mathcal{H}_R}}  \left(\mathcal{R}_{\ell}^{\text{cml}}(f) - \hat{\mathcal{R}}^{\text{cml}}_{\mcs}(f))\right)}_{\ps} + \underbrace{\mathop{\sup\limits_{\mathcal{H}_R}}  \left(\hat{\mathcal{R}}^{\text{cml}}_{\mcs}(f) - \tilde{\mathcal{R}}^{\text{cml}}_{\mcs}(f)\right)}_{\textcolor{orange}{\textcircled{3}}}\\
	\end{aligned}
	\end{equation}
	
	\noindent \textcolor{blue}{\textbf{Step 1}}. First of all, following the proof of Thm.\ref{thm3} and Thm.\ref{theo1}, for any $\delta \in (0,1)$, with probability at least $1 - \delta$, we have
	\begin{equation}
	\begin{aligned}
	\ps &= \mathop{\sup\limits_{\mathcal{H}_R}}  \left(\mathcal{R}_{\ell}^{\text{cml}}(f) - \hat{\mathcal{R}}^{\text{cml}}_{\mcs}(f)\right) \\
	&\lesssim \phi\cdot \frac{\Max(\lambda, \sqrt{R \cdot d})}{M} \cdot \sqrt{\frac{1}{\tilde{N}}} + \phi \cdot \frac{R}{M} \cdot \sqrt{\frac{\log2/\delta}{2}} \cdot \sqrt{\frac{1}{\tilde{N}}} .
	\end{aligned}\label{SBCML:eq43}
	\end{equation}
	\noindent \textcolor{blue}{\textbf{Step 2}}.
	In order to clarify $\textcolor{orange}{\textcircled{3}}$ thoroughly, recall that we have
	
	\begin{equation}
	\begin{aligned}
	\hat{\mathcal{R}}^{\text{cml}}_{\mcs}(f) &= 
	\frac{1}{M} \sum_{u_i \in \mcu} \sum_{j=1}^{n_i^+} \sum_{k=1}^{n_i^-} \hat{\mathbb{P}}^{(i)}_{jk}  \cdot \ell^{(i)}(v_j^+, v_k^-)\\
	\tilde{\mathcal{R}}^{\text{cml}}_{\mcs}(f) &= 
	\frac{1}{M} \sum_{u_i \in \mcu} \sum_{j=1}^{n_i^+} \sum_{k=1}^{n_i^-}  \tilde{\mathbb{P}}^{(i)}_{jk} \cdot \ell^{(i)}(v_j^+, v_k^-)\\
	\end{aligned}
	\end{equation}  
	where $\tilde{\mathbb{P}}^{(i)}_{jk}=\mathbb{P}(\bm{v}_j^+, \bm{v}_k^-)$ represents the probability that item $v_k^-$ is sampled as a negative instance with respect to $v_j^+$ and $\hat{\mathbb{P}}^{(i)}_{jk} = \mathbb{P}(\bm{v}_j^+, \bm{v}_k^-) \equiv \frac{1}{n_i^+n_i^-} > 0$ could be regarded as the ground truth probability.
	
	According to this, we have 
	\begin{equation}
	\begin{aligned}
	\mathop{\sup\limits_{\mathcal{H}_R}} \left(\hat{\mathcal{R}}^{\text{cml}}_{\mcs}(f) - \tilde{\mathcal{R}}^{\text{cml}}_{\mcs}(f) \right) &= \frac{1}{M} \mathop{\sup\limits_{\mathcal{H}_R}} \left( \sum_{u_i \in \mathcal{U}}\sum_{j=1}^{n_i^+} \sum_{k=1}^{n_i^-} \ell^{(i)}(v_j^+, v_k^-) \cdot \left(\hat{\mathbb{P}}^{(i)}_{jk} - \tilde{\mathbb{P}}^{(i)}_{jk}\right) \right). 
	\label{SFCML:eq54}
	\end{aligned}
	\end{equation}
	Let 
	\[
	\boldsymbol{\ell}^{(i)} = \left[\ell^{(i)}(v_1^+, v_1^-), \ell^{(i)}(v_1^+, v_2^-), \dots, \ell^{(i)}(v_{n_i^+}^+, v_{n_i^-}^-)\right]^\top,
	\]
	\[
	\hat{\boldsymbol{\mathbb{P}}}^{(i)} = \left[\hat{\mathbb{P}}^{(i)}_{11}, \dots, \hat{\mathbb{P}}^{(i)}_{21}, \dots, \hat{\mathbb{P}}^{(i)}_{n_i^+1}, \dots, \hat{\mathbb{P}}^{(i)}_{n_i^+n_i^-}\right]^\top,
	\]
	\[
	\tilde{\boldsymbol{\mathbb{P}}}^{(i)} = \left[\tilde{\mathbb{P}}^{(i)}_{11}, \dots, \tilde{\mathbb{P}}^{(i)}_{21}, \dots, \tilde{\mathbb{P}}^{(i)}_{n_i^+1}, \dots, \tilde{\mathbb{P}}^{(i)}_{n_i^+n_i^-}\right]^\top,
	\]
	and then $\textcolor{orange}{\textcircled{3}}$ could be reformulated as 
	\begin{equation}
	\begin{aligned}
	\mathop{\sup\limits_{\mathcal{H}_R}} \left(\hat{\mathcal{R}}^{\text{cml}}_{\mcs}(f) - \tilde{\mathcal{R}}^{\text{cml}}_{\mcs}(f) \right) &= \frac{1}{M} \cdot \mathop{\sup\limits_{\mathcal{H}_R}} \left( \sum_{u_i \in \mathcal{U}}\sum_{j=1}^{n_i^+} \sum_{k=1}^{n_i^-} \ell^{(i)}(v_j^+, v_k^-) \cdot \left(\hat{\mathbb{P}}^{(i)}_{jk} - \tilde{\mathbb{P}}^{(i)}_{jk}\right) \right) \\
	& \overset{\textcolor{blue}{(***)}}{\le} \frac{1}{M} \mathop{\sup\limits_{\mathcal{H}_R}} \sum_{u_i \in \mathcal{U}} \left( \left\|\boldsymbol{\ell}^{(i)}\right\|_{\infty} \cdot \left\|\hat{\mathbb{P}}^{(i)} - \tilde{\mathbb{P}}^{(i)} \right\|_1\right) \\
	&= \frac{2\left(\lambda + 4R\right)}{M} \sum_{u_i \in \mcu} \frac{1}{2} \cdot \left\|\hat{\mathbb{P}}^{(i)} - \tilde{\mathbb{P}}^{(i)} \right\|_1 \\
	& \lesssim \frac{\left(\lambda + 4R\right)}{M} \cdot \sum_{u_i \in \mcu} D_{TV}(\hat{\mathbb{P}}^{(i)}, \tilde{\mathbb{P}}^{(i)})
	\label{SFCML:eq55}
	\end{aligned}
	\end{equation}
	where again $\textcolor{blue}{(***)}$ is achieved by  the Cauchy-Buniakowsky-Schwarz Inequality $\left\|\boldsymbol{x}^\top \boldsymbol{y}\right\| \le \left\|\boldsymbol{x}\right\| \cdot \left\|\boldsymbol{y}\right\|$ here; $D_{TV}(\hat{\mathbb{P}}^{(i)}, \tilde{\mathbb{P}}^{(i)}) = \frac{1}{2} \cdot \left\|\hat{\mathbb{P}}^{(i)} - \tilde{\mathbb{P}}^{(i)} \right\|_1$ is the \textit{Total Variance}(TV), which reflects the discrepancy between sampling-strategy-induced distribution and the ground truth distribution.
	
	\noindent \textcolor{blue}{\textbf{Step 3}}. Taking \textcolor{blue}{\textbf{Step 1}} and \textcolor{blue}{\textbf{Step 2}} into account, for any $\delta \in (0,1)$, with probability at least $1 - \delta$, the following holds:
	\begin{equation}
	\begin{aligned}
	\tilde{\Phi}(\mathcal{S}) &\lesssim \phi \cdot \frac{\Max(\lambda, \sqrt{R \cdot d})}{M} \cdot \sqrt{\frac{1}{\tilde{N}}} \\
	& + \phi \cdot \frac{R}{M} \cdot \sqrt{\frac{\log2/\delta}{2} } \cdot \sqrt{\frac{1}{\tilde{N}}} \\
	& + \frac{\left(\lambda + 4R\right)}{M} \cdot \sum_{u_i \in \mcu} D_{TV}(\hat{\mathbb{P}}^{(i)}, \tilde{\mathbb{P}}^{(i)})
	\end{aligned}
	\end{equation}
	Therefore, for any $\delta \in (0,1)$, with probability at least $1 - \delta$, we have
	\begin{equation}
	\begin{aligned}
	\mathcal{R}_{\ell}^{\text{cml}}(f) \lesssim& \ \ \ \tilde{\mathcal{R}}^{\text{cml}}_{\mcs}(f) \\
	&+ \phi \cdot \frac{\Max(\lambda, \sqrt{R \cdot d})}{M} \cdot \sqrt{\frac{1}{\tilde{N}}} \\
	& + \frac{\left(\lambda + 4R\right)}{M} \cdot \sum_{u_i \in \mcu} D_{TV}(\hat{\mathbb{P}}^{(i)}, \tilde{\mathbb{P}}^{(i)})\\
	& + \phi \cdot \frac{R}{M} \cdot \sqrt{\frac{\log2/\delta}{2} } \cdot \sqrt{\frac{1}{\tilde{N}}}
	\end{aligned}
	\end{equation}
	This completed the proof.
\end{pf}

\clearpage

\section{Additional Experiment results}

\subsection{Details of evaluation metrics}\label{app_metrics}

In some typical recommendation systems, users often care about the top-$K$ items in recommendation lists, so the most relevant items should be ranked first as much as possible. Motivated by this, we evaluate the performance of competitors and our algorithm with the following extensively adopted six metrics, including:
\begin{itemize}
	\item \textbf{Precision} (P@$K$) counts the proportion that the ground-truth items are among the Top-$K$ recommended list.
	\[
	\text{P@}K = \frac{1}{M} \sum_{u_i \in \mathcal{U}}\frac{|I_{u_i} \cap R_{u_i}|}{K}
	\]
	where $I_{u_i}$ is the set of ground-truth items of user $u_i$; $R_{u_i}$ is the top-$K$ recommendation list for user $u_i$; and $|\cdot|$ means the size of set.
	\item \textbf{Recall} (R@$K$) is defined as the number of the ground-truth items in top-$K$ recommendation list divided by the amount of totally ground-truth items. This reflects the ability of model to find the relevant items.
	\[
	\text{R@}K = \frac{1}{M} \sum_{u_i \in \mathcal{U}} \frac{|I_{u_i} \cap R_{u_i}|}{|I_{u_i}|}
	\]
	\item \textbf{Normalized Discounted Cumulative Gain} (NDCG@$K$) counts the ground-truth items in the top-$K$ recommendation list with a position weighting strategy, i.e., assigning larger value on top items than bottom ones.
	\[
	\text{NDCG@}K = \frac{1}{M}\sum_{u_i \in \mathcal{U}} \frac{\text{DCG}_{u_i}\text{@}K}{\text{IDCG}_{u_i}\text{@}K}
	\]
	Specifically, the $\text{DCG}_{u_i}\text{@}K$ and $\text{IDCG}_{u_i}\text{@}K$ are defined as:
	\[
	\text{DCG}_{u_i}\text{@}K = \sum_{j = 1} ^ {|R_{u_i}|}\frac{1 \cdot \mathbb{I}(R_{u_i, j} \in I_{u_i})}{\log_2(j + 1)},
	\]
	\[
	\text{IDCG}_{u_i}\text{@}K = \sum_{k = 1}^{\min(K,|I_{u_i}|)} \frac{1}{\log_2(k + 1)},
	\]
	where $R_{u_i, j}$ respresents the $j$-th item in the top-$K$ recommendation list; $\mathbb{I}(\cdot)$ is an indicator function that returns one if the statement is true and returns zero, otherwise.
	\item \textbf{Mean Average Precision} (MAP) is an extension of Average Precision(AP). AP is the average of precision values at all positions where ground-truth items are found.
	\[
	\text{AP}_{u_i} = \frac{1}{|I_{u_i}|}\sum_{j = 1}^{|\hat{R}_{u_i}|}\frac{|I_{u_i} \cap \hat{R}_{u_i, 1: j}| \cdot \mathbb{I}(j \in I_{u_i})}{rank_j^{u_i}}
	\] 
	\[
	\text{MAP} = \frac{1}{M} \sum_{u_i \in \mathcal{U}} \text{AP}_{u_i}
	\]
	where different from $R_{u_i}$, $\hat{R}_{u_i}$ is the recommendation rankings in terms of all items for user $u_i$; $\hat{R}_{u_i, 1:j}$ represents the top-$j$ recommendation list for user $u_i$; and $rank_j^{u_i}$ means the ranking of item $j$ in $\hat{R}_{u_i}$.
	\item \textbf{Mean Reciprocal Rank} (MRR) takes the rank of each recommended item into account. It is the average of reciprocal ranks of the desired item:
	\[
	\text{MRR} = \frac{1}{M} \sum_{u_i\in \mathcal{U}} \sum_{j = 1}^{|\hat{R}_{u_i}|} \frac{1}{rank_j^{u_i}} \cdot \mathbb{I}(\hat{R}_{u_i, j} \in I_{u_i})
	\]
	\item \textbf{Area Under ROC Curve} (AUC) is the probability that a ground-truth item has a higher rank than a negative item. 
	\[
	\text{AUC}= \frac{1}{M}\sum_{u_i\in\mathcal{U}}\text{AUC}_{u_i}
	\]
	\[
	\text{AUC}_{u_i}= \frac{\sum_{j\in \mathcal{V}}rank_j^{u_i}-\frac{|I_{u_i}|\left(1+|I_{u_i}|\right)}{2}}{N\left(N-|I_{u_i}|\right)
	}
	\]
\end{itemize}

Note that, for all the above metrics, the higher the metric is, the better the performance the algorithm achieves.

\clearpage
\subsection{Adverse evidence of sampling-based CML} \label{add_neg_nums}
\begin{figure*}[h]
	\centering
	 
			\subfigure[P@3]{
				\includegraphics[width=0.3\columnwidth]{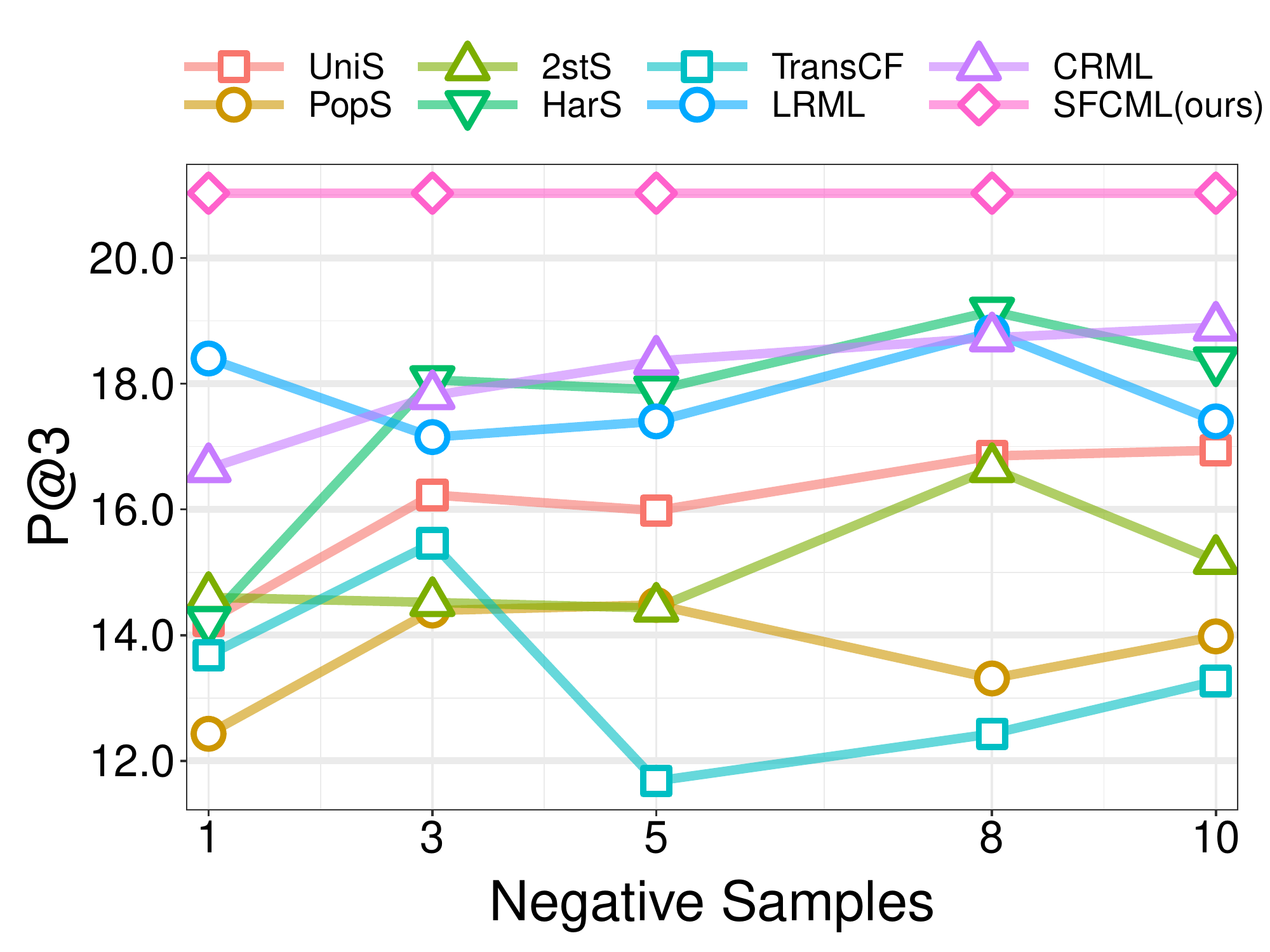}
				\label{ml-1m/P@3}
			}
			\subfigure[R@3]{
				\includegraphics[width=0.3\columnwidth]{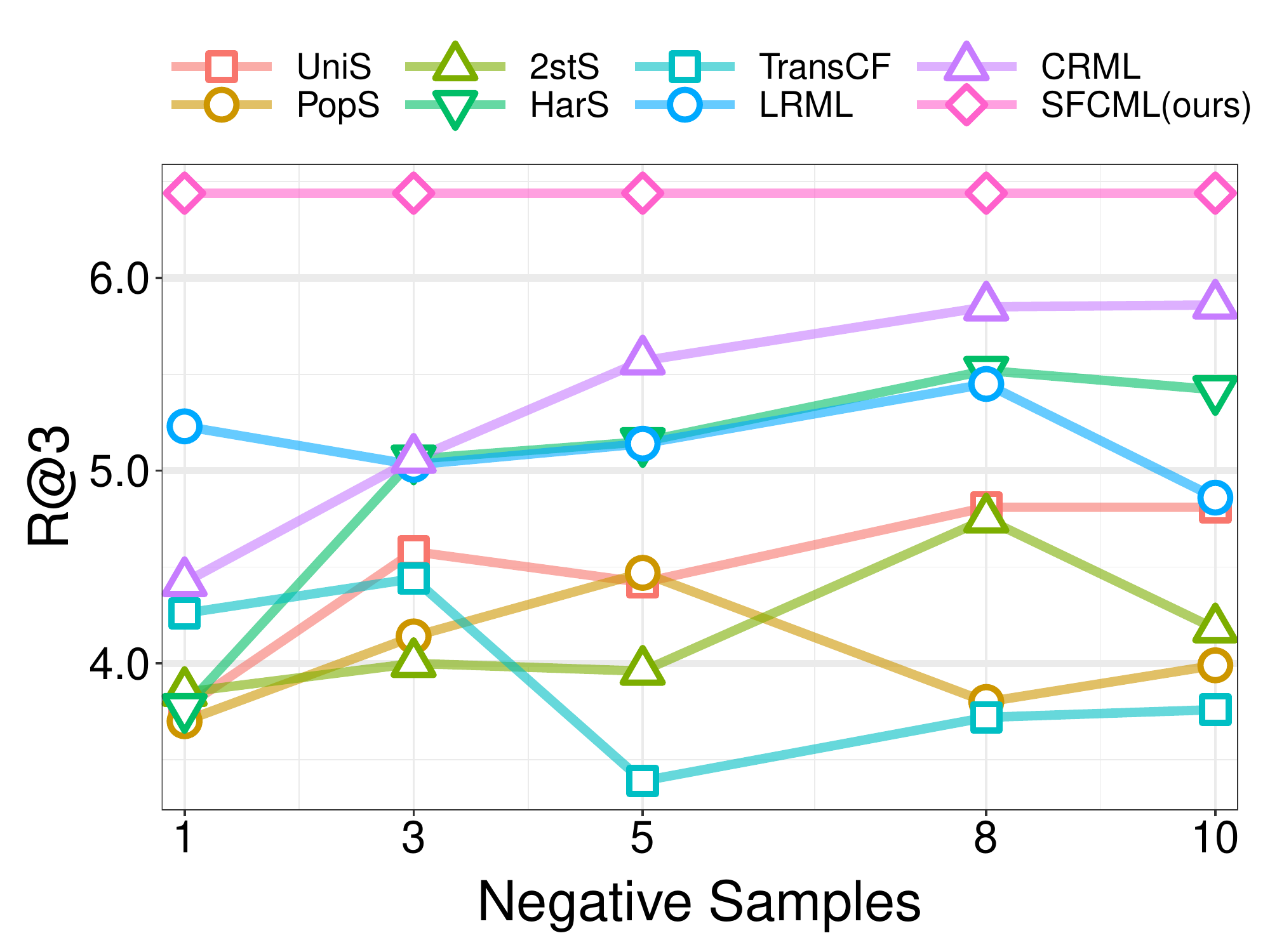}
				\label{ml-1m/R@3}
			}
			\subfigure[NDCG@3]{
				\includegraphics[width=0.3\columnwidth]{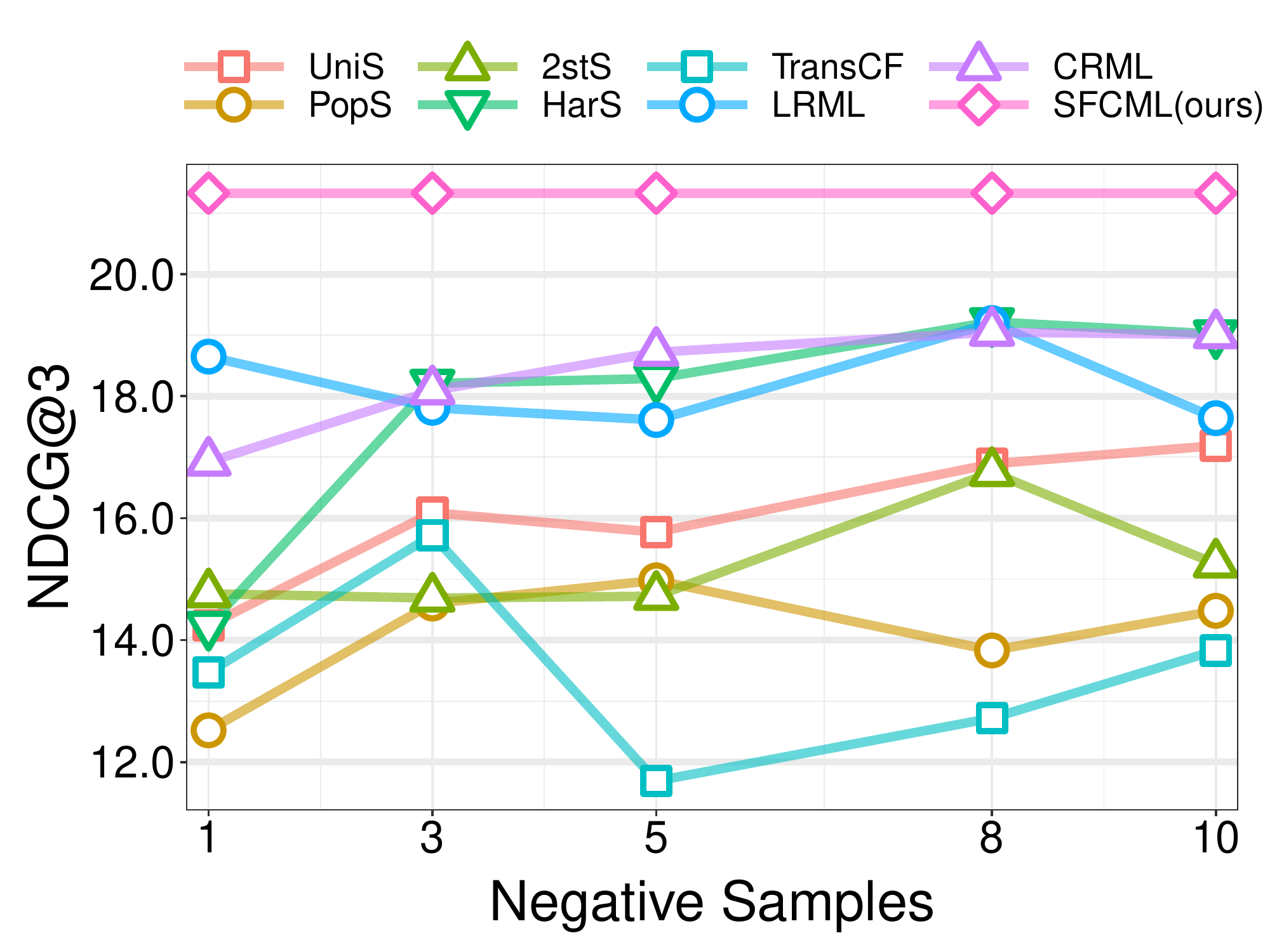}
				\label{ml-1m/NDCG@3}
			} \\ 
			\subfigure[P@5]{
				\includegraphics[width=0.3\columnwidth]{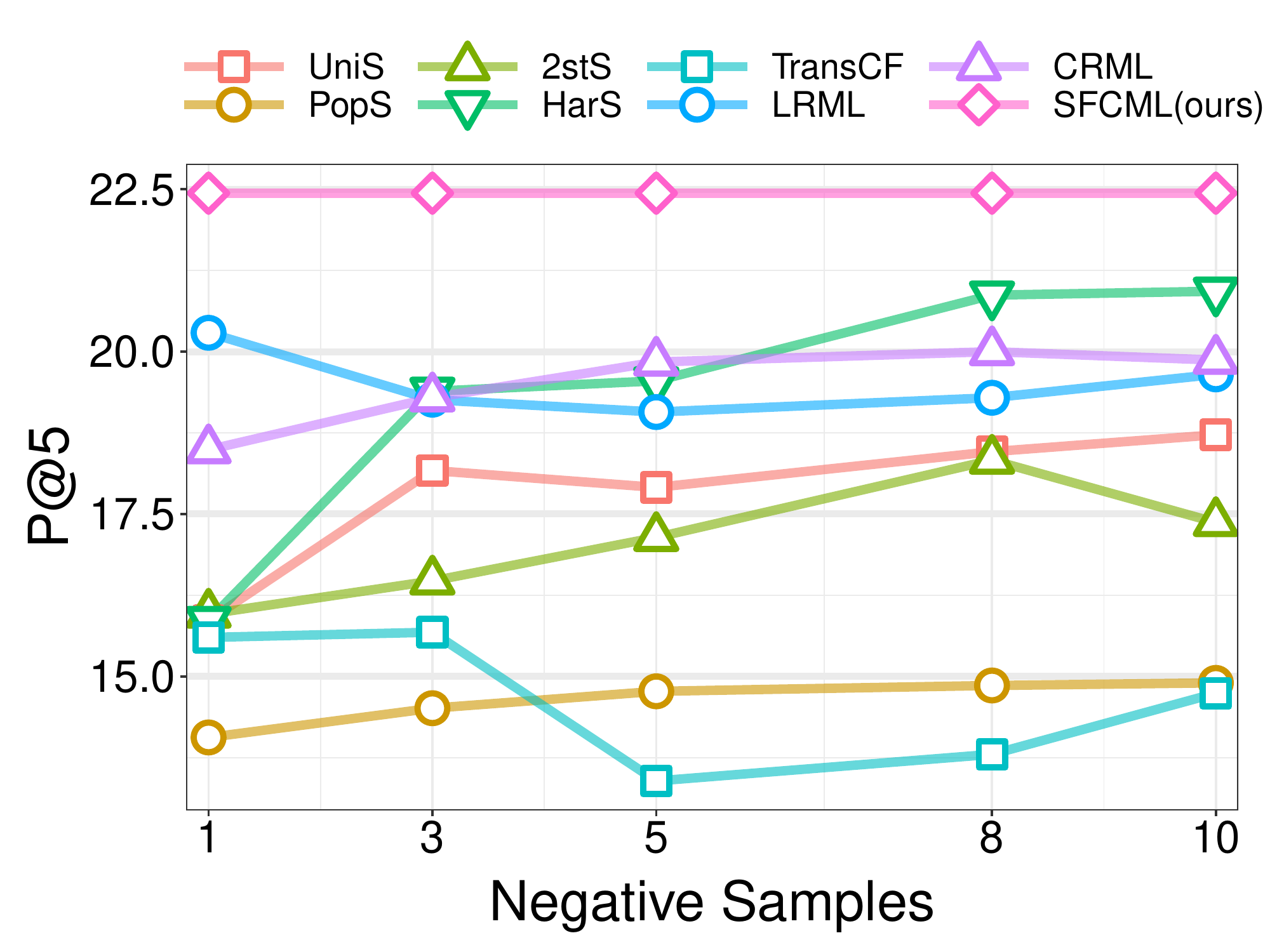}
				\label{ml-1m/P@5}
			}
		\subfigure[NDCG@5]{
			\includegraphics[width=0.3\columnwidth]{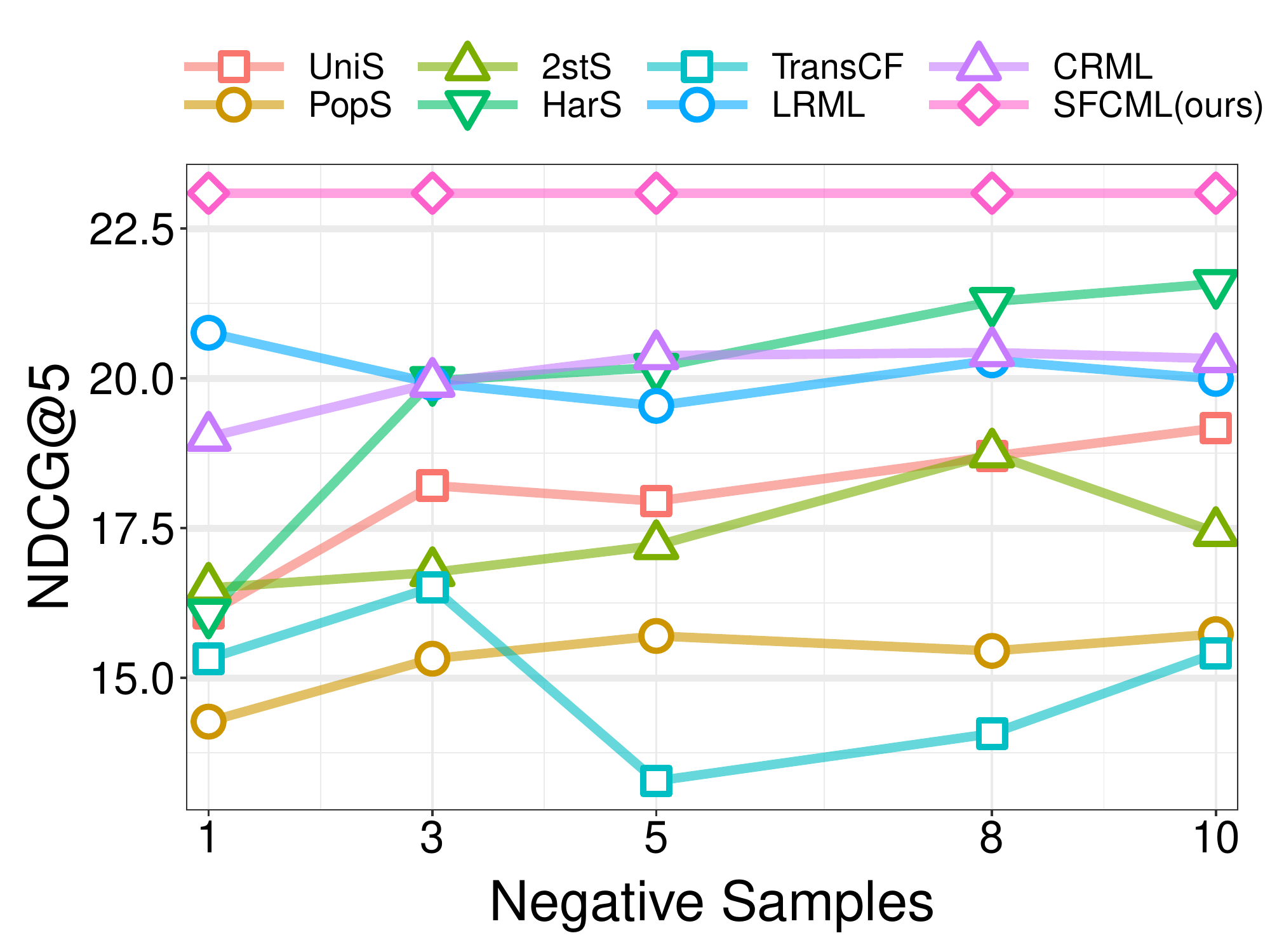}
			\label{ml-1m/NDCG@5}
		}
		\subfigure[P@10]{
			\includegraphics[width=0.3\columnwidth]{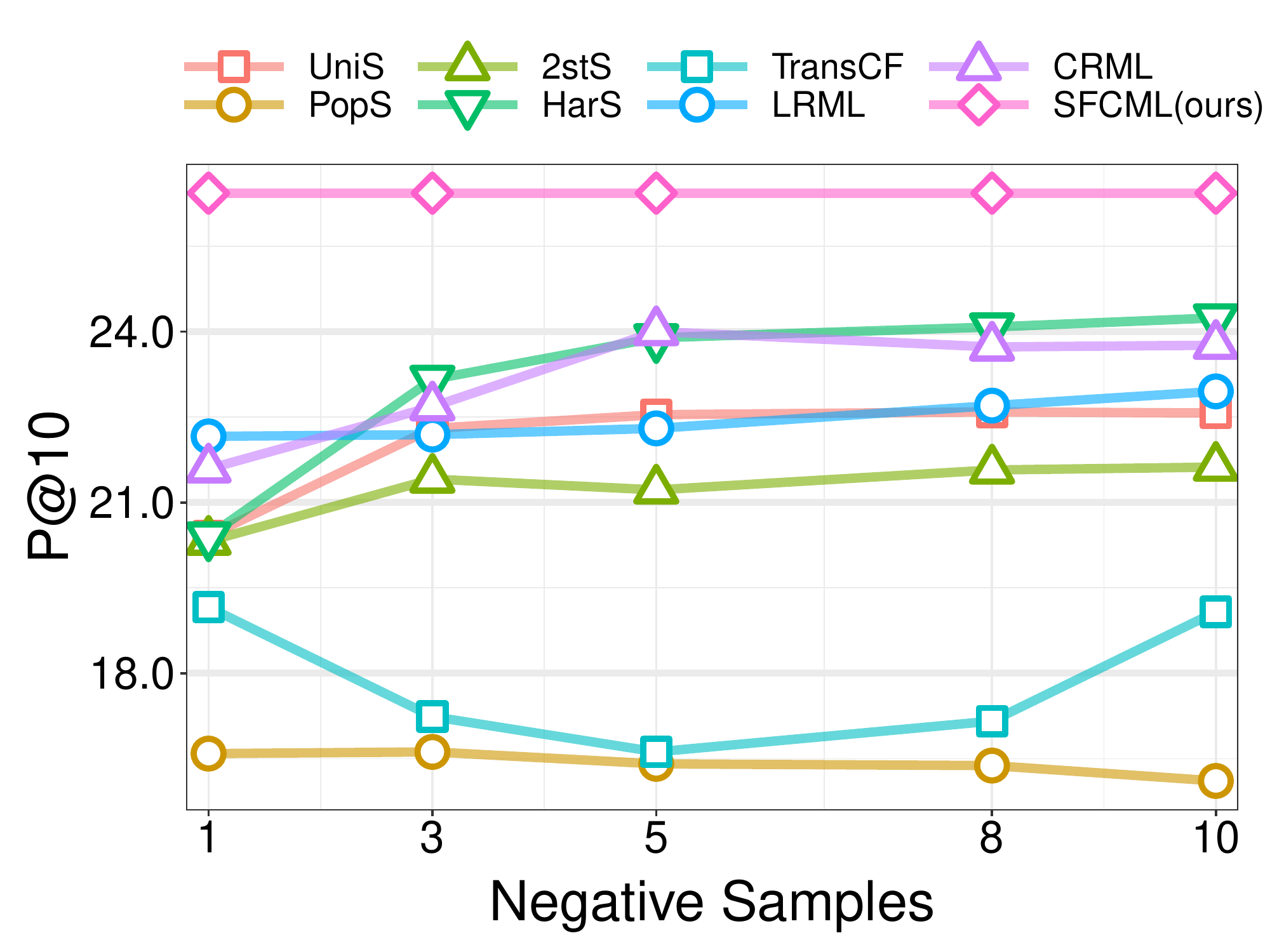}
			\label{ml-1m/P@10}
		} \\
		\subfigure[R@10]{
			\includegraphics[width=0.3\columnwidth]{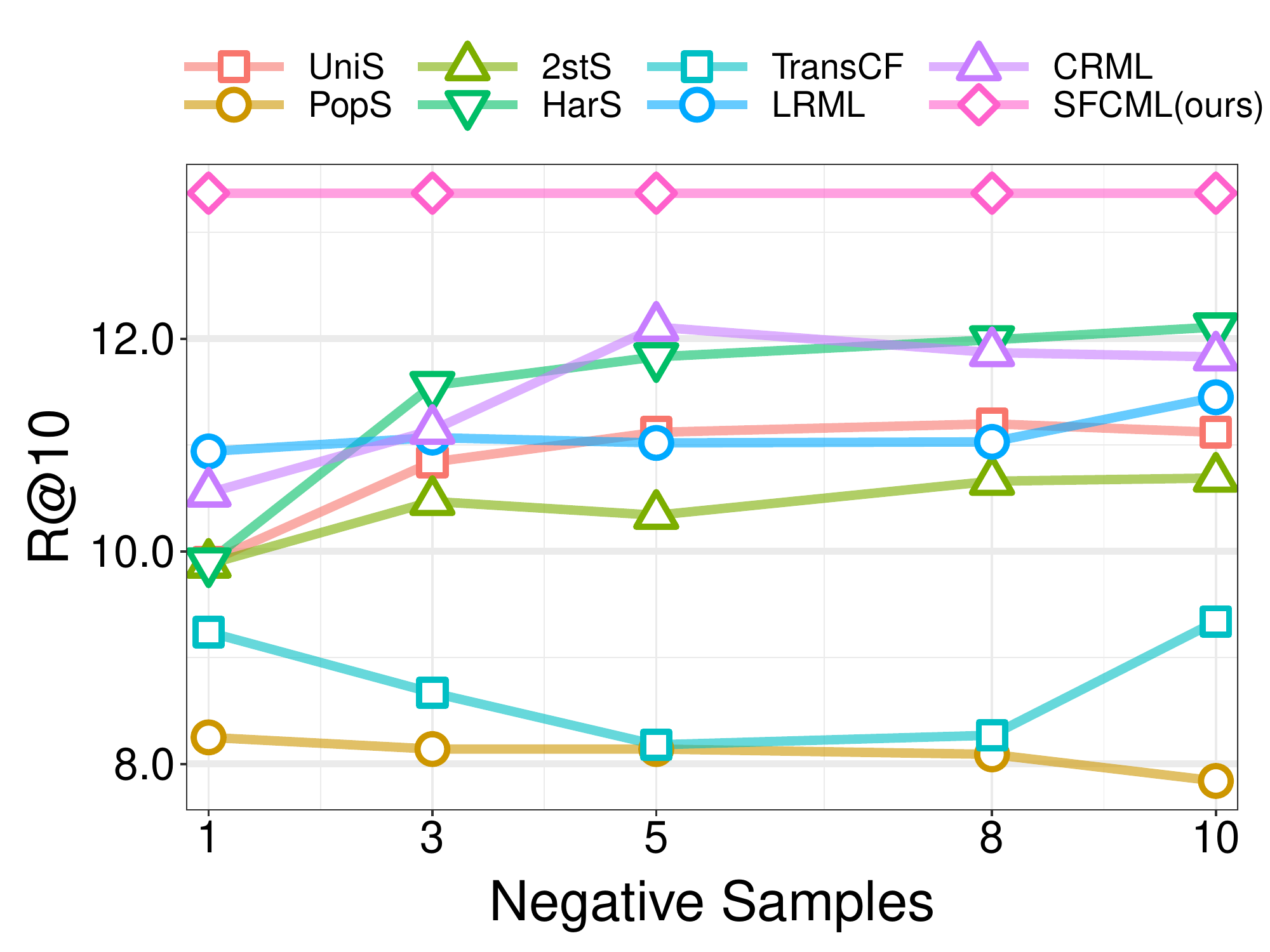}
			\label{ml-1m/R@10}
		}
	\subfigure[NDCG@10]{
		\includegraphics[width=0.3\columnwidth]{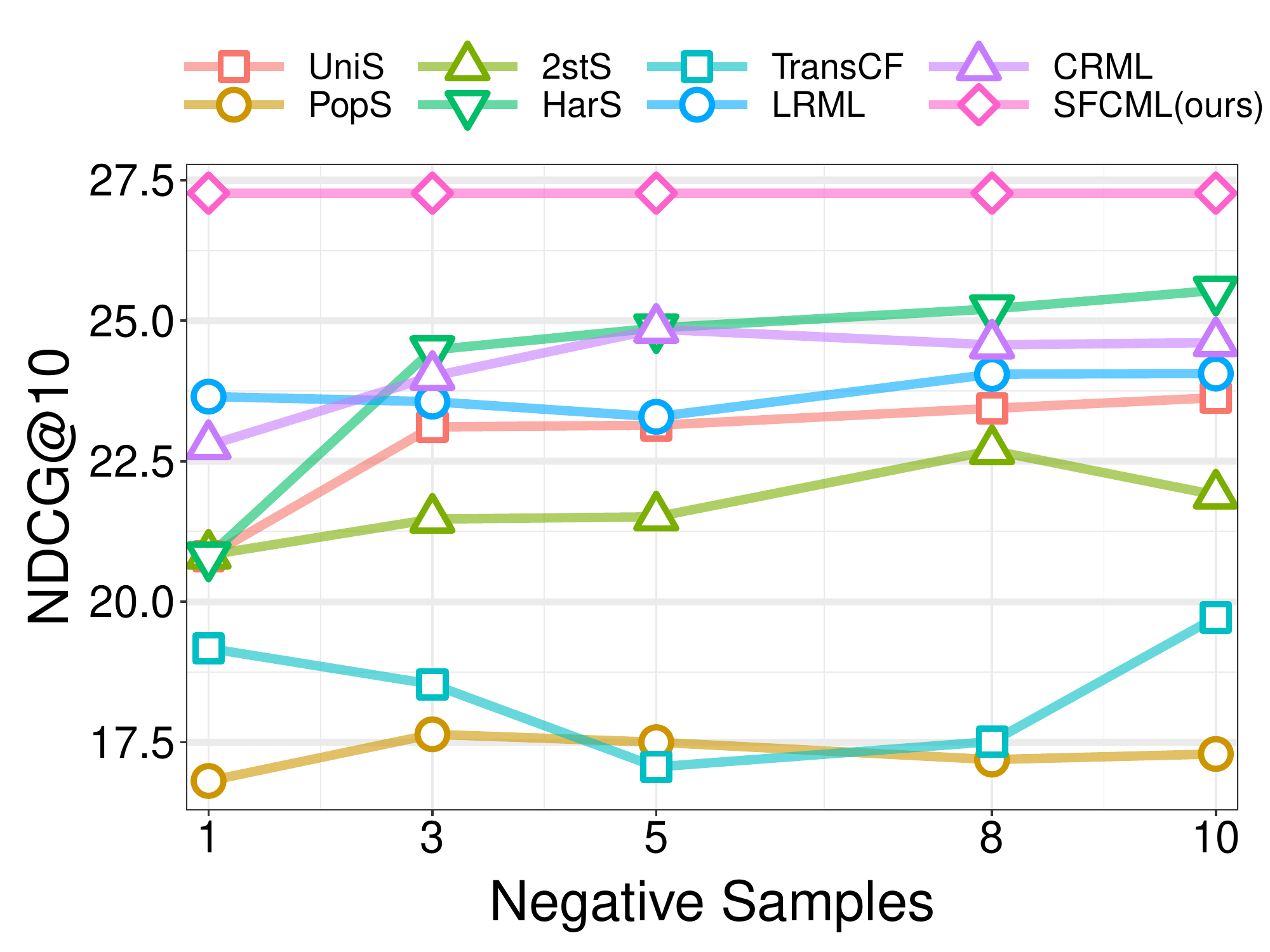}
		\label{ml-1m/NDCG@10}
	}	
	\subfigure[P@20]{
			\includegraphics[width=0.3\columnwidth]{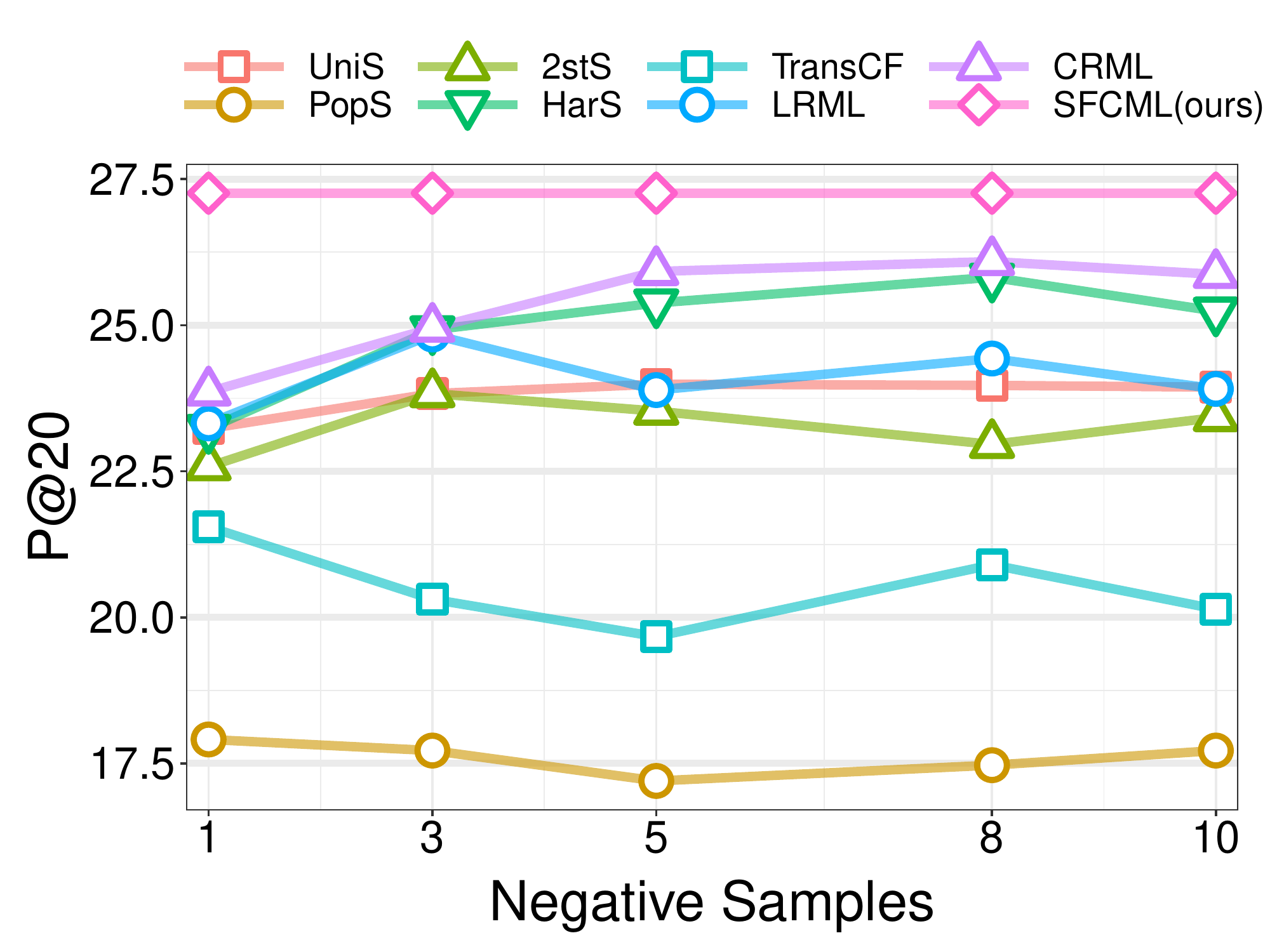}
			\label{ml-1m/P@20}
	} \\
	\subfigure[R@20]{
			\includegraphics[width=0.3\columnwidth]{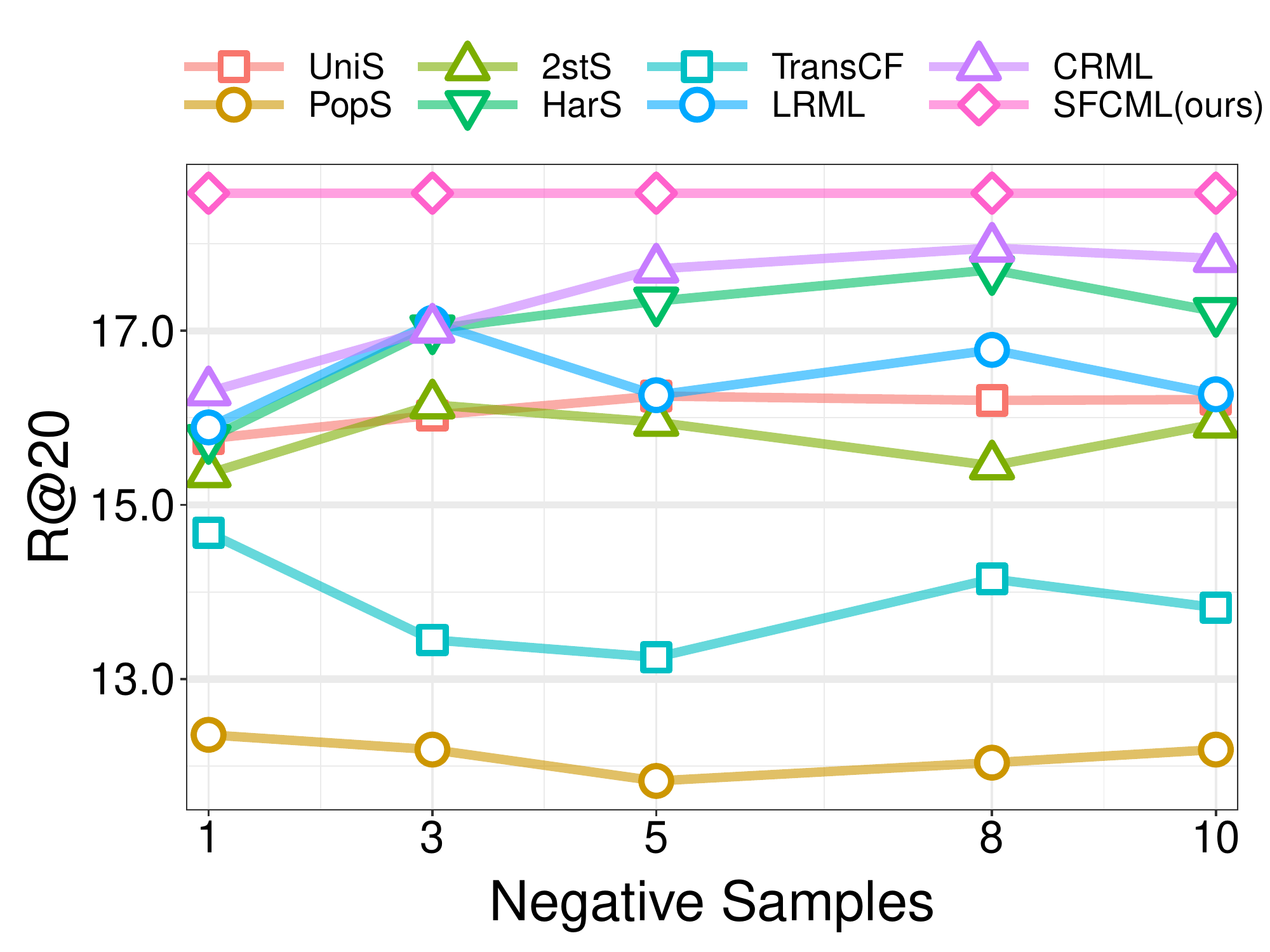}
			\label{ml-1m/R@20}
		}
	\subfigure[NDCG@20]{
		\includegraphics[width=0.3\columnwidth]{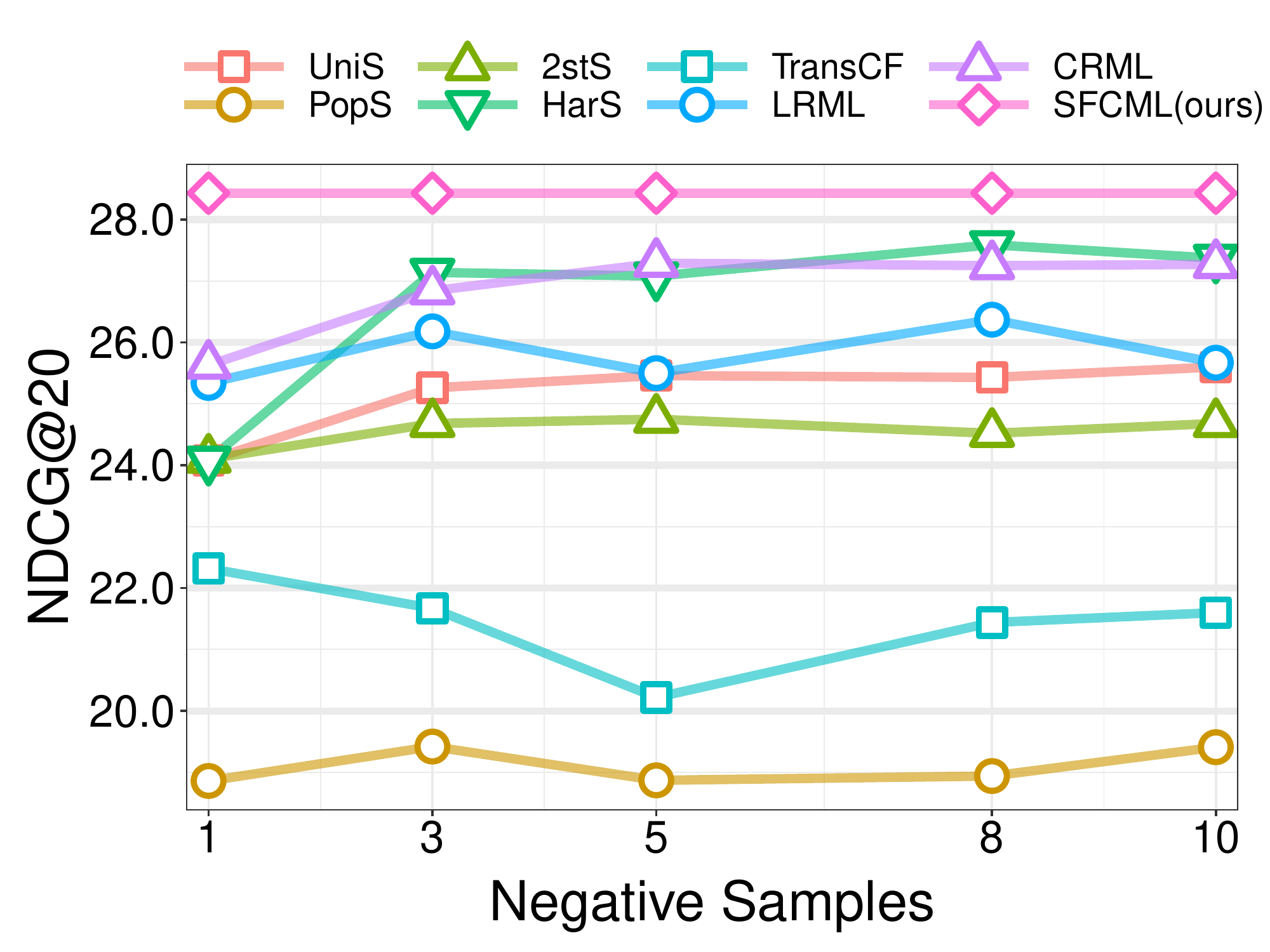}
		\label{ml-1m/NDCG@20}
	}
		\subfigure[AUC]{
				\includegraphics[width=0.3\columnwidth]{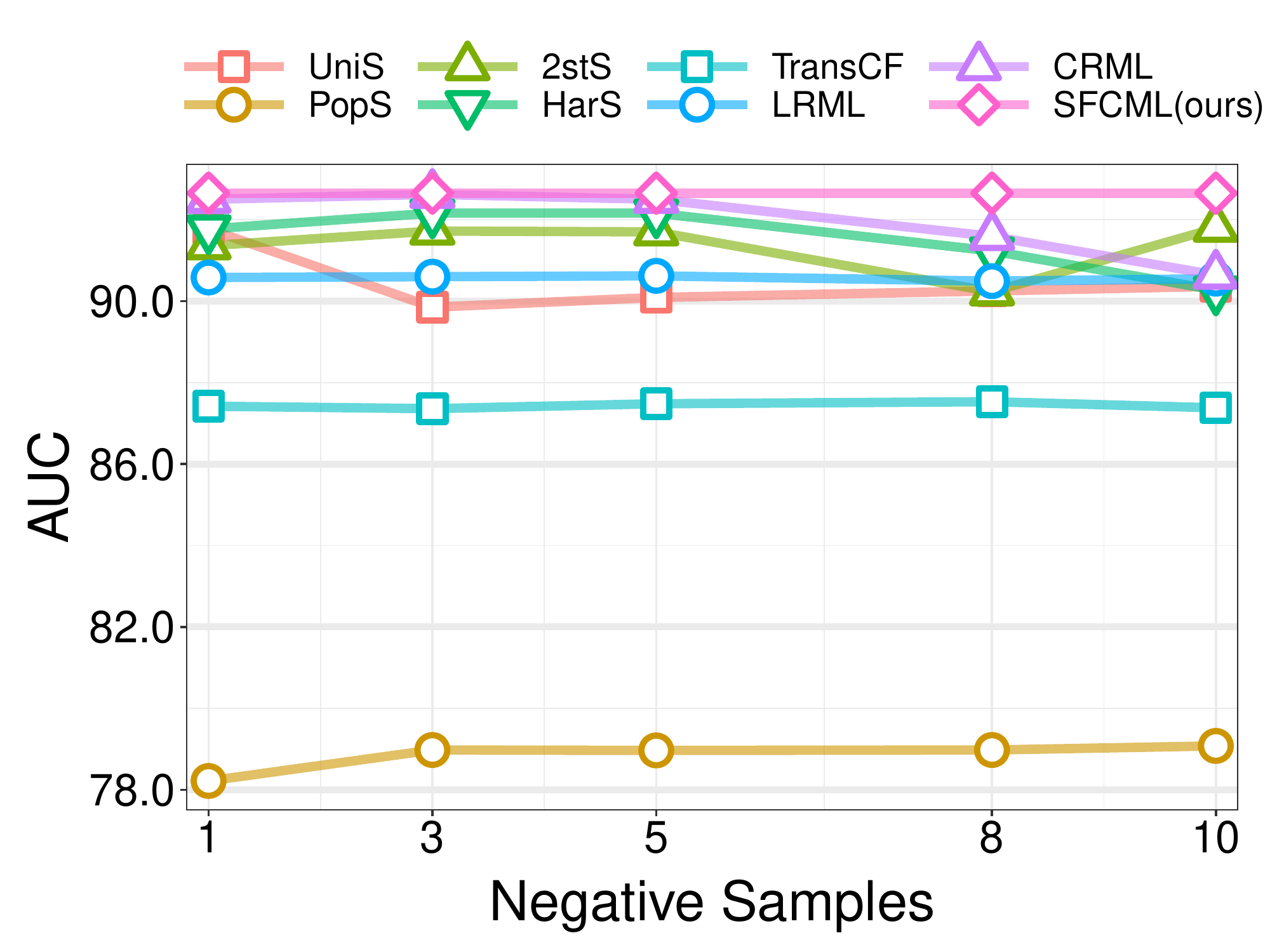}
				\label{ml-1m/AUC}
		}
		\caption{Performance comparisons on validation set of MovieLens-100k with respect to different negative sampling strategies and different sampling numbers $U \in \{1, 3, 5, 8, 10\}$.} 
	  
	\label{ml-1m/add_evidence}
\end{figure*}

\clearpage
\subsection{More evaluation results} \label{more_evaluation_results}

\subsubsection{More datasets}
Tab.\ref{results3} shows the performance comparisons with respect to two larger datasets, i.e., MovieLens-20m and Amazon-Book. 
\begin{table*}[!ht]
	\centering
	\caption{Performance comparisons on MovieLens-20m and Amazon-Book datasets, where '-' means that we cannot complete the experiments due to the out-of-memory issue. The best and second-best performance are highlighted in bold and underlined, respectively.}
	\label{results3}%
	\scalebox{1.0}{
		\begin{tabular}{c|c|ccccccccc}
			\toprule
			& Method &\cellcolor[rgb]{1.0,1.0,1.0}P@3 & R@3 & NDCG@3 & P@5 & R@5 & NDCG@5 & MAP & MRR & AUC \\
			\midrule 
			\multirow{15}[5]{*}{MovieLens-20m} & itemKNN & \cellcolor[rgb]{ .961,  .976,  .992}12.01  & \cellcolor[rgb]{ .906,  .941,  .973}3.77  & \cellcolor[rgb]{ .961,  .976,  .992}12.33  & \cellcolor[rgb]{ .961,  .976,  .992}12.44  & \cellcolor[rgb]{ .965,  .976,  .988}4.96  & \cellcolor[rgb]{ .961,  .976,  .992}12.94  & 8.04 & 26.05 & 95.82 \\
			& GMF & \cellcolor[rgb]{ .965,  .976,  .988}12.45  & \cellcolor[rgb]{ .906,  .941,  .973}3.52  & \cellcolor[rgb]{ .965,  .976,  .988}12.89  & \cellcolor[rgb]{ .965,  .976,  .988}13.40 & \cellcolor[rgb]{ .965,  .976,  .988}4.88  & \cellcolor[rgb]{ .965,  .976,  .988}13.94  & 7.94 & \cellcolor[rgb]{ .965,  .976,  .988}27.31  & \cellcolor[rgb]{ .906,  .941,  .973}96.15  \\
			& MLP & \cellcolor[rgb]{ .906,  .941,  .973}14.55  & \cellcolor[rgb]{ .843,  .906,  .961}4.02  & \cellcolor[rgb]{ .906,  .941,  .973}14.90 & \cellcolor[rgb]{ .906,  .941,  .973}15.74  & \cellcolor[rgb]{ .906,  .941,  .973}5.72  & \cellcolor[rgb]{ .906,  .941,  .973}16.26  & \cellcolor[rgb]{ .906,  .941,  .973}10.64  & \cellcolor[rgb]{ .906,  .941,  .973}30.52  & \cellcolor[rgb]{ .808,  .882,  .949}97.59  \\
			& NCF & \cellcolor[rgb]{ .843,  .906,  .961}15.79  & \cellcolor[rgb]{ .843,  .906,  .961}4.39  & \cellcolor[rgb]{ .843,  .906,  .961}16.07  & \cellcolor[rgb]{ .843,  .906,  .961}16.80 & \cellcolor[rgb]{ .843,  .906,  .961}6.21  & \cellcolor[rgb]{ .843,  .906,  .961}17.42  & \cellcolor[rgb]{ .843,  .906,  .961}11.03  & \cellcolor[rgb]{ .843,  .906,  .961}31.82  & \cellcolor[rgb]{ .843,  .906,  .961}97.49  \\
			& EHCF & \cellcolor[rgb]{ .808,  .882,  .949}\underline{17.04}  & \cellcolor[rgb]{ .808,  .882,  .949}\underline{5.39}  & \cellcolor[rgb]{ .741,  .843,  .933}\textbf{17.43}  & \cellcolor[rgb]{ .808,  .882,  .949}\underline{17.66}  & \cellcolor[rgb]{ .808,  .882,  .949}\underline{7.19}  & \cellcolor[rgb]{ .741,  .843,  .933}\textbf{18.34}  & \cellcolor[rgb]{ .808,  .882,  .949}\underline{12.66}  & \cellcolor[rgb]{ .808,  .882,  .949}\underline{34.61}  & 96.20 \\
			\cmidrule{2-11}   & UniS & \cellcolor[rgb]{1.0,1.0,1.0}10.11 & 2.57 & \cellcolor[rgb]{ .961,  .976,  .992}10.38  & \cellcolor[rgb]{ .961,  .976,  .992}10.86  & \cellcolor[rgb]{ .961,  .976,  .992}3.72  & \cellcolor[rgb]{ .961,  .976,  .992}11.35  & 7.38 & 23.17 & \cellcolor[rgb]{ .808,  .882,  .949}97.71  \\
			& PopS & \cellcolor[rgb]{1.0,1.0,1.0}10.31 & \cellcolor[rgb]{ .906,  .941,  .973}3.32  & \cellcolor[rgb]{ .961,  .976,  .992}10.68  & 10.38 & \cellcolor[rgb]{ .965,  .976,  .988}4.29  & 10.96 & 6.40 & 23.68 & 89.89 \\
			& 2stS & \cellcolor[rgb]{ .965,  .976,  .988}12.70 & \cellcolor[rgb]{ .906,  .941,  .973}3.50 & \cellcolor[rgb]{ .965,  .976,  .988}13.01  & \cellcolor[rgb]{ .961,  .976,  .992}13.47  & \cellcolor[rgb]{ .965,  .976,  .988}4.90 & \cellcolor[rgb]{ .965,  .976,  .988}14.04  & \cellcolor[rgb]{ .961,  .976,  .992}9.00 & \cellcolor[rgb]{ .965,  .976,  .988}27.76  & 94.90 \\
			& HarS & \cellcolor[rgb]{ .965,  .976,  .988}12.60 & \cellcolor[rgb]{ .961,  .976,  .992}3.31  & 12.95 & \cellcolor[rgb]{ .965,  .976,  .988}13.49  & \cellcolor[rgb]{ .965,  .976,  .988}4.75  & \cellcolor[rgb]{ .965,  .976,  .988}14.09  & \cellcolor[rgb]{ .965,  .976,  .988}9.51 & \cellcolor[rgb]{ .965,  .976,  .988}27.68  & \cellcolor[rgb]{ .808,  .882,  .949}97.60 \\
			& TransCF & \cellcolor[rgb]{1.0,1.0,1.0}7.60 & 2.19 & 7.77 & 8.13 & 3.09 & 8.41 & 5.75 & 18.78 & \cellcolor[rgb]{ .906,  .941,  .973}96.06  \\
			& LRML & \cellcolor[rgb]{ .965,  .976,  .988}12.57  & \cellcolor[rgb]{ .906,  .941,  .973}3.48  & \cellcolor[rgb]{ .965,  .976,  .988}12.96  & \cellcolor[rgb]{ .965,  .976,  .988}13.22  & \cellcolor[rgb]{ .965,  .976,  .988}4.65  & \cellcolor[rgb]{ .965,  .976,  .988}13.84  & 7.71 & \cellcolor[rgb]{ .965,  .976,  .988}27.08  & \cellcolor[rgb]{ .906,  .941,  .973}96.05  \\
			& CRML & \cellcolor[rgb]{ .906,  .941,  .973}14.94  & \cellcolor[rgb]{ .906,  .941,  .973}4.01  & \cellcolor[rgb]{ .906,  .941,  .973}15.33  & \cellcolor[rgb]{ .906,  .941,  .973}16.01  & \cellcolor[rgb]{ .906,  .941,  .973}5.73  & \cellcolor[rgb]{ .906,  .941,  .973}16.68 & \cellcolor[rgb]{ .843,  .906,  .961}11.00 & \cellcolor[rgb]{ .906,  .941,  .973}31.23  & \cellcolor[rgb]{ .808,  .882,  .949}\underline{97.98}  \\
					& NaiveCML & \cellcolor[rgb]{1.0,1.0,1.0}- & - & - & - & - & - & - & - & - \\
			\cmidrule{2-11}
			& SFCML(ours) & \cellcolor[rgb]{ .741,  .843,  .933}\textbf{17.16} & \cellcolor[rgb]{ .741,  .843,  .933}\textbf{5.78} & \cellcolor[rgb]{ .741,  .843,  .933}\textbf{17.43} & \cellcolor[rgb]{ .741,  .843,  .933}\textbf{17.90} & \cellcolor[rgb]{ .741,  .843,  .933}\textbf{7.52} & \cellcolor[rgb]{ .808,  .882,  .949}\underline{18.26} & \cellcolor[rgb]{ .741,  .843,  .933}\textbf{14.69} & \cellcolor[rgb]{ .741,  .843,  .933}\textbf{35.49} & \cellcolor[rgb]{ .741,  .843,  .933}\textbf{98.07} \\
			\midrule
			\multirow{15}[5]{*}{Amazon-Book} & itemKNN & \cellcolor[rgb]{ .996,  .984,  .98}1.87 & \cellcolor[rgb]{ .996,  .984,  .98}0.83 & \cellcolor[rgb]{ .996,  .984,  .98}1.92 & \cellcolor[rgb]{ .996,  .984,  .98}1.73 & 1.10 & \cellcolor[rgb]{ .996,  .984,  .98}1.87 & \cellcolor[rgb]{ .996,  .984,  .98}1.29 & \cellcolor[rgb]{ .973,  .918,  .918}5.25 & 71.64 \\
			& GMF & \cellcolor[rgb]{1.0,1.0,1.0}0.90 & 0.33 & 0.90 & 0.97 & 0.48 & 0.97 & 0.66 & \cellcolor[rgb]{ .996,  .984,  .98}3.08 & \cellcolor[rgb]{ .996,  .984,  .98}83.11 \\
			& MLP & \cellcolor[rgb]{1.0,1.0,1.0}0.91 & 0.32 & 0.91 & 1.04 & \cellcolor[rgb]{ .996,  .984,  .98}0.51 & \cellcolor[rgb]{ .996,  .984,  .98}1.03 & 0.85 & \cellcolor[rgb]{ .996,  .984,  .98}3.47 & \cellcolor[rgb]{ .961,  .882,  .882}89.15 \\
			& NCF & \cellcolor[rgb]{ .996,  .984,  .98}1.37 & \cellcolor[rgb]{ .996,  .984,  .98}0.44 & \cellcolor[rgb]{ .996,  .984,  .98}1.40 & \cellcolor[rgb]{ .996,  .984,  .98}1.52 & \cellcolor[rgb]{ .996,  .984,  .98}0.72 & \cellcolor[rgb]{ .996,  .984,  .98}1.56 & \cellcolor[rgb]{ .996,  .984,  .98}1.04 & \cellcolor[rgb]{ .996,  .984,  .98}4.56 & \cellcolor[rgb]{ .961,  .882,  .882}89.30 \\
			& EHCF & \cellcolor[rgb]{ .961,  .882,  .882}3.11 & \cellcolor[rgb]{ .961,  .882,  .882}1.09 & \cellcolor[rgb]{ .961,  .882,  .882}3.24 & \cellcolor[rgb]{ .961,  .882,  .882}3.31 & \cellcolor[rgb]{ .961,  .882,  .882}1.69 & \cellcolor[rgb]{ .973,  .918,  .918}3.50 & \cellcolor[rgb]{ .961,  .882,  .882}2.01 & \cellcolor[rgb]{ .961,  .882,  .882}8.66 & \cellcolor[rgb]{ .996,  .984,  .98}83.26 \\
			\cmidrule{2-11}      & UniS & \cellcolor[rgb]{ .996,  .984,  .98}1.82 & \cellcolor[rgb]{ .996,  .984,  .98}0.67 & \cellcolor[rgb]{ .996,  .984,  .98}1.93 & \cellcolor[rgb]{ .996,  .984,  .98}1.81 & \cellcolor[rgb]{ .996,  .984,  .98}0.98 & 1.97 & \cellcolor[rgb]{ .973,  .918,  .918}1.26 & \cellcolor[rgb]{ .973,  .918,  .918}5.69 & \cellcolor[rgb]{ .945,  .835,  .835}\underline{91.56}  \\
			& PopS & \cellcolor[rgb]{ .961,  .882,  .882}3.20 & \cellcolor[rgb]{ .961,  .882,  .882}1.27 & \cellcolor[rgb]{ .961,  .882,  .882}3.32 & \cellcolor[rgb]{ .961,  .882,  .882}3.33 & \cellcolor[rgb]{ .961,  .882,  .882}1.87 & \cellcolor[rgb]{ .973,  .918,  .918}3.49 & \cellcolor[rgb]{ .961,  .882,  .882}2.06 & \cellcolor[rgb]{ .961,  .882,  .882}8.83 & \cellcolor[rgb]{ .961,  .882,  .882}88.73 \\
			& 2stS & \cellcolor[rgb]{ .961,  .882,  .882}3.26 & \cellcolor[rgb]{ .961,  .882,  .882}1.18 & \cellcolor[rgb]{ .961,  .882,  .882}3.43 & \cellcolor[rgb]{ .961,  .882,  .882}3.08 & \cellcolor[rgb]{ .961,  .882,  .882}1.62 & \cellcolor[rgb]{ .973,  .918,  .918}3.41 & \cellcolor[rgb]{ .973,  .918,  .918}1.80 & \cellcolor[rgb]{ .961,  .882,  .882}8.28 & \cellcolor[rgb]{ .996,  .984,  .98}82.32 \\
			& HarS & \cellcolor[rgb]{ .961,  .882,  .882}3.43 & \cellcolor[rgb]{ .961,  .882,  .882}1.26 & \cellcolor[rgb]{ .961,  .882,  .882}3.53 & \cellcolor[rgb]{ .961,  .882,  .882}3.52 & \cellcolor[rgb]{ .961,  .882,  .882}1.87 & \cellcolor[rgb]{ .961,  .882,  .882}3.72 & \cellcolor[rgb]{ .961,  .882,  .882}2.25 & \cellcolor[rgb]{ .961,  .882,  .882}9.46 & \cellcolor[rgb]{ .945,  .835,  .835}91.31  \\
			& TransCF & \cellcolor[rgb]{ .973,  .918,  .918}2.25 & \cellcolor[rgb]{ .973,  .918,  .918}0.84 & \cellcolor[rgb]{ .973,  .918,  .918}2.33 & \cellcolor[rgb]{ .973,  .918,  .918}2.31 & \cellcolor[rgb]{ .973,  .918,  .918}1.22 & \cellcolor[rgb]{ .973,  .918,  .918}2.43 & \cellcolor[rgb]{ .973,  .918,  .918}1.38 & \cellcolor[rgb]{ .973,  .918,  .918}6.50 & \cellcolor[rgb]{ .973,  .918,  .918}87.16 \\
			& LRML & \cellcolor[rgb]{1.0,1.0,1.0}0.47 & 0.17 & 0.46 & 0.40 & 0.17 & 0.42 & 0.29 & \cellcolor[rgb]{ .996,  .984,  .98}1.56 & 77.23 \\
			& CRML & \cellcolor[rgb]{ .945,  .835,  .835}\underline{3.77}  & \cellcolor[rgb]{ .945,  .835,  .835}\underline{1.34}  & \cellcolor[rgb]{ .945,  .835,  .835}\underline{3.96}  & \cellcolor[rgb]{ .945,  .835,  .835}\underline{3.88}  & \cellcolor[rgb]{ .945,  .835,  .835}\underline{2.10} & \cellcolor[rgb]{ .945,  .835,  .835}\underline{4.17}  & \cellcolor[rgb]{ .945,  .835,  .835}\underline{2.34}  & \cellcolor[rgb]{ .945,  .835,  .835}\underline{9.98}  & \cellcolor[rgb]{ .961,  .882,  .882}89.82 \\
			& NaiveCML & \cellcolor[rgb]{1.0,1.0,1.0}- & - & - & - & - & - & - & - & - \\
			\cmidrule{2-11}      & SFCML(ours) & \cellcolor[rgb]{ .929,  .788,  .788}\textbf{4.13} & \cellcolor[rgb]{ .929,  .788,  .788}\textbf{1.61} & \cellcolor[rgb]{ .929,  .788,  .788}\textbf{4.24} & \cellcolor[rgb]{ .929,  .788,  .788}\textbf{4.05} & \cellcolor[rgb]{ .929,  .788,  .788}\textbf{2.26} & \cellcolor[rgb]{ .929,  .788,  .788}\textbf{4.30} & \cellcolor[rgb]{ .929,  .788,  .788}\textbf{2.79} & \cellcolor[rgb]{ .929,  .788,  .788}\textbf{11.14} & \cellcolor[rgb]{ .929,  .788,  .788}\textbf{92.66} \\
			\bottomrule
		\end{tabular}%
	}
\end{table*}%

%\begin{figure}[]
%	\centering
%	 
%		\includegraphics[width=1.0\textwidth]{major_revision/heatmap_polar/heatmap_MovieLens-100k}
%		\caption{ in terms of $K=\{10, 20\}$. Please see Appendix.\ref{more_evaluation_results} for more results. }
%		\label{ab}
%	  
%\end{figure}

\subsubsection{More $K$ values}
\begin{figure*}[h]
	\centering

%		\subfigure[MovieLens-100k]{
%			\centering
%			\includegraphics[width=0.48\textwidth, height=0.25\textwidth]{major_revision/heatmap_polar/heatmap_MovieLens-100k}
%			\label{heatmap_MovieLens-100k}
%		}
		\subfigure[CiteULike]{
			\centering
			\includegraphics[width=0.48\textwidth, height=0.25\textwidth]{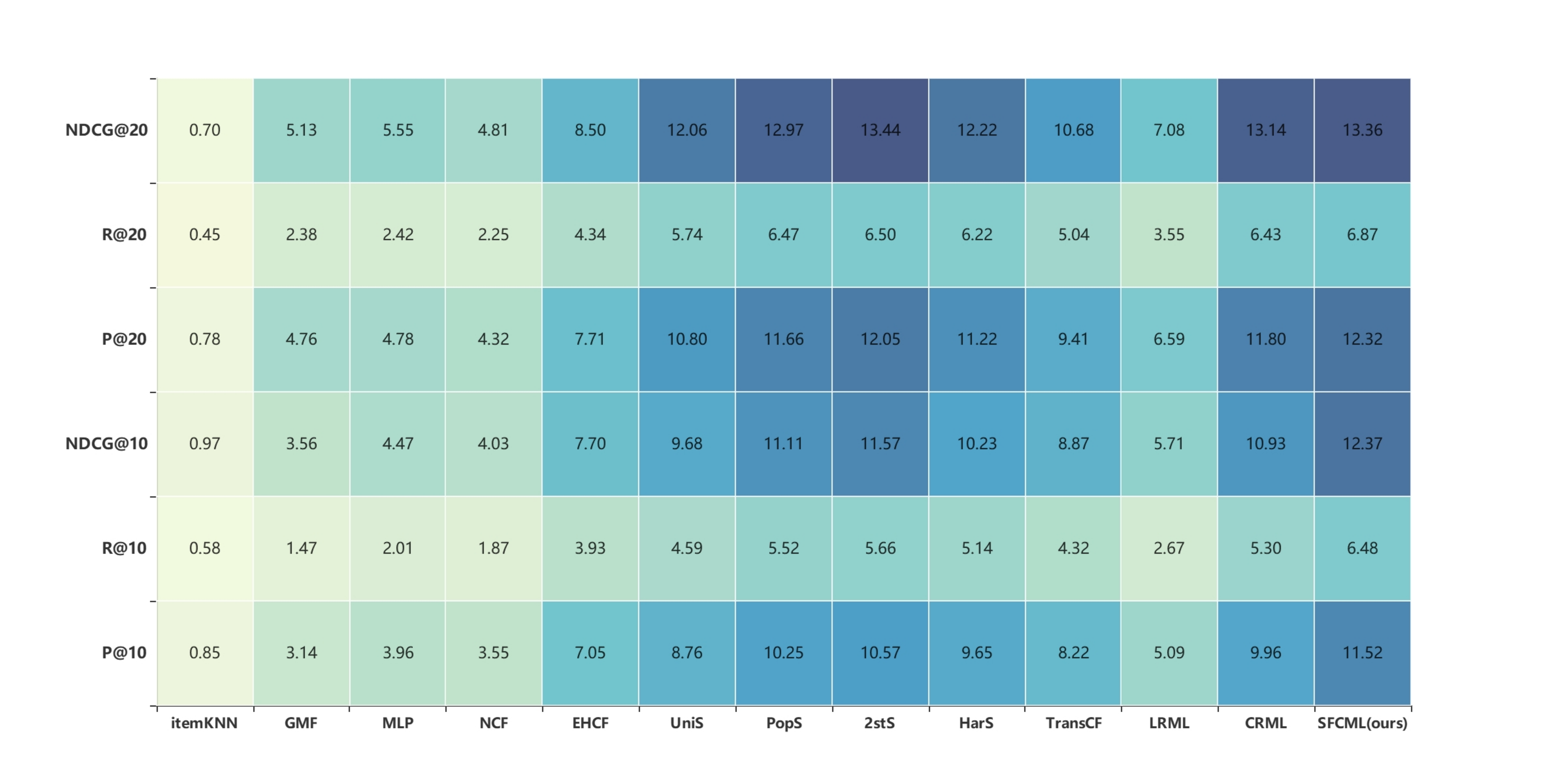}
			\label{heatmap_CiteULike}
		}
	\subfigure[MovieLens-1m]{
		\centering
		\includegraphics[width=0.48\textwidth, height=0.25\textwidth]{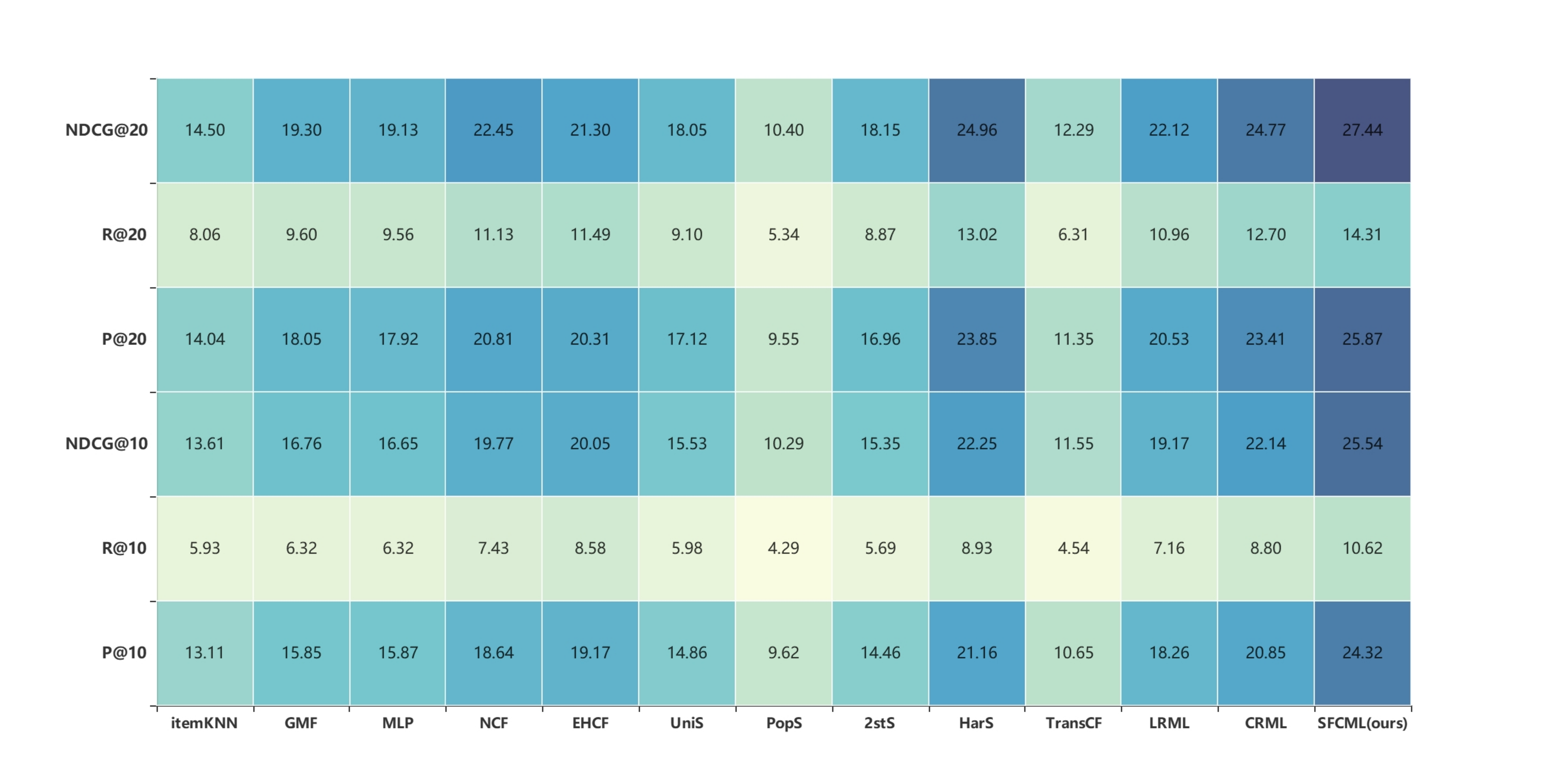}
		\label{more_ml_1m}
	}
	\subfigure[Steam-200k]{
		\centering
		\includegraphics[width=0.48\textwidth, height=0.25\textwidth]{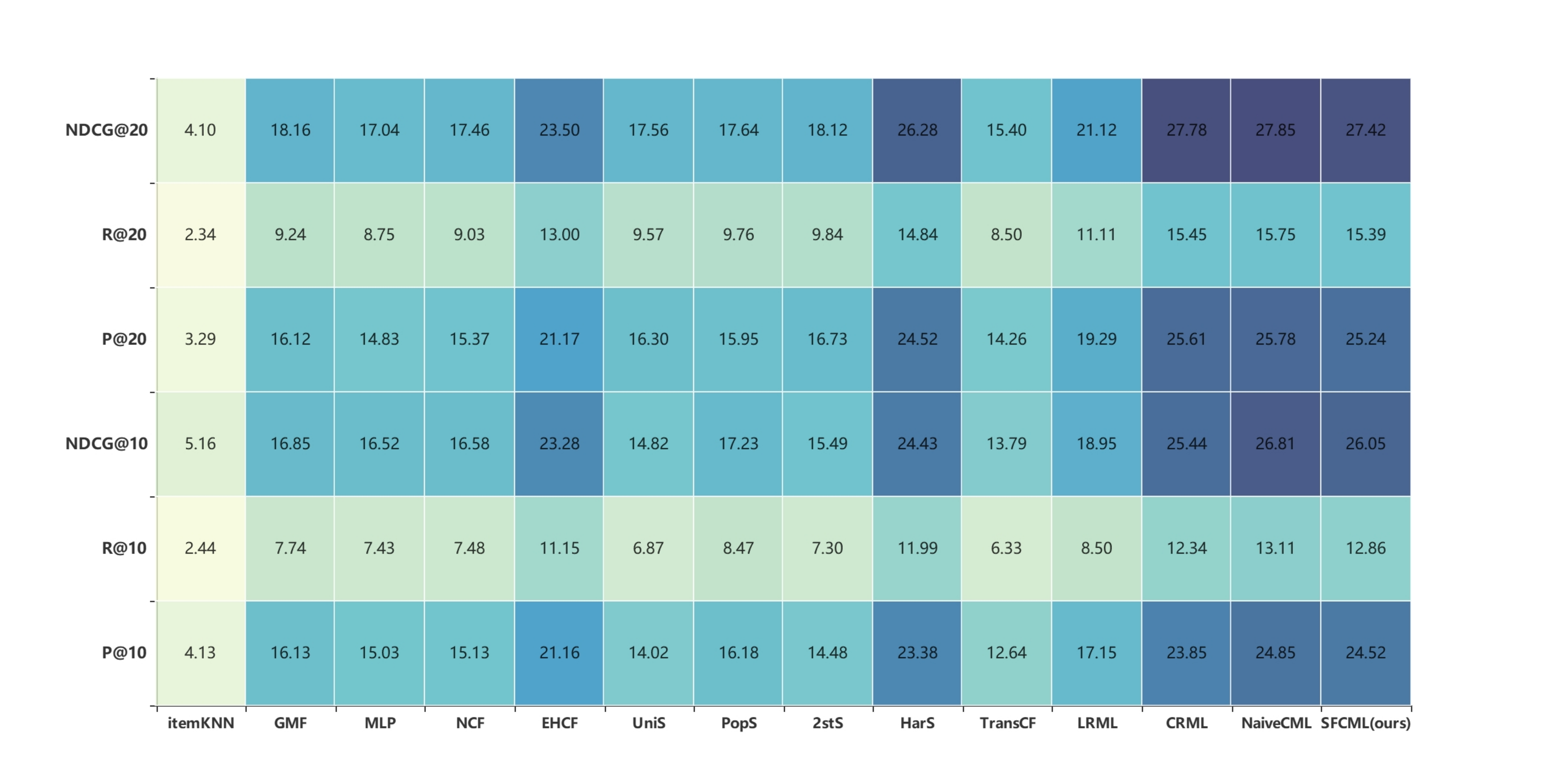}
		\label{ab.sub.200k}
	}
		\subfigure[Anime]{
			\centering
		\includegraphics[width=0.48\textwidth, height=0.25\textwidth]{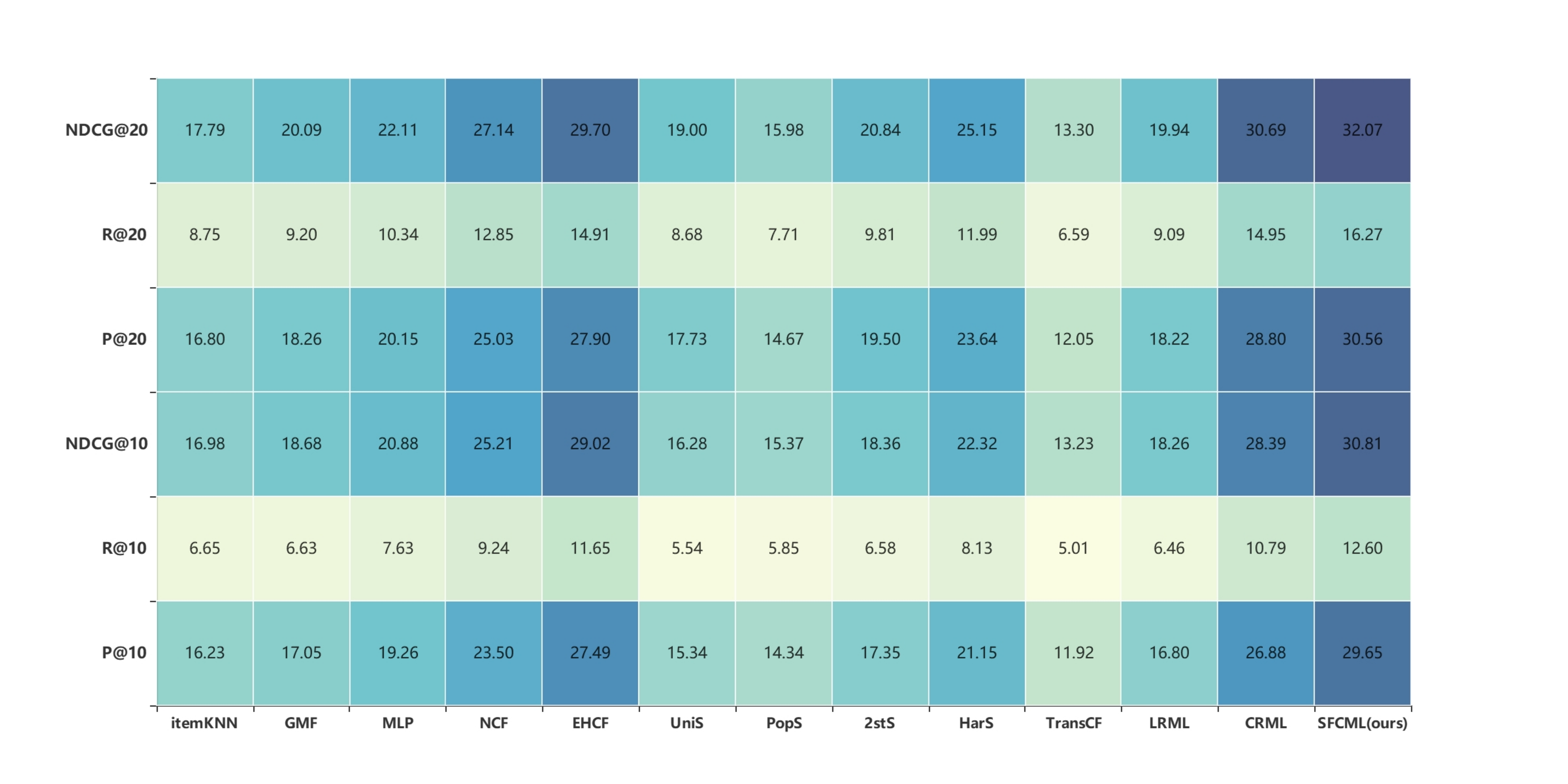}
			\label{more_anime_heatmap}
		}
		\subfigure[MovieLens-20m]{
			\centering
			\includegraphics[width=0.48\textwidth, height=0.25\textwidth]{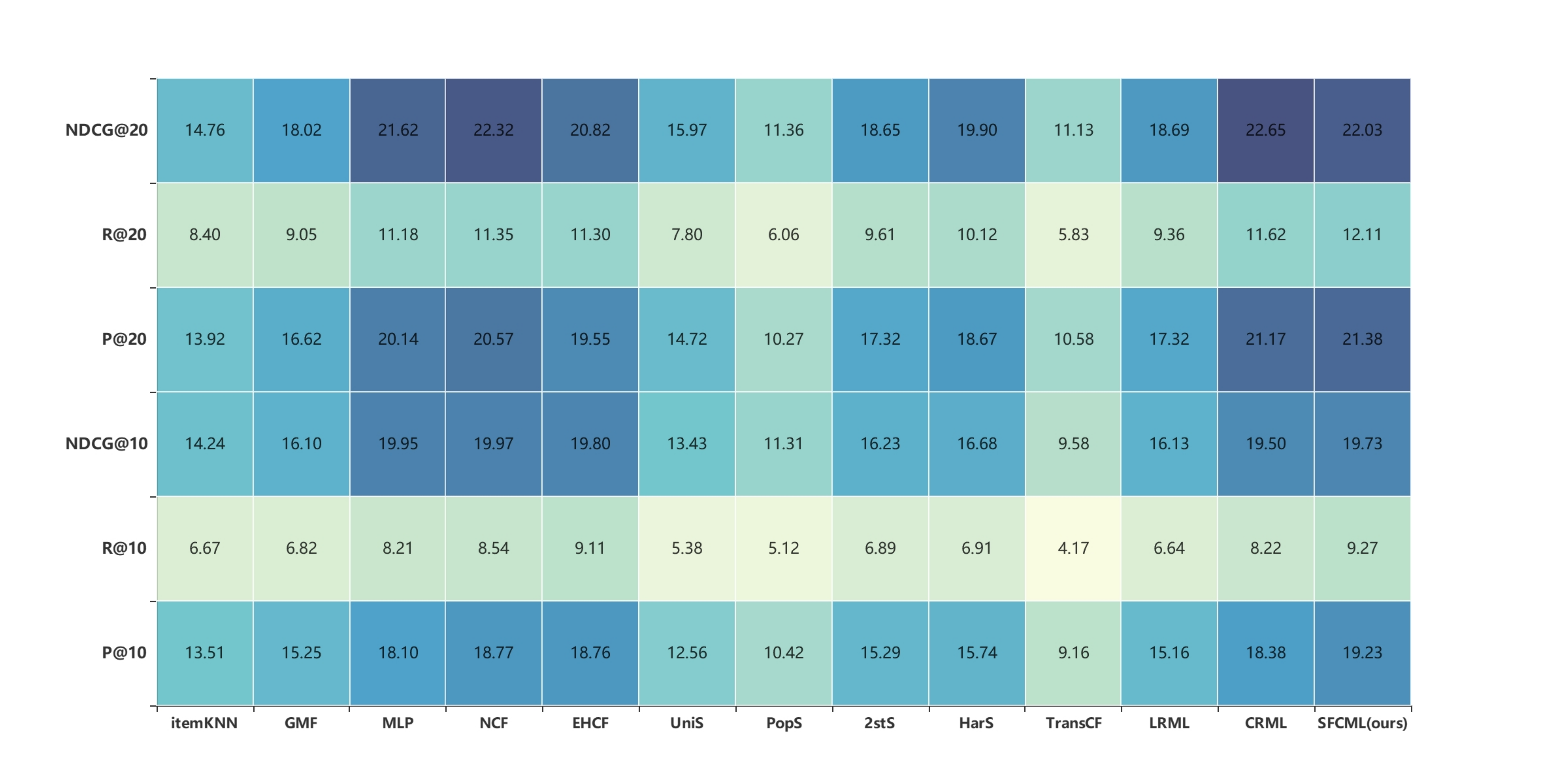}
			\label{more_20m_heatmap}
		}
		\subfigure[Amazon-Book]{
			\centering
			\includegraphics[width=0.48\textwidth, height=0.25\textwidth]{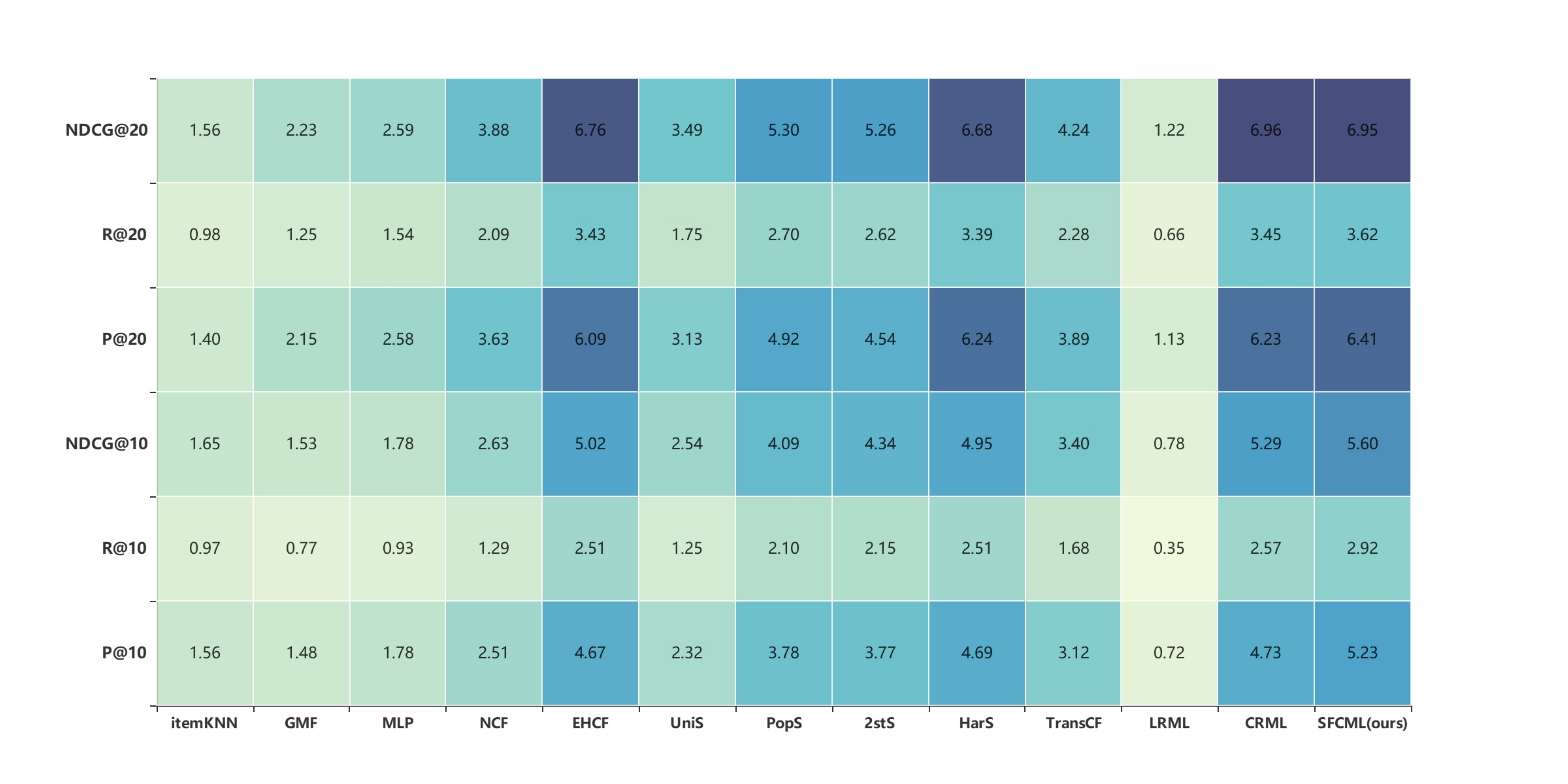}
			\label{more_book_heatmap}
		}
		\caption{Performance comparisons on CiteULike, Steam-200k, MovieLens-1m, Anime, MovieLens-20m and Amazon-Book datasets with respect to $K\in \{10, 20\}$.}
		\label{more_results_of_k_10_20}
	  
\end{figure*}

In order to prove the effectiveness of our proposed SFCML algorithm, we further report the performance of SFCML and its competitors with larger $K \in \{10, 20\}$. See Fig.\ref{more_results_of_k_10_20}.

\clearpage
\newpage

\subsection{Fine-grained performance visualization}
\subsubsection{Fine-grained AUC comparison}\label{auc: app}

\begin{figure*}[h]
	\centering
		\subfigure[MovieLens-1m-3]{
			\includegraphics[width=0.235\textwidth]{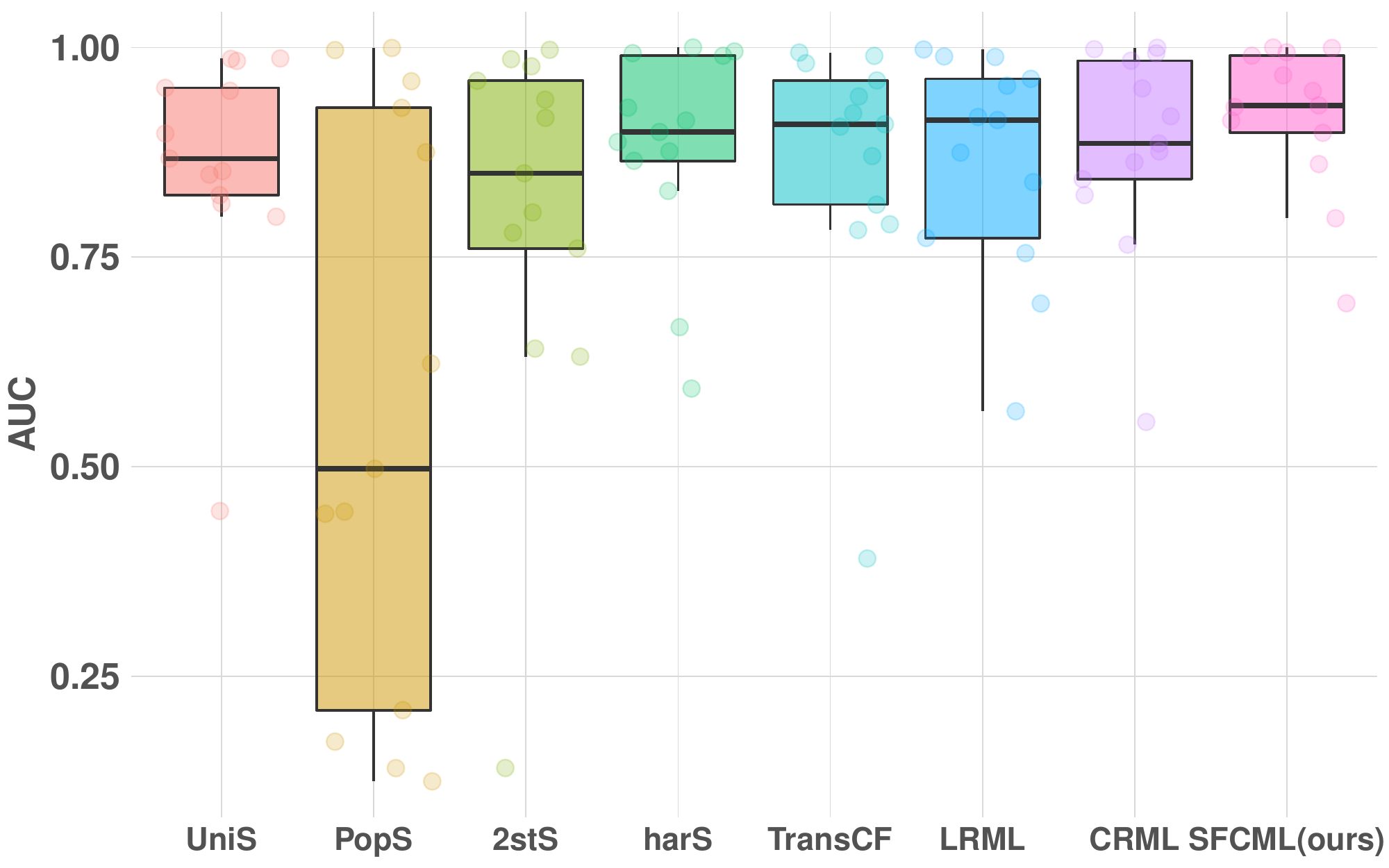}
			\label{main_ml-3}
		}
		\subfigure[MovieLens-1m-4]{
			\includegraphics[width=0.235\textwidth]{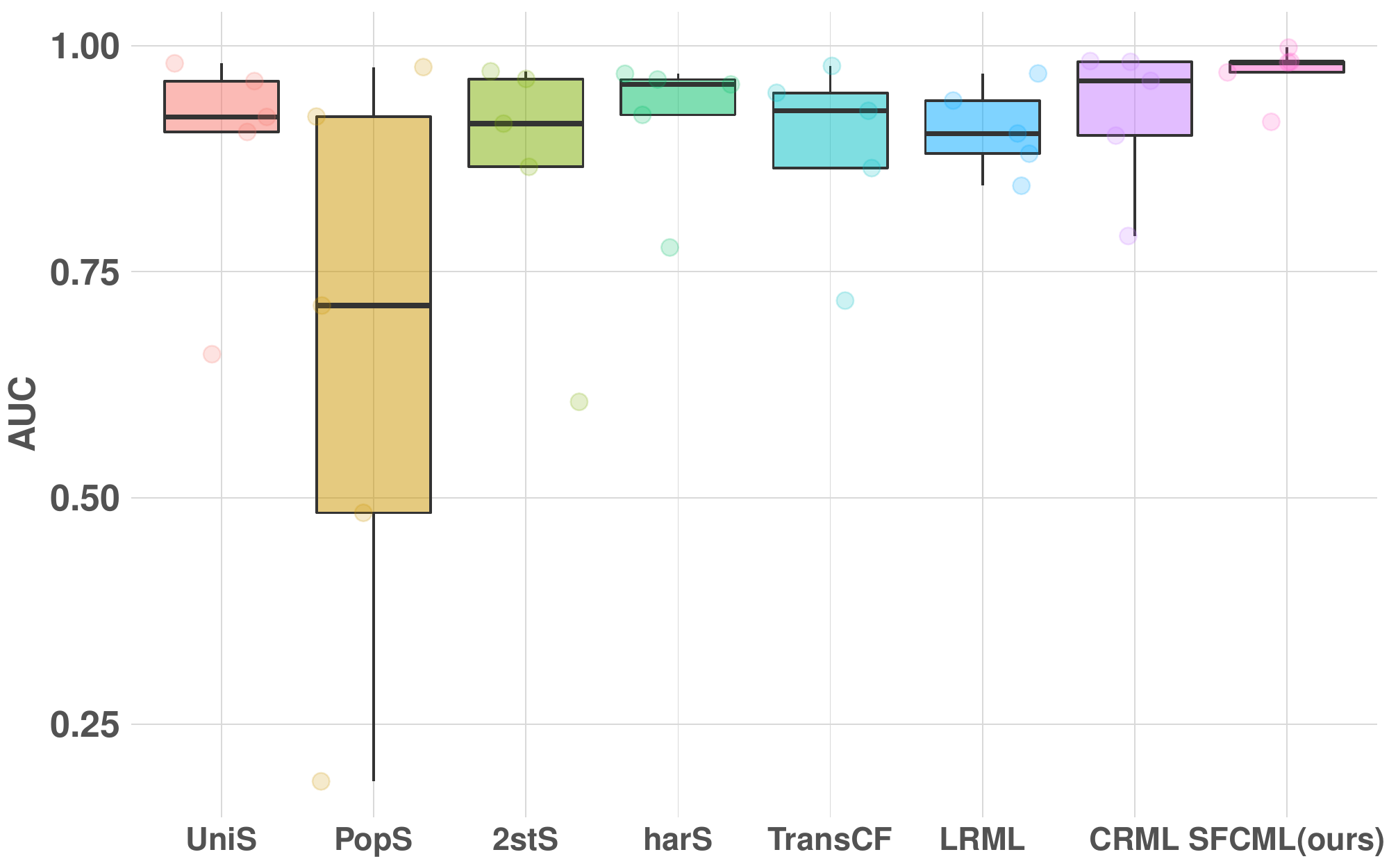}
			\label{main_ml-3-4}
		}
		\subfigure[MovieLens-1m-5]{
			\includegraphics[width=0.235\columnwidth]{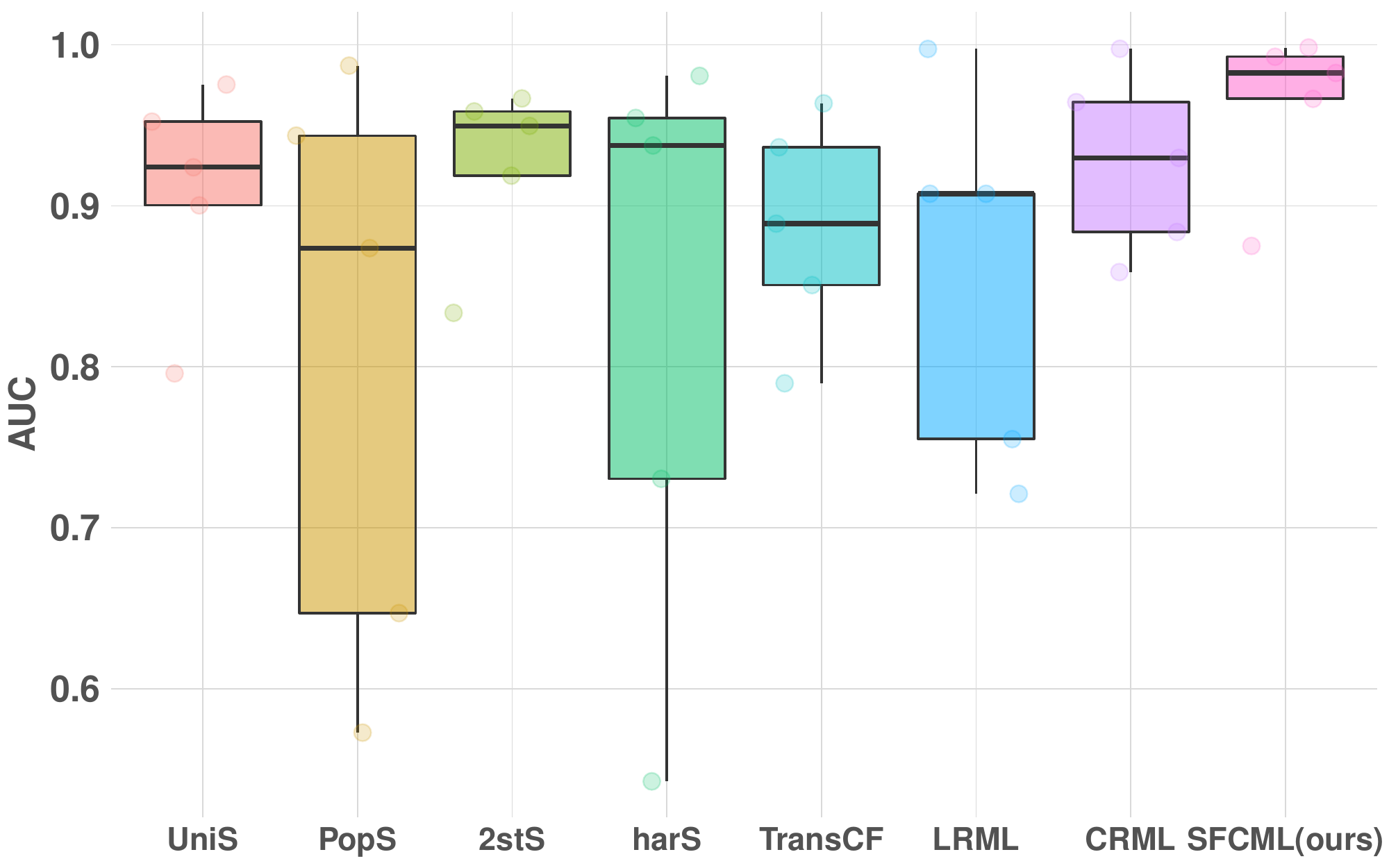}
			\label{ml-2}
		}
		\subfigure[MovieLens-1m-6]{
			\includegraphics[width=0.235\columnwidth]{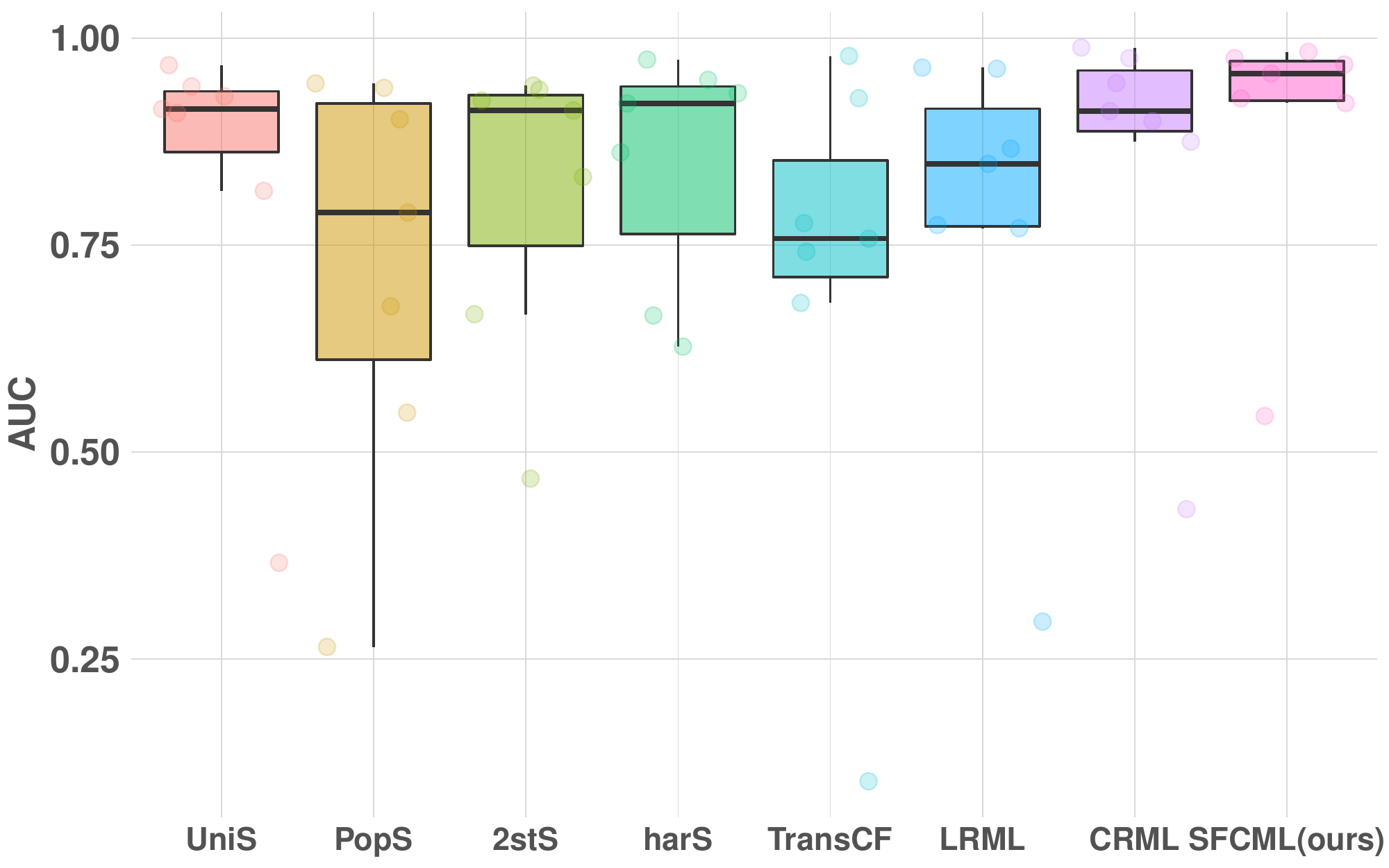}
			\label{ml-3}
		}
		\caption{Fine-grained AUC performance in terms of four users on MovieLens-1m dataset. }
		\label{auc:app_ml-1m}
	  
\end{figure*}

\begin{figure*}[h]
	\centering
	  
			\subfigure[Anime-3]{
			\includegraphics[width=0.235\columnwidth]{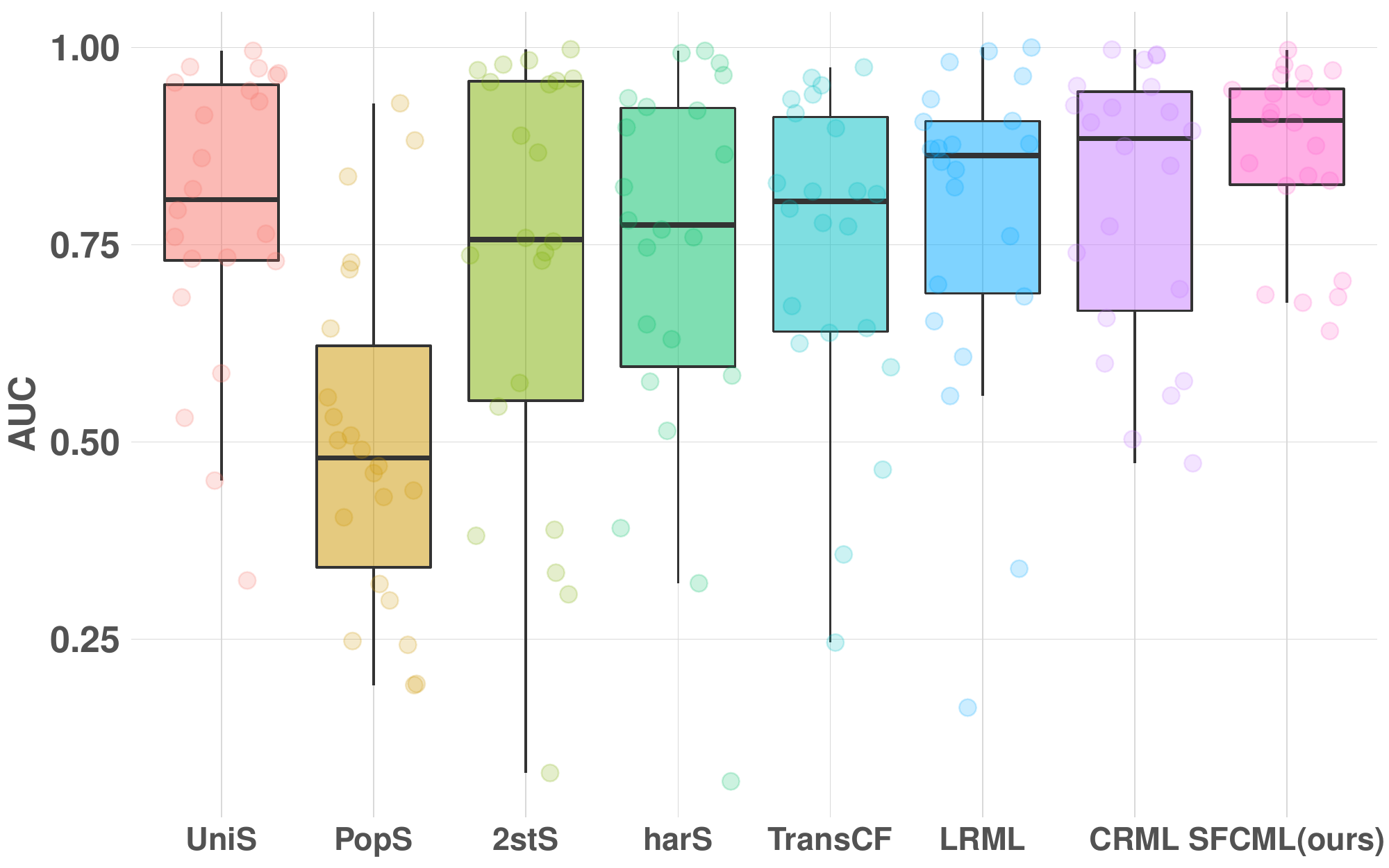}
			\label{Anime-3}
		}
		\subfigure[Anime-4]{
			\includegraphics[width=0.235\columnwidth]{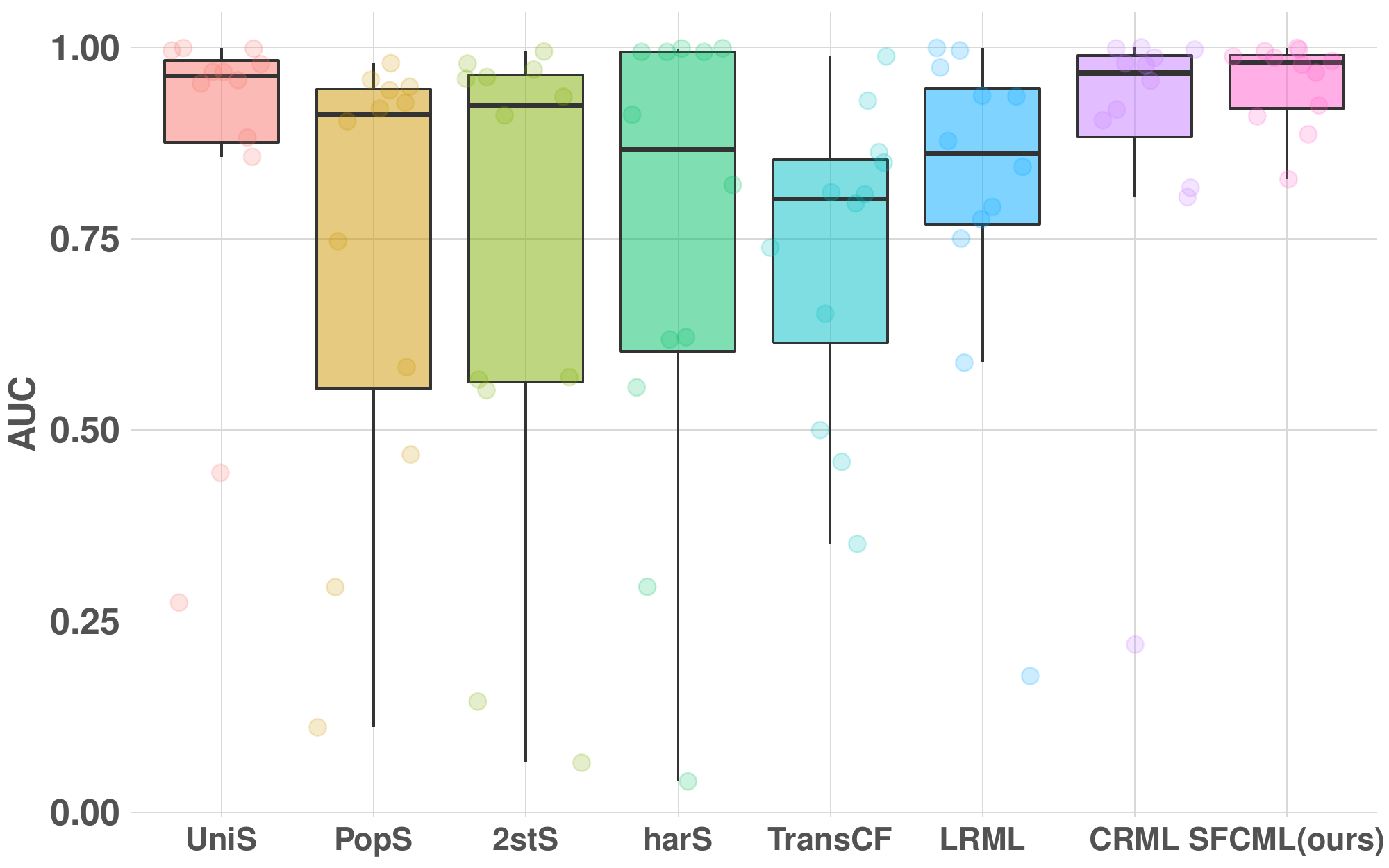}
			\label{Anime-4}
		}
		\subfigure[Anime-5]{
			\includegraphics[width=0.235\columnwidth]{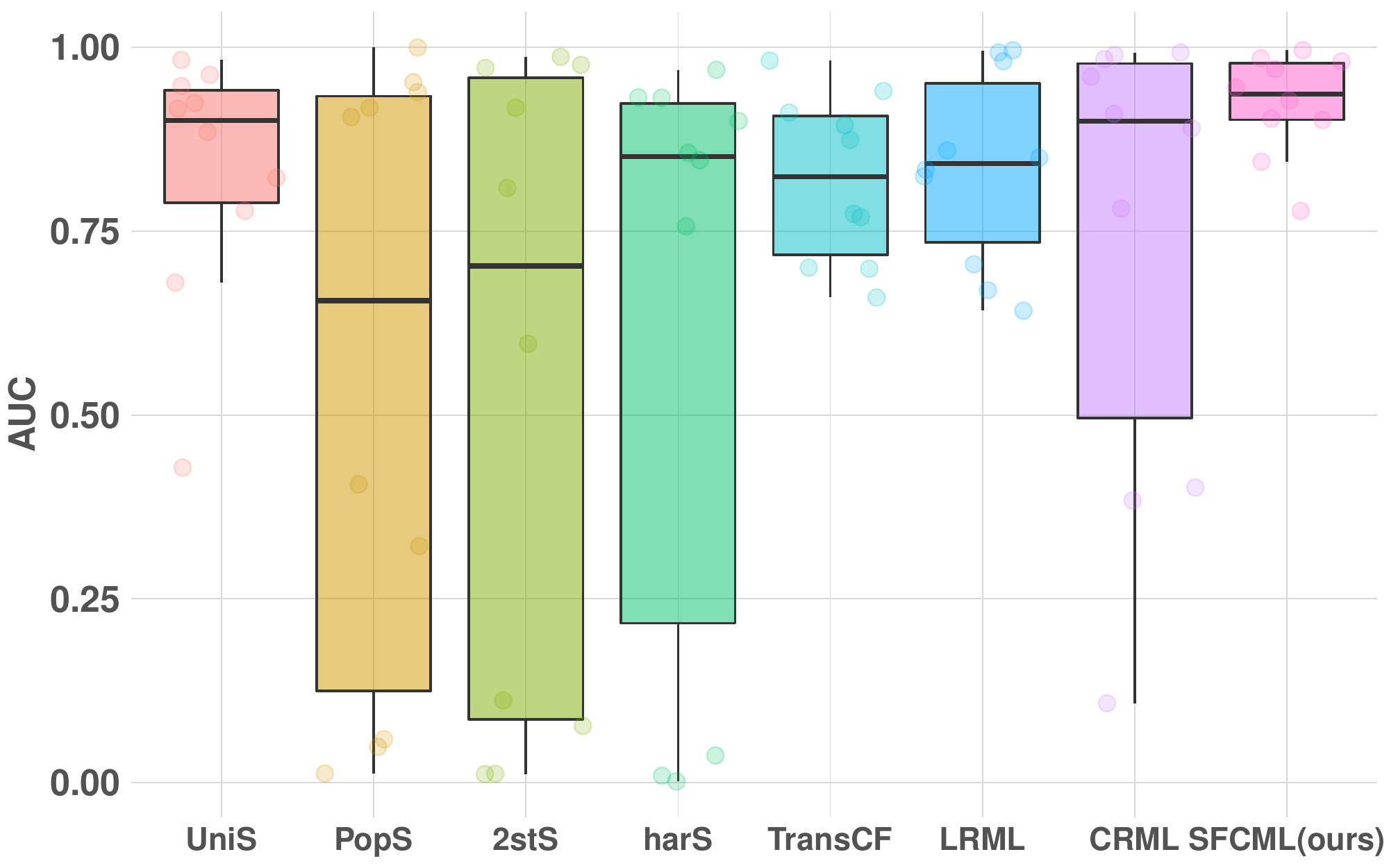}
			\label{Anime-5}
		}
		\subfigure[Anime-6]{
			\includegraphics[width=0.235\columnwidth]{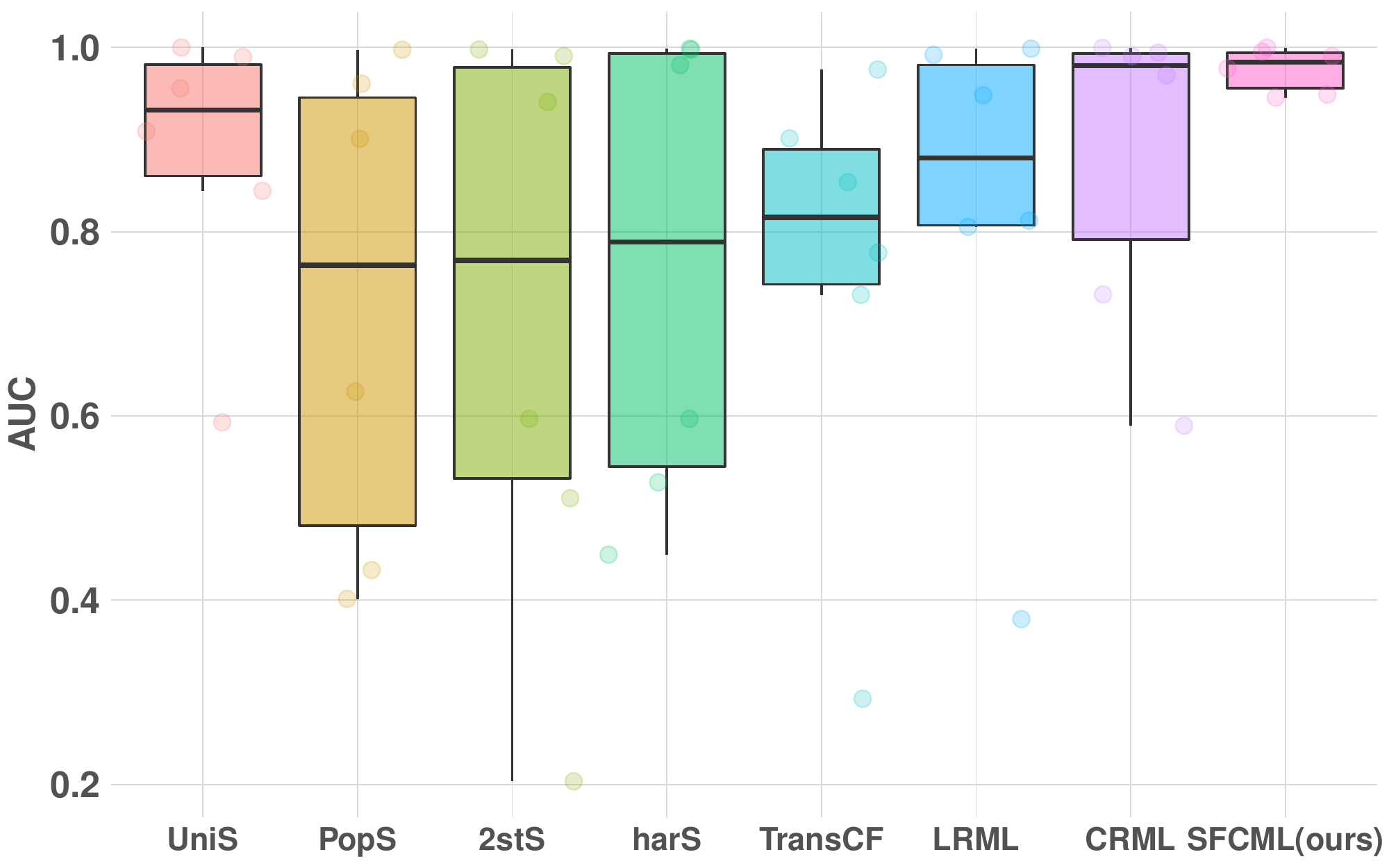}
			\label{Anime-6}
		}
		\caption{Fine-grained AUC performance with respect to four users on Anime dataset.}
		\label{auc:app_Anime}
	  
\end{figure*}

Please see Fig.\ref{auc:app_ml-1m} and Fig.\ref{auc:app_Anime}.

\subsubsection{Visualization of score density}
\begin{figure*}[!h]
	\centering
	 
		\subfigure[Anime-1]{
			\includegraphics[width=0.31\columnwidth]{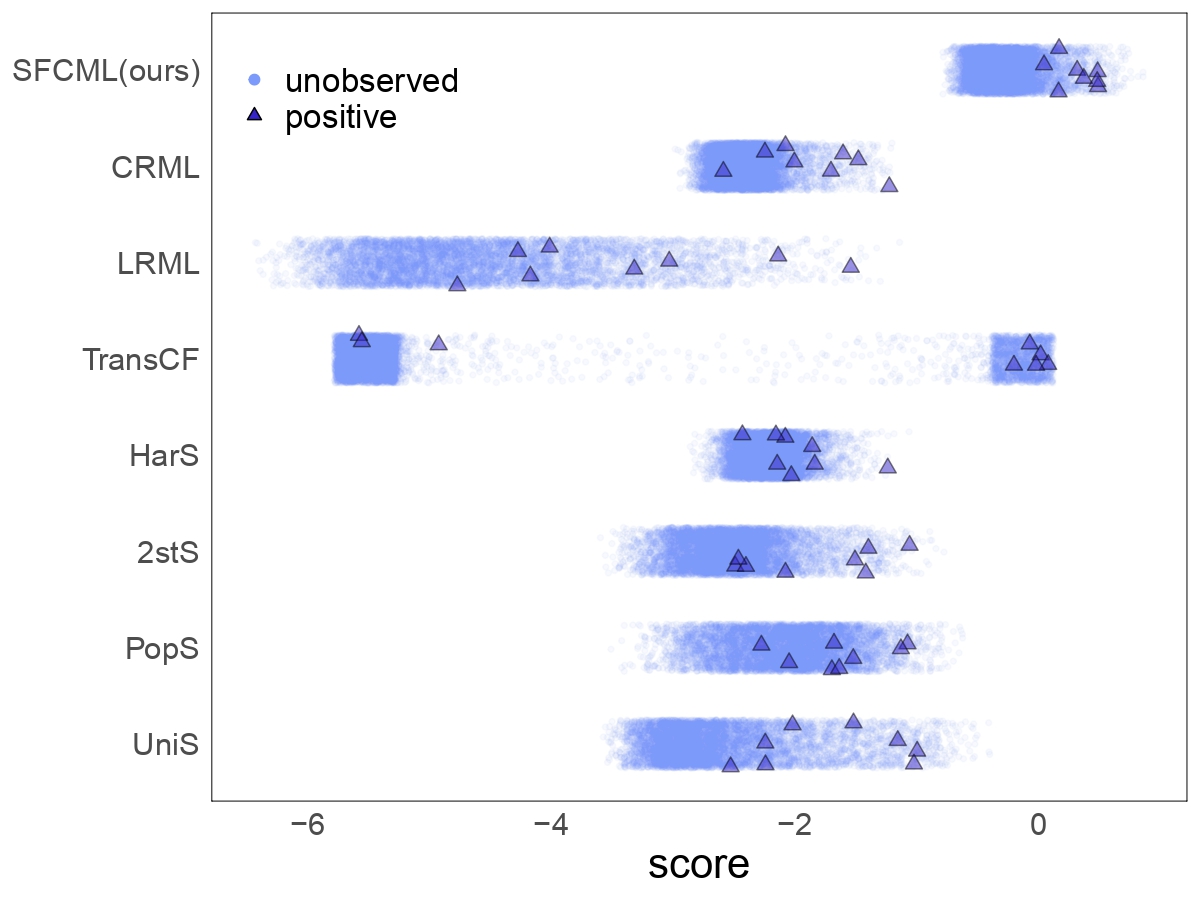}
			\label{Anime_1}
		}
		\subfigure[Anime-2]{
			\includegraphics[width=0.31\columnwidth]{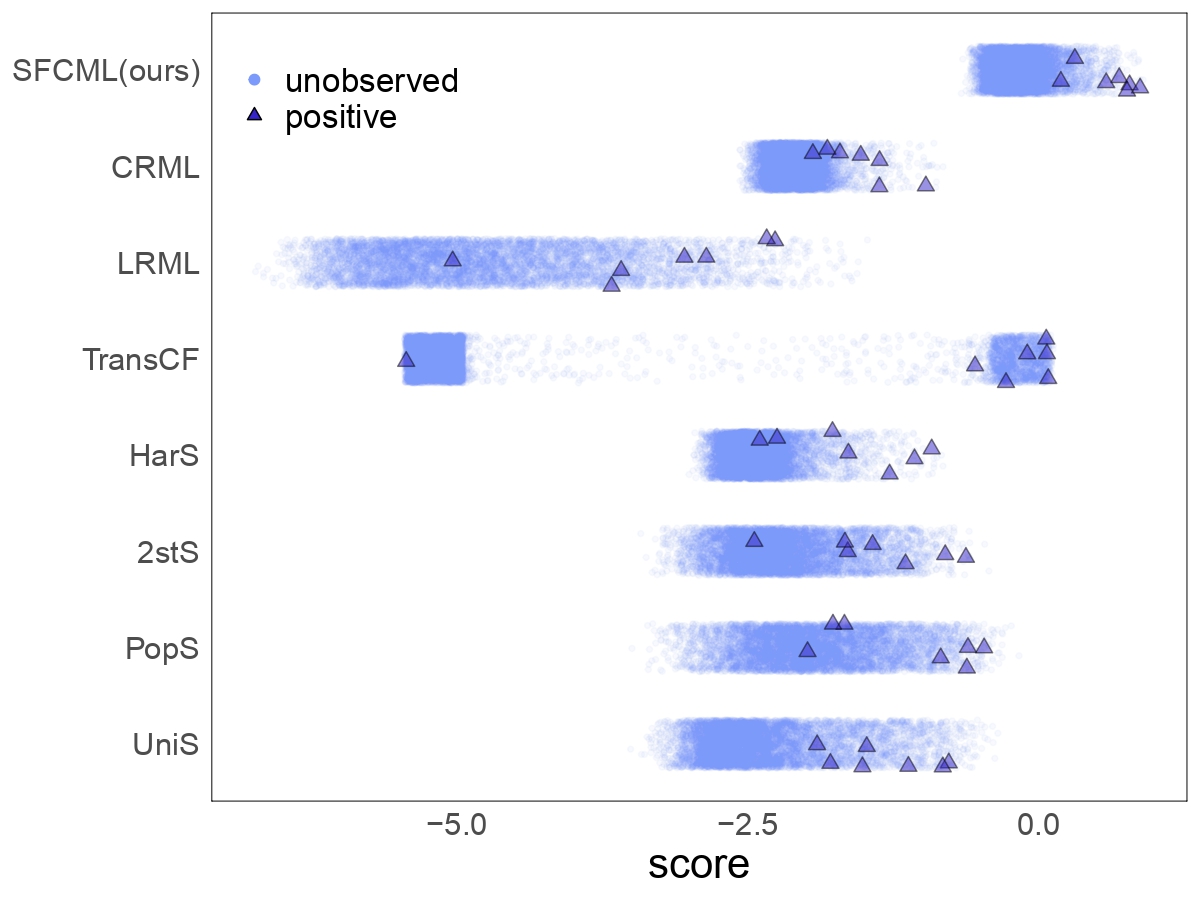}
			\label{Anime_2}
		}
		\subfigure[Anime-3]{
			\includegraphics[width=0.31\columnwidth]{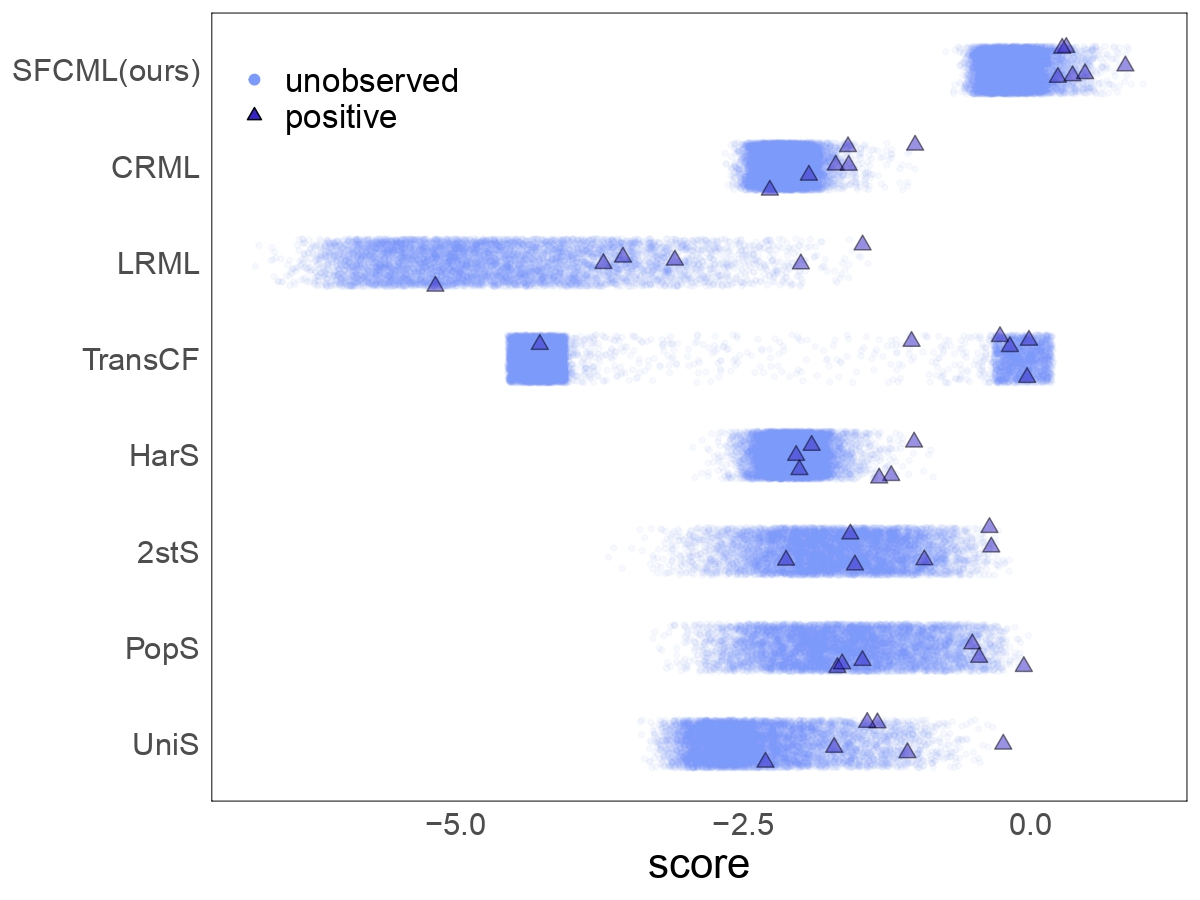}
			\label{Anime_3}
		}
%		\subfigure[Anime-3]{
%			\includegraphics[width=0.32\textwidth]{major_revision/density/Anime/user_32693}
%			\label{Anime_3}
%		}
%	\subfigure[Anime-3]{
%		\includegraphics[width=0.32\textwidth]{major_revision/density/Anime/user_38232}
%		\label{Anime_3}
%	}
		\caption{The graphical visualization of score distribution of positive and unobserved items on Anime.}
		\label{jetter_Anime}
	  
\end{figure*}

\clearpage
\newpage
\subsection{Sensitivity analysis of preference thresholds}\label{app_pre}
%  
%To figure out the rationality and the influence of $t$, we conduct the sensitive analysis on ml-100k data with different preference threshold $t \in \{1, 2, 3, 4, 5\}$. More results are reported in Tab.\ref{a}-\ref{n}.
%  

\begin{minipage}{\textwidth}
	\begin{minipage}[h]{0.5\textwidth}
		\centering
		\makeatletter\def\@captype{table}\makeatother\caption{The empirical results of P@3 with respect to different preference thresholds $t$ on MovieLens-100k. The best and second-best are highlighted in bold and underlined, respectively.}
		\label{a}
		\scalebox{0.8}{
\begin{tabular}{c|c|ccccc}
			\toprule
			Method & Trend & \cellcolor[rgb]{ 1.0,  1.0,  1.0} t=1 & t=2 & t=3 & t=4 & t=5 \\
			\midrule
			itemKNN &  \begin{minipage}[!t]{0.1\columnwidth}
				\centering
				\raisebox{-.5\height}{\includegraphics[width=\textwidth, height=0.3\textwidth]{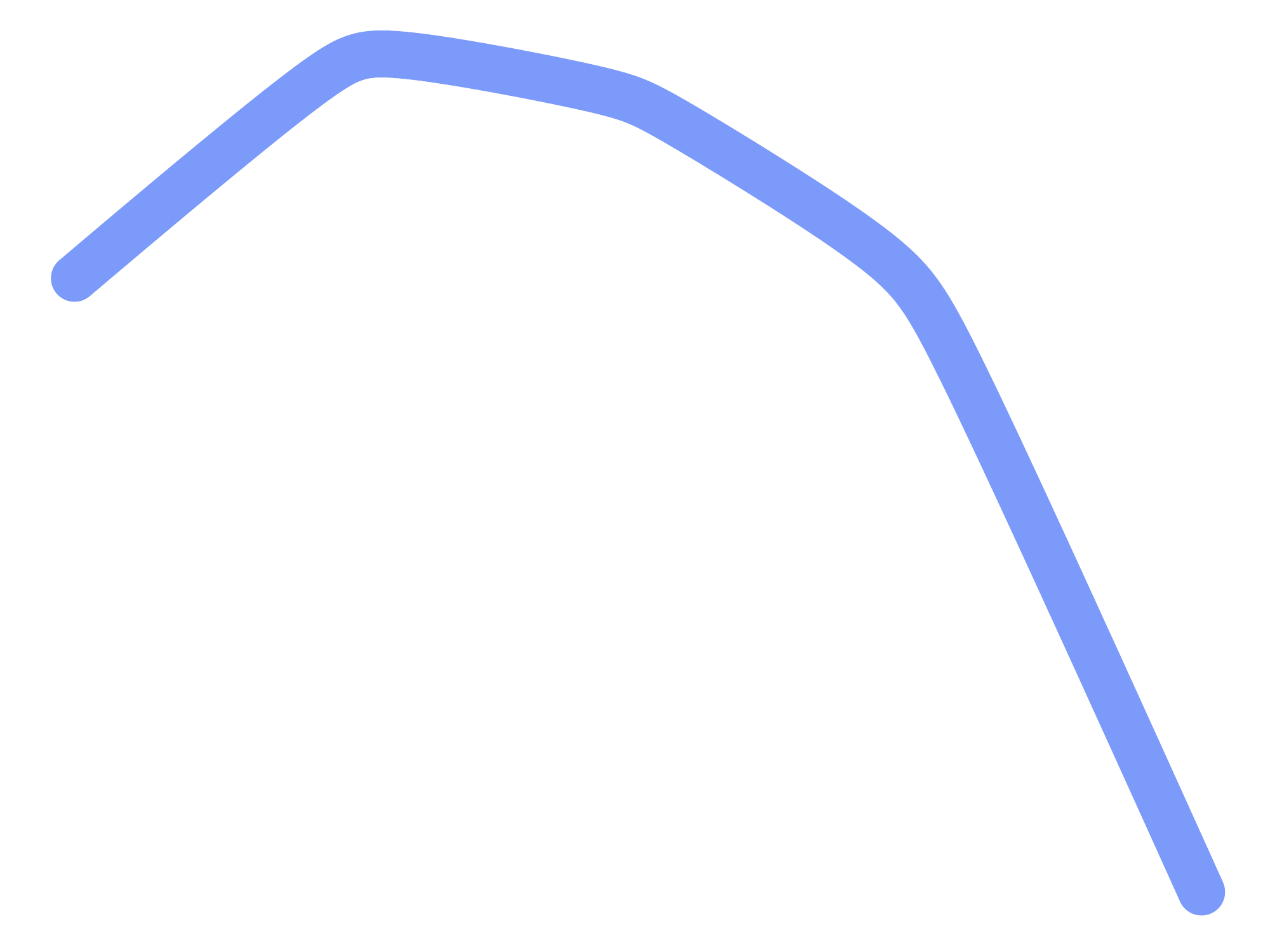}}
			\end{minipage} & \cellcolor[rgb]{ 1.0,  1.0,  1.0} 11.63 & 15.34 & 14.59 & 11.35 & 1.32 \\
			GMF & \begin{minipage}[!t]{0.1\columnwidth}
				\centering
				\raisebox{-.5\height}{\includegraphics[width=\textwidth, height=0.3\textwidth]{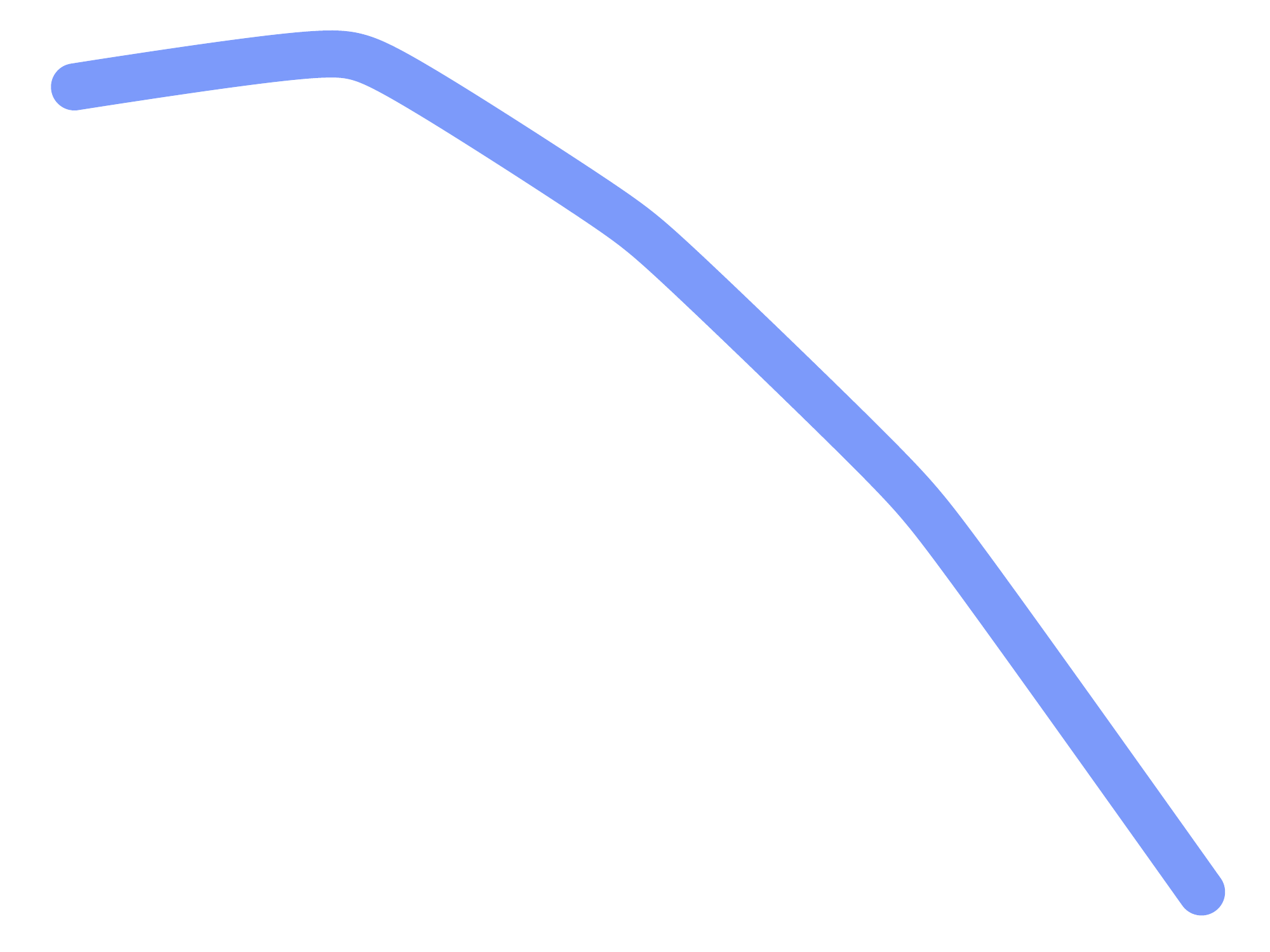}}
			\end{minipage}  & \cellcolor[rgb]{ 1,  .988,  .984}18.77 & \cellcolor[rgb]{ 1,  .988,  .984}19.09 & \cellcolor[rgb]{ .996,  .949,  .925}17.25 & \cellcolor[rgb]{ .996,  .949,  .925}14.35 & \cellcolor[rgb]{ .988,  .925,  .882}10.23  \\
			MLP & \begin{minipage}[!t]{0.1\columnwidth}
				\centering
				\raisebox{-.5\height}{\includegraphics[width=\textwidth, height=0.3\textwidth]{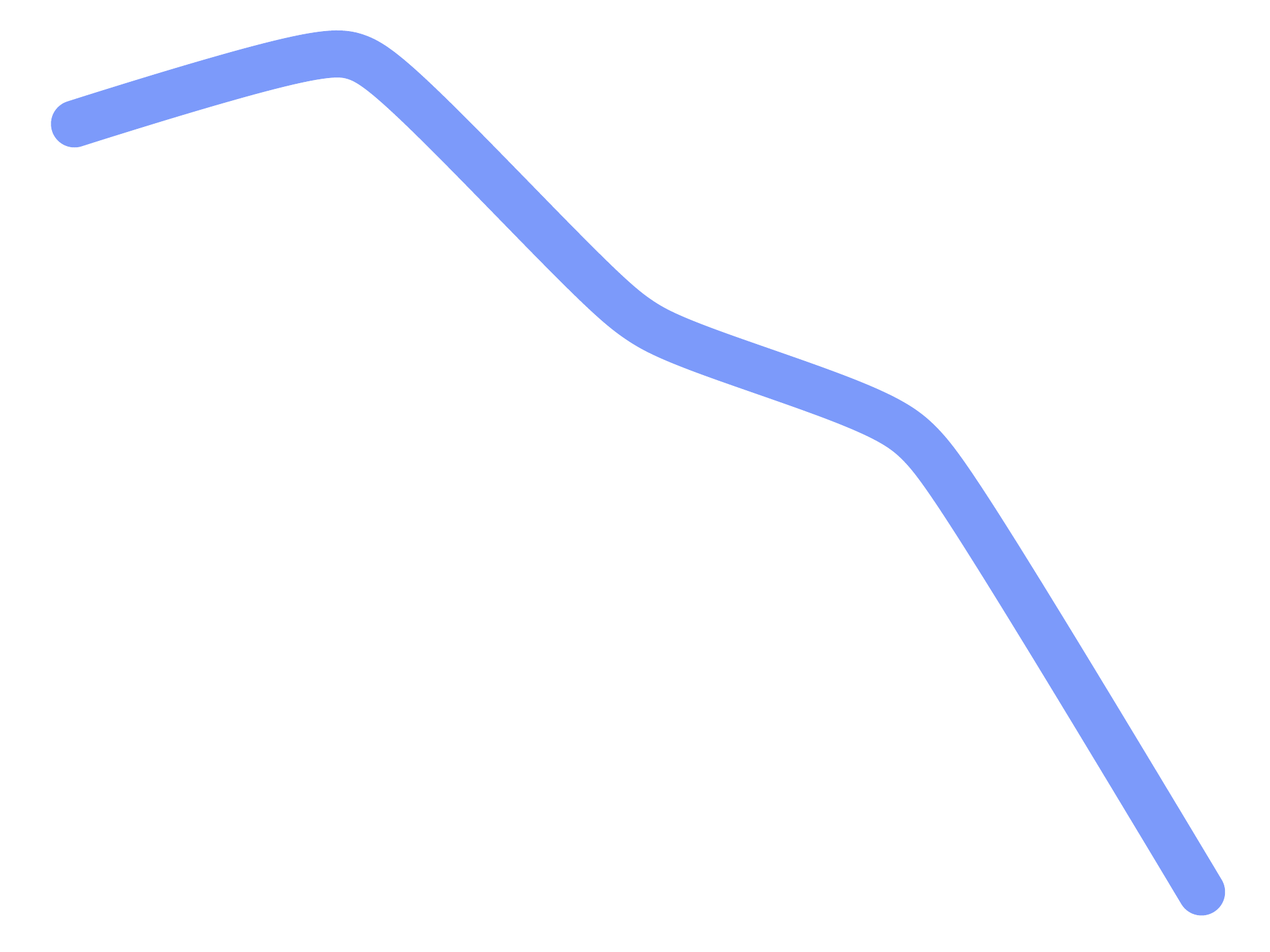}}
			\end{minipage}  & \cellcolor[rgb]{ 1,  .988,  .984}18.52 & \cellcolor[rgb]{ 1,  .988,  .984}19.26 & \cellcolor[rgb]{ .996,  .949,  .925}16.36 & \cellcolor[rgb]{ .996,  .949,  .925}14.98 & \cellcolor[rgb]{ .988,  .925,  .882}10.02  \\
			NCF & \begin{minipage}[!t]{0.1\columnwidth}
				\centering
				\raisebox{-.5\height}{\includegraphics[width=\textwidth, height=0.3\textwidth]{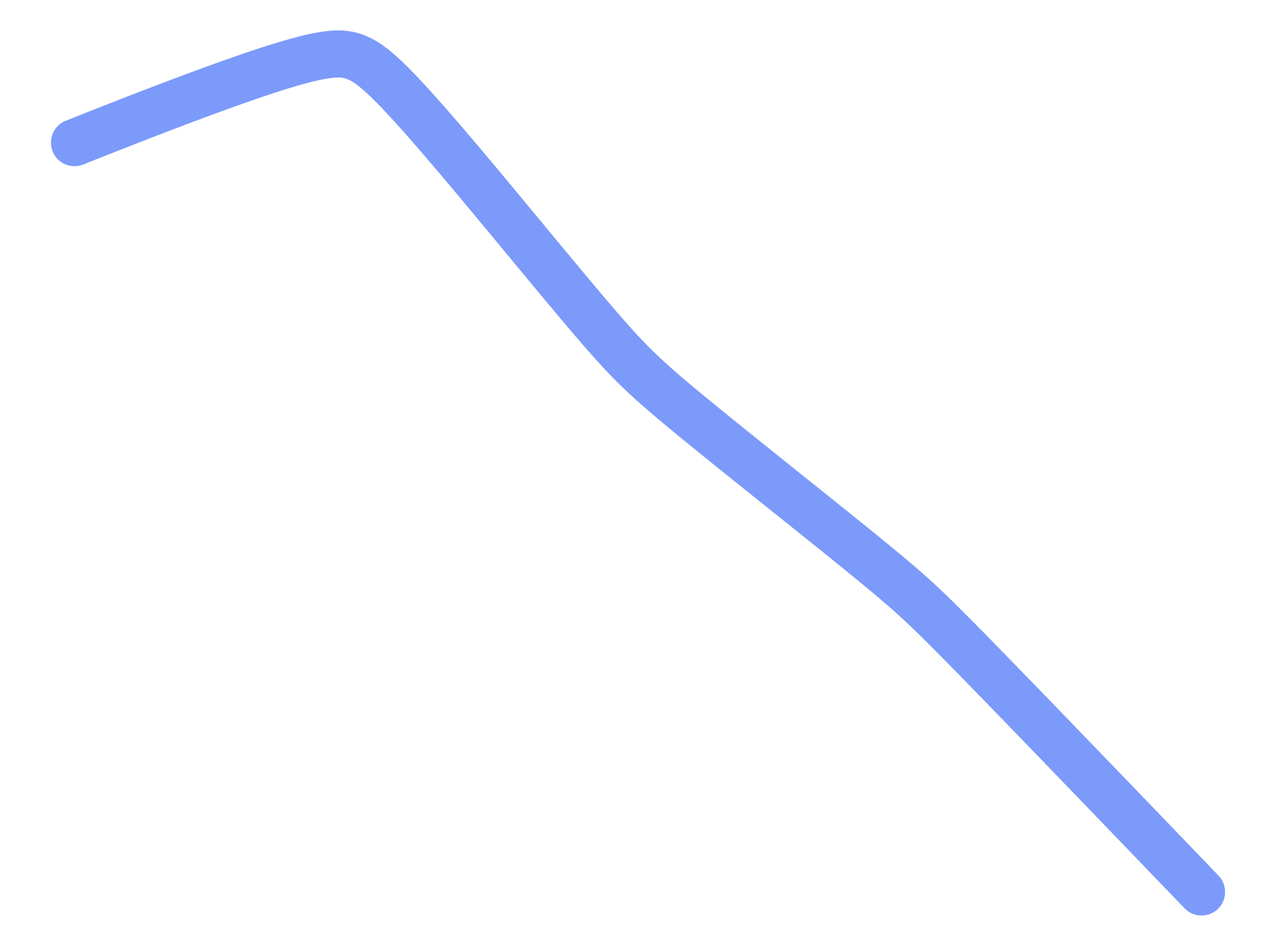}}
			\end{minipage}  & \cellcolor[rgb]{ .988,  .925,  .882}25.27  & \cellcolor[rgb]{ .996,  .89,  .839}27.00 & \cellcolor[rgb]{ .988,  .925,  .882}20.67  & \cellcolor[rgb]{ .988,  .925,  .882}15.94  & \cellcolor[rgb]{ .988,  .925,  .882}10.07  \\
\bottomrule
		\end{tabular}%
	}
	\end{minipage}
	\hspace{0.05cm}
	\begin{minipage}[h]{0.5\textwidth}
		\centering
		\makeatletter\def\@captype{table}\makeatother\caption{The empirical results of R@3 with respect to different preference thresholds $t$ on MovieLens-100k. The best and second-best are highlighted in bold and underlined, respectively.}
		\label{b}   
		\scalebox{0.8}{
			\begin{tabular}{c|c|ccccc}
			\toprule
			Method & Trend & \cellcolor[rgb]{ 1.0,  1.0,  1.0} t=1 & t=2 & t=3 & t=4 & t=5 \\
			\midrule
			itemKNN & \begin{minipage}[!t]{0.1\columnwidth}
				\centering
				\raisebox{-.5\height}{\includegraphics[width=\textwidth, height=0.3\textwidth]{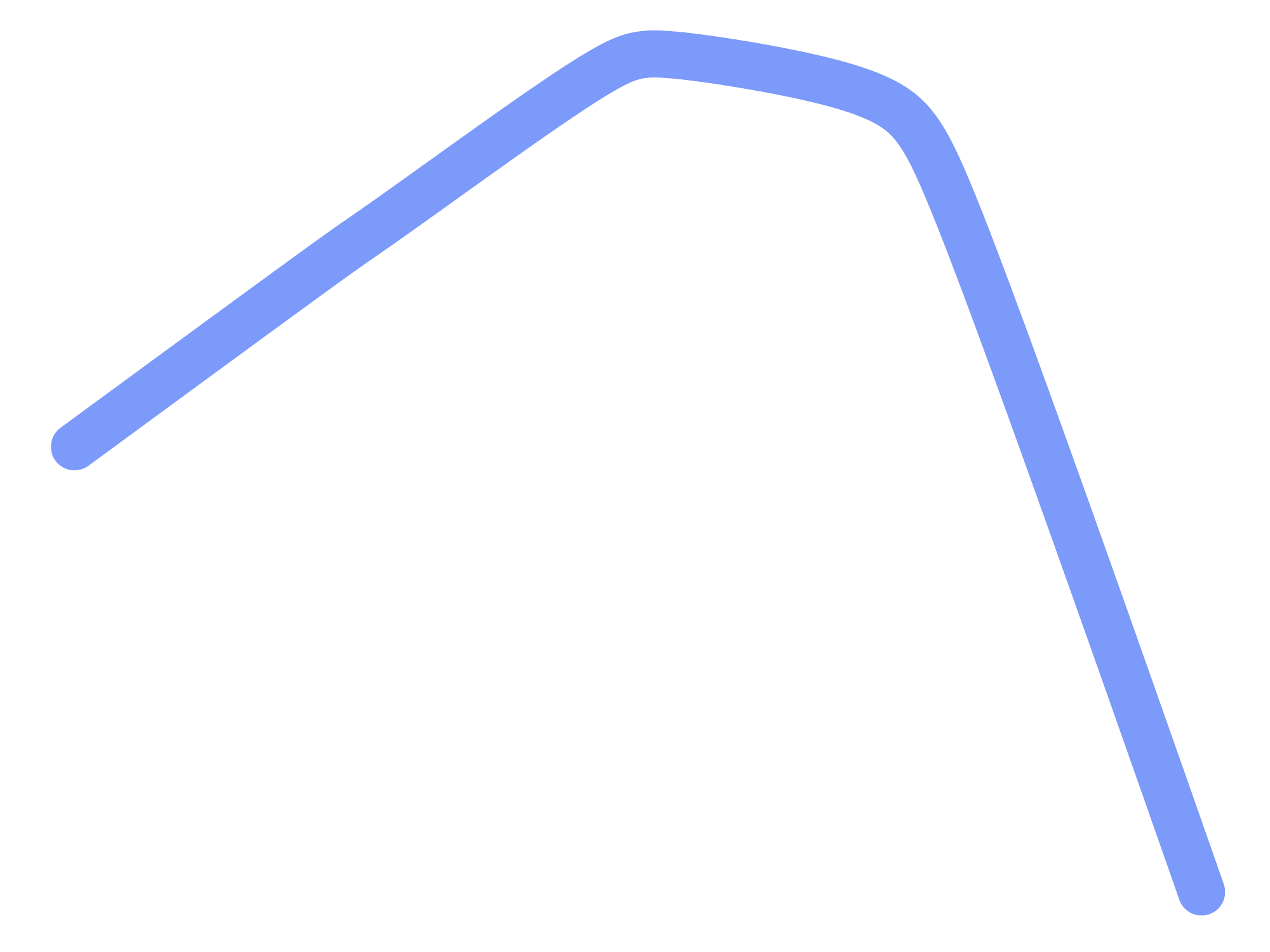}}
			\end{minipage}  & \cellcolor[rgb]{ 1.0,  1.0,  1.0} 1.77 & 2.19 & 2.57 & 2.41 & 0.86 \\
			GMF & \begin{minipage}[!t]{0.1\columnwidth}
				\centering
				\raisebox{-.5\height}{\includegraphics[width=\textwidth, height=0.3\textwidth]{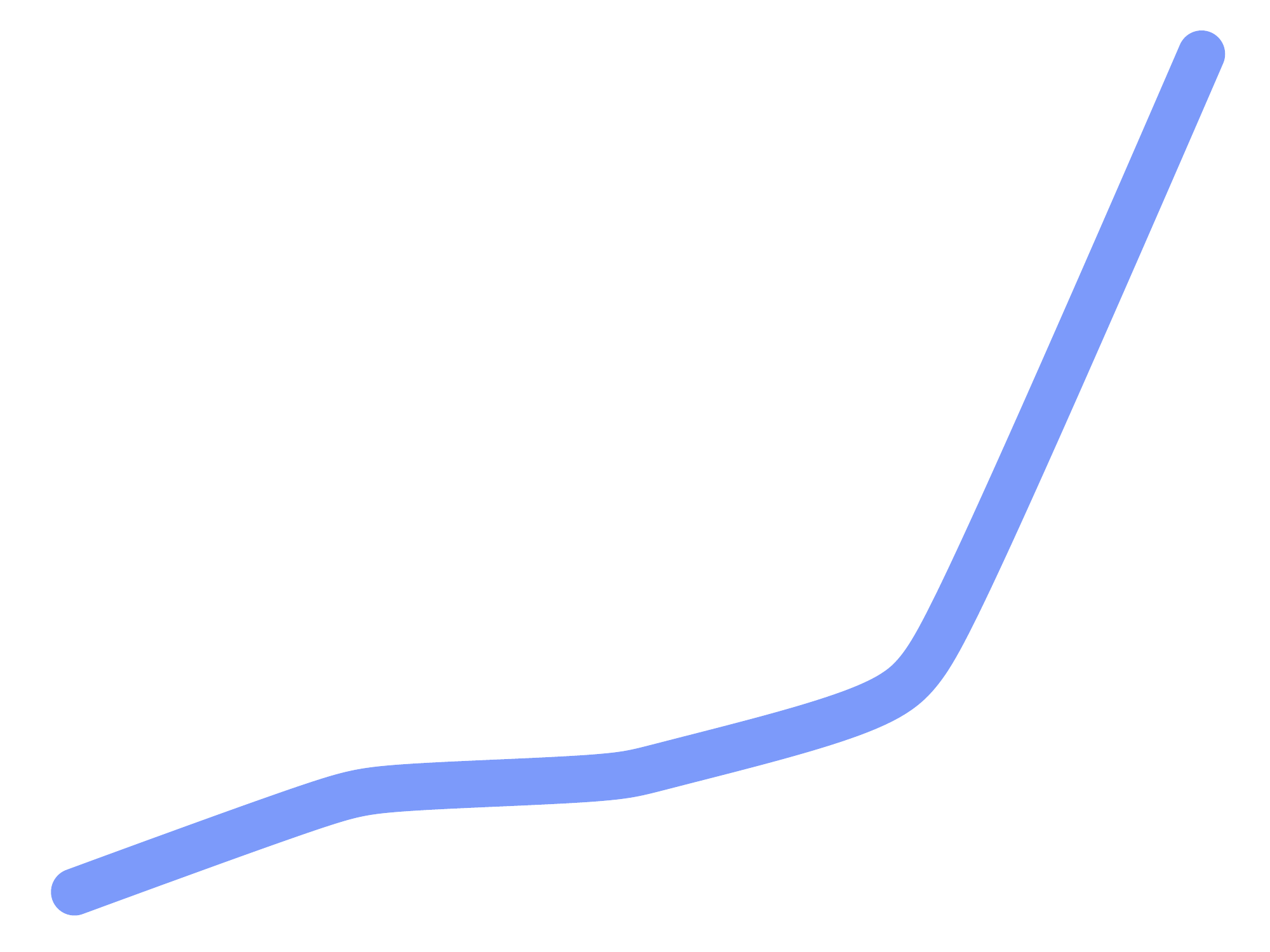}}
			\end{minipage}  & \cellcolor[rgb]{ 1.0,  1.0,  1.0} 2.85 & 3.08 & 3.13 & \cellcolor[rgb]{ .996,  .949,  .925}3.37 & \cellcolor[rgb]{ .988,  .925,  .882}4.81  \\
			MLP & \begin{minipage}[!t]{0.1\columnwidth}
				\centering
				\raisebox{-.5\height}{\includegraphics[width=\textwidth, height=0.3\textwidth]{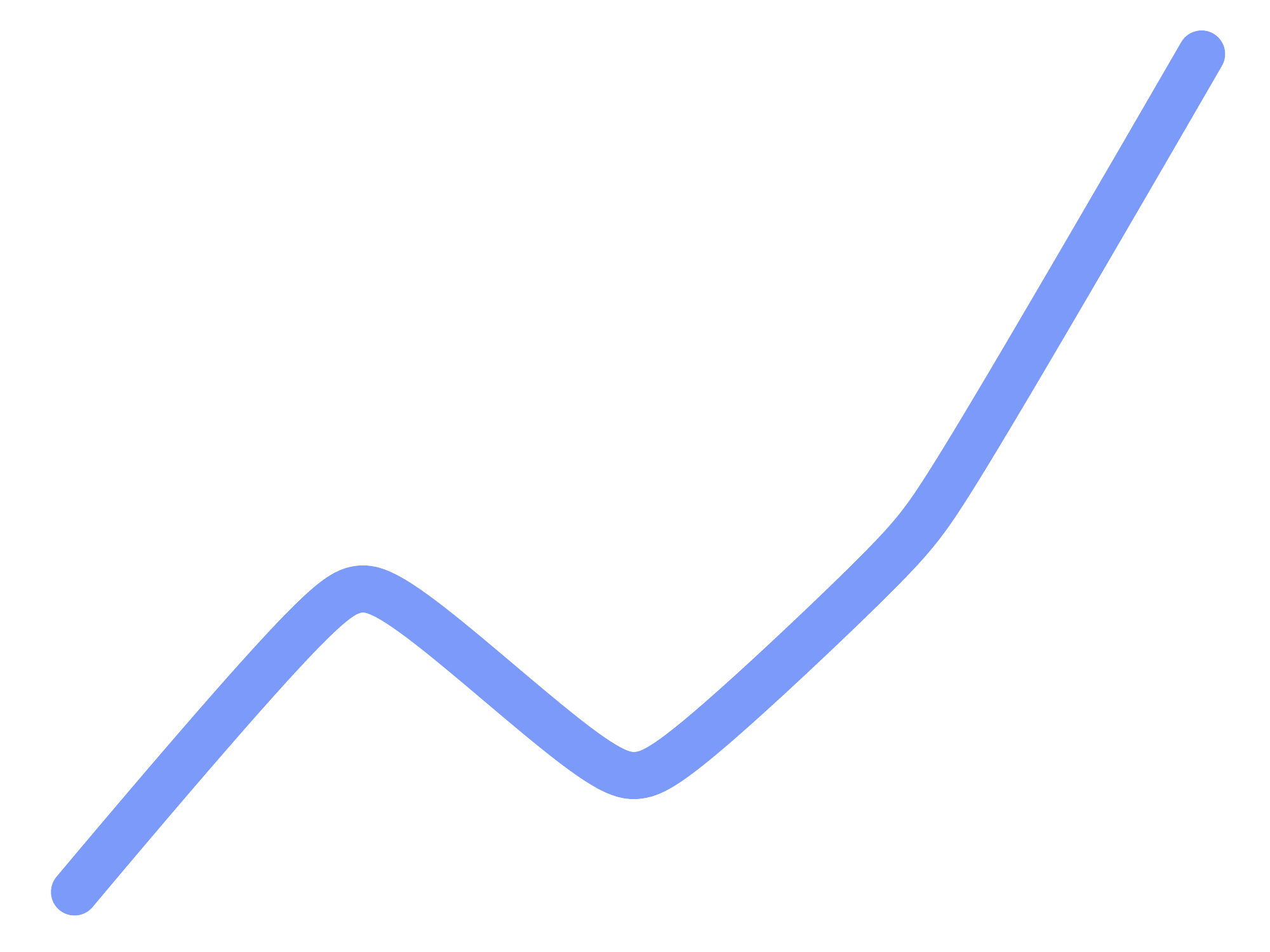}}
			\end{minipage}  & \cellcolor[rgb]{ .996,  .949,  .925}3.24 & \cellcolor[rgb]{ .996,  .949,  .925}3.81 & \cellcolor[rgb]{ .996,  .949,  .925}3.46 & \cellcolor[rgb]{ .996,  .949,  .925}3.93 & \cellcolor[rgb]{ .988,  .925,  .882}4.82 \\
			NCF &  \begin{minipage}[!t]{0.1\columnwidth}
				\centering
				\raisebox{-.5\height}{\includegraphics[width=\textwidth, height=0.3\textwidth]{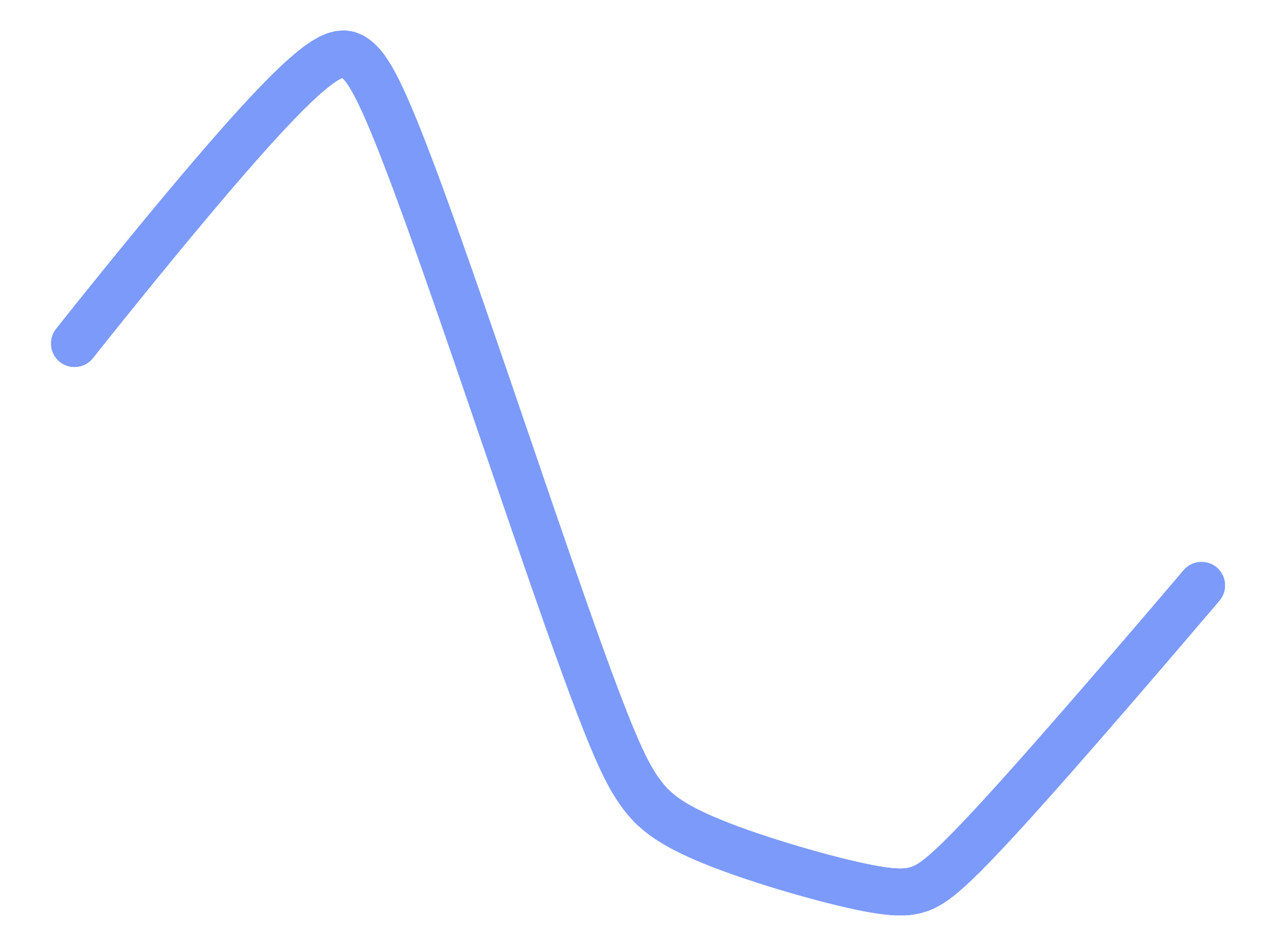}}
			\end{minipage} & \cellcolor[rgb]{ .996,  .89,  .839}5.26 & \cellcolor[rgb]{ .996,  .89,  .839}5.86 & \cellcolor[rgb]{ .988,  .925,  .882}4.32  & \cellcolor[rgb]{ .988,  .925,  .882}4.11 & \cellcolor[rgb]{ .988,  .925,  .882}4.75  \\
\bottomrule
		\end{tabular}%
	}
	\end{minipage}
\end{minipage}

\vspace{0.3cm}
\begin{minipage}{\textwidth}
\begin{minipage}[h]{0.5\textwidth}
	\centering
	\makeatletter\def\@captype{table}\makeatother\caption{The empirical results of NDCG@3 with respect to different preference thresholds $t$ on MovieLens-100k. The best and second-best are highlighted in bold and underlined, respectively.}
	\label{c}   
	\scalebox{0.8}{
		\begin{tabular}{c|c|ccccc}
			\toprule
			Method & Trend & \cellcolor[rgb]{ 1.0,  1.0,  1.0} t=1 & t=2 & t=3 & t=4 & t=5 \\
			\midrule
			itemKNN & \begin{minipage}[!t]{0.1\columnwidth}
				\centering
				\raisebox{-.5\height}{\includegraphics[width=\textwidth, height=0.3\textwidth]{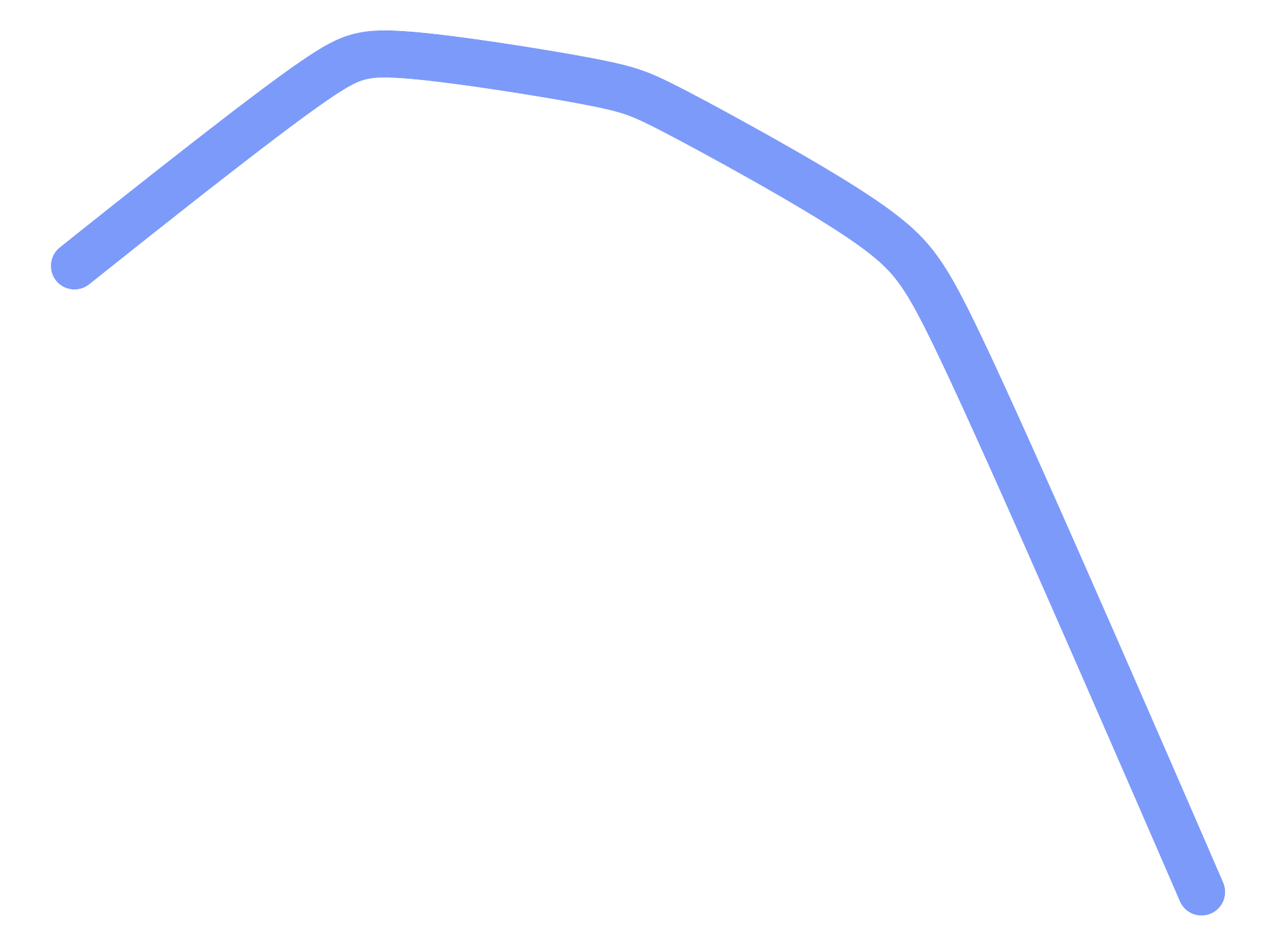}}
			\end{minipage}  &\cellcolor[rgb]{ 1.0,  1.0,  1.0}  11.67 & 15.11 & 14.51 & 11.57 & 1.33 \\
			GMF &  \begin{minipage}[!t]{0.1\columnwidth}
				\centering
				\raisebox{-.5\height}{\includegraphics[width=\textwidth, height=0.3\textwidth]{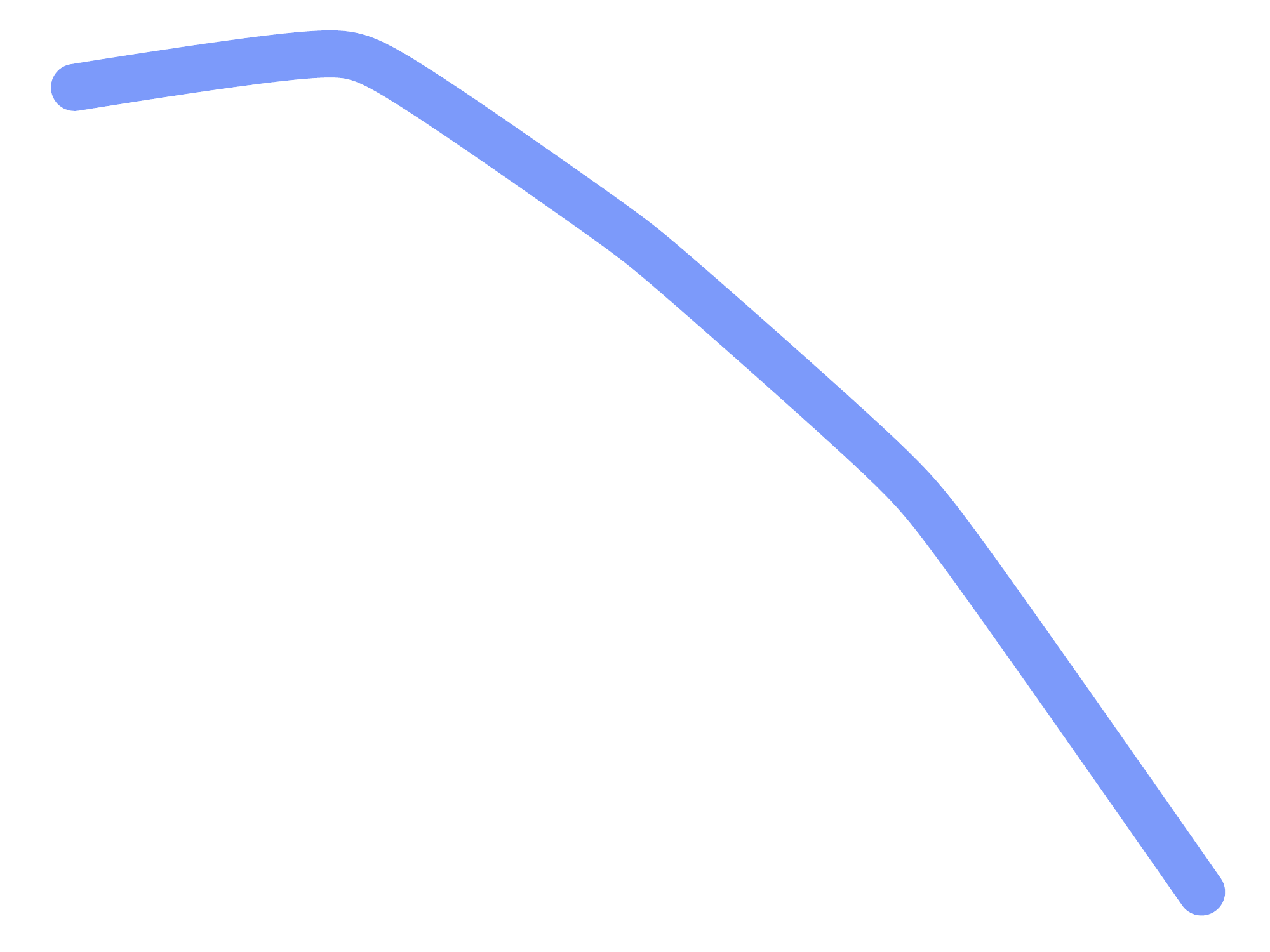}}
			\end{minipage} & \cellcolor[rgb]{ .984,  .992,  .976}19.79  & \cellcolor[rgb]{ .945,  .969,  .929}20.13  & \cellcolor[rgb]{ .984,  .992,  .976}18.05  & \cellcolor[rgb]{ .984,  .992,  .976}15.20 & \cellcolor[rgb]{ .914,  .953,  .89}10.79  \\
			MLP & \begin{minipage}[!t]{0.1\columnwidth}
				\centering
				\raisebox{-.5\height}{\includegraphics[width=\textwidth, height=0.3\textwidth]{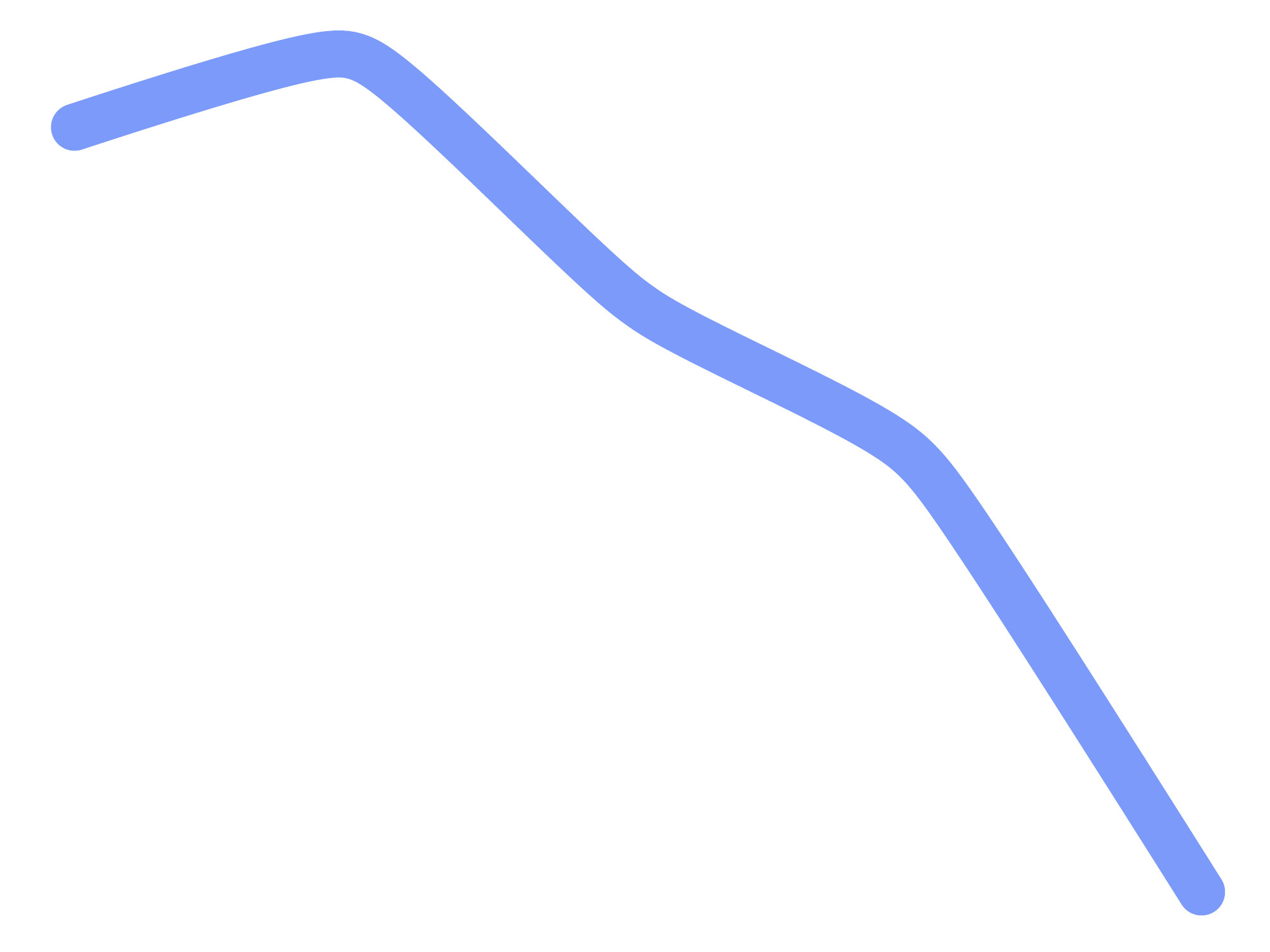}}
			\end{minipage}  & \cellcolor[rgb]{ .984,  .992,  .976}19.23  & \cellcolor[rgb]{ .945,  .969,  .929}20.00 & \cellcolor[rgb]{ .984,  .992,  .976}17.30 & \cellcolor[rgb]{ .984,  .992,  .976}15.57  & \cellcolor[rgb]{ .914,  .953,  .89}10.90 \\
			NCF &  \begin{minipage}[!t]{0.1\columnwidth}
				\centering
				\raisebox{-.5\height}{\includegraphics[width=\textwidth, height=0.3\textwidth]{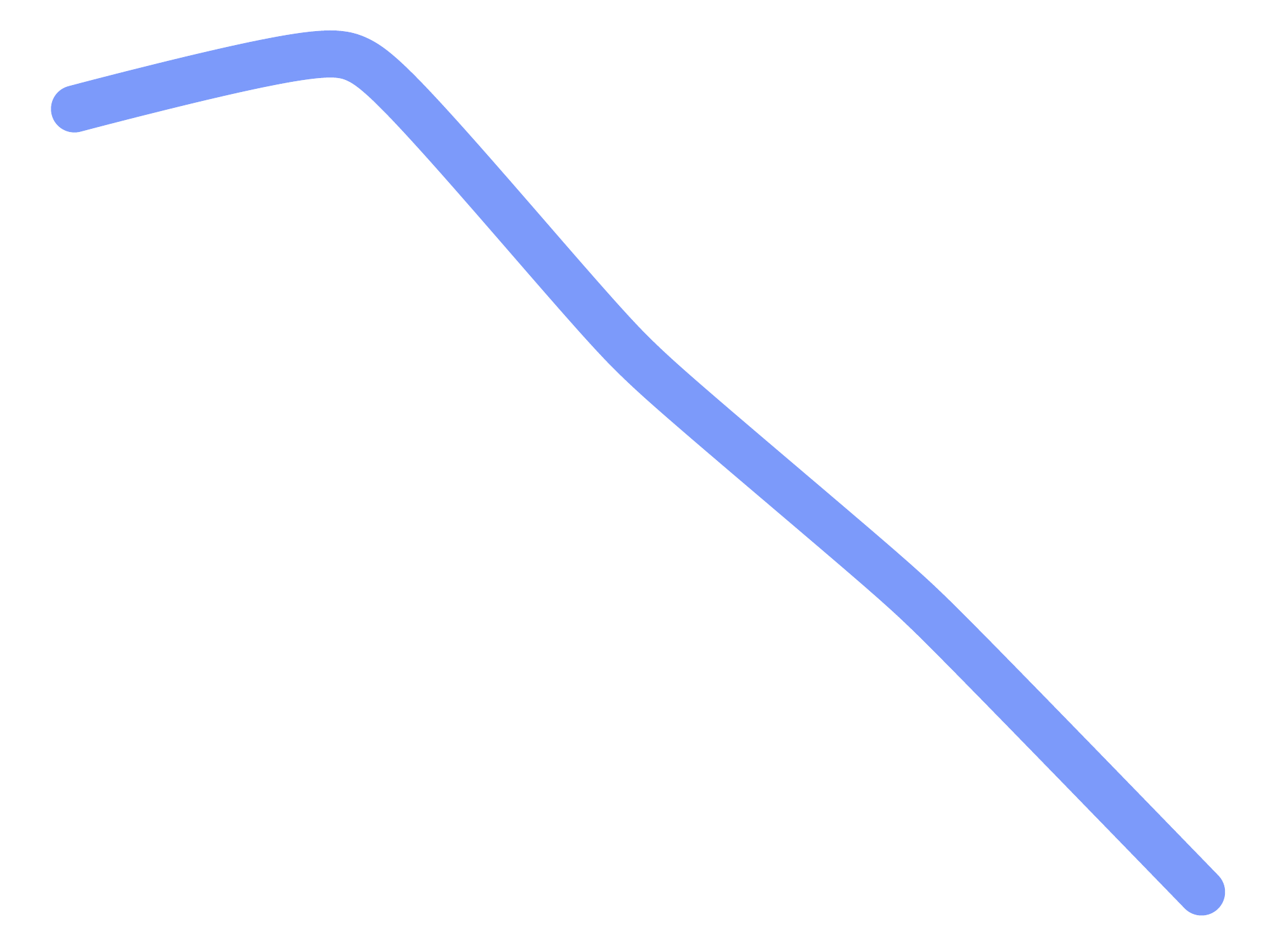}}
			\end{minipage} & \cellcolor[rgb]{ .914,  .953,  .89}26.66  & \cellcolor[rgb]{ .886,  .937,  .851}27.66  & \cellcolor[rgb]{ .945,  .969,  .929}21.65  & \cellcolor[rgb]{ .945,  .969,  .929}16.75  & \cellcolor[rgb]{ .886,  .937,  .851}11.00 \\
\bottomrule
		\end{tabular}%
	}
\end{minipage}
\hspace{0.05cm}
\begin{minipage}[h]{0.5\textwidth}
	\centering
	\makeatletter\def\@captype{table}\makeatother\caption{The empirical results of P@5 with respect to different preference thresholds $t$ on MovieLens-100k. The best and second-best are highlighted in bold and underlined, respectively.}
	\label{d}   
	\scalebox{0.8}{
		\begin{tabular}{c|c|ccccc}
			\toprule
			Method & \multicolumn{1}{c|}{Trend} & t=1 & t=2 & t=3 & t=4 & t=5 \\
			\midrule
			itemKNN & \begin{minipage}[!t]{0.1\columnwidth}
				\centering
				\raisebox{-.5\height}{\includegraphics[width=\textwidth, height=0.3\textwidth]{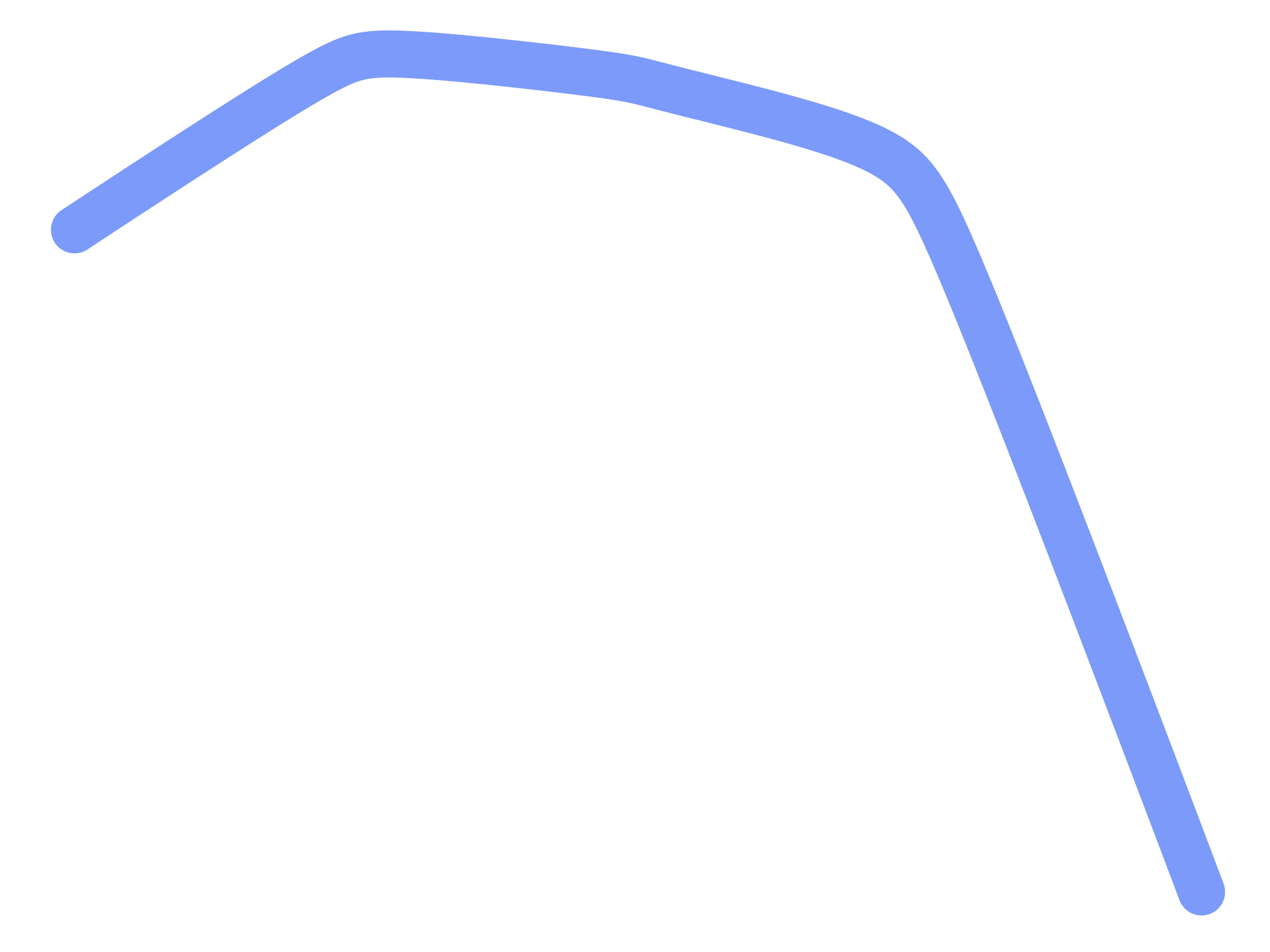}}
			\end{minipage}  & 12.25 & 15.08 & 14.70 & 12.96 & 1.38 \\
			GMF & \begin{minipage}[!t]{0.1\columnwidth}
				\centering
				\raisebox{-.5\height}{\includegraphics[width=\textwidth, height=0.3\textwidth]{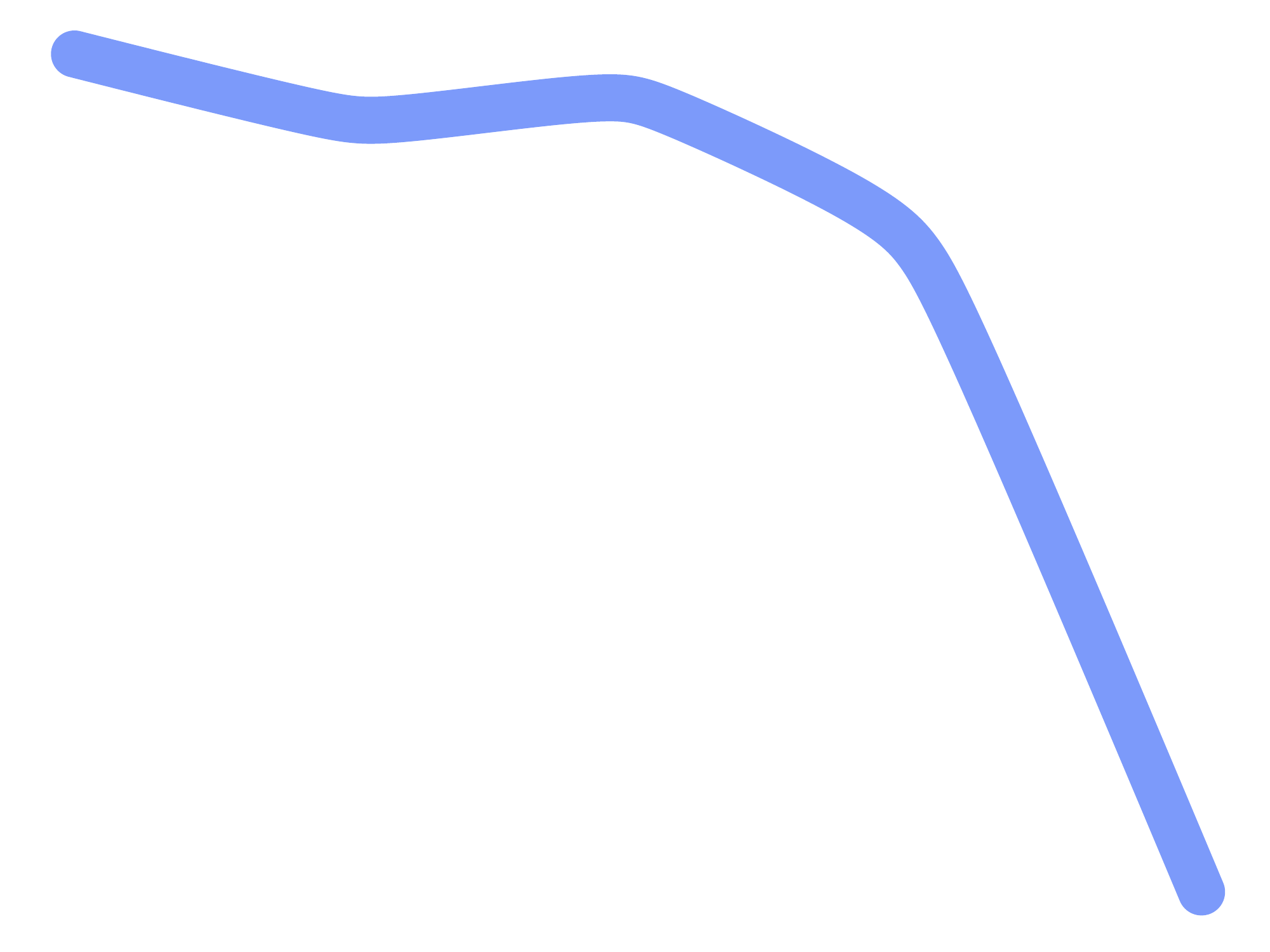}}
			\end{minipage}  & \cellcolor[rgb]{ .945,  .969,  .929}17.80 & \cellcolor[rgb]{ .984,  .992,  .976}17.35  & \cellcolor[rgb]{ .984,  .992,  .976}17.48  & \cellcolor[rgb]{ .984,  .992,  .976}16.43  & \cellcolor[rgb]{ .914,  .953,  .89}12.06  \\
			MLP & \begin{minipage}[!t]{0.1\columnwidth}
				\centering
				\raisebox{-.5\height}{\includegraphics[width=\textwidth, height=0.3\textwidth]{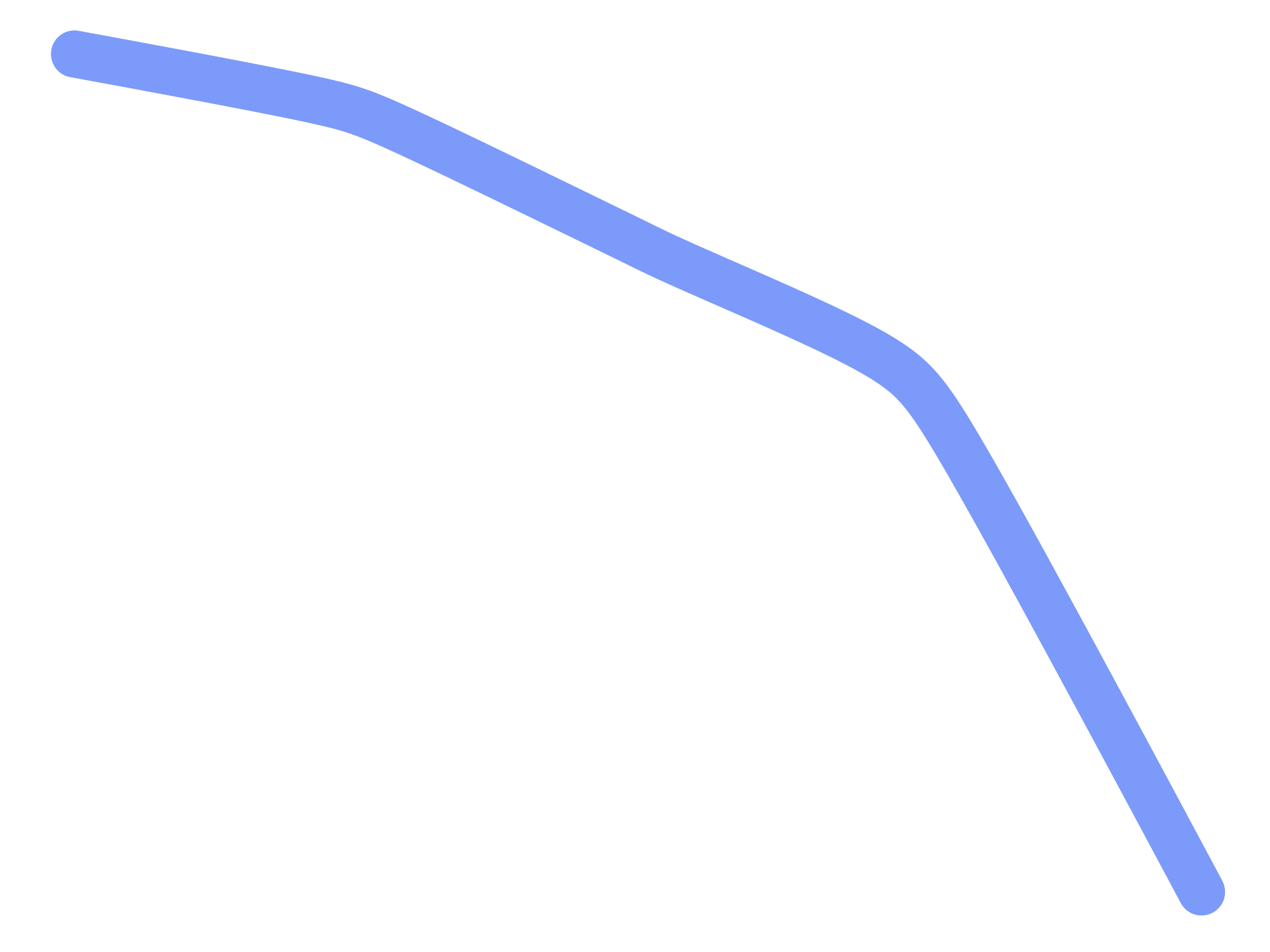}}
			\end{minipage}  & \cellcolor[rgb]{ .945,  .969,  .929}18.16  & \cellcolor[rgb]{ .984,  .992,  .976}17.71  & 16.66 & 15.51 & \cellcolor[rgb]{ .945,  .969,  .929}11.53 \\
			NCF & \begin{minipage}[!t]{0.1\columnwidth}
				\centering
				\raisebox{-.5\height}{\includegraphics[width=\textwidth, height=0.3\textwidth]{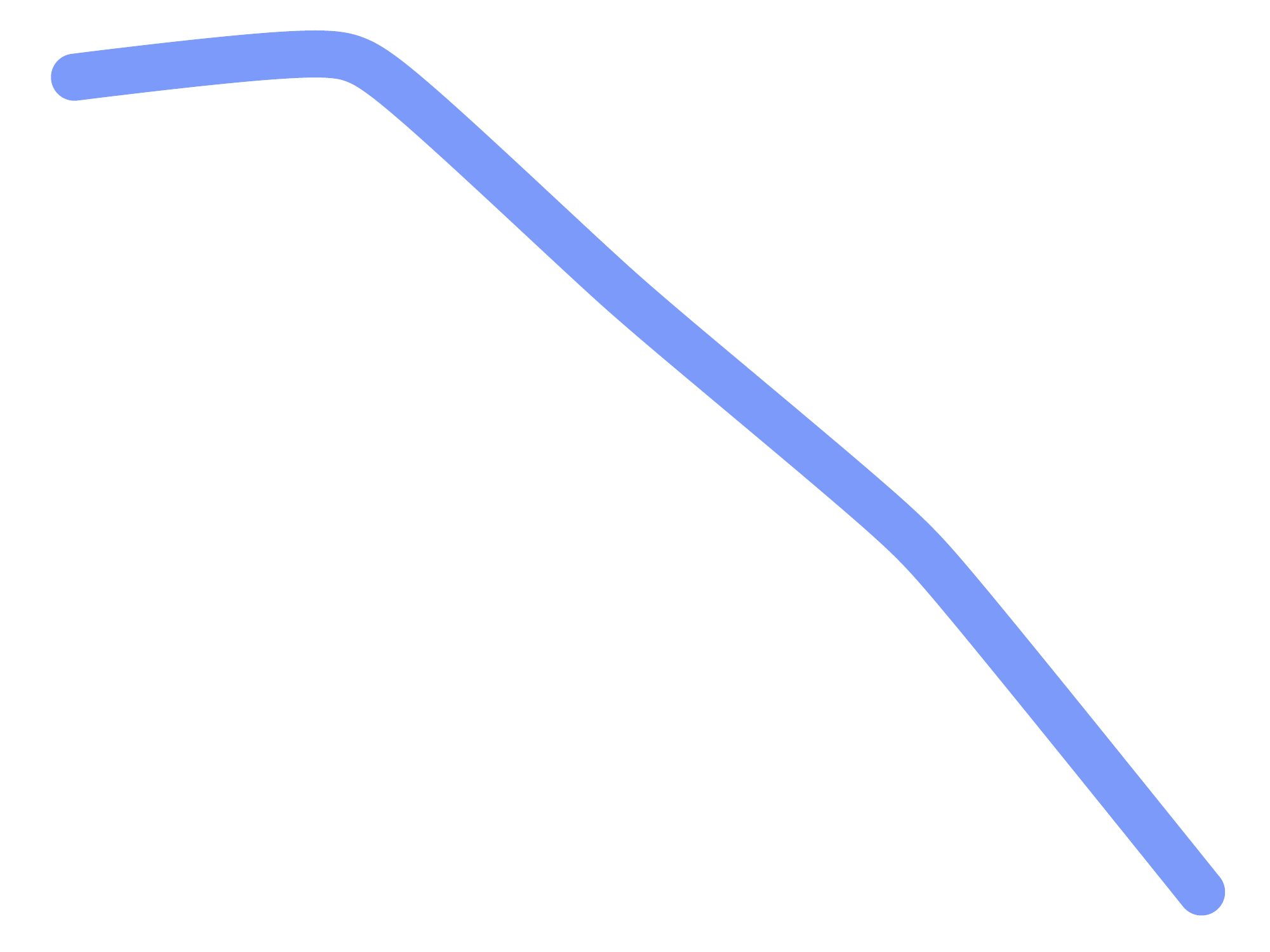}}
			\end{minipage}  & \cellcolor[rgb]{ .914,  .953,  .89}24.39  & \cellcolor[rgb]{ .914,  .953,  .89}24.65  & \cellcolor[rgb]{ .945,  .969,  .929}20.93  & \cellcolor[rgb]{ .945,  .969,  .929}17.26  & \cellcolor[rgb]{ .914,  .953,  .89}12.06  \\
\bottomrule
		\end{tabular}%
	}
\end{minipage}
\end{minipage}

\vspace{0.3cm}
\begin{minipage}{\textwidth}
	\begin{minipage}[h]{0.5\textwidth}
		\centering
		\makeatletter\def\@captype{table}\makeatother\caption{The empirical results of R@5 with respect to different preference thresholds $t$ on MovieLens-100k. The best and second-best are highlighted in bold and underlined, respectively.}
		\label{e}   
		\scalebox{0.8}{
				\begin{tabular}{c|c|ccccc}
				\toprule
				Method & Trend & \cellcolor[rgb]{ 1.0,  1.0,  1.0} t=1 & t=2 & t=3 & t=4 & t=5 \\
				\midrule
				itemKNN & \begin{minipage}[!t]{0.1\columnwidth}
					\centering
					\raisebox{-.5\height}{\includegraphics[width=\textwidth, height=0.3\textwidth]{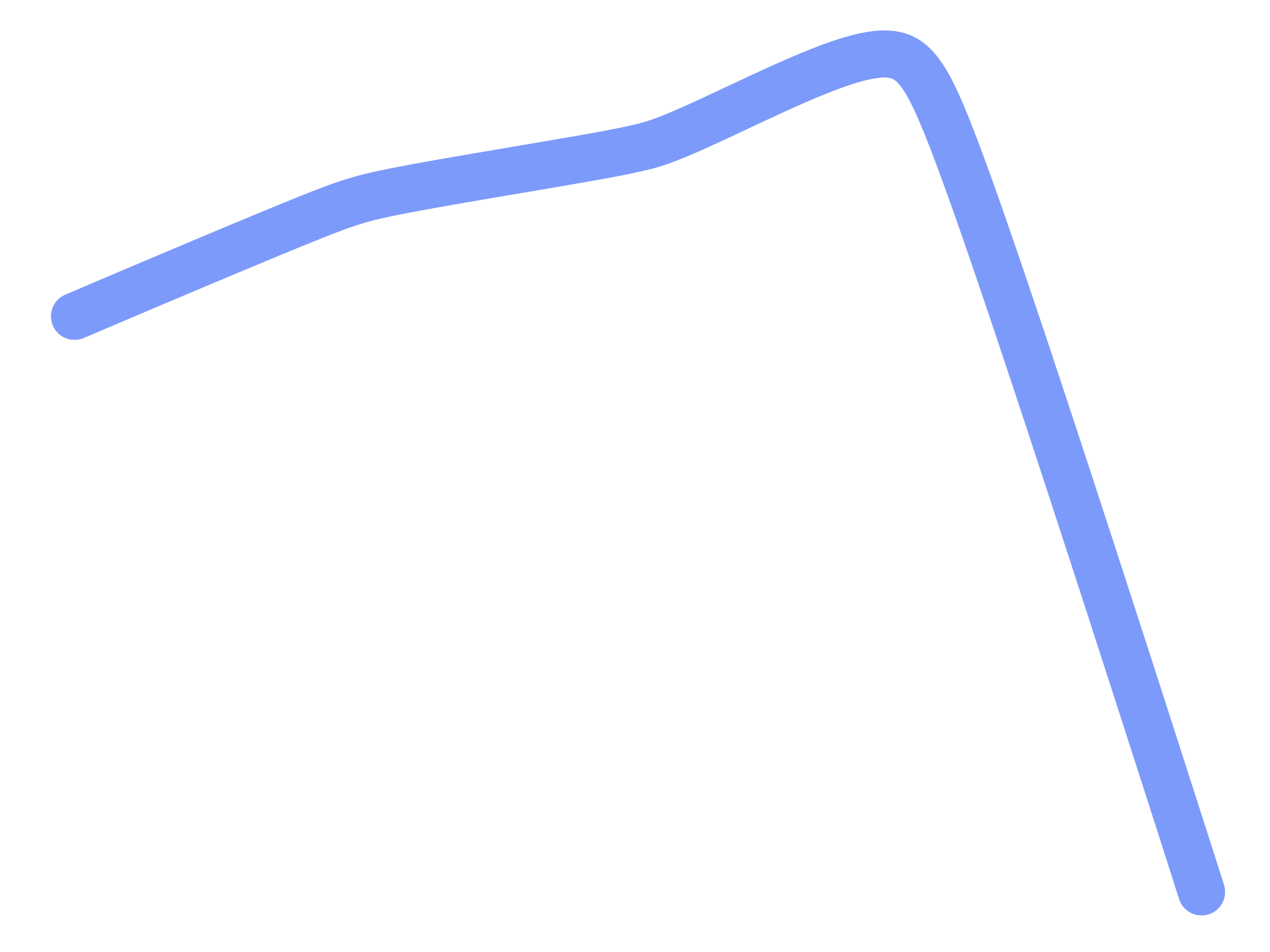}}
				\end{minipage}  &\cellcolor[rgb]{ 1.0,  1.0,  1.0} 3.14 & 3.60 & 3.81 & 4.11 & 0.85 \\
				GMF & \begin{minipage}[!t]{0.1\columnwidth}
					\centering
					\raisebox{-.5\height}{\includegraphics[width=\textwidth, height=0.3\textwidth]{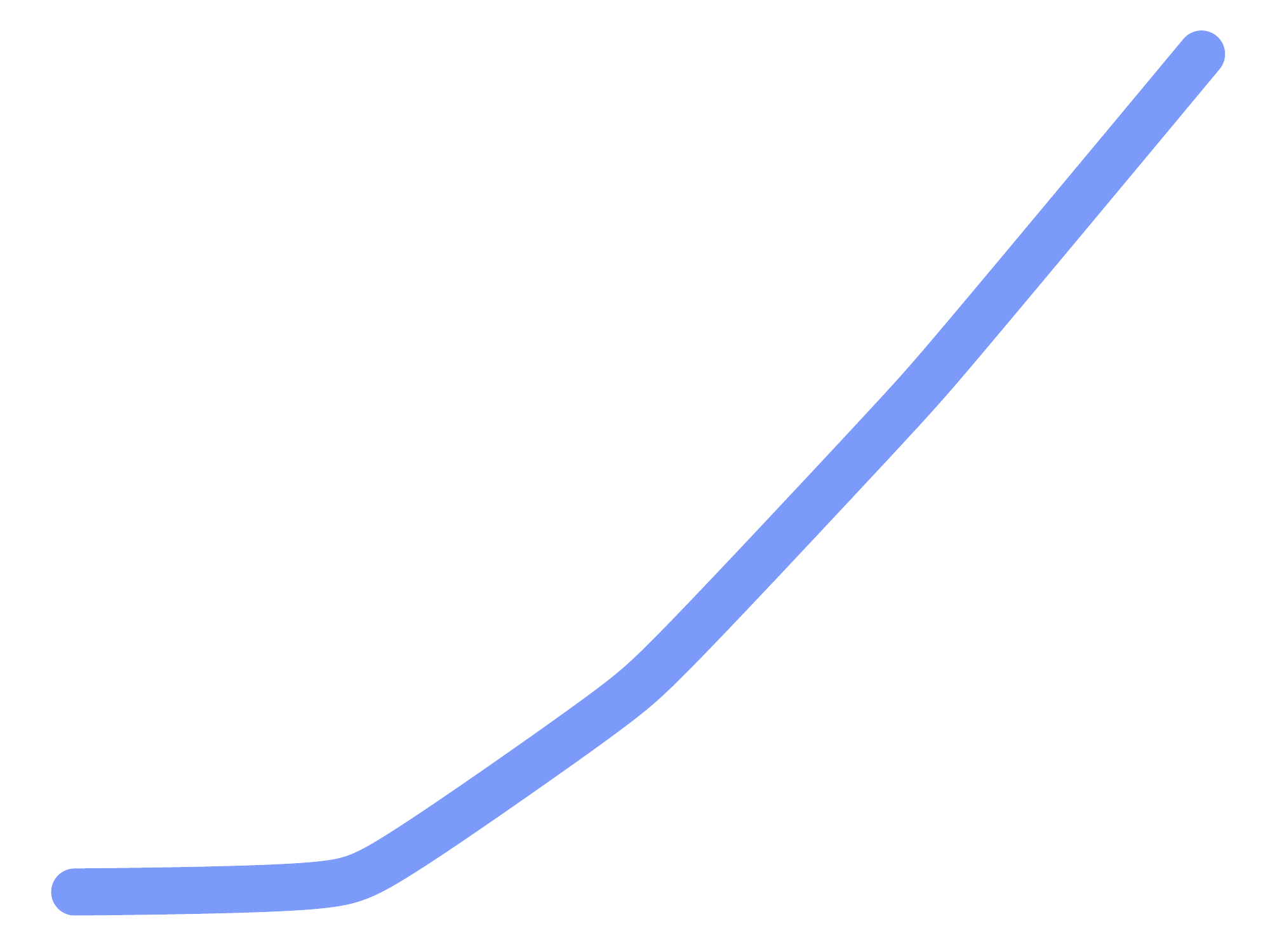}}
				\end{minipage}  & \cellcolor[rgb]{ 1.0,  1.0,  1.0}4.40 & 4.44 & 4.97 & 5.79 & \cellcolor[rgb]{ .965,  .976,  .988}6.72  \\
				MLP & \begin{minipage}[!t]{0.1\columnwidth}
					\centering
					\raisebox{-.5\height}{\includegraphics[width=\textwidth, height=0.3\textwidth]{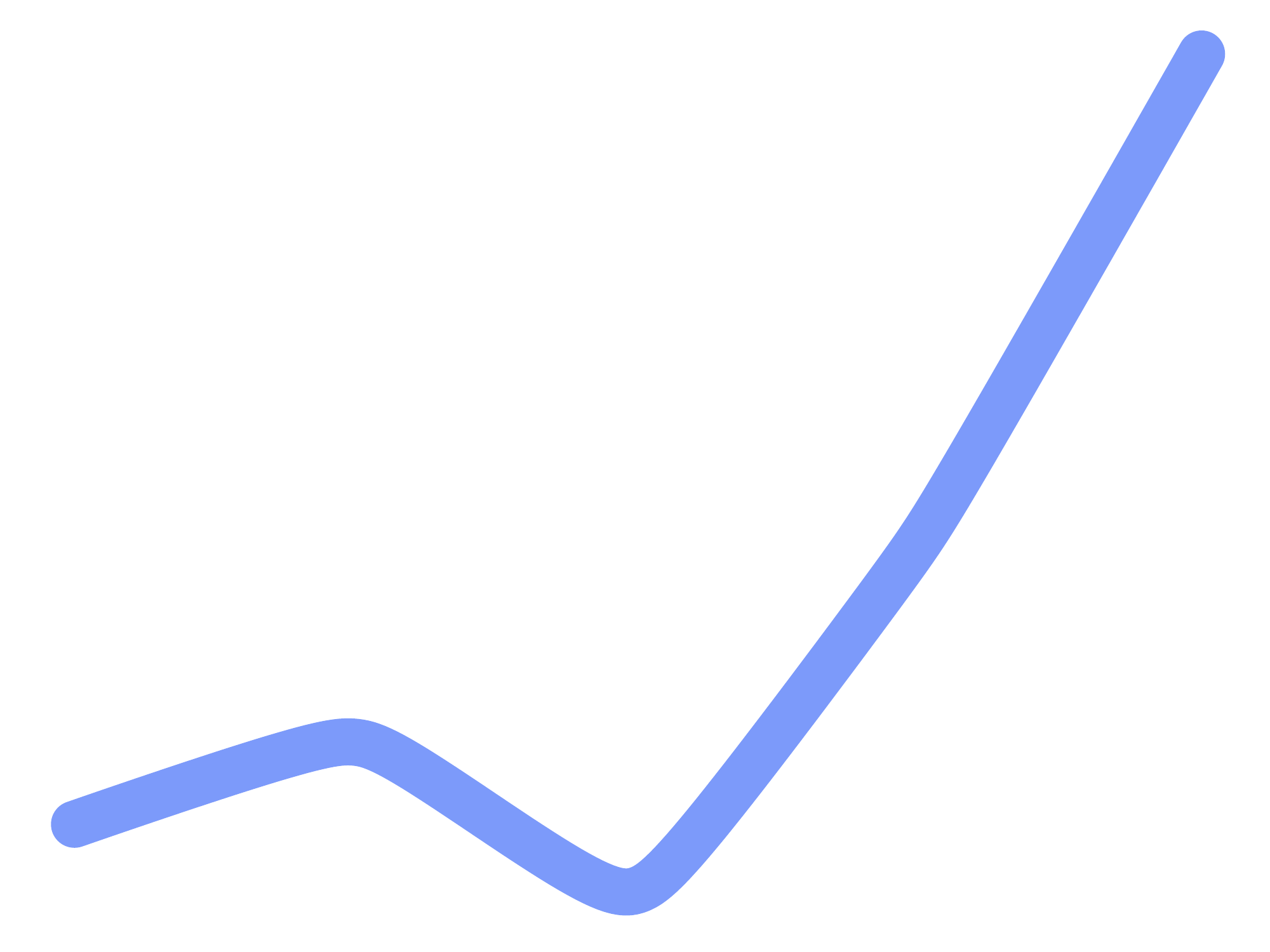}}
				\end{minipage}  & \cellcolor[rgb]{ 1.0,  1.0,  1.0}5.18 & 5.33 & 5.06 & 5.70 & \cellcolor[rgb]{ .965,  .976,  .988}6.59  \\
				NCF & \begin{minipage}[!t]{0.1\columnwidth}
					\centering
					\raisebox{-.5\height}{\includegraphics[width=\textwidth, height=0.3\textwidth]{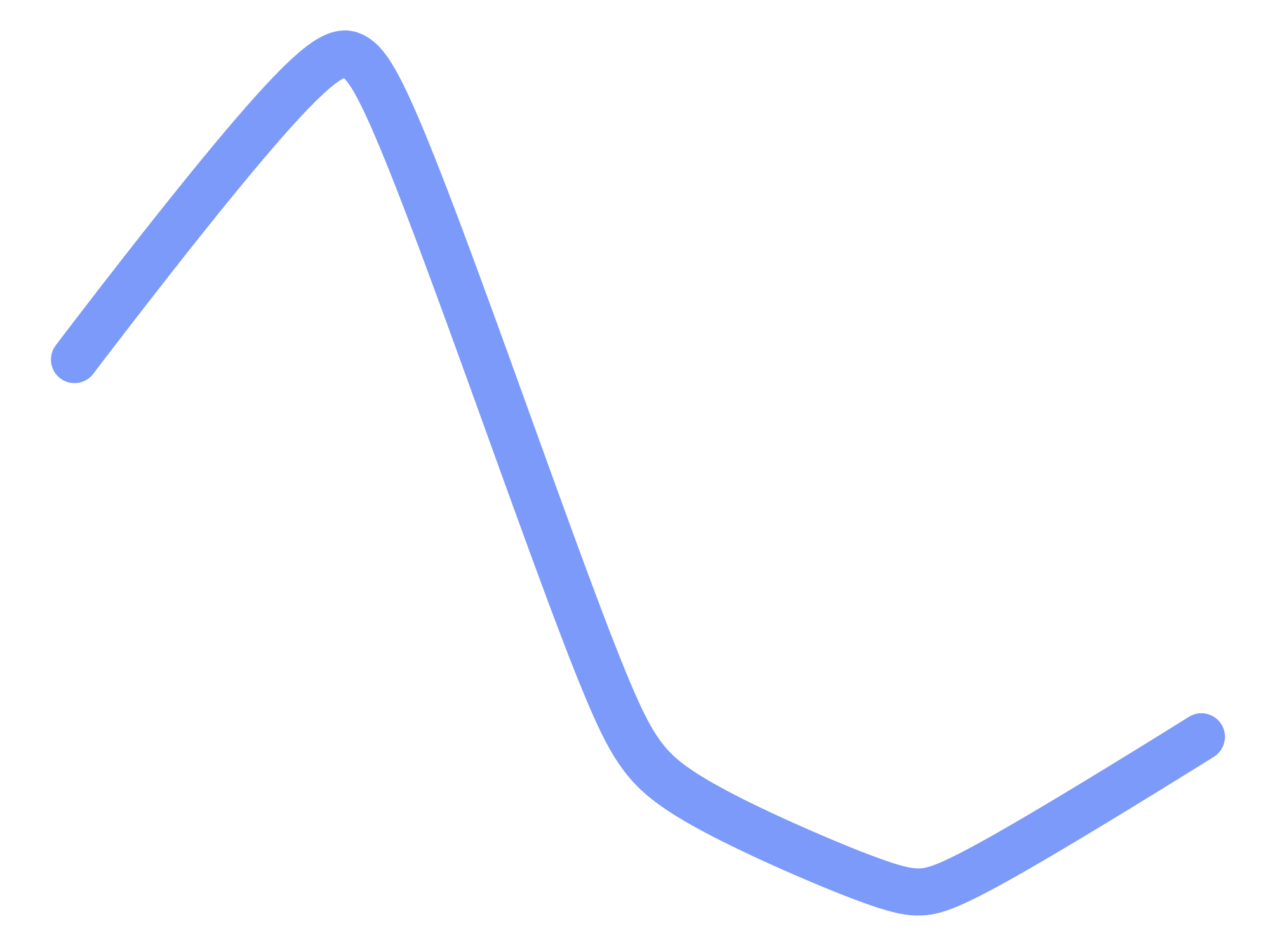}}
				\end{minipage}  & \cellcolor[rgb]{ .906,  .941,  .973}7.65  & \cellcolor[rgb]{ .906,  .941,  .973}8.33  & \cellcolor[rgb]{ .965,  .976,  .988}6.77  & \cellcolor[rgb]{ .906,  .941,  .973}6.45  & \cellcolor[rgb]{ .965,  .976,  .988}6.80 \\
				EHCF & \begin{minipage}[!t]{0.1\columnwidth}
					\centering
					\raisebox{-.5\height}{\includegraphics[width=\textwidth, height=0.3\textwidth]{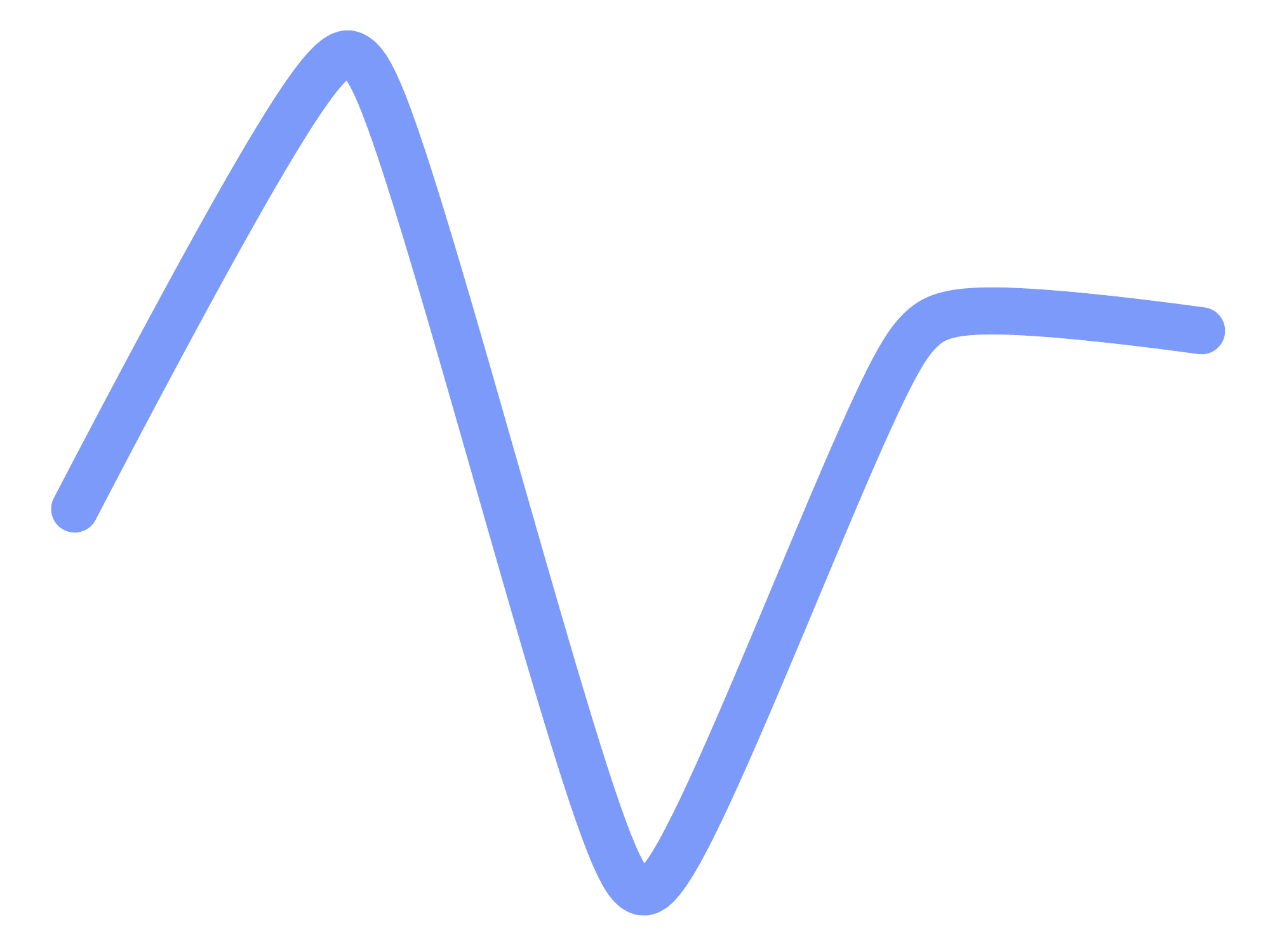}}
				\end{minipage}  & \cellcolor[rgb]{ .843,  .906,  .961}8.67  & \cellcolor[rgb]{ .843,  .906,  .961}9.05  & \cellcolor[rgb]{ .843,  .906,  .961}8.35  & \cellcolor[rgb]{ .843,  .906,  .961}8.82  & \cellcolor[rgb]{ .808,  .882,  .949}\underline{8.82}  \\
\bottomrule
			\end{tabular}%
		}
	\end{minipage}
	\hspace{0.05cm}
	\begin{minipage}[h]{0.5\textwidth}
		\centering
		\makeatletter\def\@captype{table}\makeatother\caption{The empirical results of NDCG@5 with respect to different preference thresholds $t$ on MovieLens-100k. The best and second-best are highlighted in bold and underlined, respectively.}
		\label{f}   
		\scalebox{0.8}{
			\begin{tabular}{c|c|ccccc}
				\toprule
				Method & Trend & \cellcolor[rgb]{1.0,1.0,1.0}t=1 & t=2 & t=3 & t=4 & t=5 \\
				\midrule
				itemKNN & \begin{minipage}[!t]{0.1\columnwidth}
					\centering
					\raisebox{-.5\height}{\includegraphics[width=\textwidth, height=0.3\textwidth]{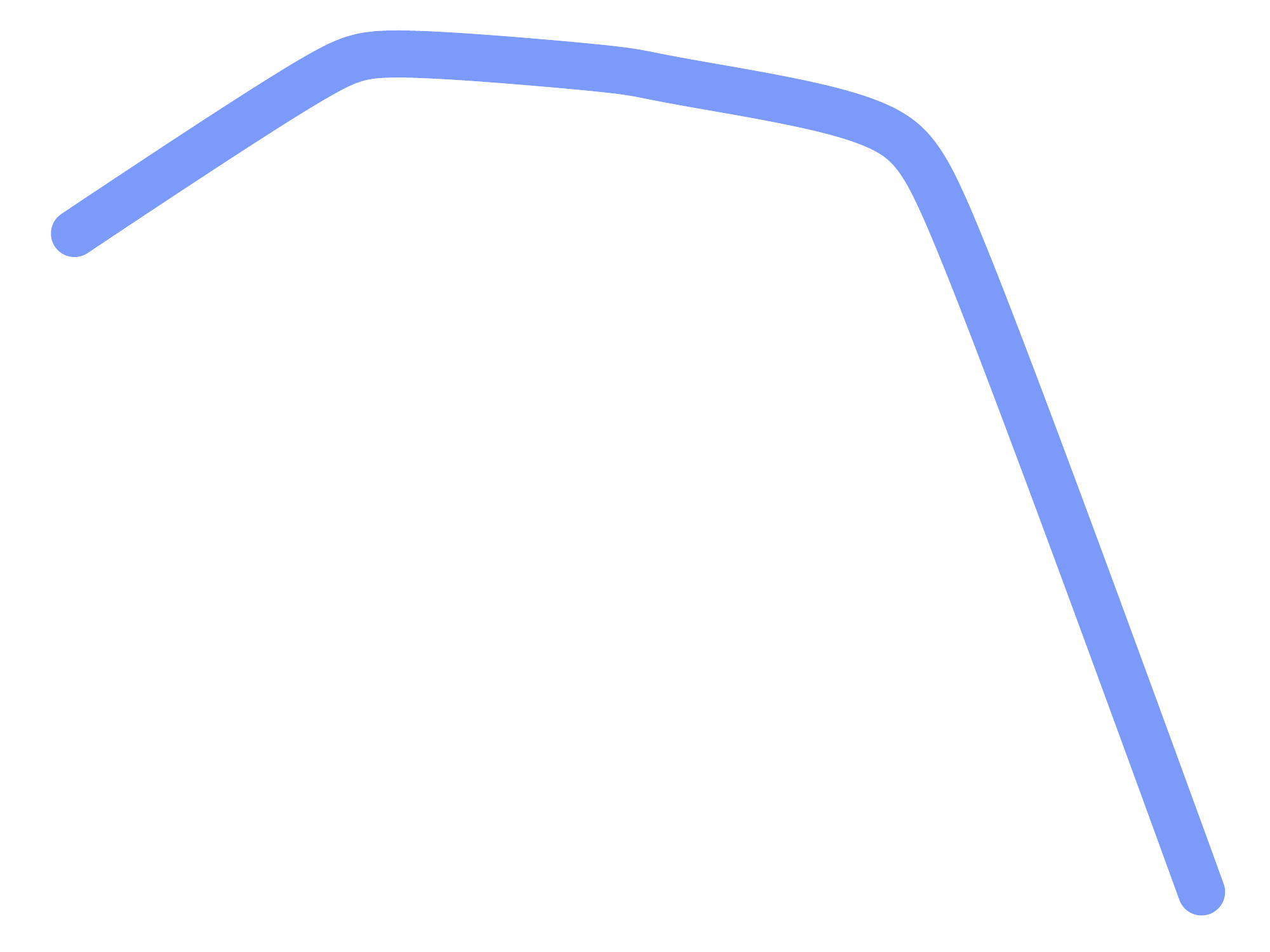}}
				\end{minipage}  & \cellcolor[rgb]{1.0,1.0,1.0}12.23 & 15.17 & 14.92 & 13.45 & 1.18 \\
				GMF & \begin{minipage}[!t]{0.1\columnwidth}
					\centering
					\raisebox{-.5\height}{\includegraphics[width=\textwidth, height=0.3\textwidth]{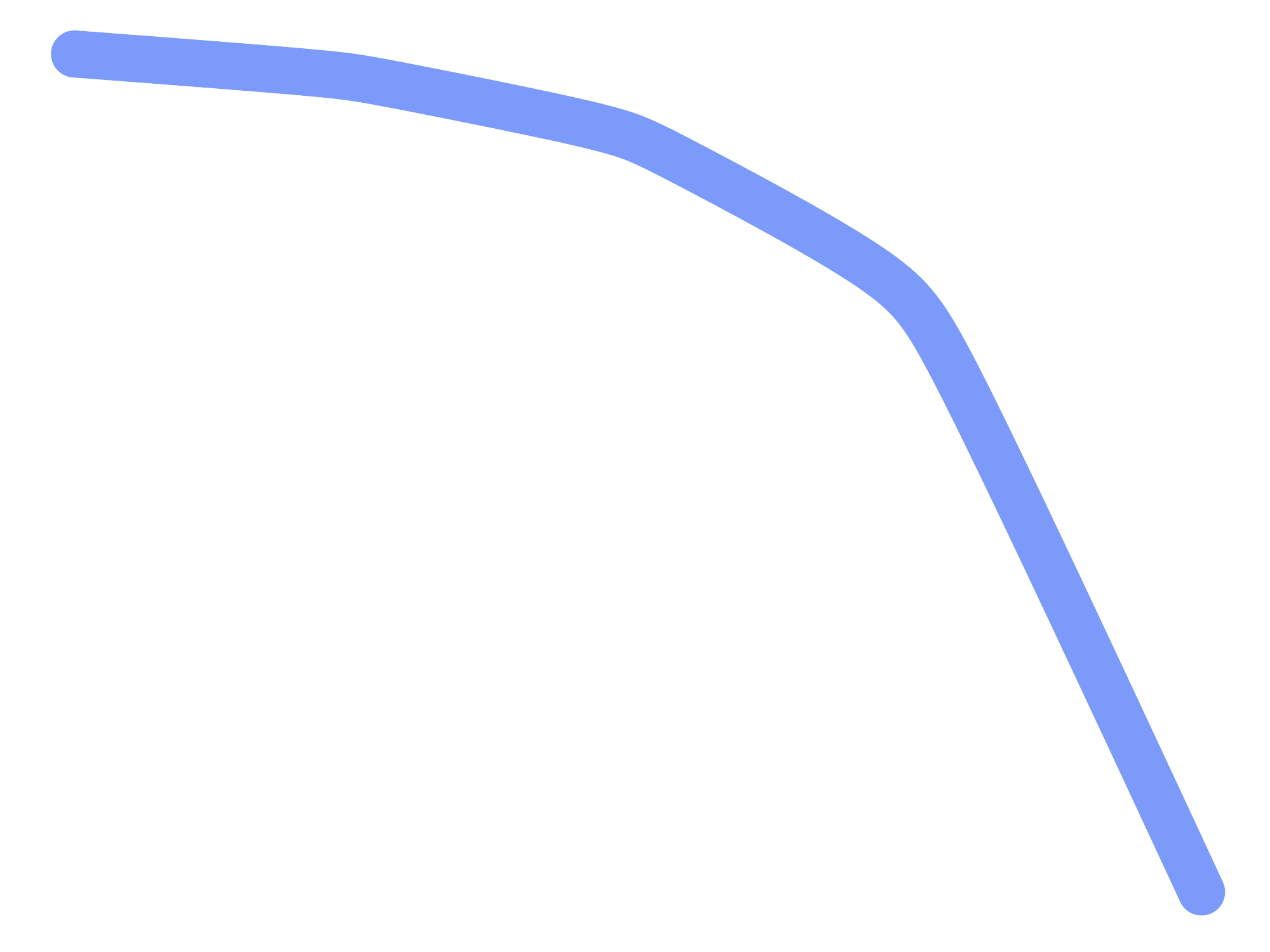}}
				\end{minipage}  & \cellcolor[rgb]{1.0,1.0,1.0}19.05 & 18.88 & 18.44 & 17.21 & \cellcolor[rgb]{ .906,  .941,  .973}13.08  \\
				MLP &\begin{minipage}[!t]{0.1\columnwidth}
					\centering
					\raisebox{-.5\height}{\includegraphics[width=\textwidth, height=0.3\textwidth]{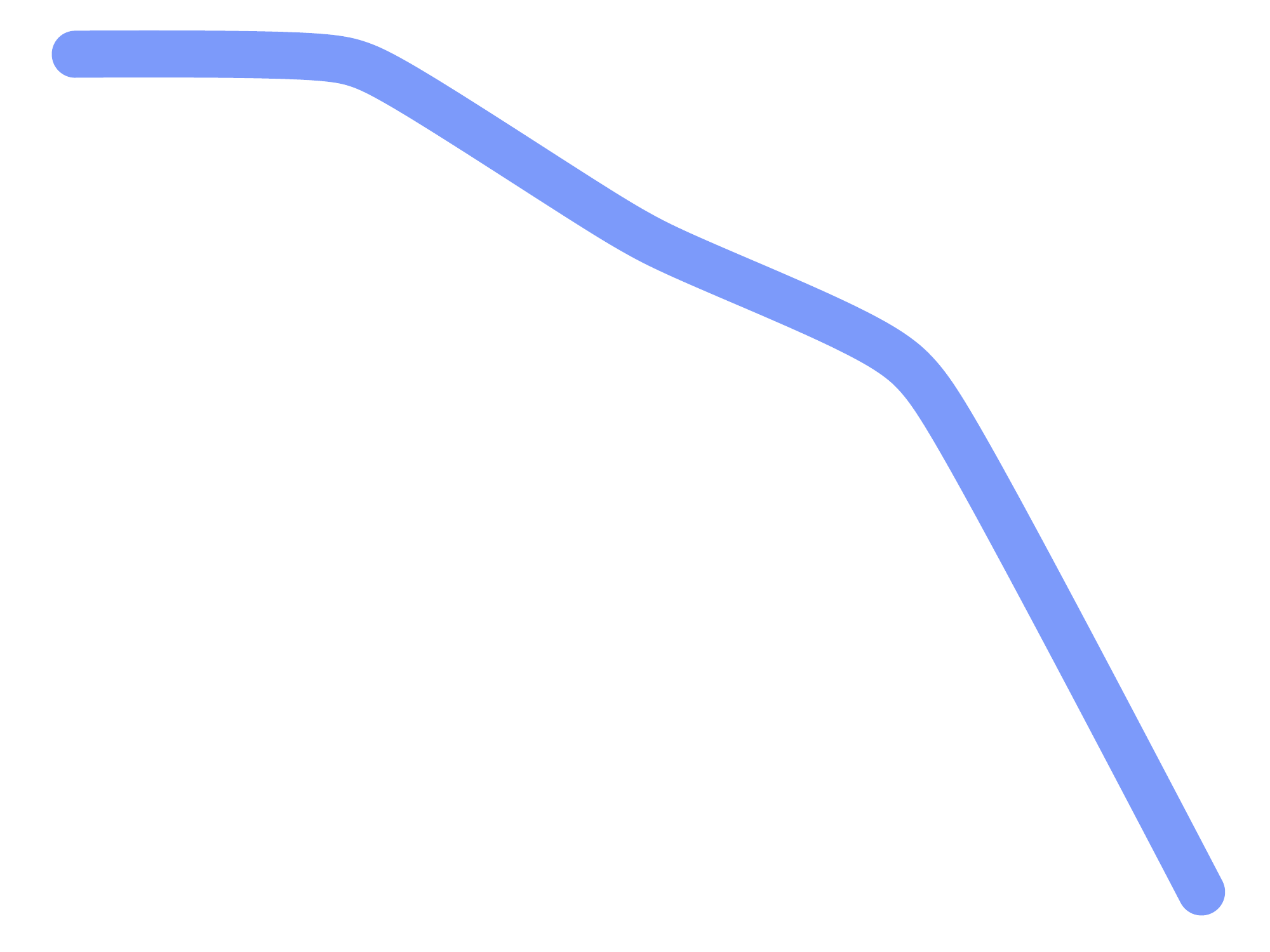}}
				\end{minipage}   & \cellcolor[rgb]{1.0,1.0,1.0}18.98 & 18.91 & 17.63 & 16.54 & \cellcolor[rgb]{ .965,  .976,  .988}12.67  \\
				NCF & \begin{minipage}[!t]{0.1\columnwidth}
					\centering
					\raisebox{-.5\height}{\includegraphics[width=\textwidth, height=0.3\textwidth]{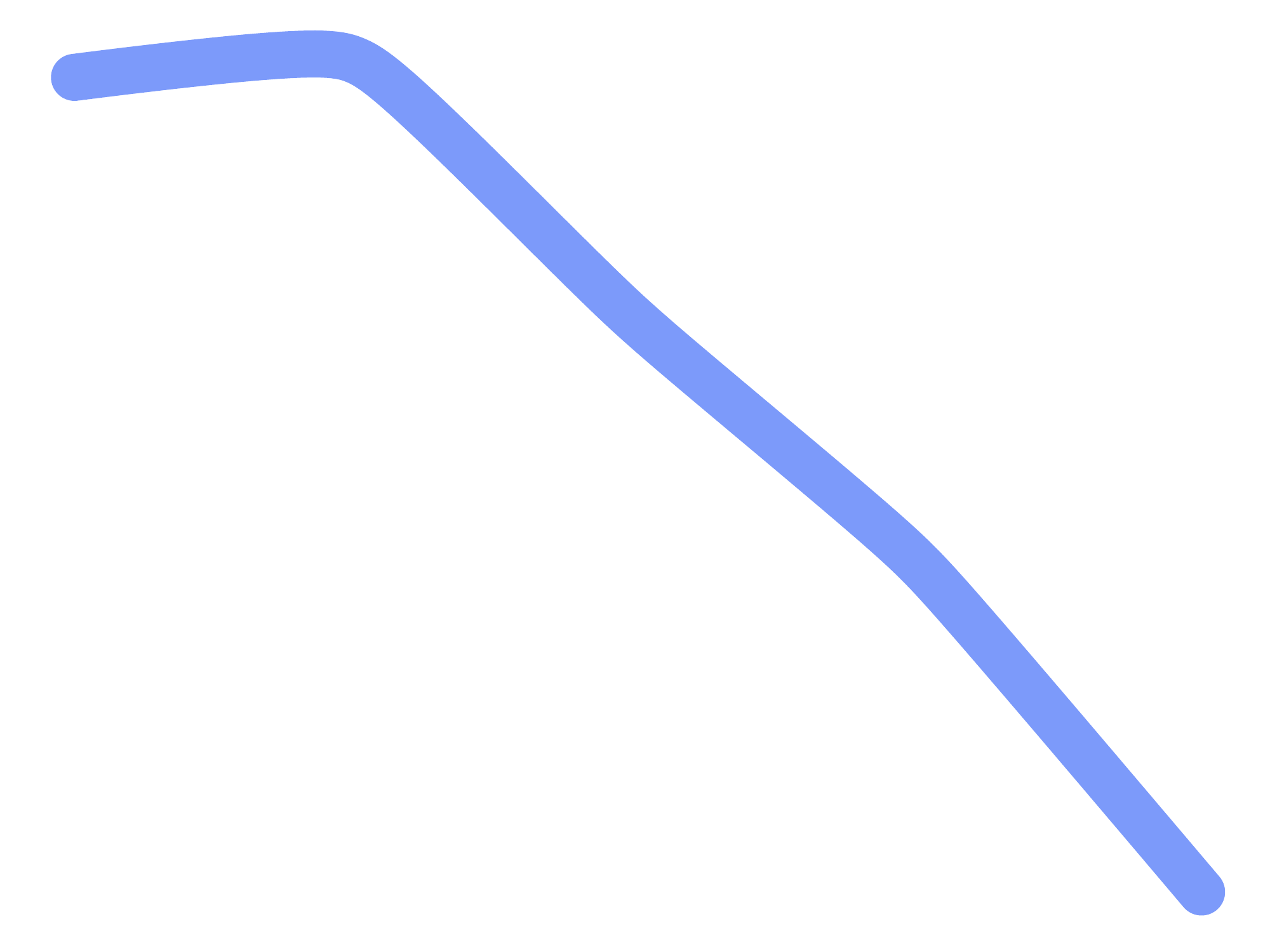}}
				\end{minipage}  & \cellcolor[rgb]{ .906,  .941,  .973}25.81  & \cellcolor[rgb]{ .843,  .906,  .961}26.07  & \cellcolor[rgb]{ .965,  .976,  .988}22.01  & \cellcolor[rgb]{ .965,  .976,  .988}18.25  & \cellcolor[rgb]{ .906,  .941,  .973}13.18  \\
				EHCF & \begin{minipage}[!t]{0.1\columnwidth}
					\centering
					\raisebox{-.5\height}{\includegraphics[width=\textwidth, height=0.3\textwidth]{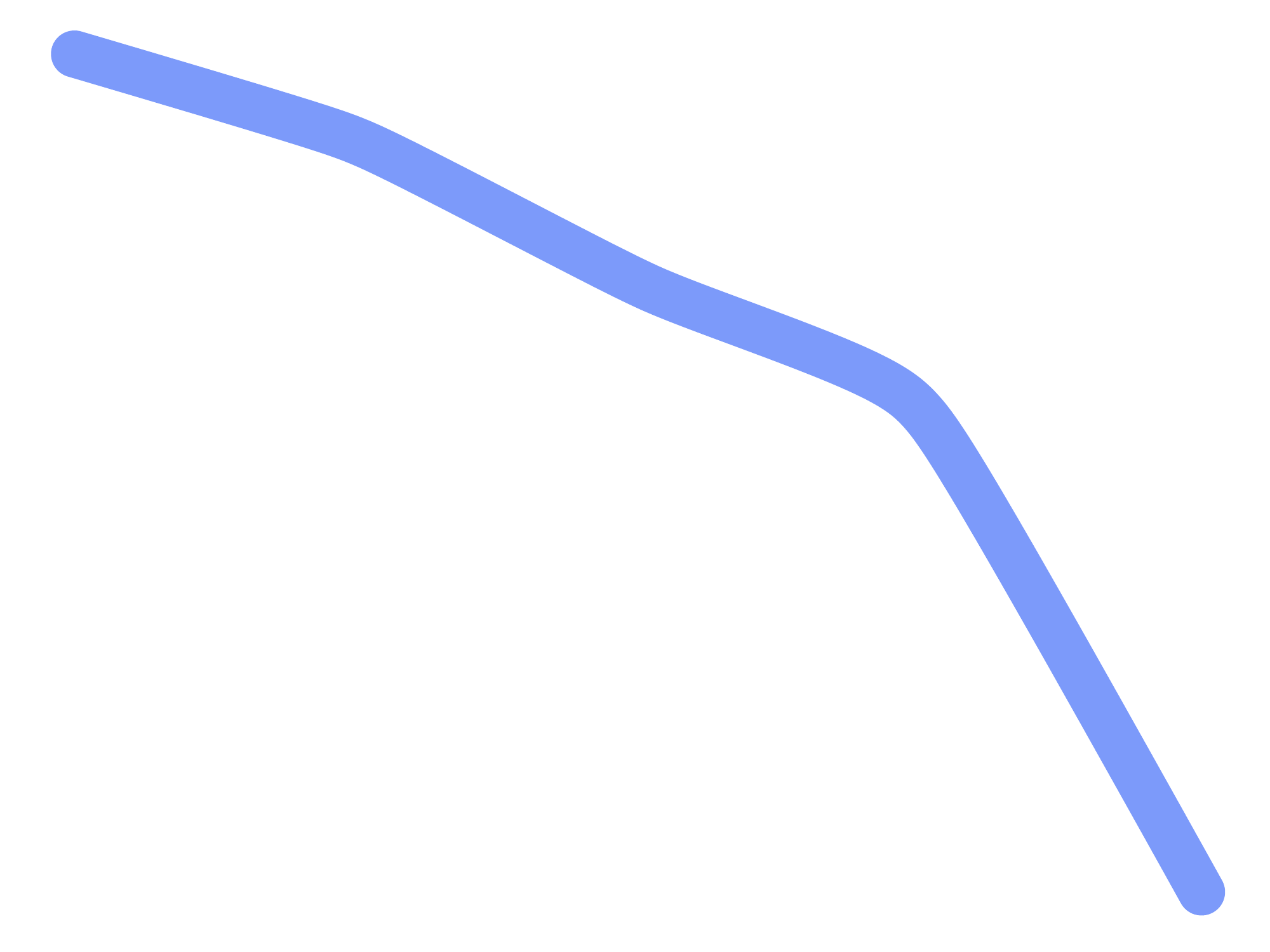}}
				\end{minipage}  & \cellcolor[rgb]{ .843,  .906,  .961}27.00 & \cellcolor[rgb]{ .906,  .941,  .973}25.80 & \cellcolor[rgb]{ .906,  .941,  .973}23.84  & \cellcolor[rgb]{ .843,  .906,  .961}22.08  & \cellcolor[rgb]{ .808,  .882,  .949}\underline{15.41}  \\
\bottomrule
			\end{tabular}%
		}
	\end{minipage}
\end{minipage}

\vspace{0.3cm}
\begin{minipage}{\textwidth}
	\begin{minipage}[h]{0.5\textwidth}
		\centering
		\makeatletter\def\@captype{table}\makeatother\caption{The empirical results of P@10 with respect to different preference thresholds $t$ on MovieLens-100k. The best and second-best are highlighted in bold and underlined, respectively.}
		\label{g}   
		\scalebox{0.8}{
			\begin{tabular}{c|c|ccccc}
				\toprule
				Method & Trend & \cellcolor[rgb]{ 1.0,  1.0,  1.0}t=1 & t=2 & t=3 & t=4 & t=5 \\
				\midrule
				itemKNN & \begin{minipage}[!t]{0.1\columnwidth}
					\centering
					\raisebox{-.5\height}{\includegraphics[width=\textwidth, height=0.3\textwidth]{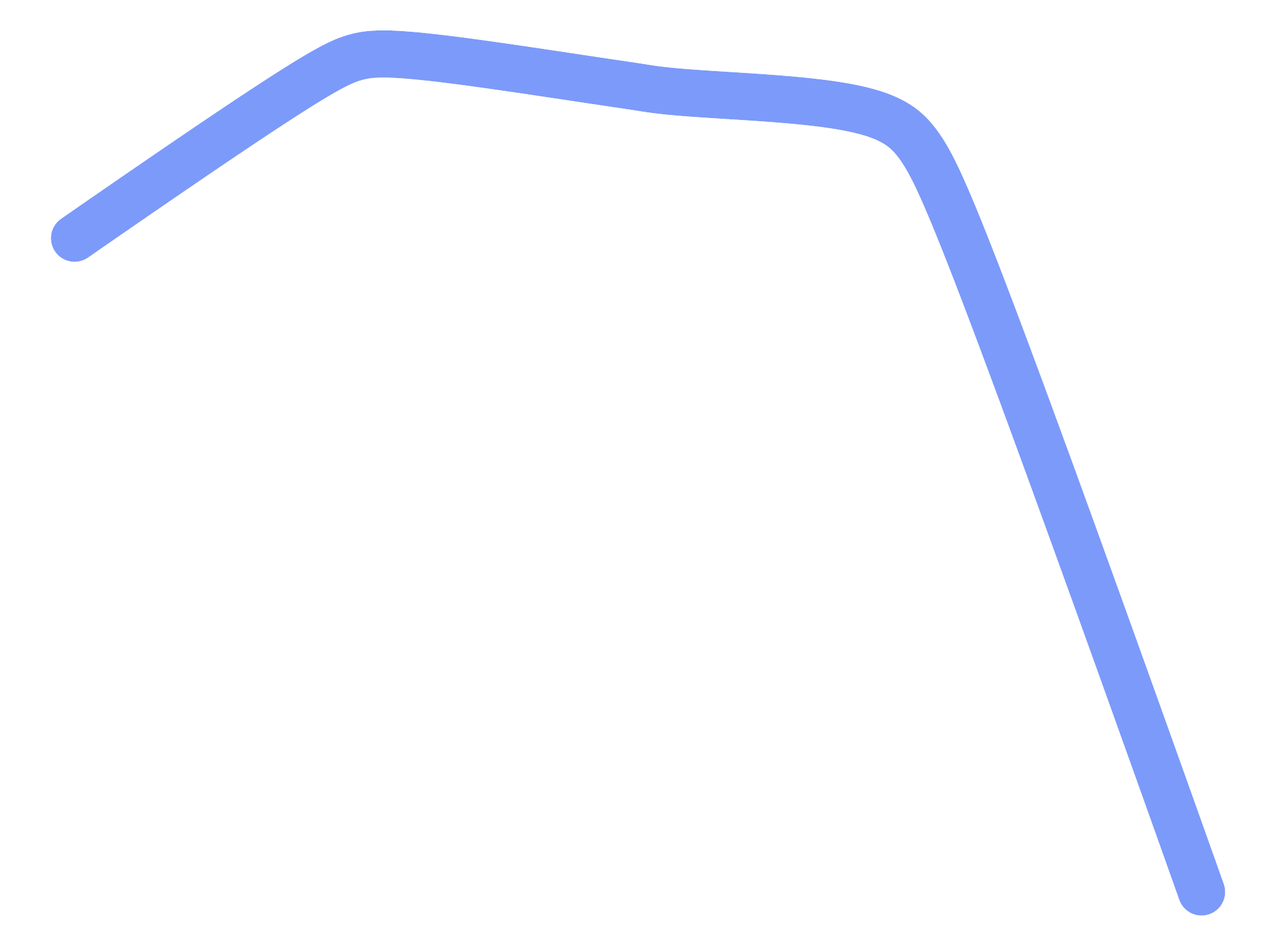}}
				\end{minipage}  & \cellcolor[rgb]{ 1.0,  1.0,  1.0}15.09 & 18.97 & 18.33 & 17.05 & 1.09 \\
				GMF & \begin{minipage}[!t]{0.1\columnwidth}
					\centering
					\raisebox{-.5\height}{\includegraphics[width=\textwidth, height=0.3\textwidth]{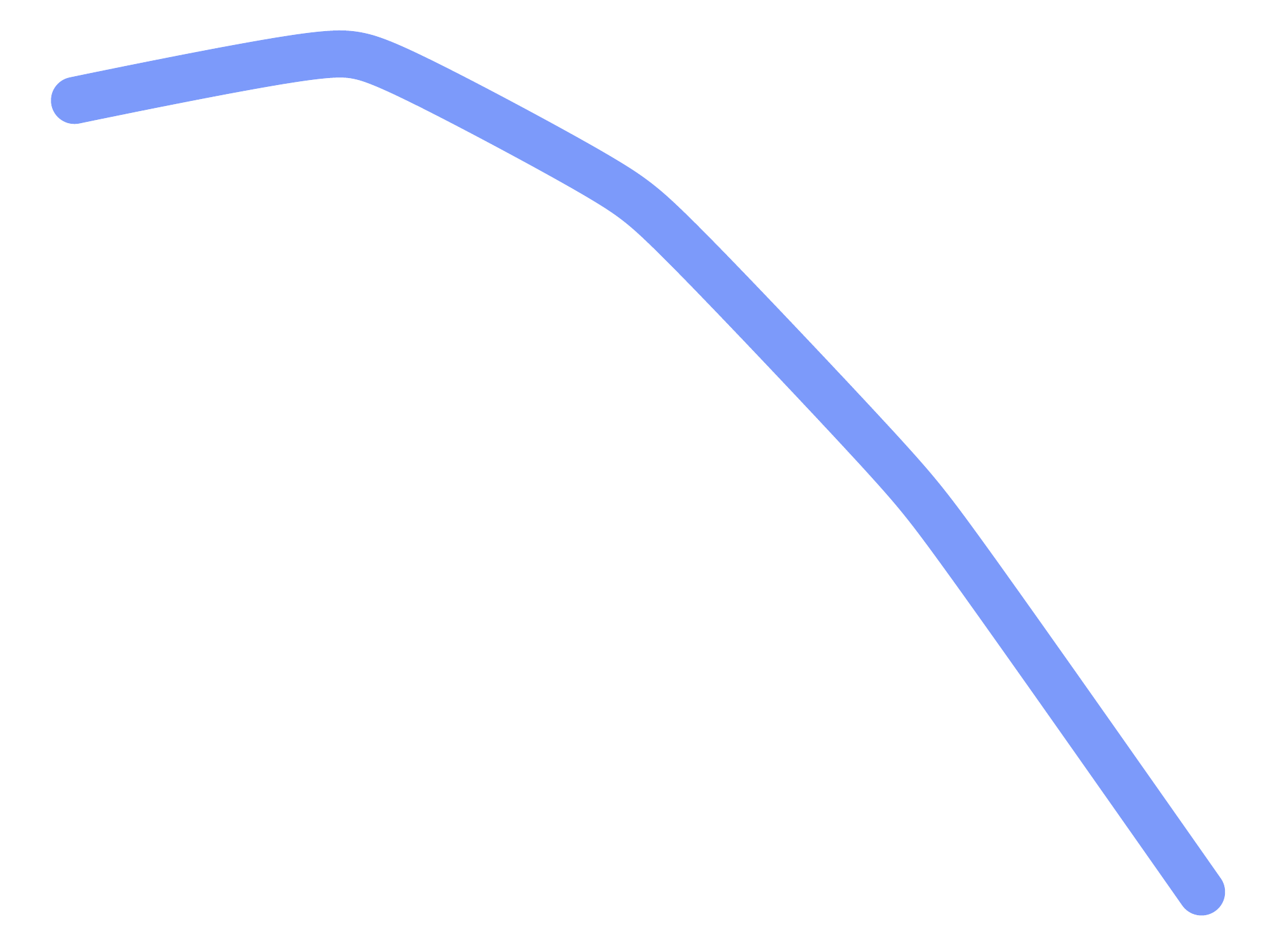}}
				\end{minipage}   & \cellcolor[rgb]{ .984,  .992,  .976}20.70 & \cellcolor[rgb]{ .984,  .992,  .976}21.00 & \cellcolor[rgb]{ .984,  .992,  .976}20.02  & \cellcolor[rgb]{ .984,  .992,  .976}18.06  & \cellcolor[rgb]{ .945,  .969,  .929}15.43 \\
				MLP & \begin{minipage}[!t]{0.1\columnwidth}
					\centering
					\raisebox{-.5\height}{\includegraphics[width=\textwidth, height=0.3\textwidth]{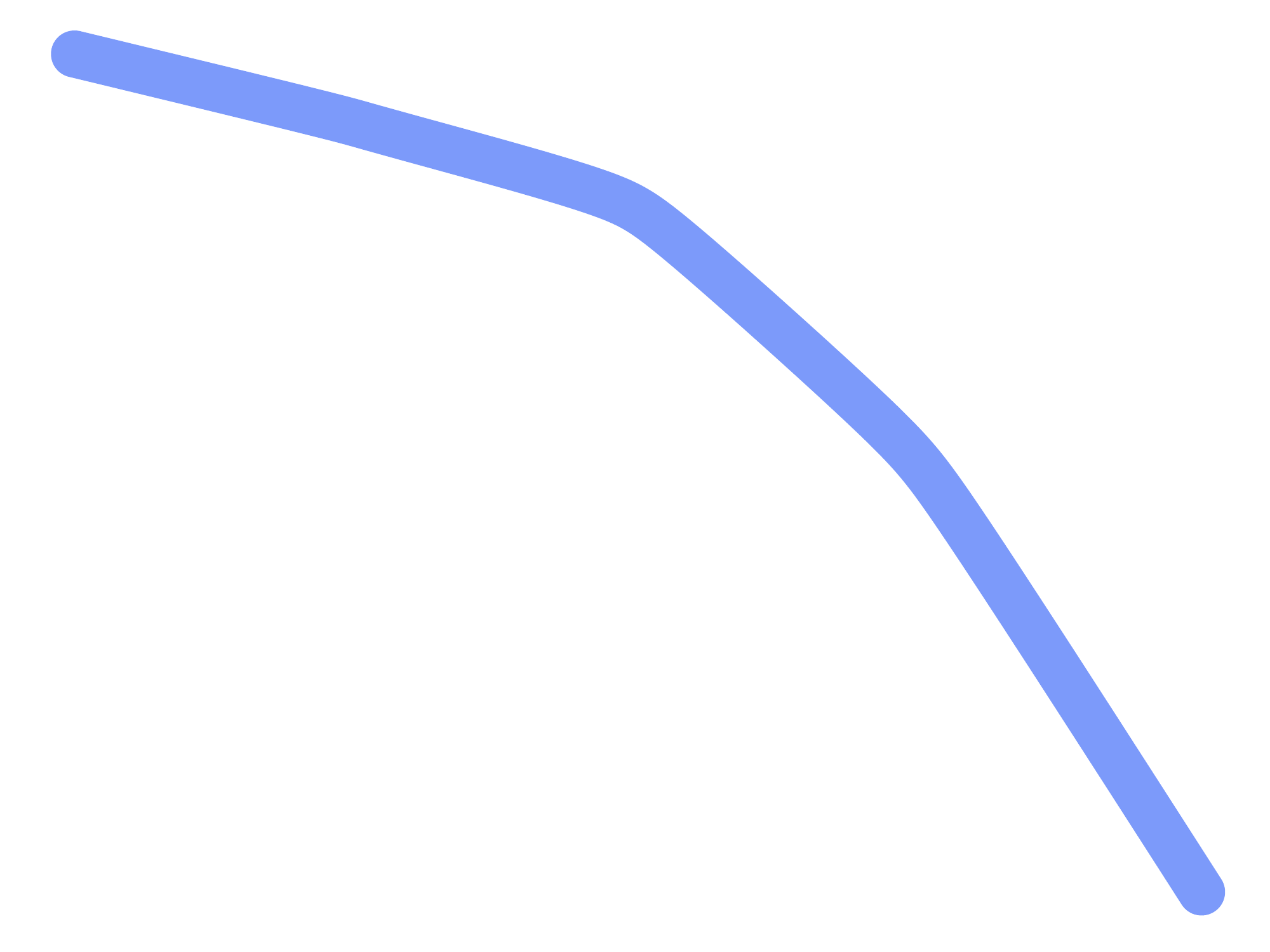}}
				\end{minipage}   & \cellcolor[rgb]{ .984,  .992,  .976}20.47  & \cellcolor[rgb]{ .984,  .992,  .976}19.97  & 19.34 & 17.51 & \cellcolor[rgb]{ .984,  .992,  .976}14.42  \\
				NCF & \begin{minipage}[!t]{0.1\columnwidth}
					\centering
					\raisebox{-.5\height}{\includegraphics[width=\textwidth, height=0.3\textwidth]{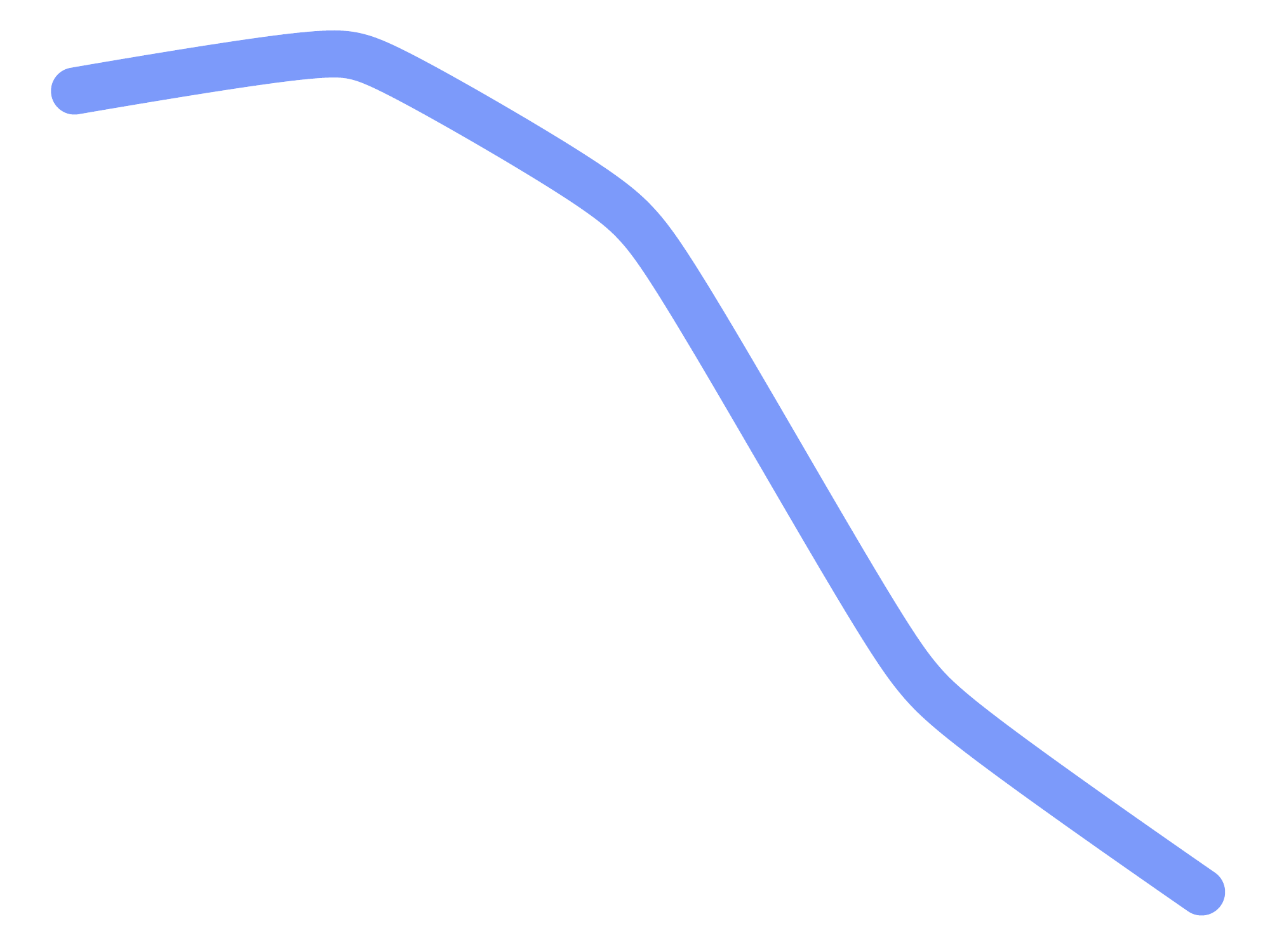}}
				\end{minipage}   & \cellcolor[rgb]{ .945,  .969,  .929}25.75  & \cellcolor[rgb]{ .945,  .969,  .929}26.19  & \cellcolor[rgb]{ .945,  .969,  .929}24.04  & 18.27 & \cellcolor[rgb]{ .945,  .969,  .929}15.58  \\
\bottomrule
			\end{tabular}%
		}
	\end{minipage}
	\hspace{0.05cm}
	\begin{minipage}[h]{0.5\textwidth}
		\centering
		\makeatletter\def\@captype{table}\makeatother\caption{The empirical results of R@10 with respect to different preference thresholds $t$ on MovieLens-100k. The best and second-best are highlighted in bold and underlined, respectively.}
		\label{h}   
		\scalebox{0.8}{
			\begin{tabular}{c|c|ccccc}
				\toprule
				Method & Trend & \cellcolor[rgb]{ 1.0,  1.0,  1.0}t=1 & t=2 & t=3 & t=4 & t=5 \\
				\midrule
				itemKNN & \begin{minipage}[!t]{0.1\columnwidth}
					\centering
					\raisebox{-.5\height}{\includegraphics[width=\textwidth, height=0.3\textwidth]{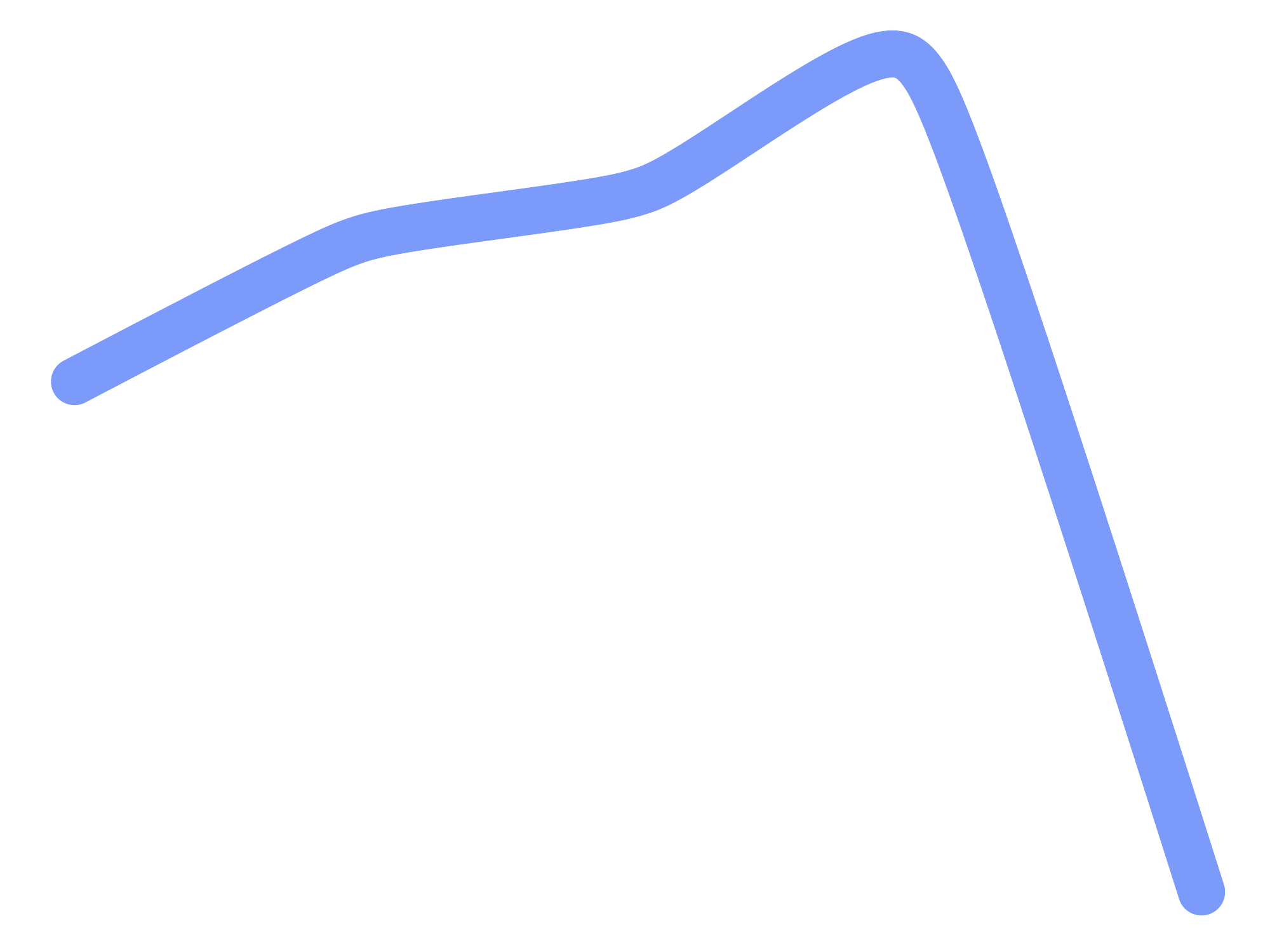}}
				\end{minipage}   & \cellcolor[rgb]{ 1.0,  1.0,  1.0}5.40 & 6.66 & 7.09 & 8.19 & 0.85 \\
				GMF & \begin{minipage}[!t]{0.1\columnwidth}
					\centering
					\raisebox{-.5\height}{\includegraphics[width=\textwidth, height=0.3\textwidth]{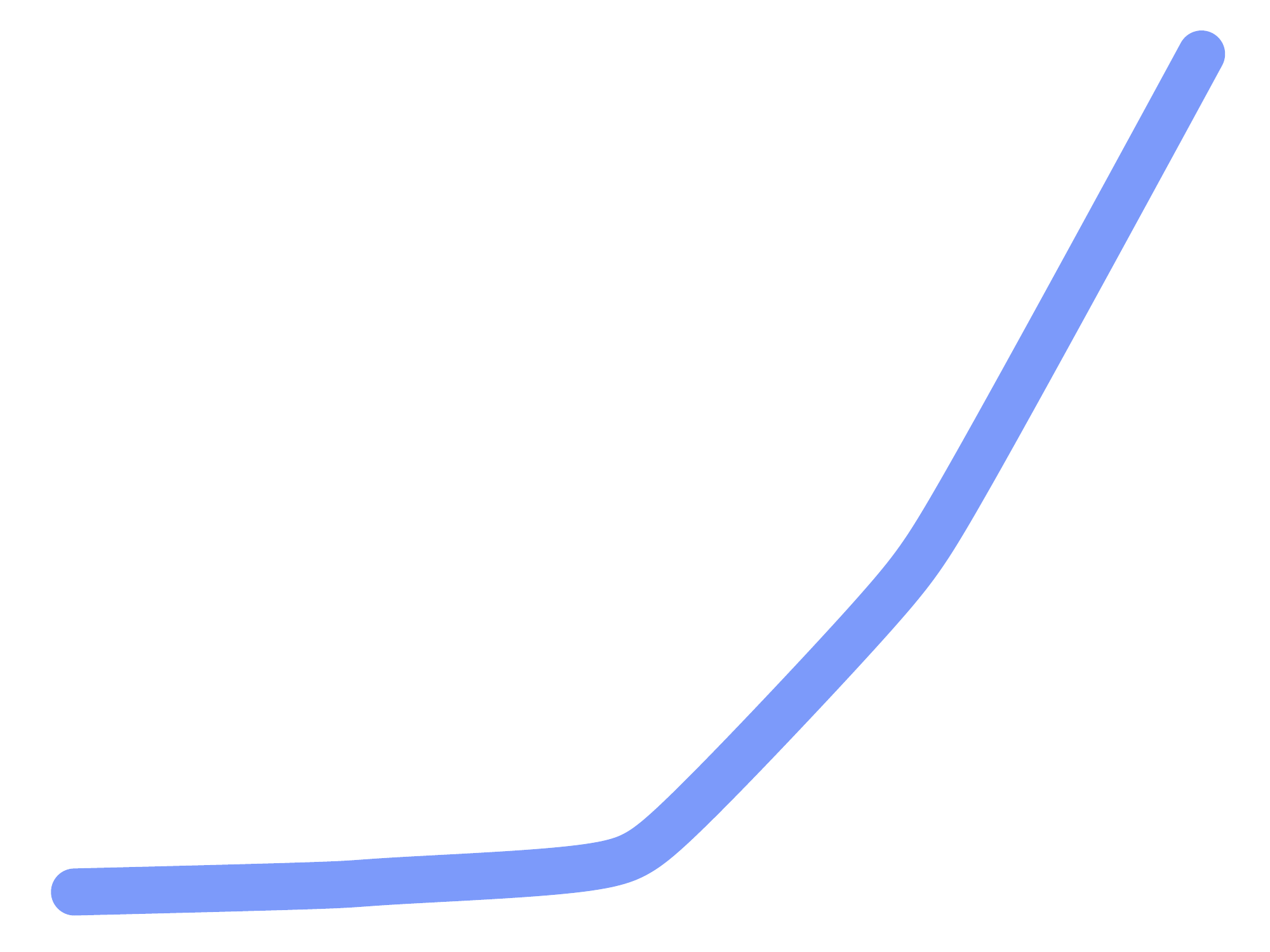}}
				\end{minipage}  & \cellcolor[rgb]{ 1.0,  1.0,  1.0}7.83 & 7.86 & 7.97 & 9.02 & \cellcolor[rgb]{ .945,  .969,  .929}10.82  \\
				MLP & \begin{minipage}[!t]{0.1\columnwidth}
					\centering
					\raisebox{-.5\height}{\includegraphics[width=\textwidth, height=0.3\textwidth]{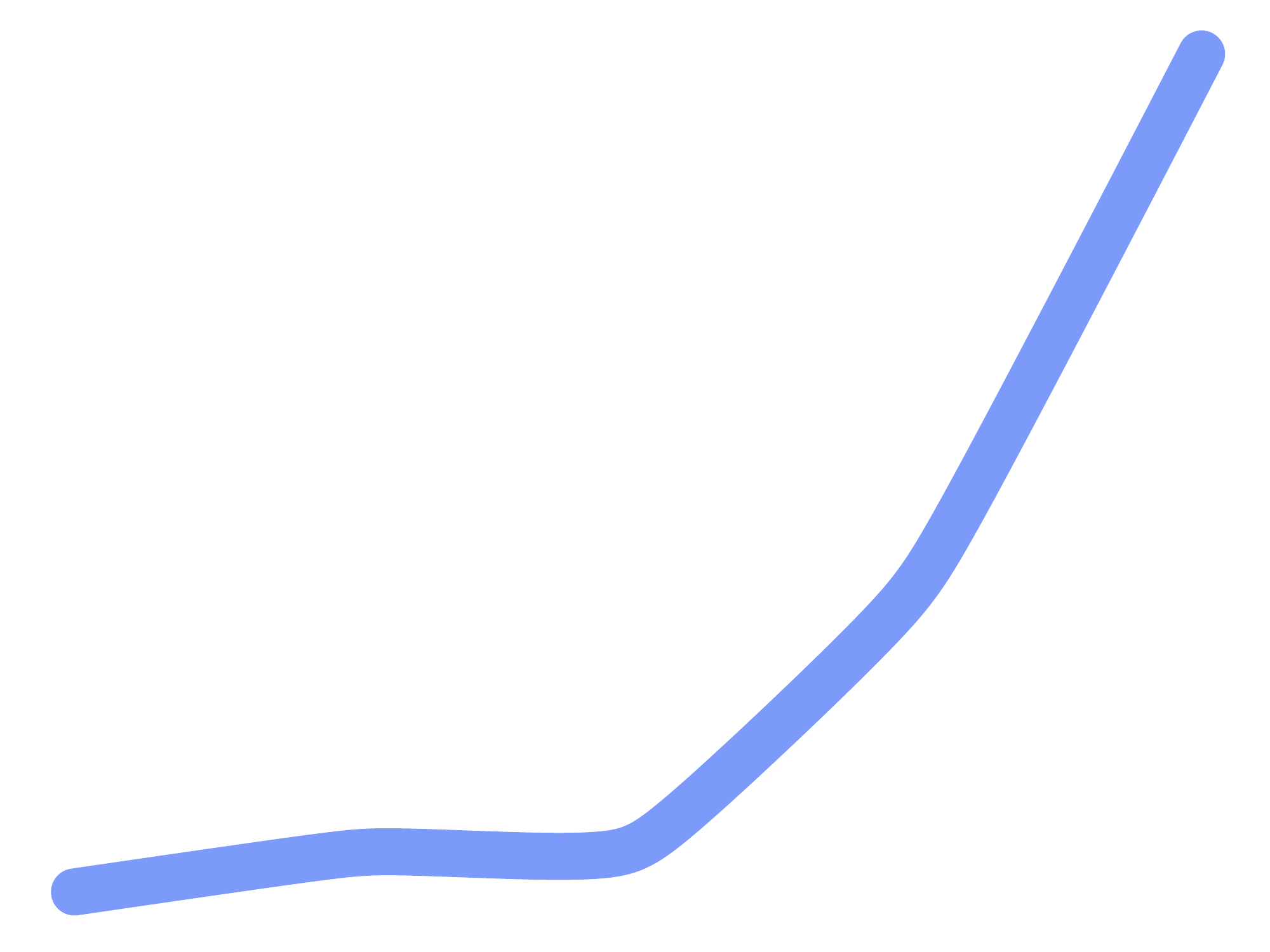}}
				\end{minipage}  & \cellcolor[rgb]{ 1.0,  1.0,  1.0}7.98 & 8.09 & 8.11 & 8.86 & \cellcolor[rgb]{ .945,  .969,  .929}10.34  \\
				NCF & \begin{minipage}[!t]{0.1\columnwidth}
					\centering
					\raisebox{-.5\height}{\includegraphics[width=\textwidth, height=0.3\textwidth]{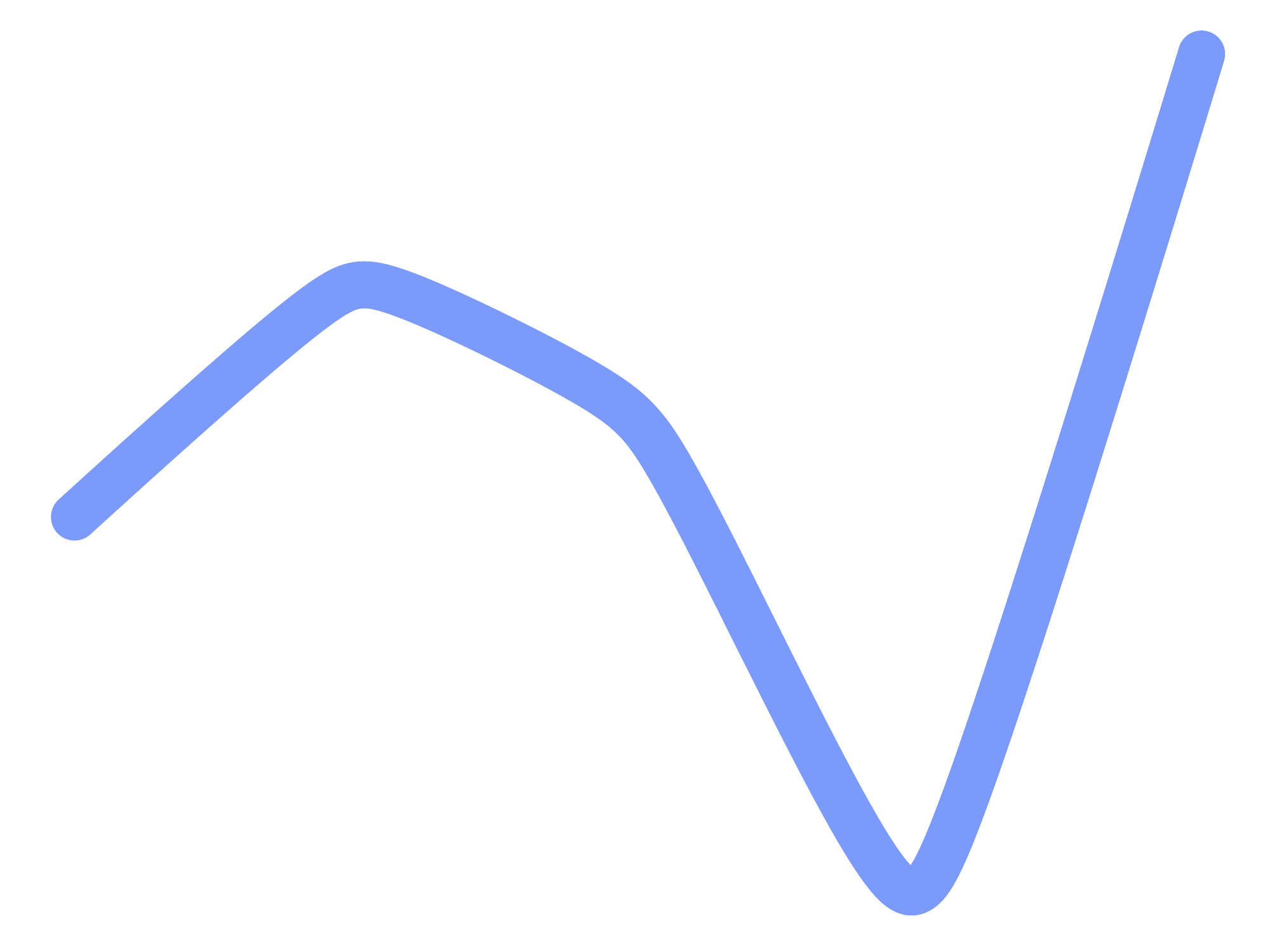}}
				\end{minipage}  & \cellcolor[rgb]{ .945,  .969,  .929}10.09  & \cellcolor[rgb]{ .945,  .969,  .929}10.60 & \cellcolor[rgb]{ .945,  .969,  .929}10.30 & 9.27 & \cellcolor[rgb]{ .886,  .937,  .851}11.11 \\
				EHCF & \begin{minipage}[!t]{0.1\columnwidth}
					\centering
					\raisebox{-.5\height}{\includegraphics[width=\textwidth, height=0.3\textwidth]{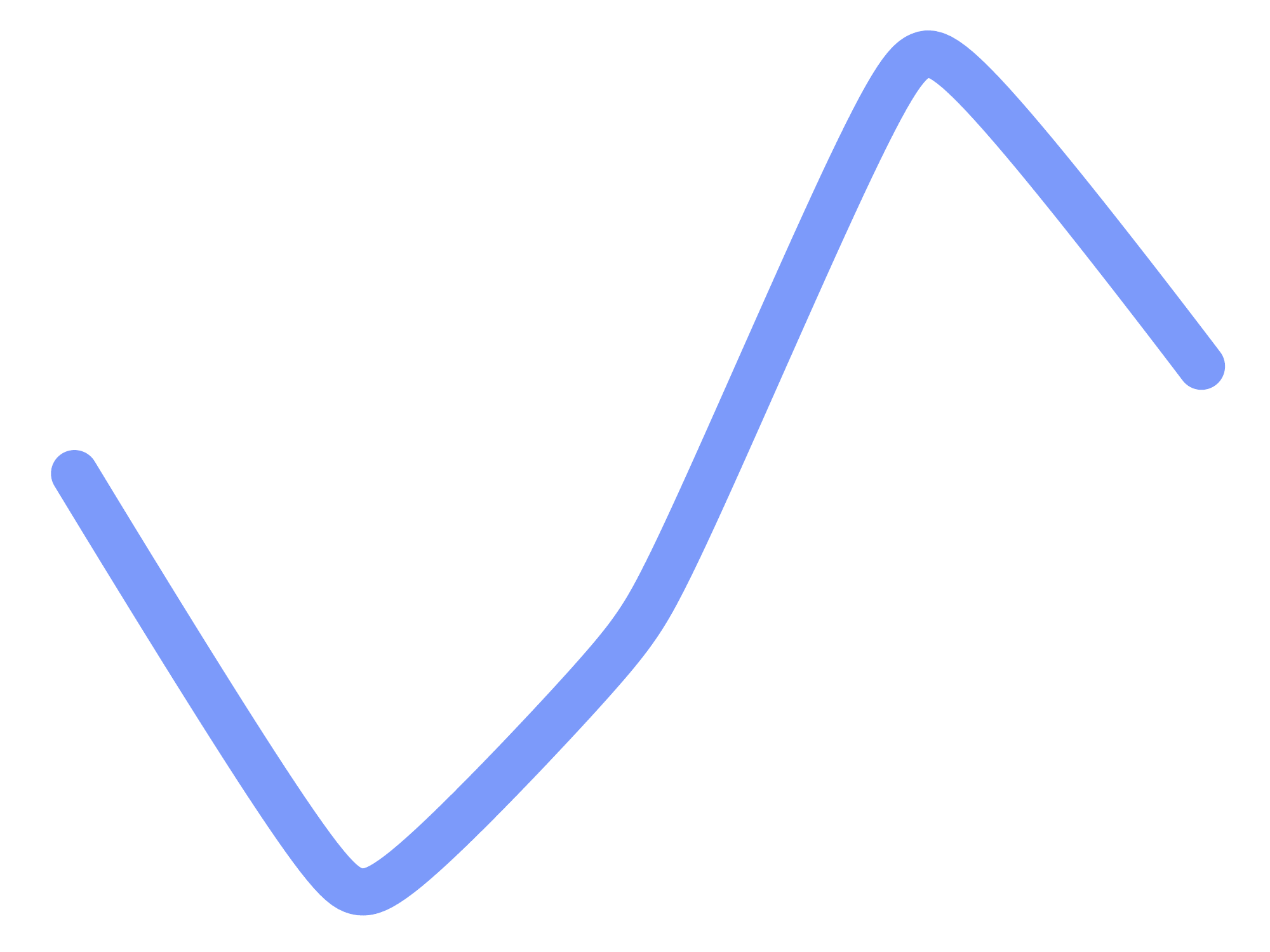}}
				\end{minipage}  & \cellcolor[rgb]{ .886,  .937,  .851}11.30 & \cellcolor[rgb]{ .945,  .969,  .929}10.91  & \cellcolor[rgb]{ .886,  .937,  .851}11.16  & \cellcolor[rgb]{ .835,  .91,  .784}\underline{11.69}  & \cellcolor[rgb]{ .886,  .937,  .851}11.40 \\
\bottomrule
			\end{tabular}%
		}
	\end{minipage}
\end{minipage}

\vspace{0.3cm}
\begin{minipage}{\textwidth}
	\begin{minipage}[h]{0.5\textwidth}
		\centering
		\makeatletter\def\@captype{table}\makeatother\caption{The empirical results of NDCG@10 with respect to different preference thresholds $t$ on MovieLens-100k. The best and second-best are highlighted in bold and underlined, respectively.}
		\label{i}   
		\scalebox{0.8}{
			\begin{tabular}{c|c|ccccc}
				\toprule
				Method & Trend & \cellcolor[rgb]{ 1.0,  1.0,  1.0}t=1 & t=2 & t=3 & t=4 & t=5 \\
				\midrule
				itemKNN & \begin{minipage}[!t]{0.1\columnwidth}
					\centering
					\raisebox{-.5\height}{\includegraphics[width=\textwidth, height=0.3\textwidth]{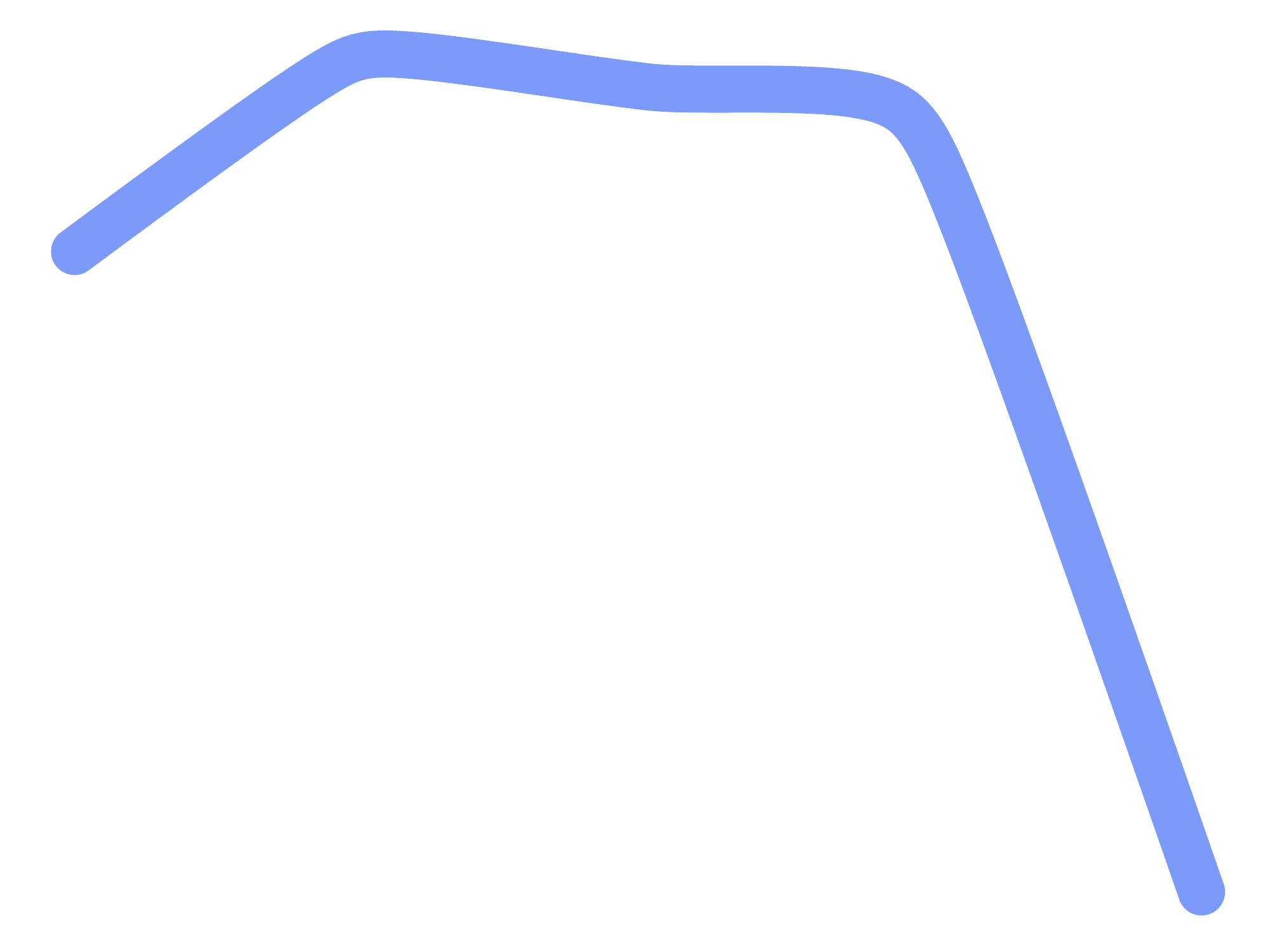}}
				\end{minipage}  & \cellcolor[rgb]{ 1.0,  1.0,  1.0}15.29 & 19.57 & 18.95 & 17.98 & 1.16 \\
				GMF & \begin{minipage}[!t]{0.1\columnwidth}
					\centering
					\raisebox{-.5\height}{\includegraphics[width=\textwidth, height=0.3\textwidth]{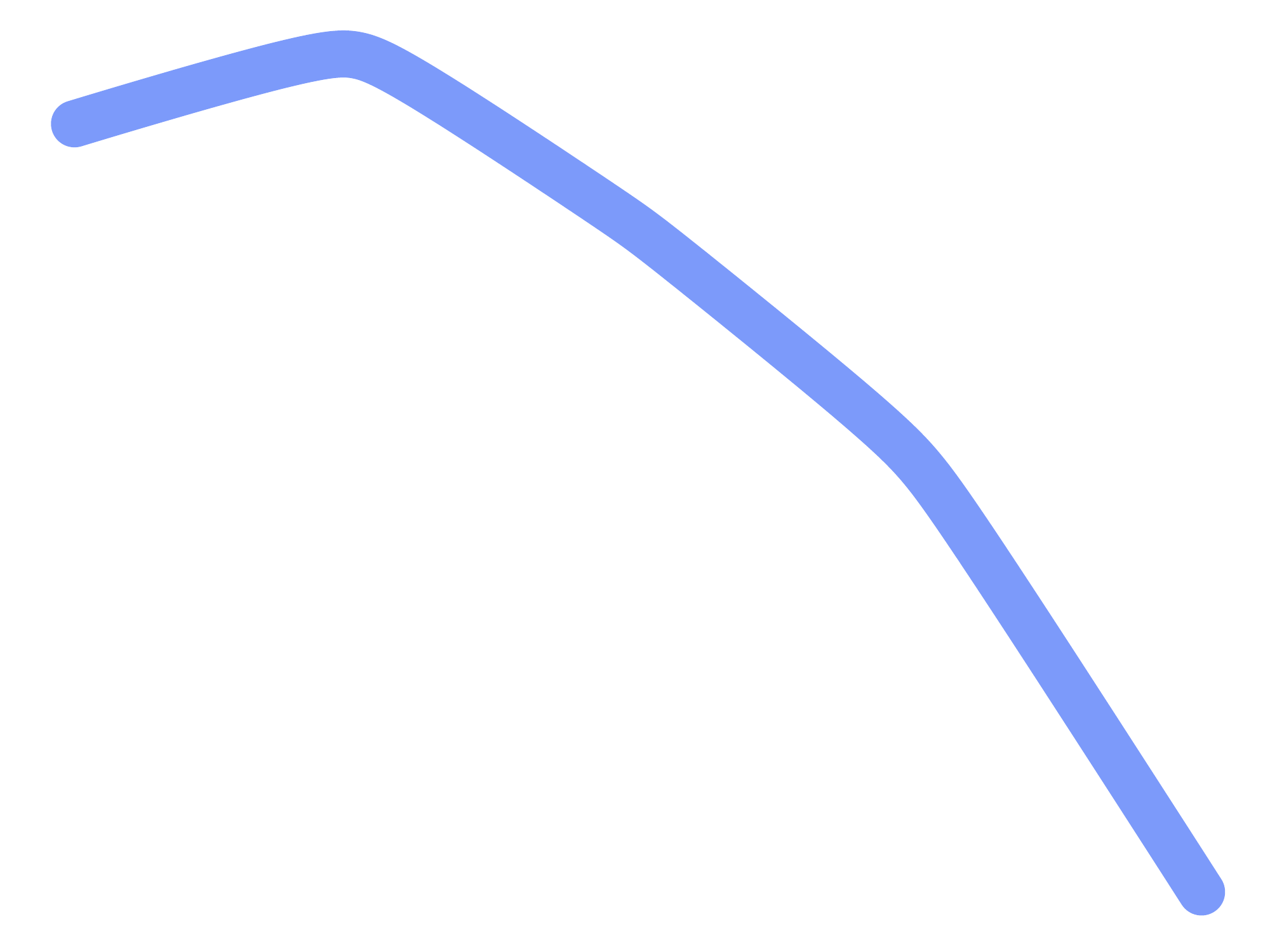}}
				\end{minipage}   & \cellcolor[rgb]{ .965,  .976,  .988}22.60 & \cellcolor[rgb]{ .906,  .941,  .973}23.15  & \cellcolor[rgb]{ .965,  .976,  .988}21.76  & \cellcolor[rgb]{ .965,  .976,  .988}19.86  & \cellcolor[rgb]{ .906,  .941,  .973}16.45  \\
				MLP & \begin{minipage}[!t]{0.1\columnwidth}
					\centering
					\raisebox{-.5\height}{\includegraphics[width=\textwidth, height=0.3\textwidth]{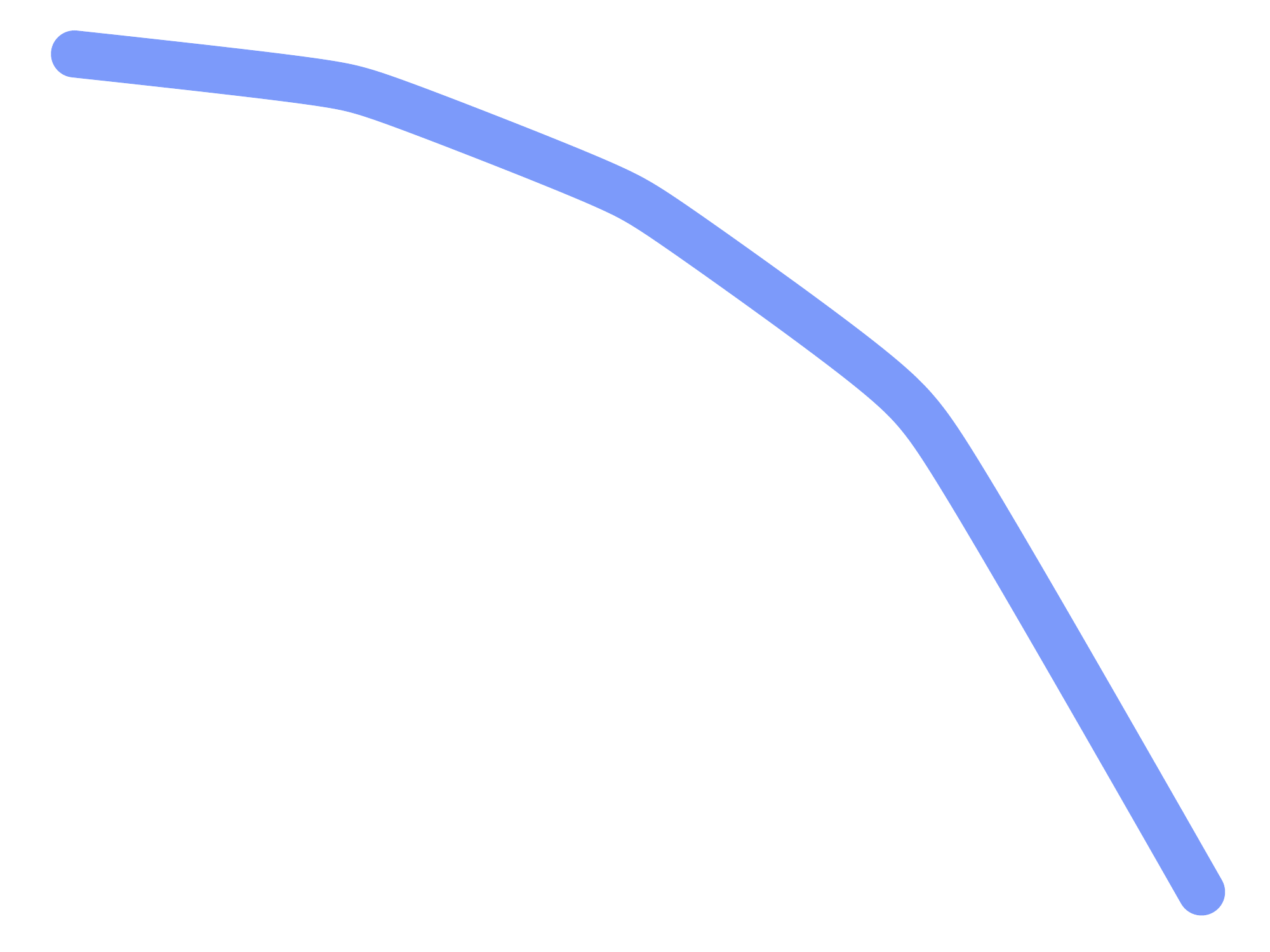}}
				\end{minipage}   & \cellcolor[rgb]{ .965,  .976,  .988}21.89  & \cellcolor[rgb]{ .965,  .976,  .988}21.62  & \cellcolor[rgb]{ .965,  .976,  .988}20.77  & \cellcolor[rgb]{ .965,  .976,  .988}19.15  & \cellcolor[rgb]{ .906,  .941,  .973}15.54  \\
				NCF & \begin{minipage}[!t]{0.1\columnwidth}
					\centering
					\raisebox{-.5\height}{\includegraphics[width=\textwidth, height=0.3\textwidth]{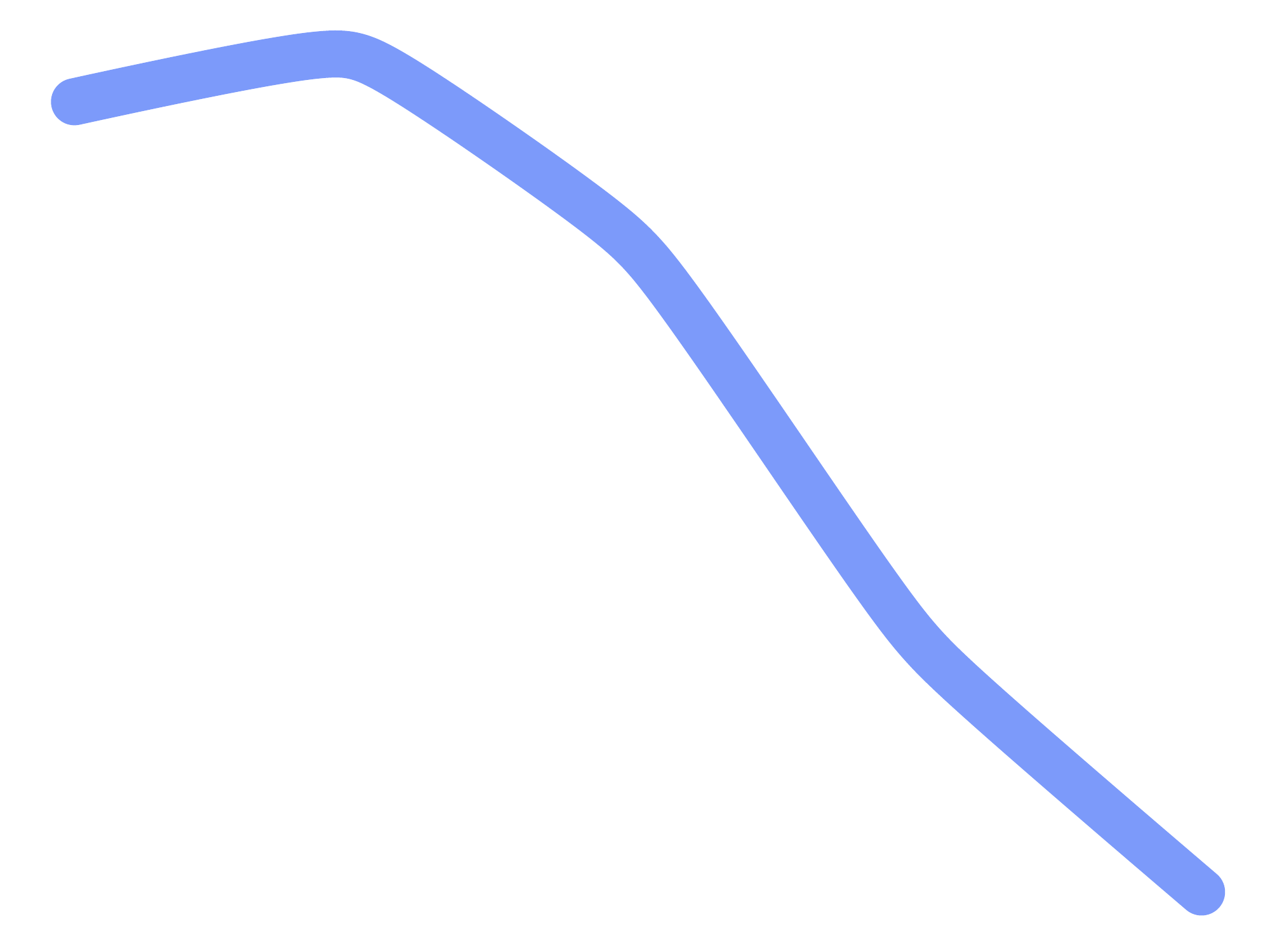}}
				\end{minipage}   & \cellcolor[rgb]{ .906,  .941,  .973}27.58  & \cellcolor[rgb]{ .843,  .906,  .961}28.20 & \cellcolor[rgb]{ .906,  .941,  .973}25.56  & \cellcolor[rgb]{ .906,  .941,  .973}20.20 & \cellcolor[rgb]{ .906,  .941,  .973}16.78  \\
\bottomrule
			\end{tabular}%
		}
	\end{minipage}
	\hspace{0.05cm}
	\begin{minipage}[h]{0.5\textwidth}
		\centering
		\makeatletter\def\@captype{table}\makeatother\caption{The empirical results of P@20 with respect to different preference thresholds $t$ on MovieLens-100k. The best and second-best are highlighted in bold and underlined, respectively.}
		\label{j}   
		\scalebox{0.8}{
			\begin{tabular}{c|c|ccccc}
				\toprule
				Method & Trend & \cellcolor[rgb]{ 1.0,  1.0,  1.0}t=1 & t=2 & t=3 & t=4 & t=5 \\
				\midrule
				itemKNN & \begin{minipage}[!t]{0.1\columnwidth}
					\centering
					\raisebox{-.5\height}{\includegraphics[width=\textwidth, height=0.3\textwidth]{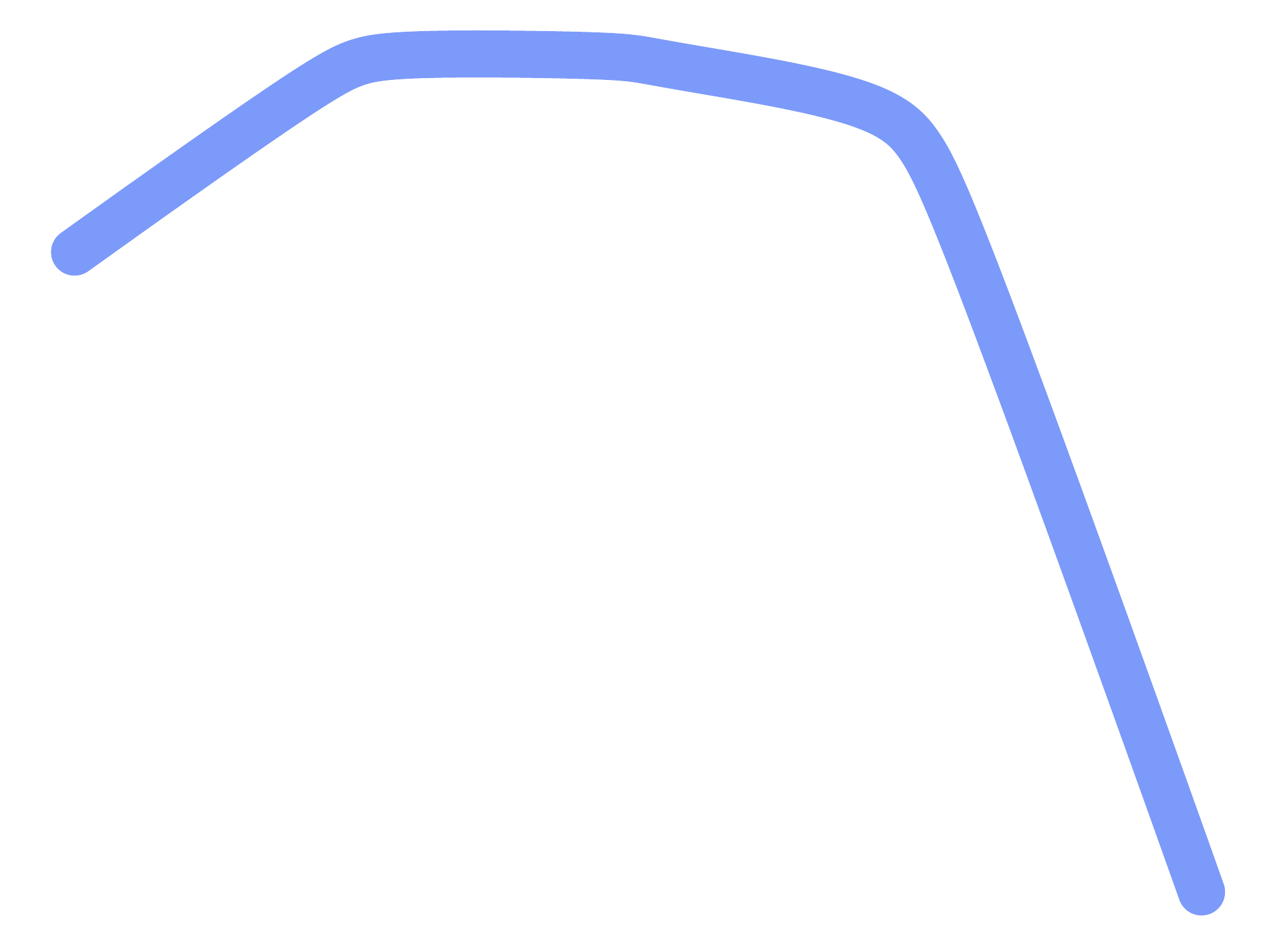}}
				\end{minipage}   & \cellcolor[rgb]{ 1.0,  1.0,  1.0}17.90 & \cellcolor[rgb]{ .906,  .941,  .973}22.94  & \cellcolor[rgb]{ .965,  .976,  .988}23.04  & \cellcolor[rgb]{ .965,  .976,  .988}20.72  & 0.87 \\
				GMF & \begin{minipage}[!t]{0.1\columnwidth}
					\centering
					\raisebox{-.5\height}{\includegraphics[width=\textwidth, height=0.3\textwidth]{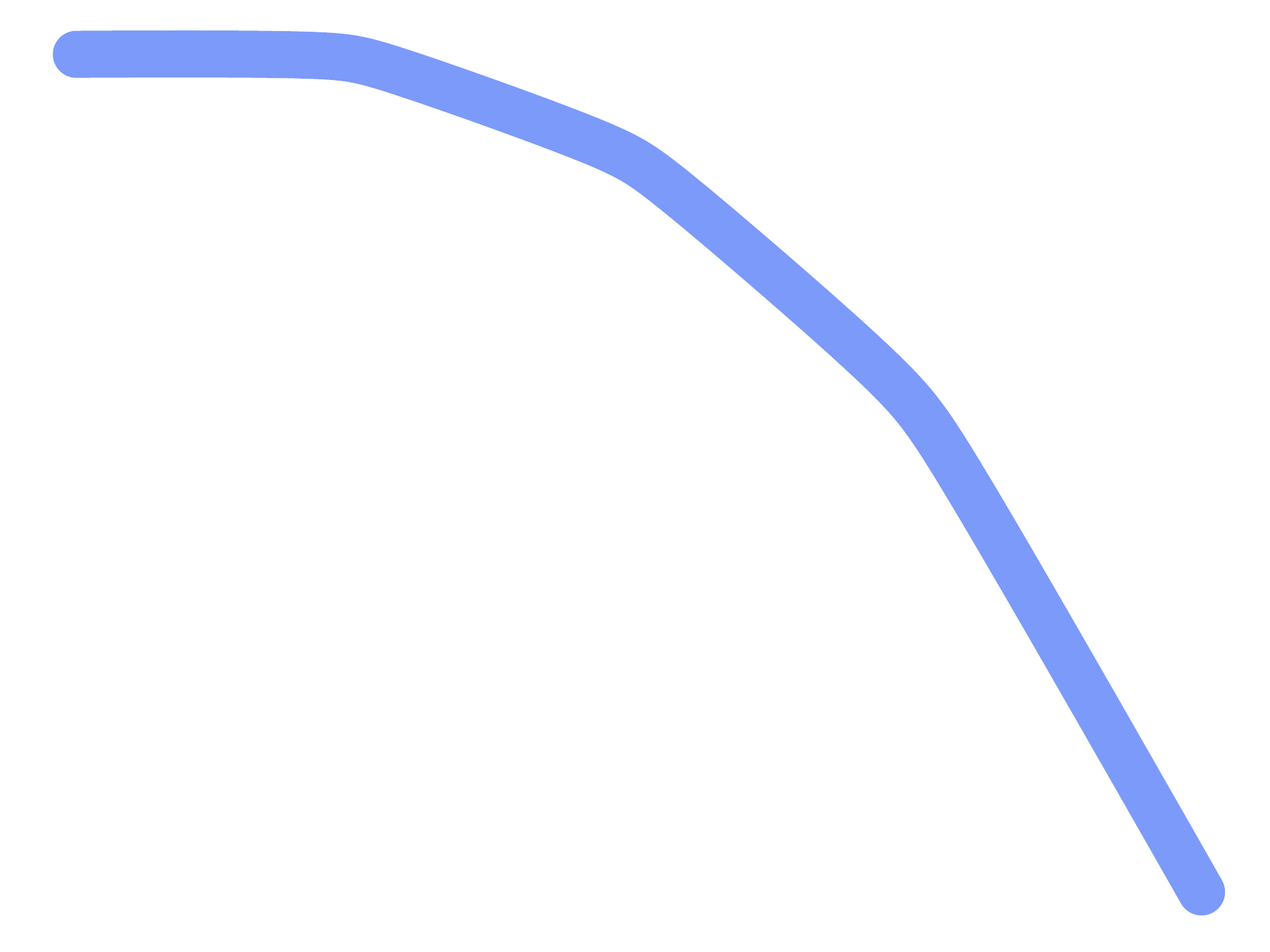}}
				\end{minipage}  & \cellcolor[rgb]{ .906,  .941,  .973}23.69  & \cellcolor[rgb]{ .906,  .941,  .973}23.65  & \cellcolor[rgb]{ .965,  .976,  .988}22.84  & \cellcolor[rgb]{ .965,  .976,  .988}20.90 & \cellcolor[rgb]{ .906,  .941,  .973}17.17  \\
				MLP &  \begin{minipage}[!t]{0.1\columnwidth}
					\centering
					\raisebox{-.5\height}{\includegraphics[width=\textwidth, height=0.3\textwidth]{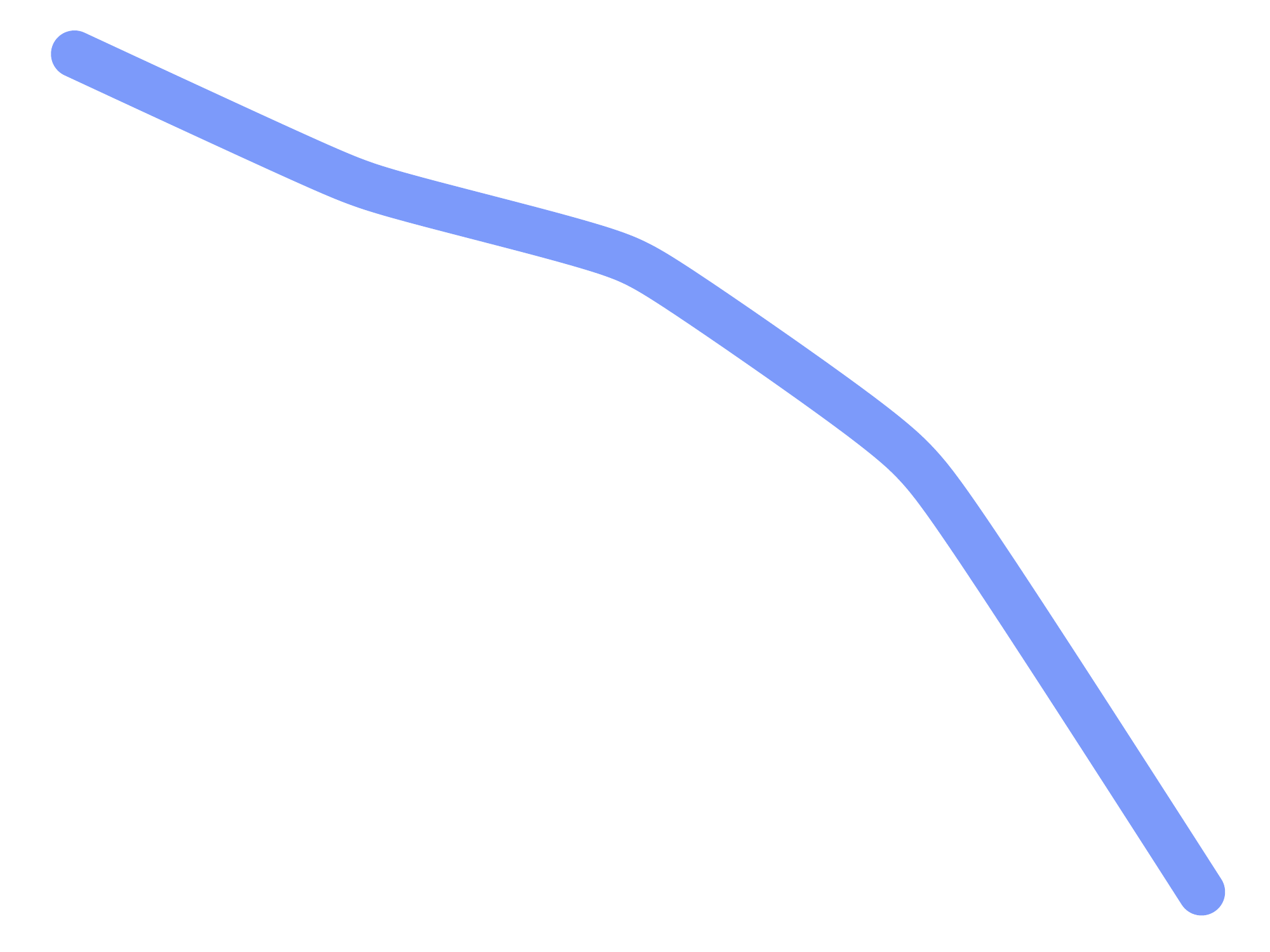}}
				\end{minipage} & \cellcolor[rgb]{ .906,  .941,  .973}23.31  & \cellcolor[rgb]{ .906,  .941,  .973}22.24  & \cellcolor[rgb]{ .965,  .976,  .988}21.55  & 19.85 & \cellcolor[rgb]{ .906,  .941,  .973}16.30 \\
				NCF & \begin{minipage}[!t]{0.1\columnwidth}
					\centering
					\raisebox{-.5\height}{\includegraphics[width=\textwidth, height=0.3\textwidth]{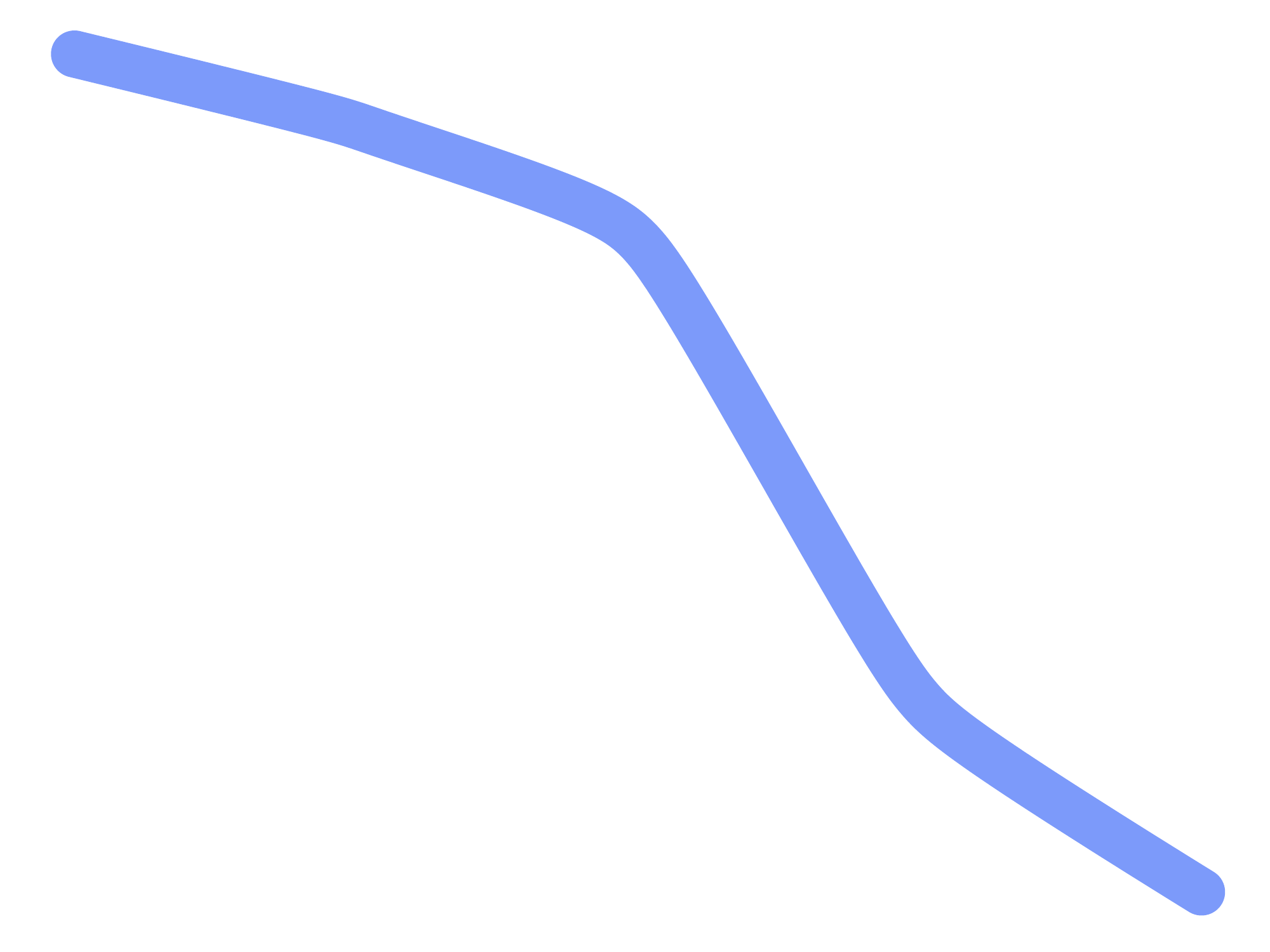}}
				\end{minipage}  & \cellcolor[rgb]{ .843,  .906,  .961}28.13  & \cellcolor[rgb]{ .906,  .941,  .973}27.38  & \cellcolor[rgb]{ .906,  .941,  .973}26.18  & \cellcolor[rgb]{ .906,  .941,  .973}21.36  & \cellcolor[rgb]{ .808,  .882,  .949}19.34  \\
\bottomrule
			\end{tabular}%
		}
	\end{minipage}
\end{minipage}

\vspace{0.3cm}
\begin{minipage}{\textwidth}
	\begin{minipage}[h]{0.5\textwidth}
		\centering
		\makeatletter\def\@captype{table}\makeatother\caption{The empirical results of R@20 with respect to different preference thresholds $t$ on MovieLens-100k. The best and second-best are highlighted in bold and underlined, respectively.}
		\label{k}   
		\scalebox{0.8}{
			\begin{tabular}{c|c|ccccc}
				\toprule
				Method & Trend & \cellcolor[rgb]{ 1.0,  1.0,  1.0} t=1 & t=2 & t=3 & t=4 & t=5 \\
				\midrule
				itemKNN & \begin{minipage}[!t]{0.1\columnwidth}
					\centering
					\raisebox{-.5\height}{\includegraphics[width=\textwidth, height=0.3\textwidth]{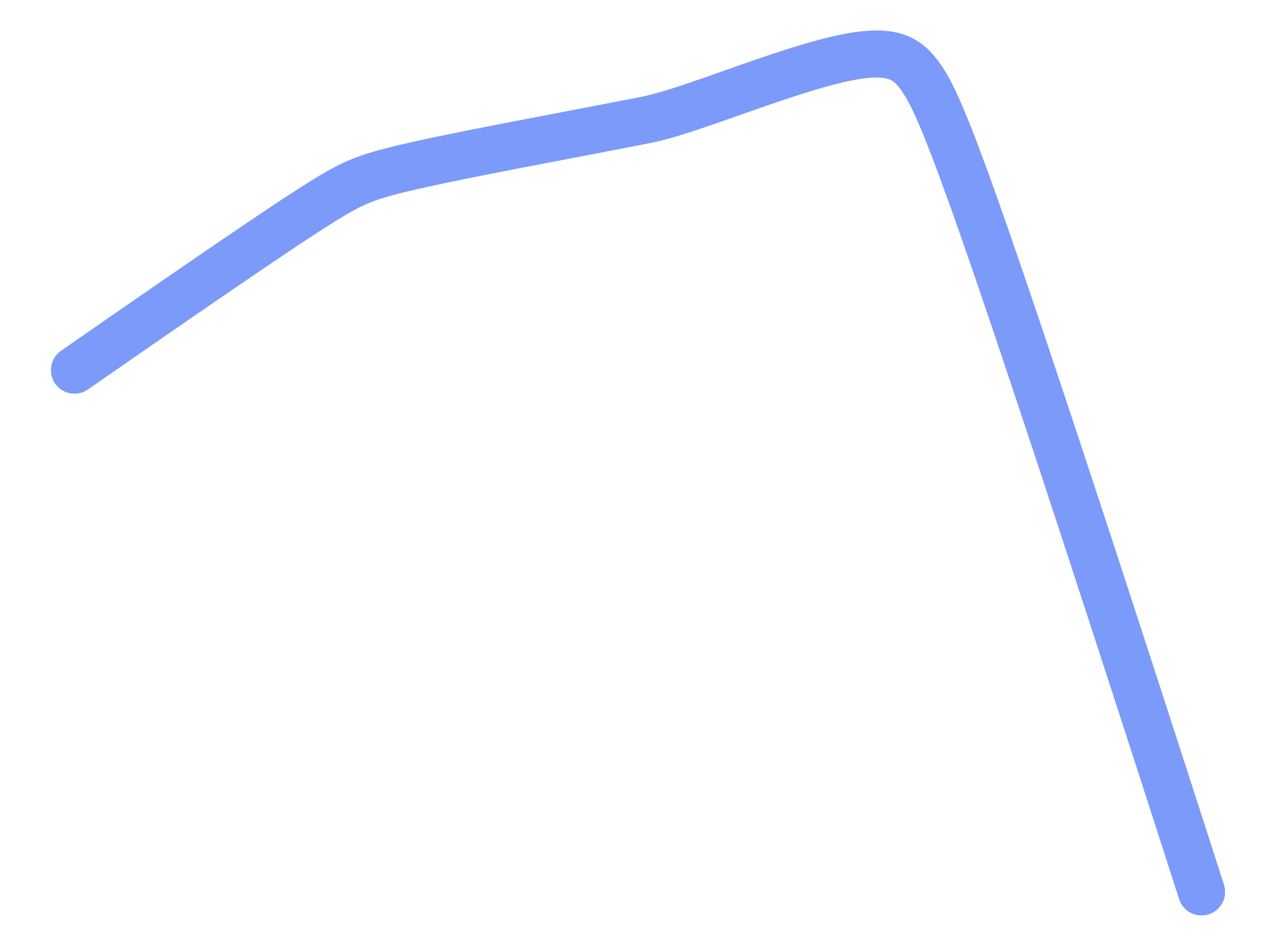}}
				\end{minipage}  & \cellcolor[rgb]{ 1.0,  1.0,  1.0}9.16 & 12.15 & \cellcolor[rgb]{ .996,  .984,  .98}13.13 & \cellcolor[rgb]{ .996,  .984,  .98}13.86 & 0.83 \\
				GMF & \begin{minipage}[!t]{0.1\columnwidth}
					\centering
					\raisebox{-.5\height}{\includegraphics[width=\textwidth, height=0.3\textwidth]{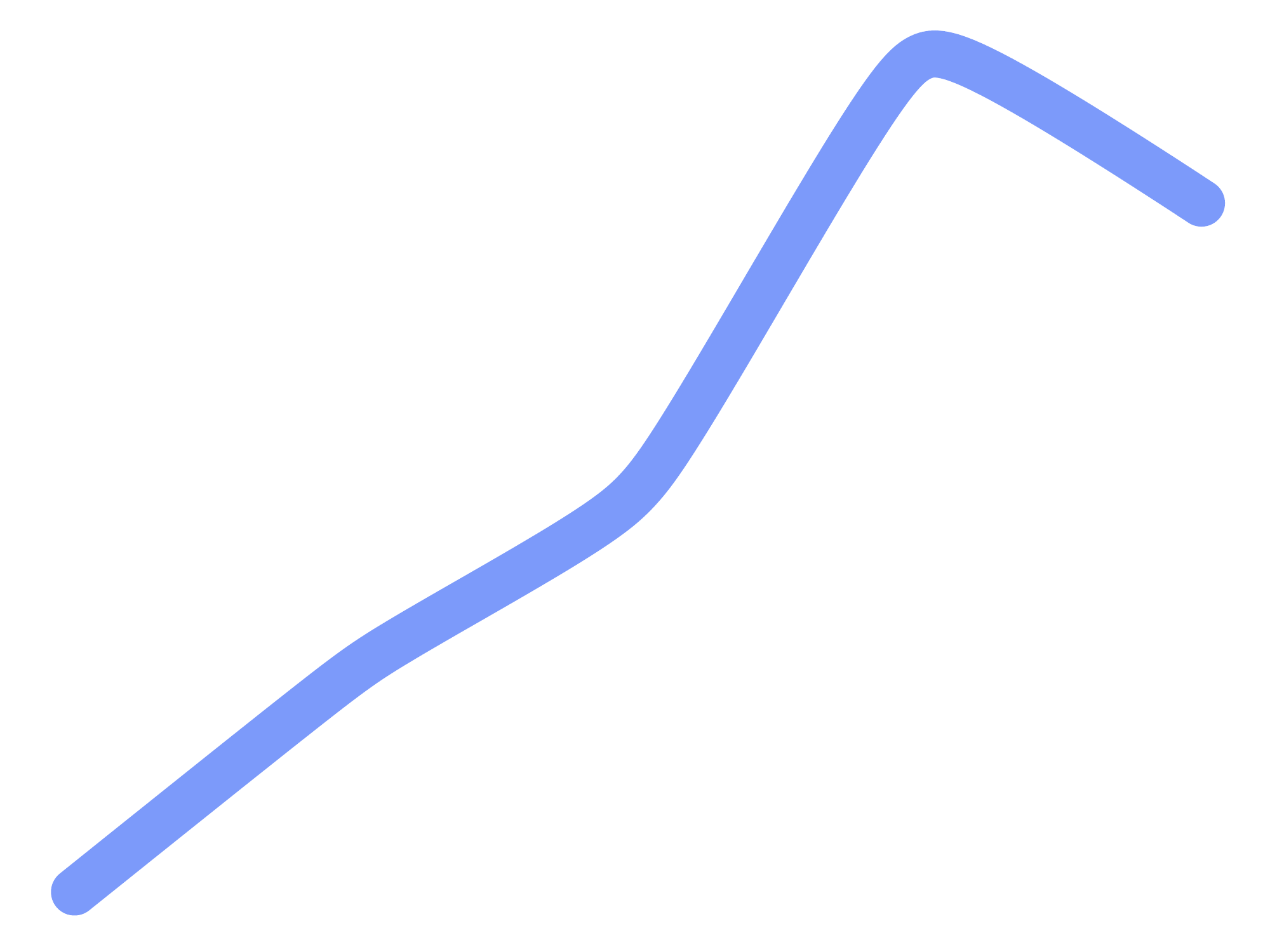}}
				\end{minipage}  & \cellcolor[rgb]{ 1.0,  1.0,  1.0}12.24 & 12.70 & \cellcolor[rgb]{ .996,  .984,  .98}13.07 & \cellcolor[rgb]{ .996,  .984,  .98}13.96 & \cellcolor[rgb]{ .973,  .918,  .918}13.66 \\
				MLP & \begin{minipage}[!t]{0.1\columnwidth}
					\centering
					\raisebox{-.5\height}{\includegraphics[width=\textwidth, height=0.3\textwidth]{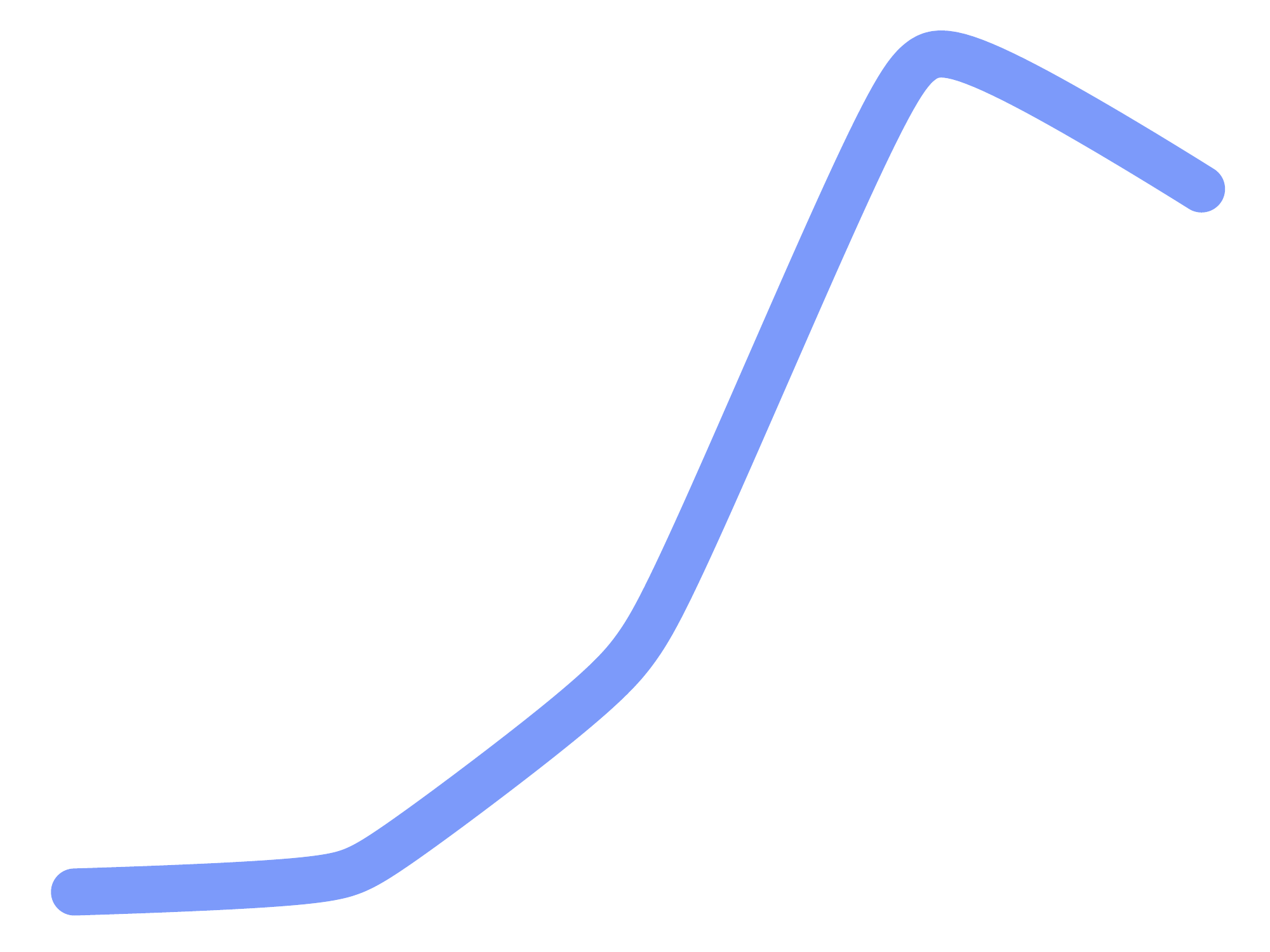}}
				\end{minipage}  & \cellcolor[rgb]{ 1.0,  1.0,  1.0}12.13 & 12.16 & 12.46 & \cellcolor[rgb]{ .996,  .984,  .98}13.24 & \cellcolor[rgb]{ .973,  .918,  .918}13.07 \\
				NCF & \begin{minipage}[!t]{0.1\columnwidth}
					\centering
					\raisebox{-.5\height}{\includegraphics[width=\textwidth, height=0.3\textwidth]{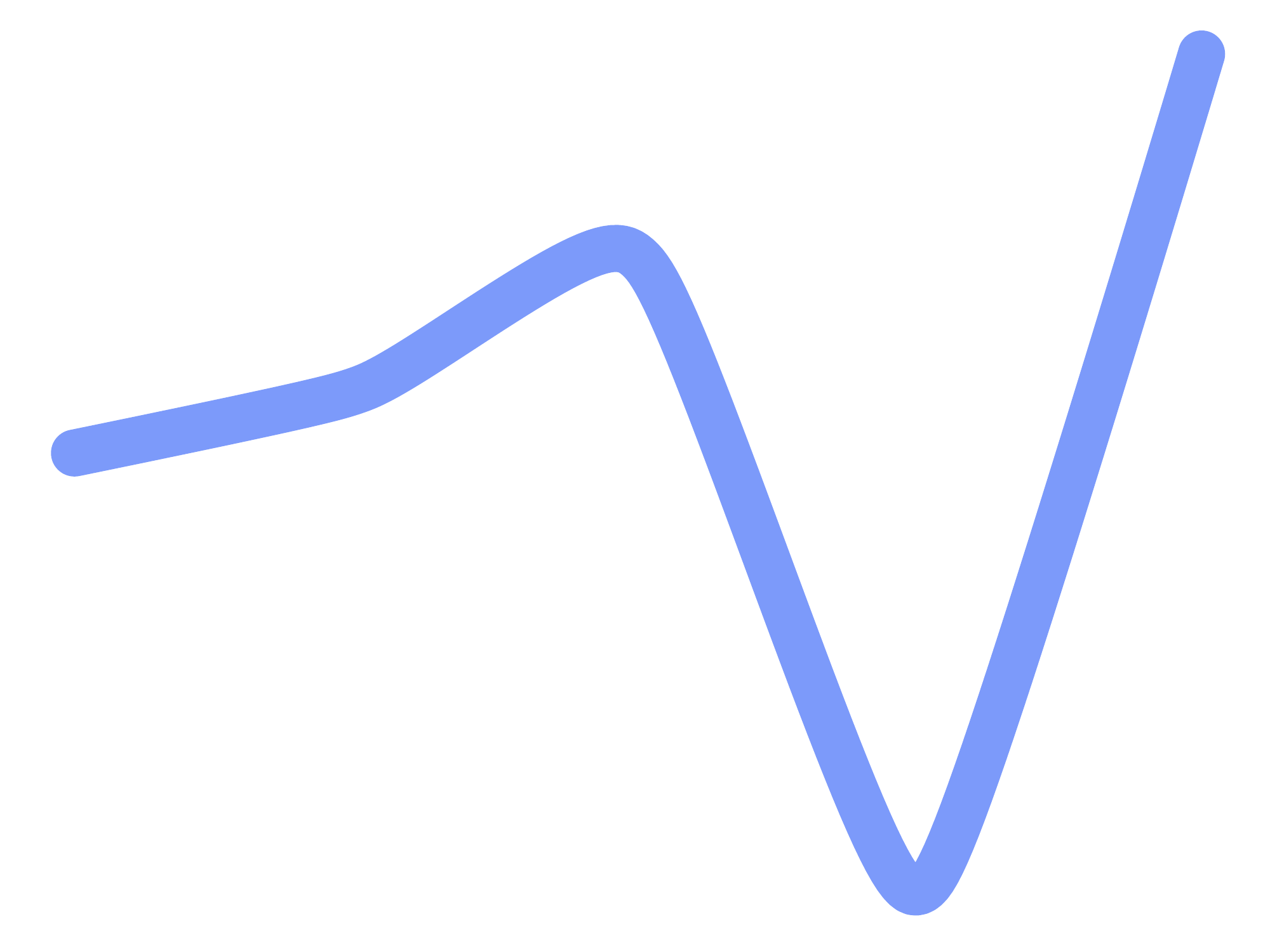}}
				\end{minipage}  & \cellcolor[rgb]{ .973,  .918,  .918}14.80 & \cellcolor[rgb]{ .973,  .918,  .918}14.88 & \cellcolor[rgb]{ .973,  .918,  .918}15.05 & \cellcolor[rgb]{ .996,  .984,  .98}14.24 & \cellcolor[rgb]{ .961,  .882,  .882}15.31 \\
\bottomrule
			\end{tabular}%
		}
	\end{minipage}
	\hspace{0.05cm}
	\begin{minipage}[h]{0.5\textwidth}
		\centering
		\makeatletter\def\@captype{table}\makeatother\caption{The empirical results of NDCG@20 with respect to different preference thresholds $t$ on MovieLens-100k. The best and second-best are highlighted in bold and underlined, respectively.}
		\label{l}   
		\scalebox{0.8}{
			\begin{tabular}{c|c|ccccc}
				\toprule
				Method & Trend & \cellcolor[rgb]{ 1.0,  1.0,  1.0}t=1 & t=2 & t=3 & t=4 & t=5 \\
				\midrule
				itemKNN & \begin{minipage}[!t]{0.1\columnwidth}
					\centering
					\raisebox{-.5\height}{\includegraphics[width=\textwidth, height=0.3\textwidth]{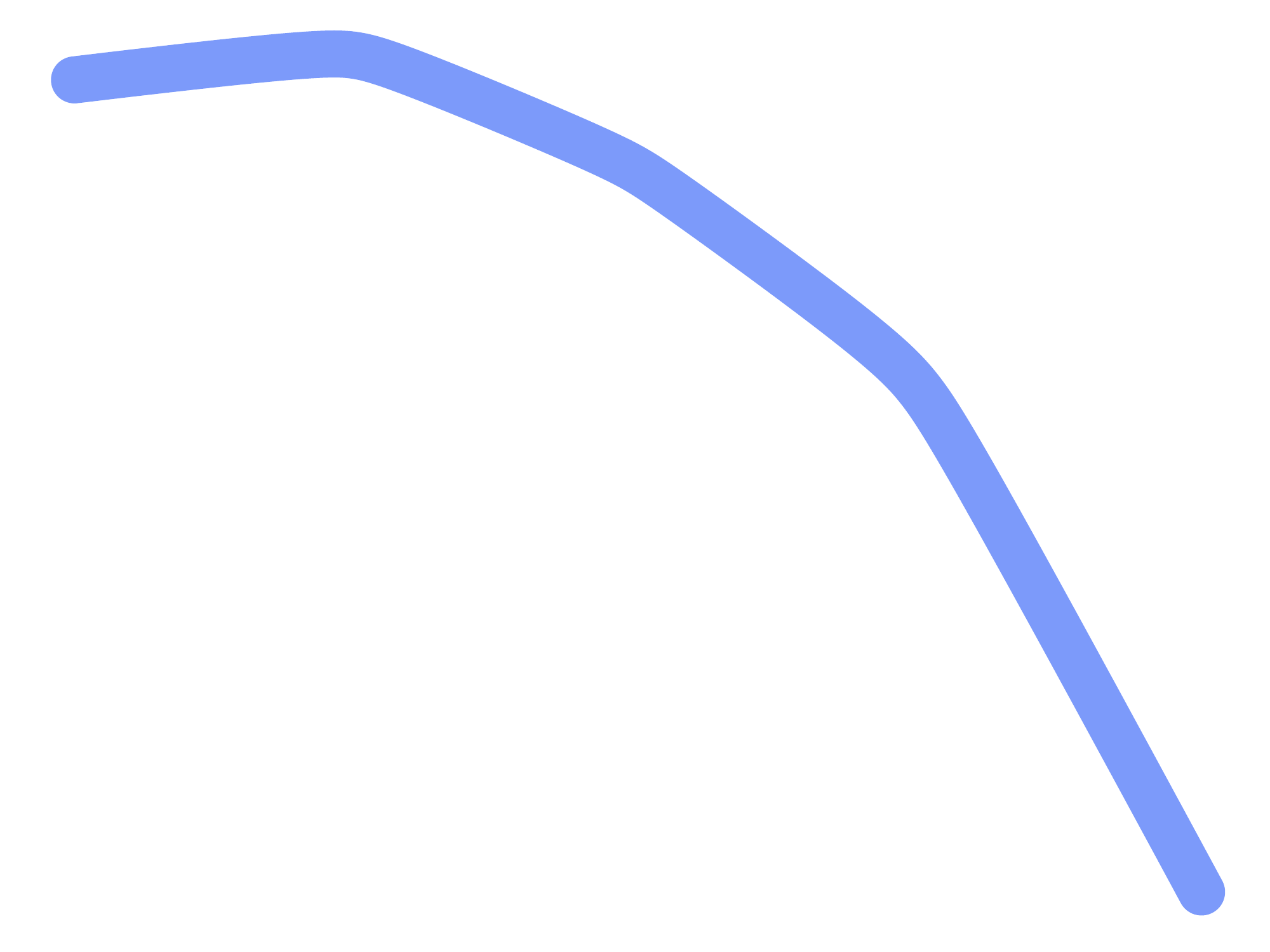}}
				\end{minipage}  & \cellcolor[rgb]{ 1.0,  1.0,  1.0}18.54 & 24.27 & 24.14 & 22.34 & 0.72 \\
				GMF & \begin{minipage}[!t]{0.1\columnwidth}
					\centering
					\raisebox{-.5\height}{\includegraphics[width=\textwidth, height=0.3\textwidth]{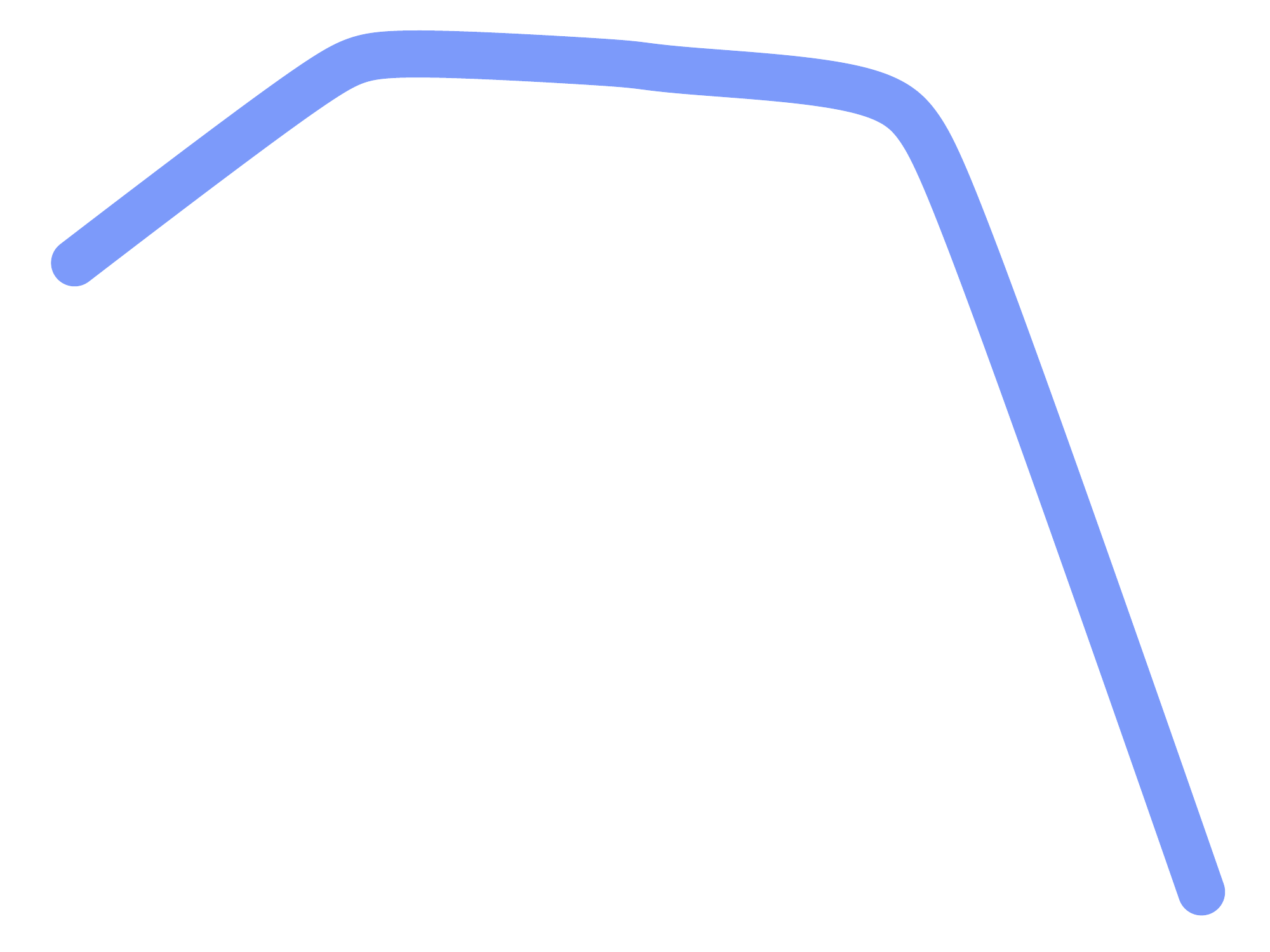}}
				\end{minipage}  & \cellcolor[rgb]{ 1.0,  1.0,  1.0}25.92 & \cellcolor[rgb]{ .996,  .984,  .98}26.15 & 25.06 & 22.99 & \cellcolor[rgb]{ .996,  .984,  .98}18.24 \\
				MLP &  \begin{minipage}[!t]{0.1\columnwidth}
					\centering
					\raisebox{-.5\height}{\includegraphics[width=\textwidth, height=0.3\textwidth]{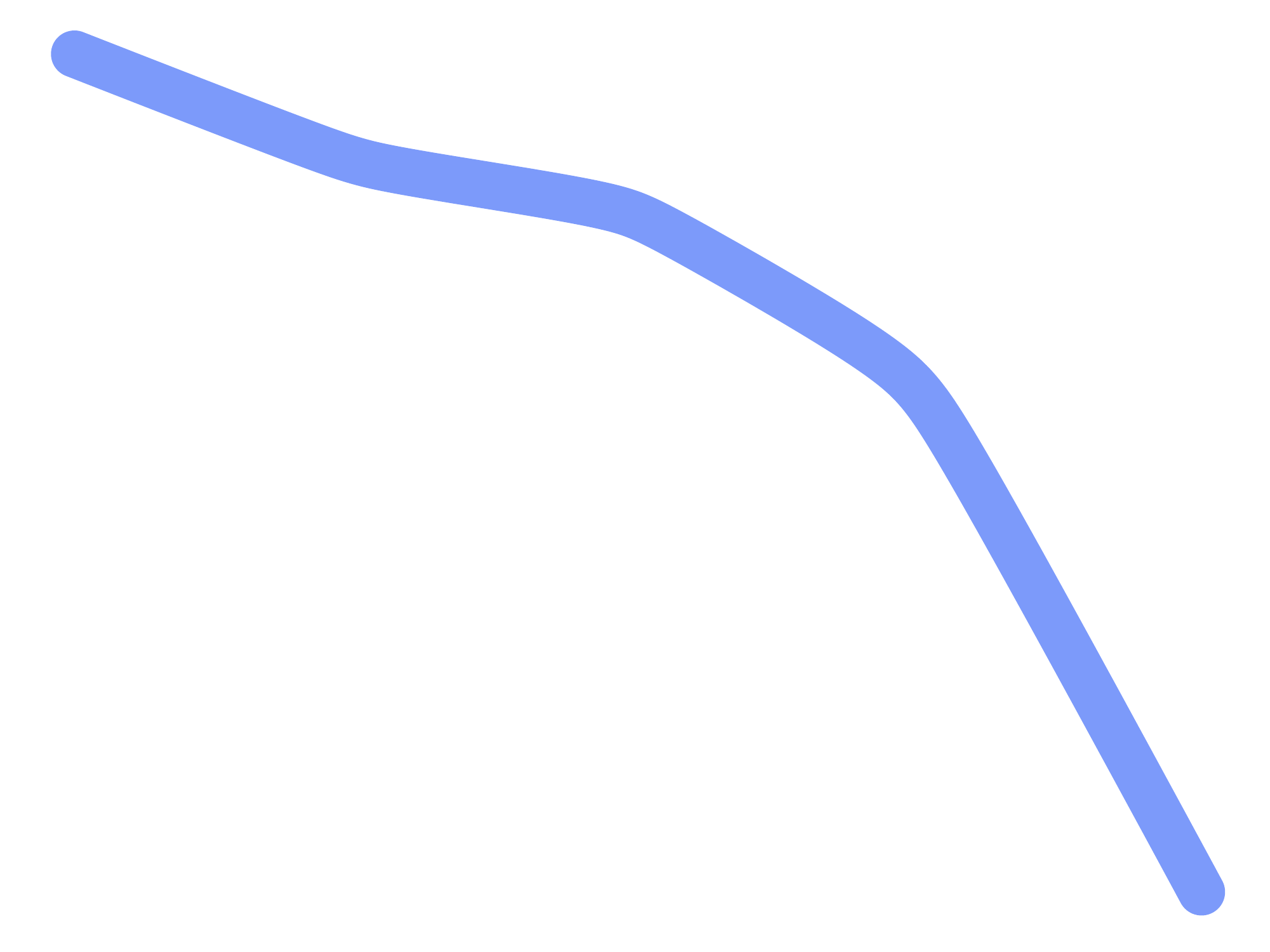}}
				\end{minipage} & \cellcolor[rgb]{ 1.0,  1.0,  1.0}24.83 & 23.85 & 23.34 & 21.72 & \cellcolor[rgb]{ .996,  .984,  .98}17.14 \\
				NCF & \begin{minipage}[!t]{0.1\columnwidth}
					\centering
					\raisebox{-.5\height}{\includegraphics[width=\textwidth, height=0.3\textwidth]{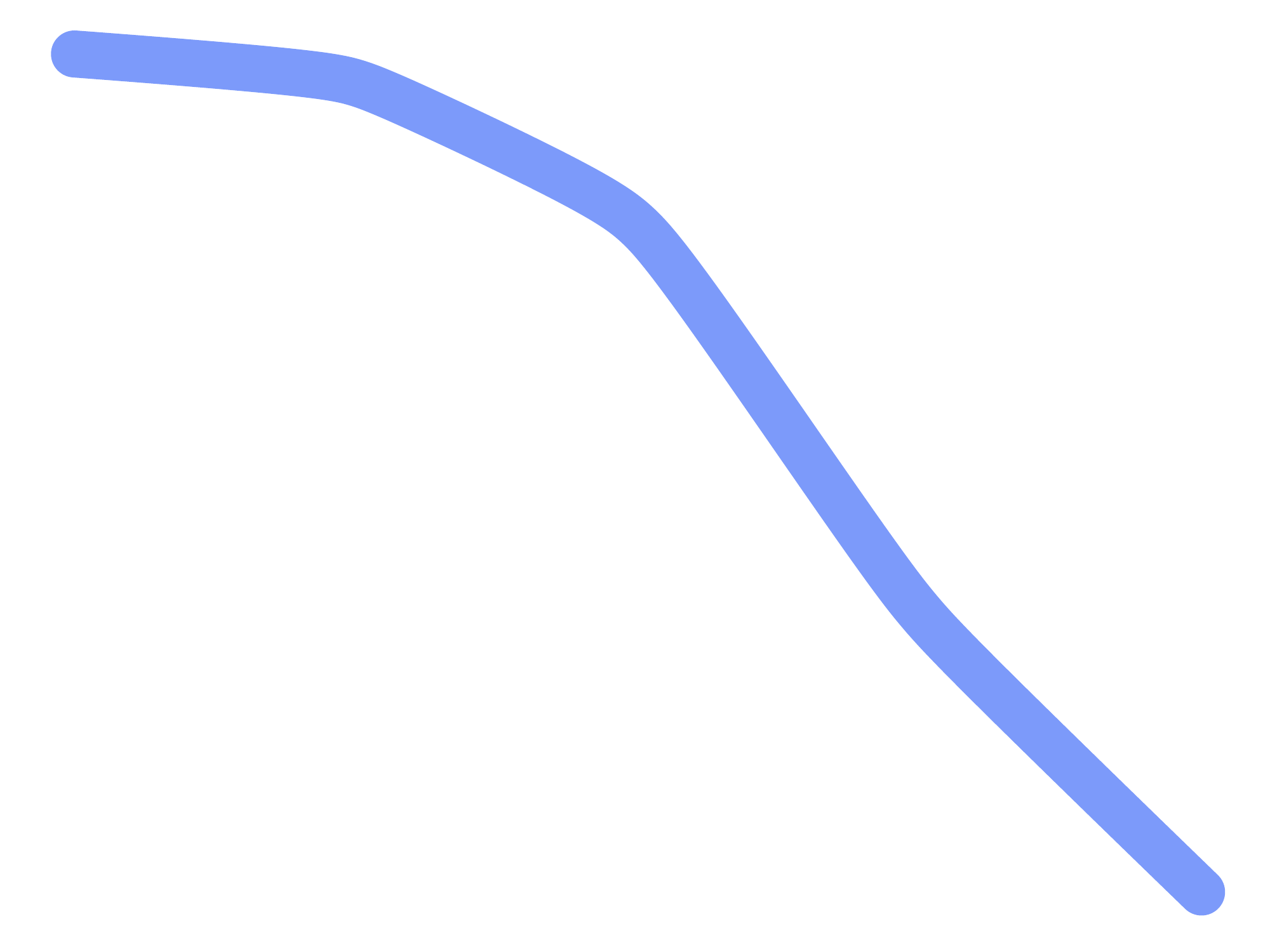}}
				\end{minipage}  & \cellcolor[rgb]{ .973,  .918,  .918}30.21 & \cellcolor[rgb]{ .973,  .918,  .918}29.85 & \cellcolor[rgb]{ .996,  .984,  .98}28.07 & \cellcolor[rgb]{ .996,  .984,  .98}23.26 & \cellcolor[rgb]{ .973,  .918,  .918}19.77 \\
				EHCF & \begin{minipage}[!t]{0.1\columnwidth}
					\centering
					\raisebox{-.5\height}{\includegraphics[width=\textwidth, height=0.3\textwidth]{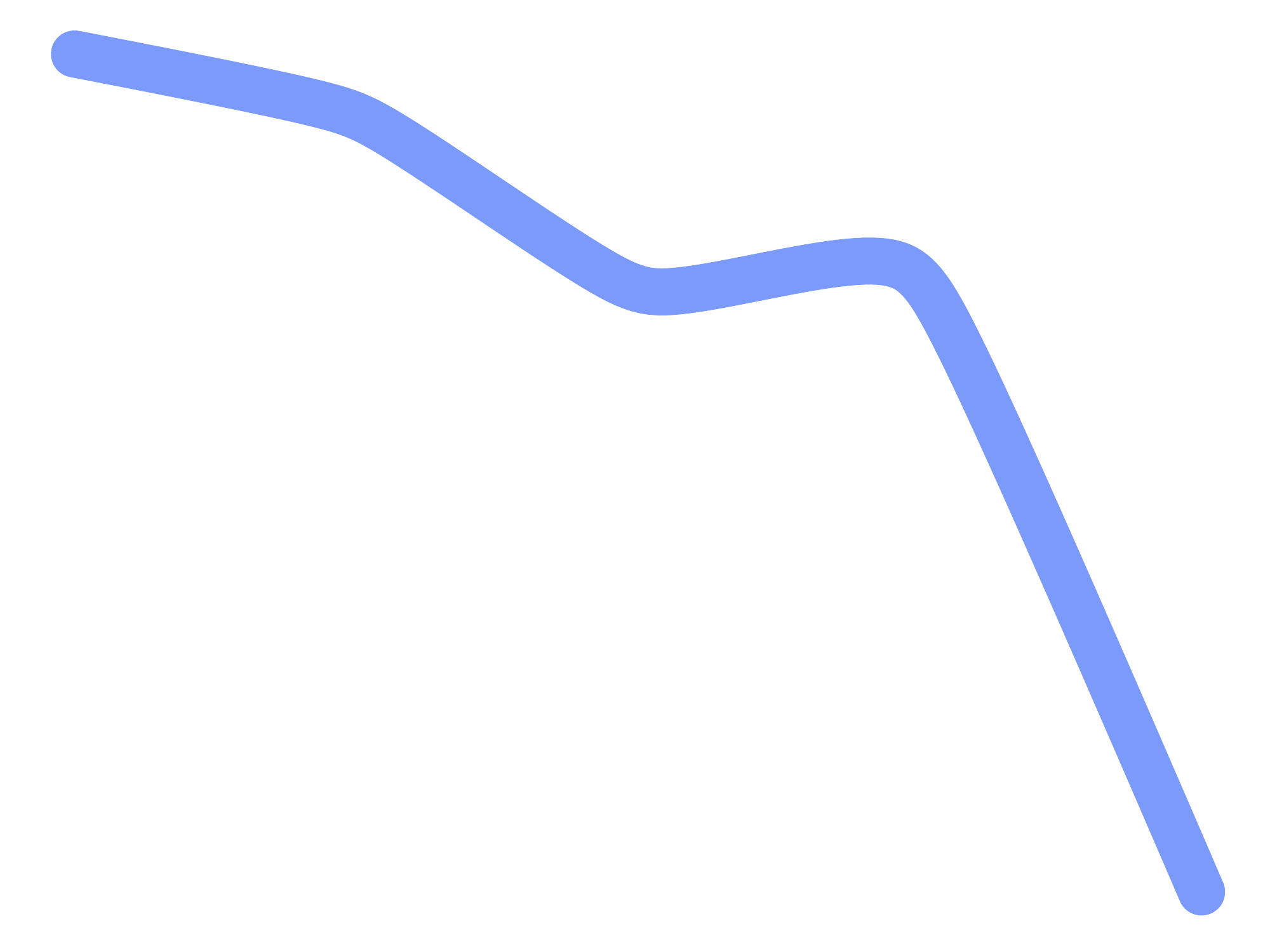}}
				\end{minipage}  & \cellcolor[rgb]{ .973,  .918,  .918}30.08 & \cellcolor[rgb]{ .973,  .918,  .918}29.31 & \cellcolor[rgb]{ .996,  .984,  .98}27.16 & \cellcolor[rgb]{ .961,  .882,  .882}27.27 & \cellcolor[rgb]{ .973,  .918,  .918}19.65 \\
\bottomrule
			\end{tabular}%
		}
	\end{minipage}
\end{minipage}

\vspace{0.3cm}
\begin{minipage}{\textwidth}
	\begin{minipage}[h]{0.5\textwidth}
		\centering
		\makeatletter\def\@captype{table}\makeatother\caption{The empirical results of MAP with respect to different preference thresholds $t$ on MovieLens-100k. The best and second-best are highlighted in bold and underlined, respectively.}
		\label{m}   
		\scalebox{0.8}{
			\begin{tabular}{c|c|ccccc}
				\toprule
				Method & Trend & \cellcolor[rgb]{ 1.0,  1.0,  1.0} t=1 & t=2 & t=3 & t=4 & t=5 \\
				\midrule
				itemKNN & \begin{minipage}[!t]{0.1\columnwidth}
					\centering
					\raisebox{-.5\height}{\includegraphics[width=\textwidth, height=0.3\textwidth]{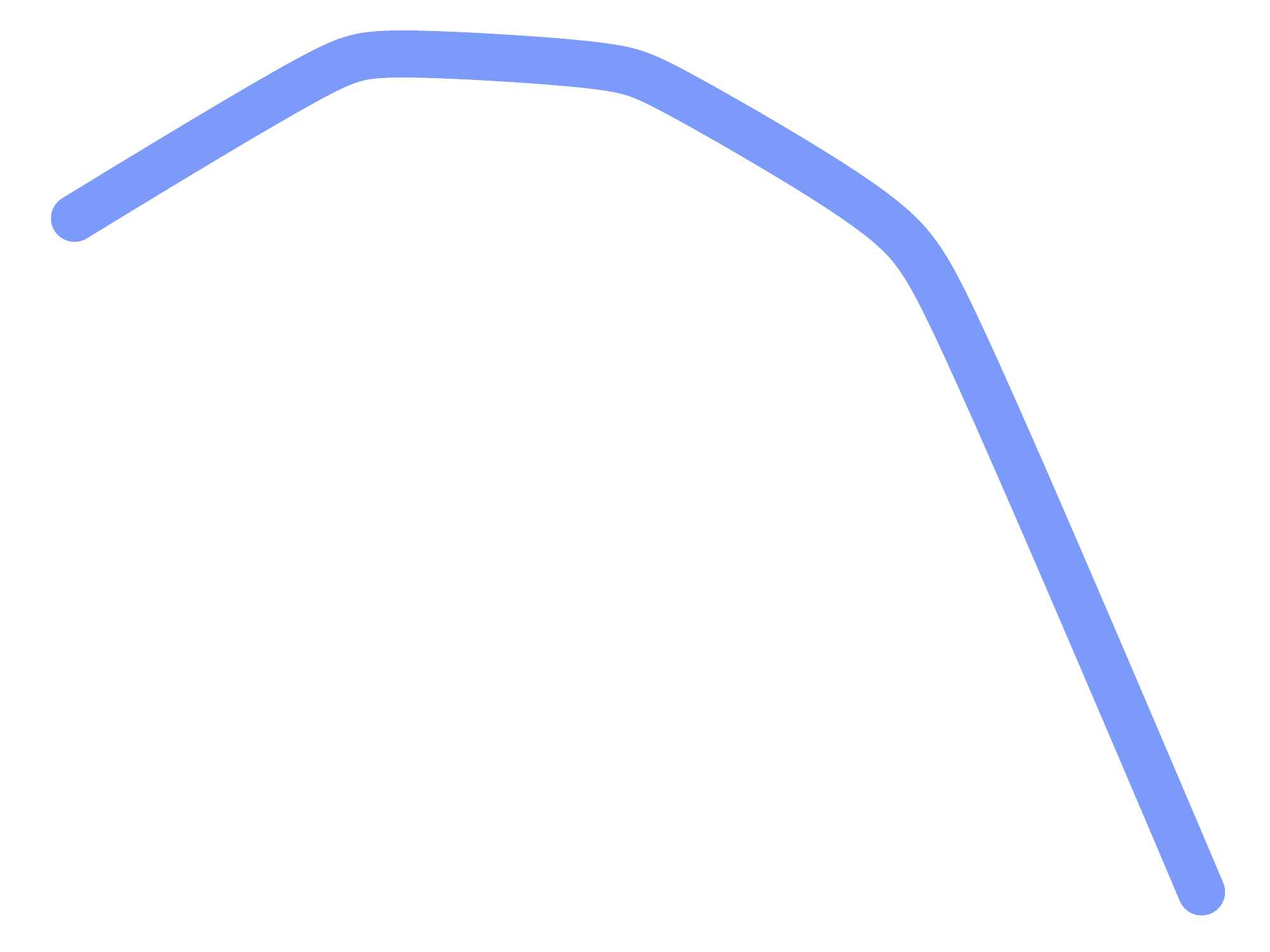}}
				\end{minipage}  & \cellcolor[rgb]{ 1.0,  1.0,  1.0} 8.81 & 10.09 & 9.96 & 8.49 & 3.43 \\
				GMF & \begin{minipage}[!t]{0.1\columnwidth}
					\centering
					\raisebox{-.5\height}{\includegraphics[width=\textwidth, height=0.3\textwidth]{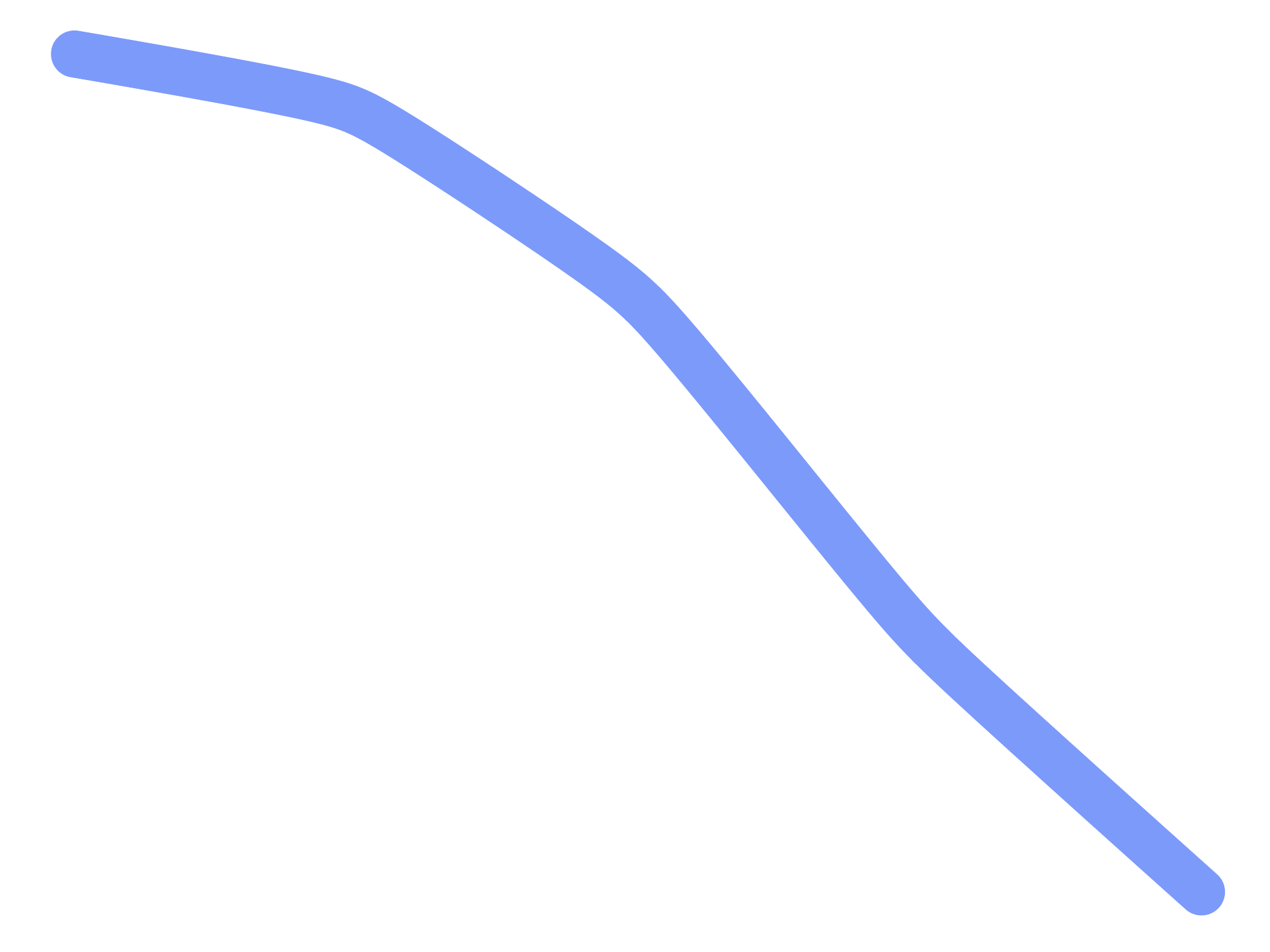}}
				\end{minipage}  & \cellcolor[rgb]{ 1.0,  1.0,  1.0} 11.24 & \cellcolor[rgb]{ .996,  .984,  .98}11.10 & \cellcolor[rgb]{ .996,  .984,  .98}10.64 & 9.82 & \cellcolor[rgb]{ .973,  .918,  .918}9.19 \\
				MLP & \begin{minipage}[!t]{0.1\columnwidth}
					\centering
					\raisebox{-.5\height}{\includegraphics[width=\textwidth, height=0.3\textwidth]{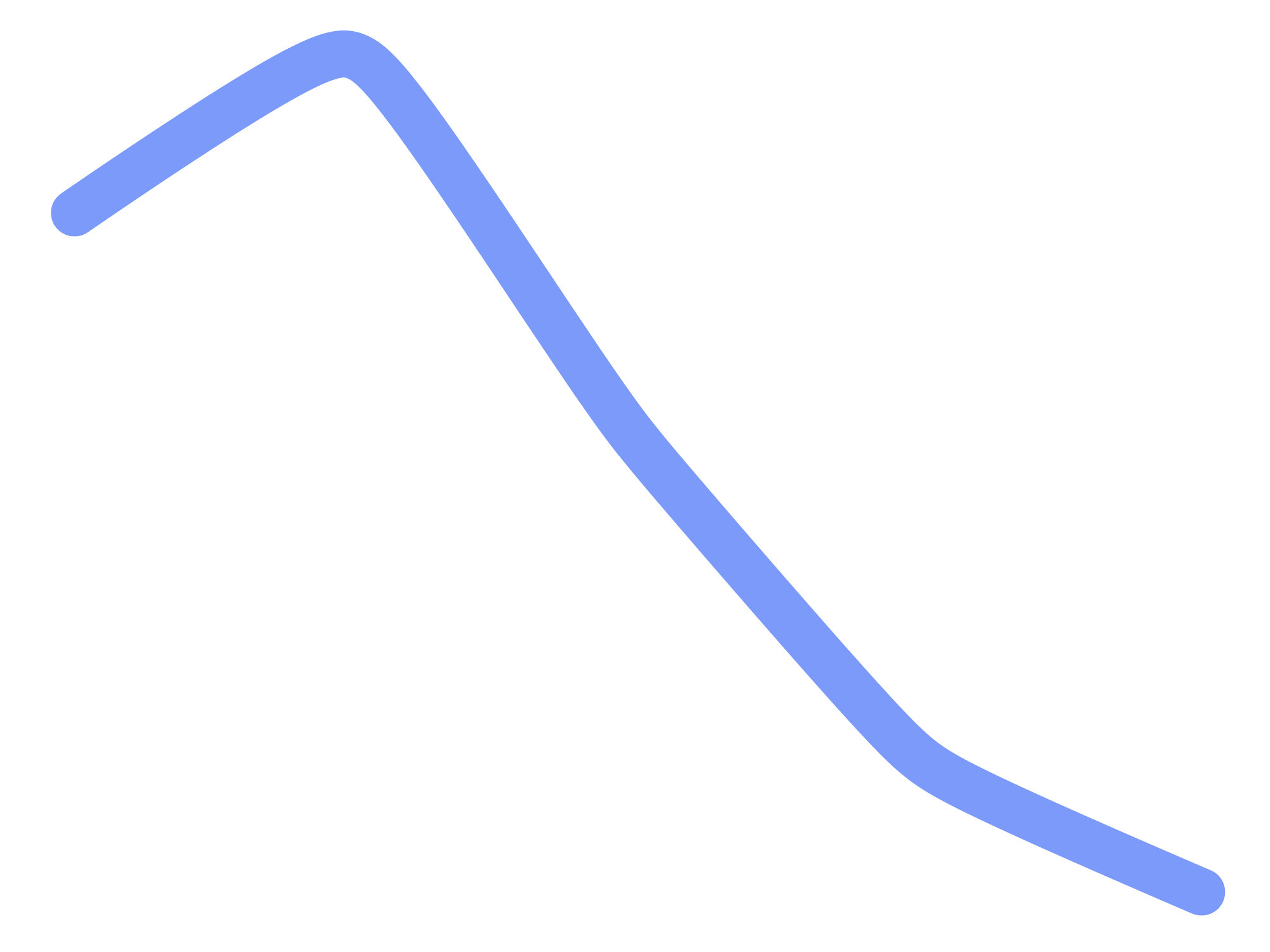}}
				\end{minipage}  & 11.52 & \cellcolor[rgb]{ .996,  .984,  .98}11.93 & \cellcolor[rgb]{ .996,  .984,  .98}10.92 & \cellcolor[rgb]{ .996,  .984,  .98}10.09 & \cellcolor[rgb]{ .961,  .882,  .882}9.74 \\
				NCF & \begin{minipage}[!t]{0.1\columnwidth}
					\centering
					\raisebox{-.5\height}{\includegraphics[width=\textwidth, height=0.3\textwidth]{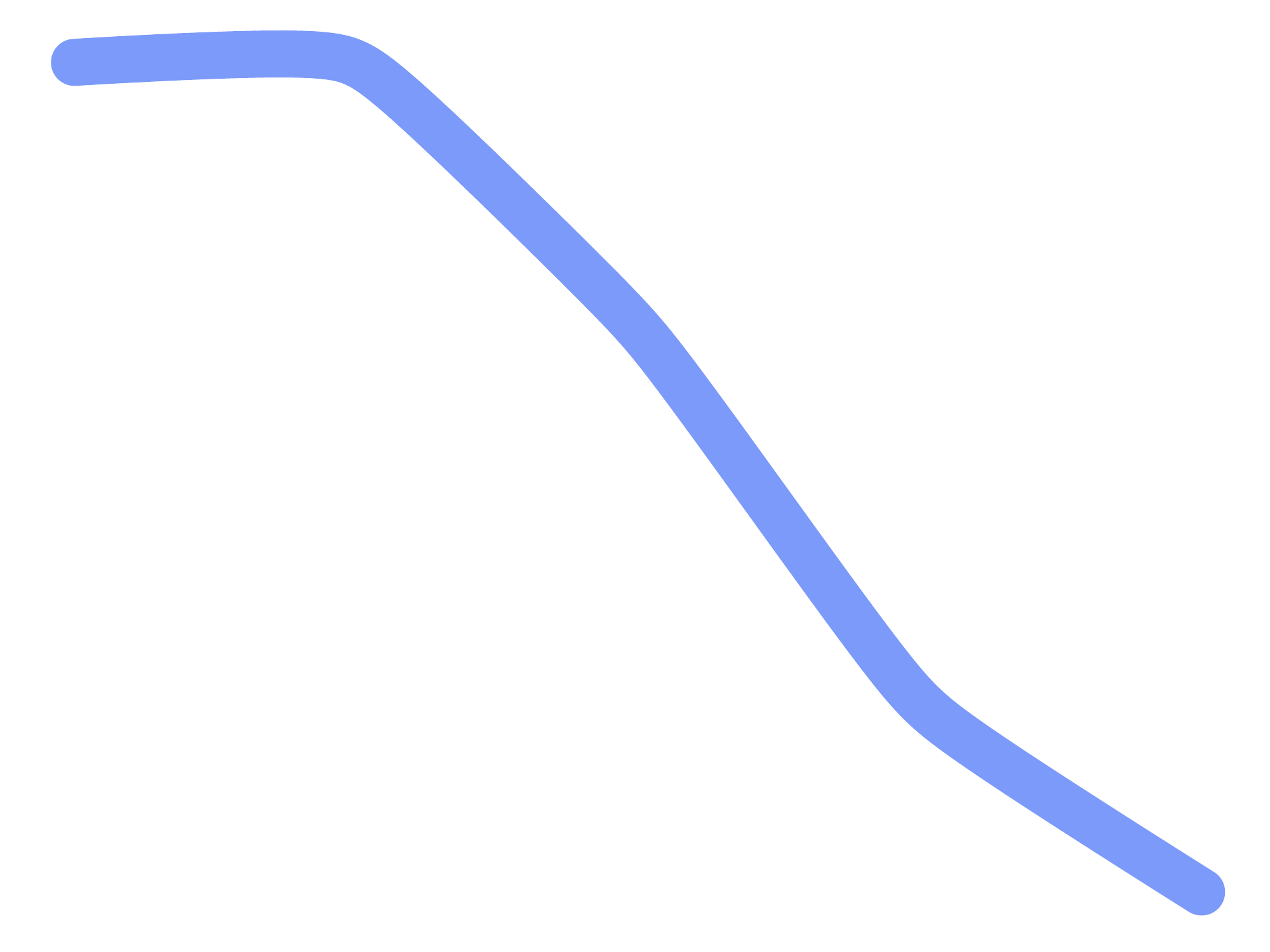}}
				\end{minipage}  & \cellcolor[rgb]{ .973,  .918,  .918}16.33 & \cellcolor[rgb]{ .973,  .918,  .918}16.32 & \cellcolor[rgb]{ .973,  .918,  .918}14.26 & \cellcolor[rgb]{ .996,  .984,  .98}11.35 & \cellcolor[rgb]{ .961,  .882,  .882}9.87 \\
\bottomrule
			\end{tabular}%
		}
	\end{minipage}
	\hspace{0.05cm}
	\begin{minipage}[h]{0.5\textwidth}
		\centering
		\makeatletter\def\@captype{table}\makeatother\caption{The empirical results of MRR with respect to different preference thresholds $t$ on MovieLens-100k. The best and second-best are highlighted in bold and underlined, respectively.}
		\label{n}   
		\scalebox{0.8}{
				\begin{tabular}{c|c|ccccc}
				\toprule
				Method & Trend & \cellcolor[rgb]{ 1.0,  1.0,  1.0} t=1 & t=2 & t=3 & t=4 & t=5 \\
				\midrule
				itemKNN & \begin{minipage}[!t]{0.1\columnwidth}
					\centering
					\raisebox{-.5\height}{\includegraphics[width=\textwidth, height=0.3\textwidth]{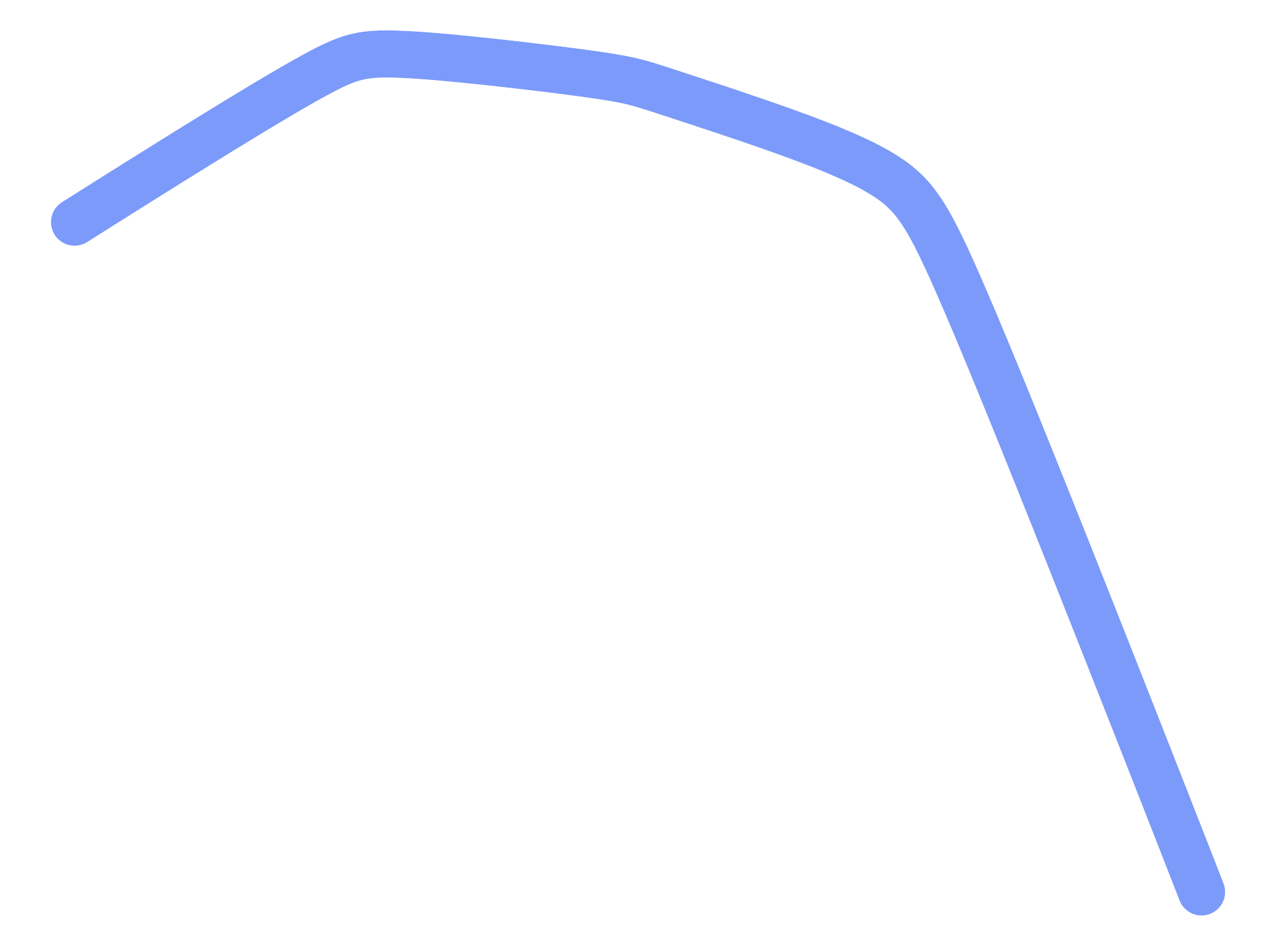}}
				\end{minipage}  & \cellcolor[rgb]{ 1.0,  1.0,  1.0} 24.27 & 28.75 & 28.05 & 24.63 & 6.10 \\
				GMF & \begin{minipage}[!t]{0.1\columnwidth}
					\centering
					\raisebox{-.5\height}{\includegraphics[width=\textwidth, height=0.3\textwidth]{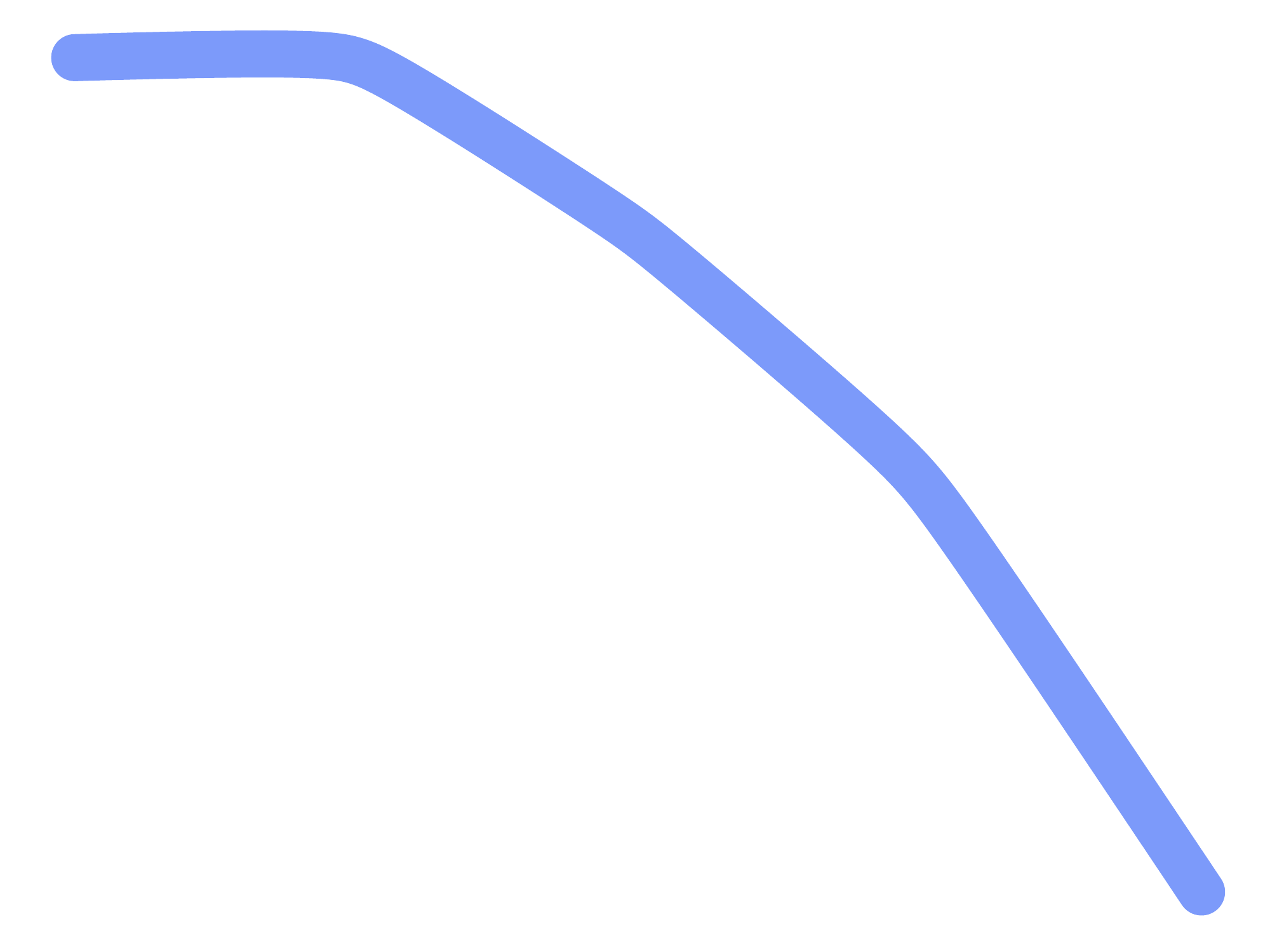}}
				\end{minipage}  & \cellcolor[rgb]{ 1.0,  1.0,  1.0} 37.44 & 37.39 & 34.76 & 31.00 & \cellcolor[rgb]{ .973,  .918,  .918}24.76 \\
				MLP & \begin{minipage}[!t]{0.1\columnwidth}
					\centering
					\raisebox{-.5\height}{\includegraphics[width=\textwidth, height=0.3\textwidth]{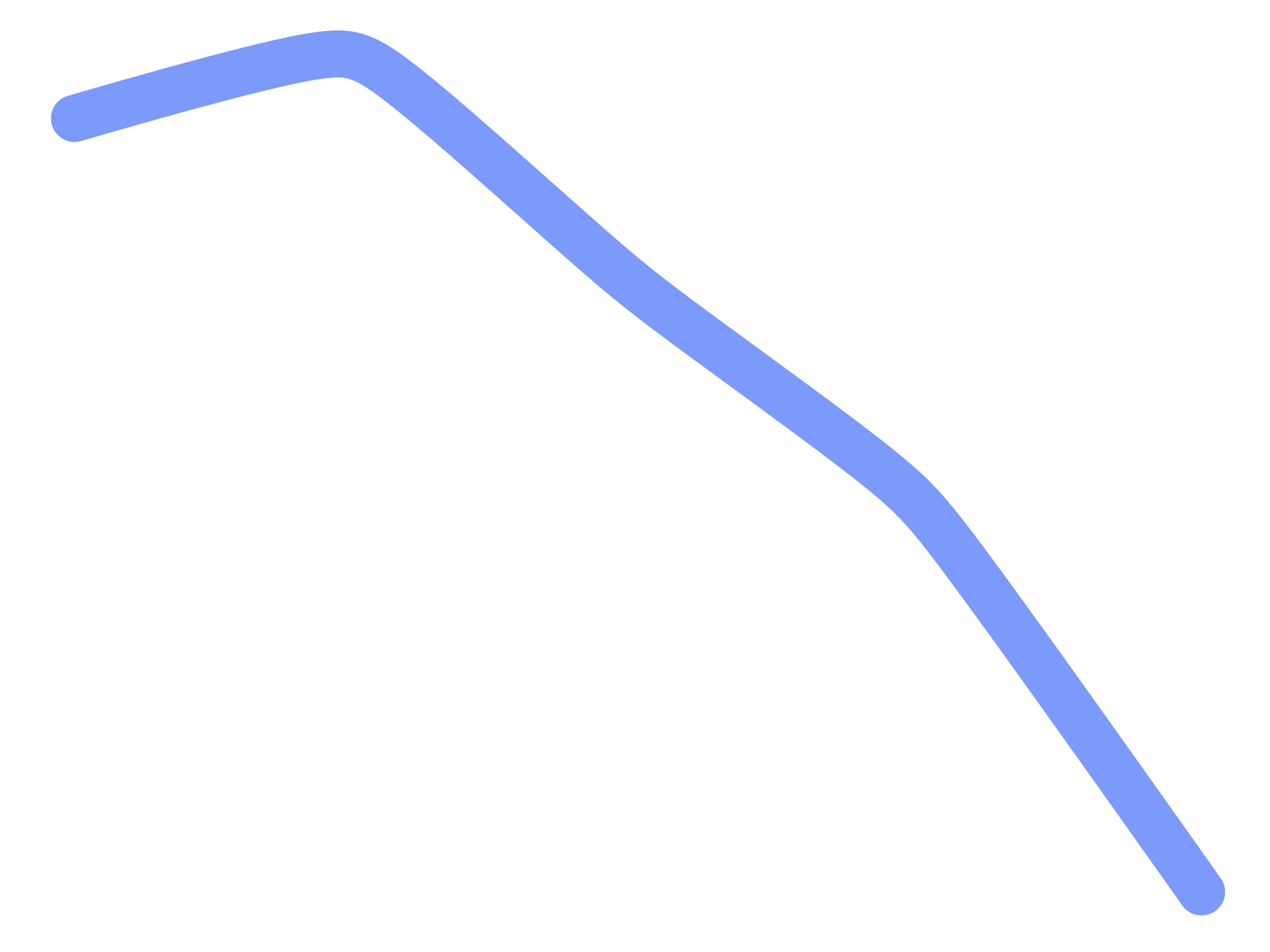}}
				\end{minipage}  &\cellcolor[rgb]{ 1.0,  1.0,  1.0} 37.99 & 38.95 & 35.36 & 31.99 & \cellcolor[rgb]{ .973,  .918,  .918}25.99 \\
				NCF & \begin{minipage}[!t]{0.1\columnwidth}
					\centering
					\raisebox{-.5\height}{\includegraphics[width=\textwidth, height=0.3\textwidth]{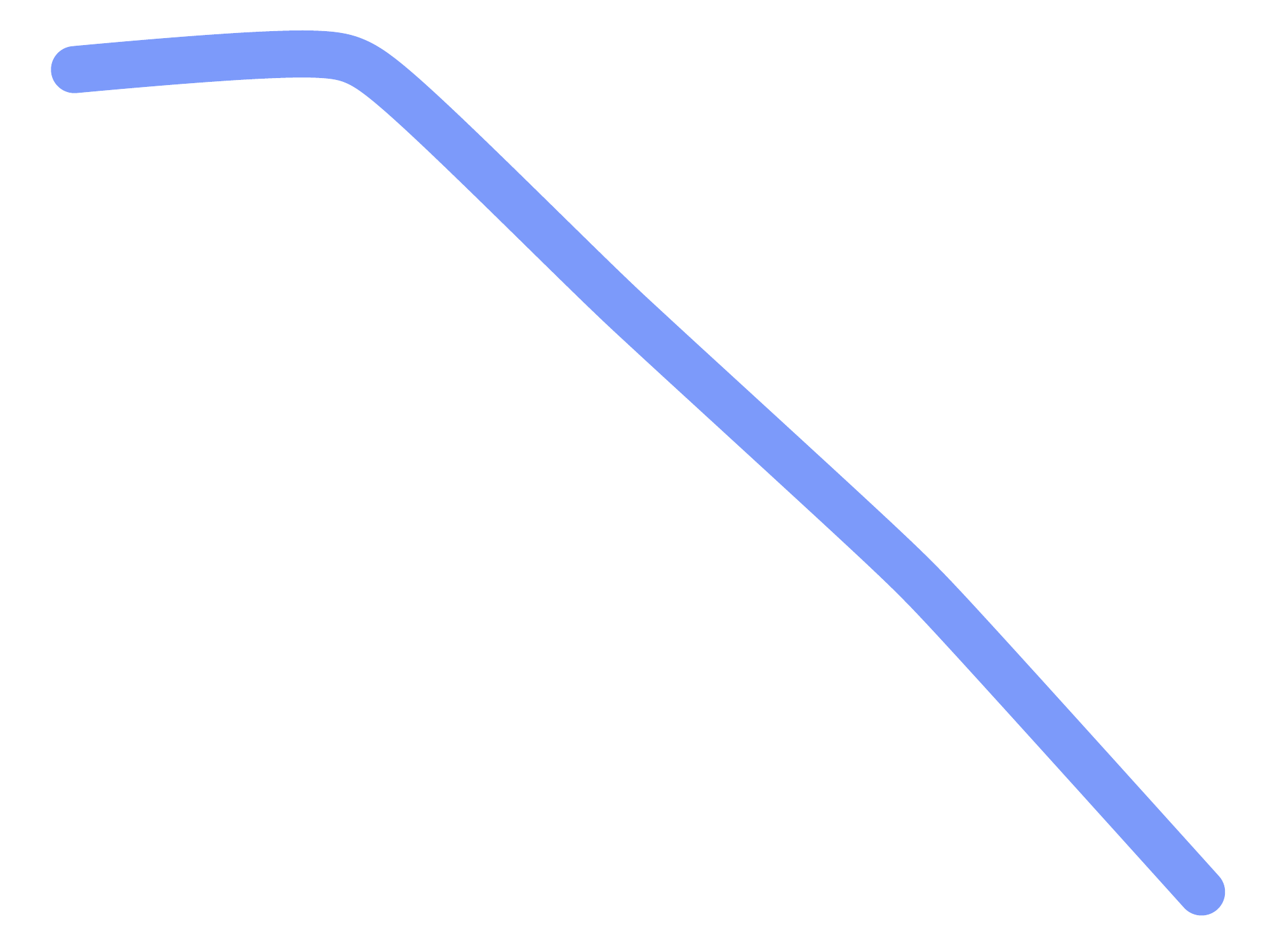}}
				\end{minipage}  & \cellcolor[rgb]{ .973,  .918,  .918} 47.99 & \cellcolor[rgb]{ .961,  .882,  .882}48.20 & 41.29 & 34.34 & \cellcolor[rgb]{ .973,  .918,  .918}26.11 \\
				EHCF & \begin{minipage}[!t]{0.1\columnwidth}
					\centering
					\raisebox{-.5\height}{\includegraphics[width=\textwidth, height=0.3\textwidth]{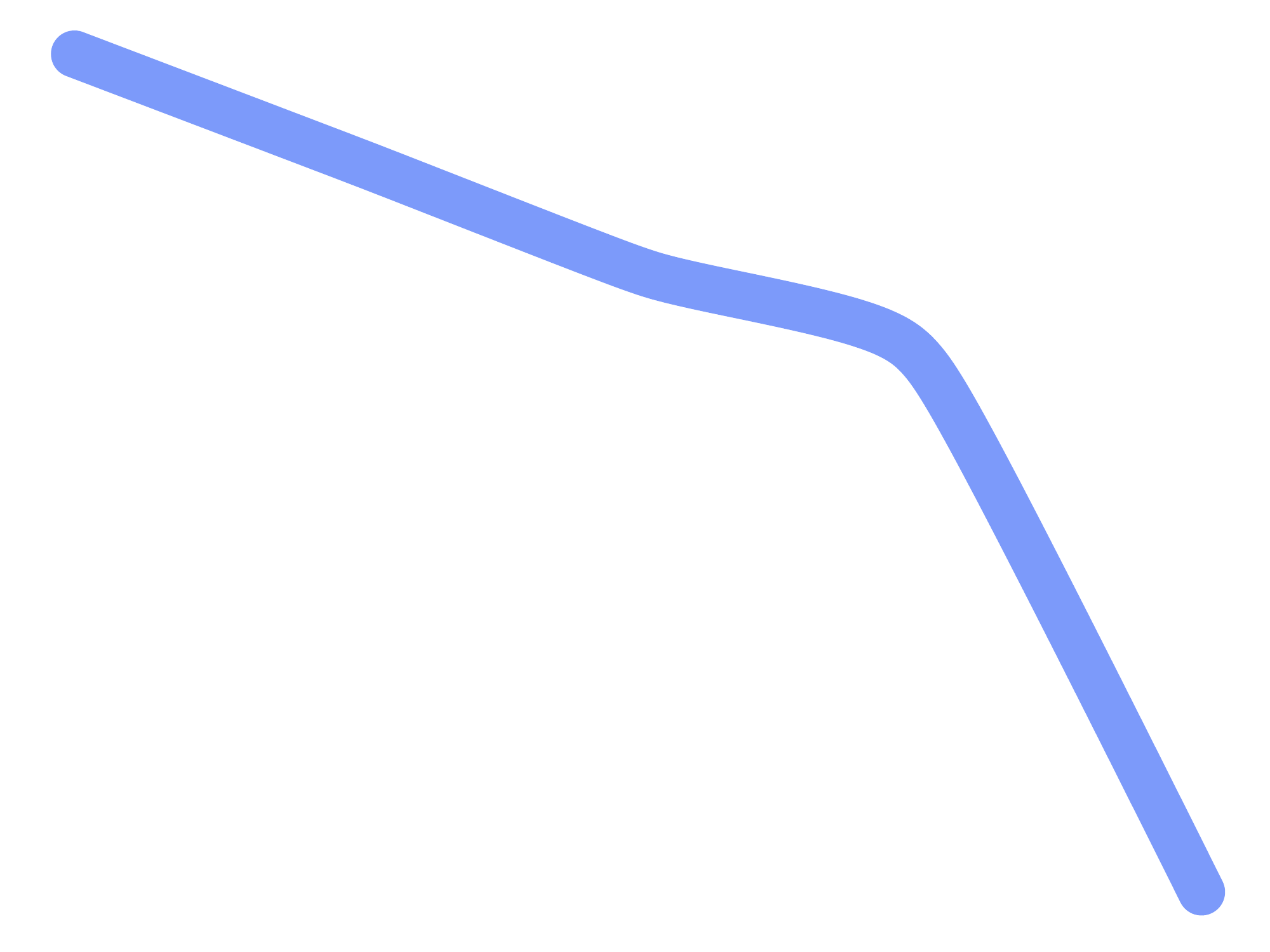}}
				\end{minipage}  & \cellcolor[rgb]{ .961,  .882,  .882} 49.02 & \cellcolor[rgb]{ .973,  .918,  .918}46.46 & \cellcolor[rgb]{ .973,  .918,  .918}43.86 & \cellcolor[rgb]{ .945,  .835,  .835}\underline{41.77}  & \cellcolor[rgb]{ .945,  .835,  .835}\underline{29.02}  \\
\bottomrule
			\end{tabular}%
		}
	\end{minipage}
	
\end{minipage}

%\clearpage
%\newpage

\subsection{Additional results of efficiency} \label{detail_eff}
\subsubsection{Additional average efficiency}

\begin{figure*}[h]
	\centering
	 
%	\subfigure[MovieLens-100k]{
%		\includegraphics[width=0.65\columnwidth]{major_revision/heatmap_polar/polar_ml-100k}
%		\label{ab.sub.ml-100k}
%	}
%	\subfigure[CiteULike]{
%		\includegraphics[width=0.65\columnwidth]{major_revision/heatmap_polar/polar_CiteULike}
%		\label{ab.sub.CiteULike}
%	}
%	\subfigure[MovieLens-1m]{
%		\includegraphics[width=0.65\columnwidth]{major_revision/heatmap_polar/polar_ml-1m}
%		\label{ab.sub.ml-1m}
%	}
		\subfigure[CiteULike]{
			\includegraphics[width=0.235\textwidth, height=0.21\textwidth]{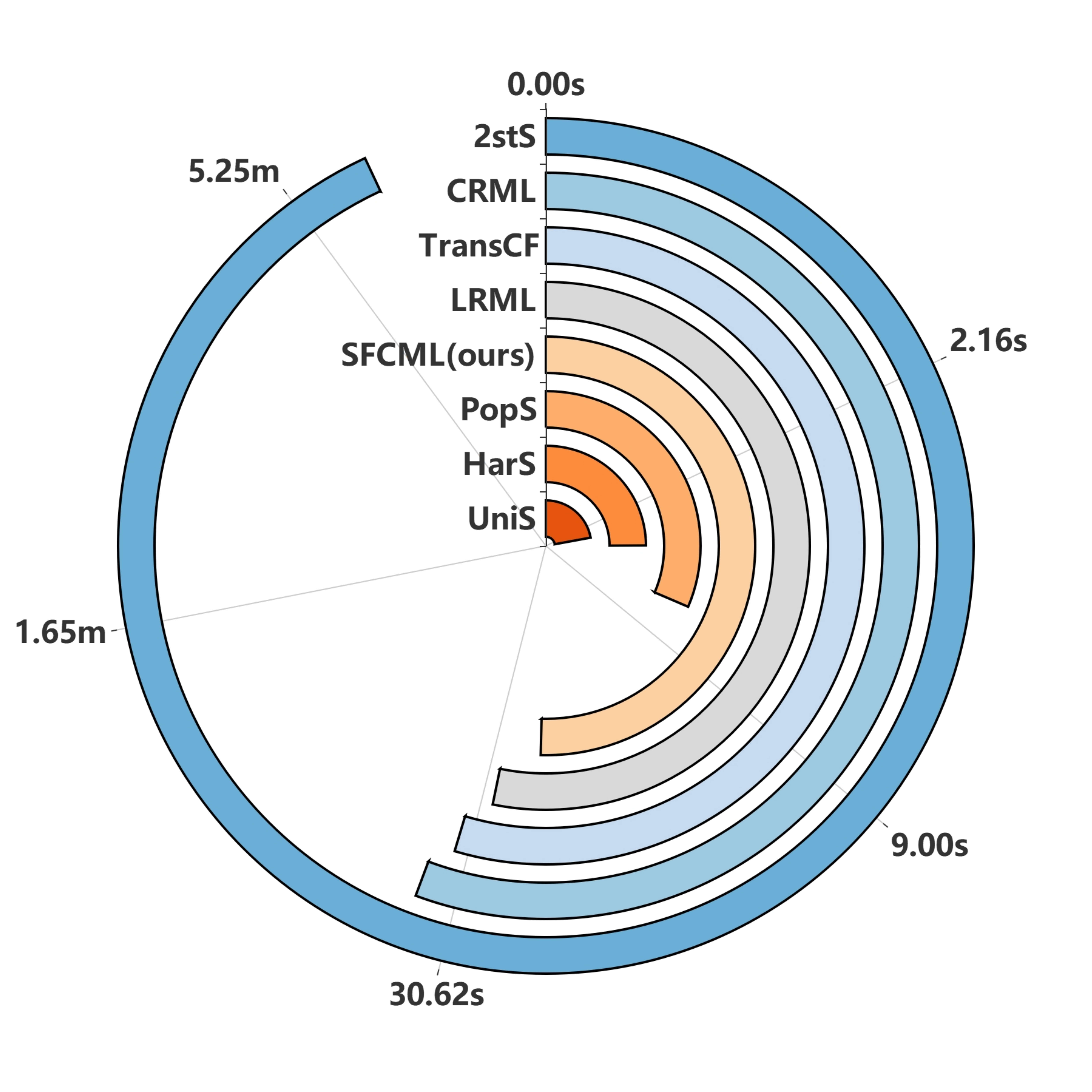}
			\label{ab.sub.CiteULike}
		}
			\subfigure[Steam-200k]{
			\includegraphics[width=0.235\textwidth, height=0.21\textwidth]{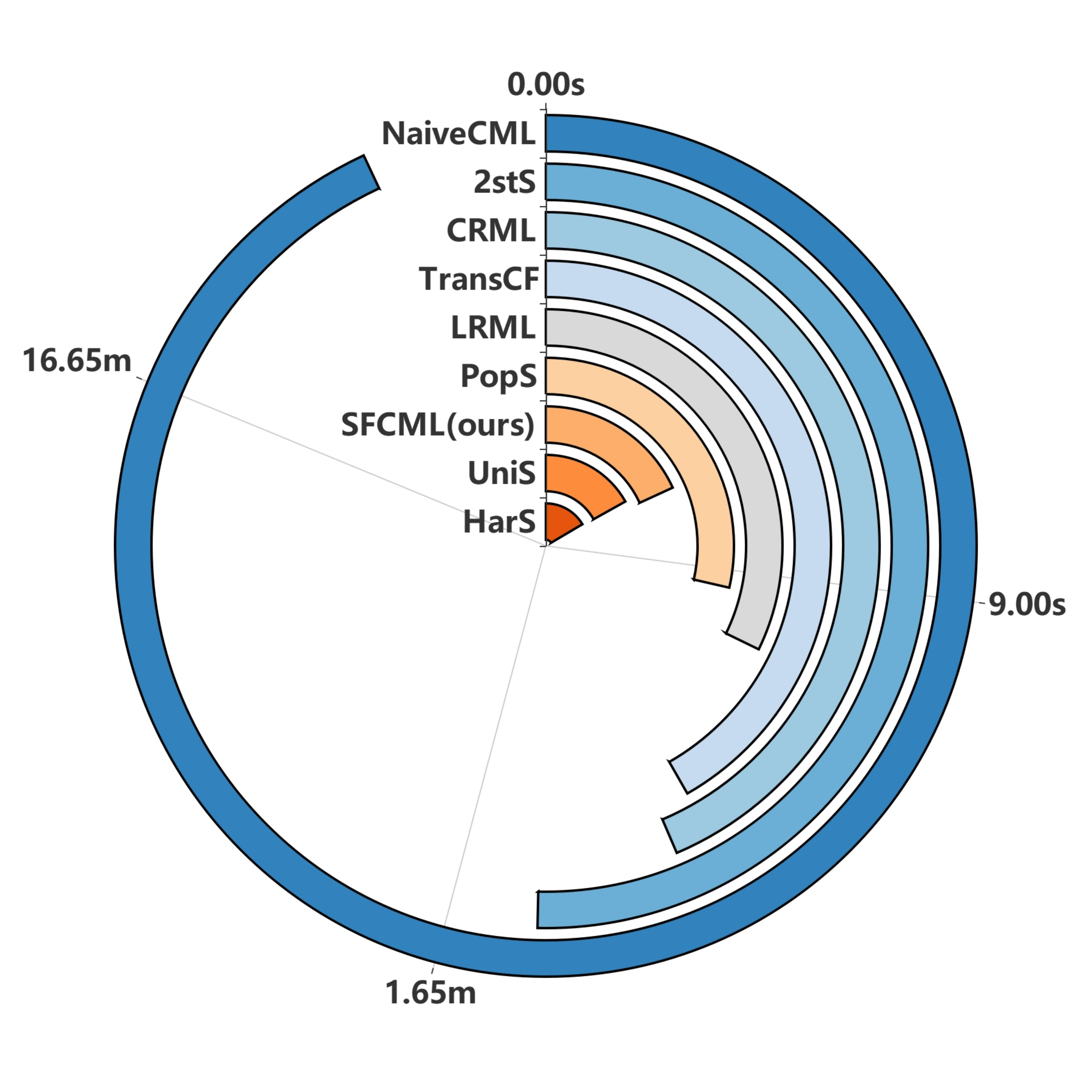}
			\label{ab.sub.Steam}
		}
		\subfigure[Anime]{
			\includegraphics[width=0.235\textwidth, height=0.21\textwidth]{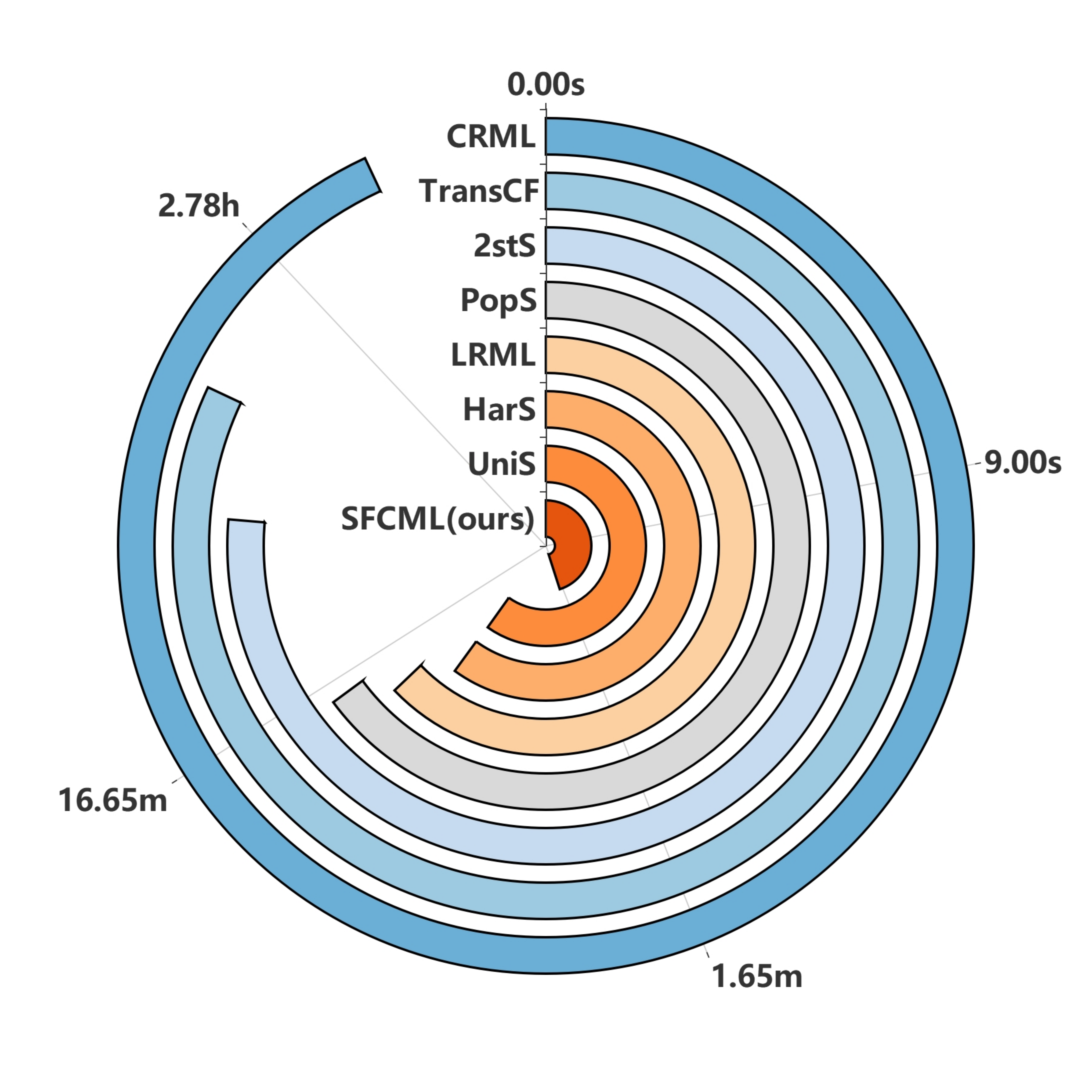}
			\label{ab.sub.Anime}
		}
		\subfigure[Amazon-Book]{
			\includegraphics[width=0.235\textwidth, height=0.21\textwidth]{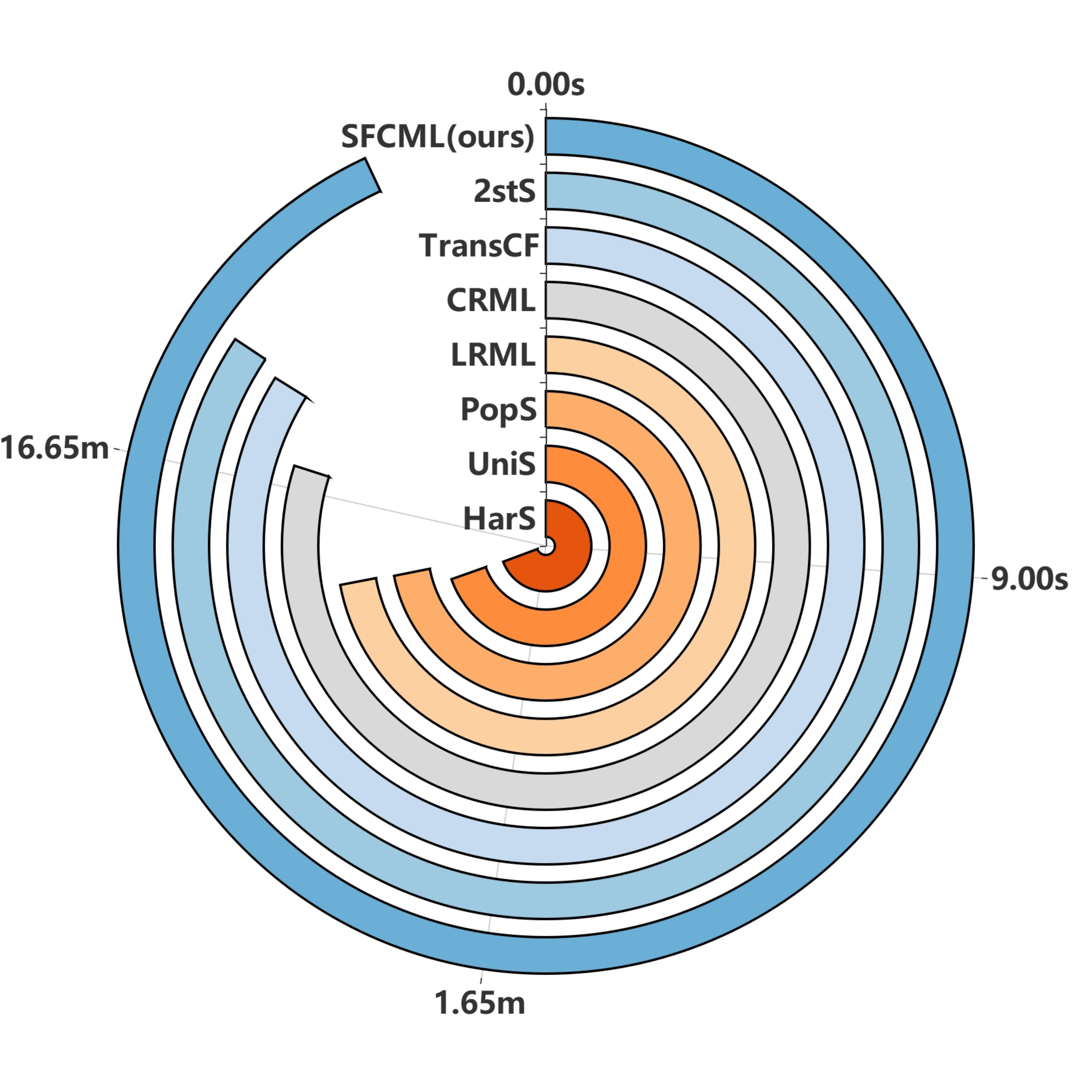}
			\label{ab.sub.Book}
		}
	\caption{Comparisons against average running time with respect to CML framework algorithms and SFCML(ours) on CiteULike, Steam-200k, Anime and Amazon-Book datasets. The method closer to the center of the circle enjoys better efficiency. Note that, here the 's', 'm', and 'h' represent the second, minute and hour, respectively.}
  
	\label{add_runtime}
\end{figure*}

\subsubsection{Running time per epoch}
% Table generated by Excel2LaTeX from sheet 'running_time_28_tf'
\begin{table}[htbp]
	\centering
	\caption{Comparisons against running time with respect to CML framework algorithms and SFCML(ours) on MovieLens-100k and Steam-200k, where '-' means that we cannot complete the experiments due to the out-of-memory issue. Note that, here the 's' and 'm' represent the second and minute, respectively. The best and second-best are highlighted in bold and underlined, respectively.}
	\scalebox{0.85}{
	\begin{tabular}{r|c|ccccccccccc}
		\toprule
		& Method & Epoch 1 & Epoch 2 & Epoch 3 & Epoch 4 & Epoch 5 & Epoch 6 & Epoch 7 & Epoch 8 & Epoch 9 & Epoch 10 & Average $\downarrow$ \\
		\midrule
		\multicolumn{1}{c|}{\multirow{9}[5]{*}{MovieLens-100k}} & UniS & 1.52s & 1.52s & 1.51s & 1.51s & 1.50s & 1.50s & 1.49s & 1.51s & 1.49s & 1.50s & \cellcolor[rgb]{ .98,  .863,  .788}\underline{1.51s} \\
		& PopS & 4.10s & 4.07s & 4.11s & 4.09s & 4.09s & 4.11s & 4.09s & 4.07s & 4.10s & 4.12s & \cellcolor[rgb]{ .996,  .949,  .925}4.10s \\
		& 2stS & 31.70s & 32.97s & 32.58s & 33.68s & 33.96s & 33.79s & 32.43s & 35.43s & 34.59s & 35.55s & 33.67s \\
		& HarS & 1.54s & 1.55s & 1.55s & 1.55s & 1.56s & 1.57s & 1.54s & 1.54s & 1.54s & 1.54s & \cellcolor[rgb]{ .996,  .89,  .839}1.55s \\
		& TransCF & 17.94s & 16.17s & 17.20s & 17.60s & 16.66s & 16.94s & 16.80s & 17.41s & 16.93s & 16.88s & \cellcolor[rgb]{ .996,  .973,  .961}17.05s \\
		& LRML & 6.67s & 6.60s & 6.46s & 6.66s & 6.49s & 6.54s & 6.61s & 6.57s & 6.66s & 6.59s & \cellcolor[rgb]{ .996,  .949,  .925}6.59s \\
		& CRML & 23.41s & 23.52s & 23.61s & 23.42s & 23.44s & 23.60s & 23.33s & 23.51s & 23.53s & 23.63s & 23.50s \\
		& NaiveCML & 6.87m & 6.86m & 6.88m & 6.87m & 6.88m & 6.86m & 6.87m & 6.88m & 6.88m & 6.88m & 6.87m \\
		\cmidrule{2-13}      & SFCML(ours) & 0.37s & 0.37s & 0.37s & 0.38s & 0.38s & 0.37s & 0.37s & 0.37s & 0.37s & 0.37s & \cellcolor[rgb]{ .973,  .796,  .678}\textbf{0.37s} \\
		\midrule
		\multicolumn{1}{c|}{\multirow{8}[2]{*}{Steam-200k}} & UniS & 3.22s & 3.25s & 3.21s & 3.25s & 3.20s & 3.18s & 3.18s & 3.19s & 3.21s & 3.20s & \cellcolor[rgb]{ .835,  .91,  .784}\underline{3.21s} \\
		& PopS & 10.38s & 10.33s & 10.38s & 10.39s & 10.34s & 10.36s & 10.35s & 10.32s & 10.35s & 10.32s & \cellcolor[rgb]{ .886,  .937,  .851}10.35s \\
		& 2stS & 1.22m & 1.23m & 1.17m & 1.19m & 1.17m & 1.17m & 1.16m & 1.24m & 1.16m & 1.19m & \cellcolor[rgb]{ .984,  .992,  .976}1.19m \\
		& HarS & 3.03s & 3.11s & 3.10s & 3.06s & 2.99s & 3.07s & 3.09s & 3.11s & 3.07s & 3.05s & \cellcolor[rgb]{ .776,  .878,  .706}\textbf{3.07s} \\
		& TransCF & 34.10s & 33.66s & 33.74s & 33.68s & 33.80s & 33.70s & 34.11s & 33.75s & 33.81s & 33.75s & \cellcolor[rgb]{ .945,  .969,  .929}33.81s \\
		& LRML & 14.37s & 14.55s & 14.55s & 14.43s & 14.55s & 14.54s & 14.37s & 14.47s & 14.31s & 14.49s & \cellcolor[rgb]{ .914,  .953,  .89}14.46s \\
		& CRML & 39.78s & 39.76s & 39.55s & 40.18s & 38.99s & 39.38s & 40.44s & 39.64s & 39.7s & 39.96s & \cellcolor[rgb]{ .945,  .969,  .929}39.74s \\
		& NaiveCML & 45.59m & 45.57m & 45.7m & 45.76m & 45.69m & 45.52m & 45.46m & 45.56m & 45.47m & 45.35m & 45.57m \\
		\cmidrule{2-13}
		& SFCML(ours) & 3.81s & 3.68s & 3.64s & 3.72s & 3.68s & 3.59s & 3.64s & 3.62s & 3.72s & 3.73s & \cellcolor[rgb]{ .835,  .91,  .784} 3.68s \\
		\bottomrule   \end{tabular}%
}
	\label{tab:eff_100k_200k}%
\end{table}%

% Table generated by Excel2LaTeX from sheet 'running_time_28_tf'
\begin{table}[htbp]
	\centering
	\caption{Comparisons against running time with respect to CML framework algorithms and SFCML(ours) on CiteULike, MovieLens-1m and Anime, where '-' means that we cannot complete the experiments due to the out-of-memory issue. Note that, here the 's', 'm' and 'h' represent the second, minute and hour, respectively. The best and second-best are highlighted in bold and underlined, respectively.}
	\scalebox{0.85}{
	\begin{tabular}{c|c|ccccccccccc}
		\toprule
		& Method & Epoch 1 & Epoch 2 & Epoch 3 & Epoch 4 & Epoch 5 & Epoch 6 & Epoch 7 & Epoch 8 & Epoch 9 & Epoch 10 & Average $\downarrow$ \\
		\midrule
		\multirow{9}[3]{*}{CiteULike} & UniS & 3.16s & 3.13s & 3.14s & 3.15s & 3.08s & 3.15s & 3.10s & 3.04s & 3.13s & 3.12s & \cellcolor[rgb]{ .741,  .843,  .933}\textbf{3.12s} \\
		& PopS & 6.49s & 6.52s & 6.50s & 6.45s & 6.49s & 6.44s & 6.46s & 6.48s & 6.53s & 6.48s & \cellcolor[rgb]{ .843,  .906,  .961}6.48s \\
		& 2stS & 6.35m & 6.60m & 6.56m & 6.56m & 6.39m & 6.34m & 6.45m & 6.33m & 6.50m & 6.27m & 6.43m \\
		& HarS & 3.92s & 3.96s & 3.98s & 3.93s & 3.93s & 3.92s & 3.97s & 3.94s & 3.91s & 3.91s & \cellcolor[rgb]{ .741,  .843,  .933}\underline{3.94s} \\
		& TransCF & 32.30s & 32.35s & 31.97s & 32.24s & 31.30s & 31.60s & 32.30s & 31.68s & 32.33s & 32.25s & \cellcolor[rgb]{ .906,  .941,  .973}32.03s \\
		& LRML & 29.29s & 28.93s & 29.06s & 29.36s & 29.35s & 29.08s & 29.24s & 29.42s & 29.34s & 29.05s & \cellcolor[rgb]{ .906,  .941,  .973}29.21s \\
		& CRML & 34.29s & 34.42s & 34.41s & 34.25s & 34.30s & 34.40s & 34.23s & 34.39s & 34.27s & 34.31s & \cellcolor[rgb]{ .906,  .941,  .973}34.33s \\
		& NaiveCML & - & - & - & - & - & - & - & - & - & - & - \\
		\cmidrule{2-13}      & SFCML(ours) & 24.21s & 24.15s & 24.22s & 24.26s & 24.20s & 24.21s & 24.17s & 24.18s & 24.20s & 24.21s & \cellcolor[rgb]{ .843,  .906,  .961}24.20s \\
		\midrule
		\multirow{9}[3]{*}{MovieLens-1m} & UniS & 11.13s & 11.52s & 11.41s & 11.59s & 11.37s & 11.11s & 11.17s & 11.61s & 11.24s & 11.39s & \cellcolor[rgb]{ .945,  .835,  .835}\underline{11.35s} \\
		& PopS & 40.82s & 40.63s & 40.41s & 40.12s & 40.51s & 40.12s & 40.38s & 40.67s & 40.54s & 40.37s & \cellcolor[rgb]{ .973,  .918,  .918}40.46s \\
		& 2stS & 4.80m & 4.85m & 4.87m & 4.78m & 4.87m & 4.83m & 4.90m & 4.89m & 4.89m & 4.80m & \cellcolor[rgb]{ .996,  .984,  .98}4.85m \\
		& HarS & 12.18s & 12.15s & 12.10s & 12.01s & 12.13s & 12.03s & 12.19s & 12.04s & 12.14s & 12.14s & \cellcolor[rgb]{ .945,  .835,  .835}12.11s \\
		& TransCF & 3.23m & 3.24m & 3.2m & 3.2m & 3.18m & 3.14m & 3.2m & 3.17m & 3.17m & 3.16m & \cellcolor[rgb]{ .996,  .984,  .98}3.19m \\
		& LRML & 1.25m & 1.24m & 1.26m & 1.26m & 1.26m & 1.23m & 1.25m & 1.25m & 1.26m & 1.24m & \cellcolor[rgb]{ .961,  .882,  .882}1.25m \\
		& CRML & 6.93m & 6.92m & 6.94m & 6.94m & 6.96m & 6.9m & 6.93m & 6.93m & 6.91m & 6.90m & \cellcolor[rgb]{ .996,  .984,  .98}6.93m \\
		& NaiveCML & - & - & - & - & - & - & - & - & - & - & - \\
		\cmidrule{2-13}      & SFCML(ours) & 5.88s & 5.89s & 5.88s & 5.87s & 5.86s & 5.90s & 5.89s & 5.87s & 5.90s & 5.85s & \cellcolor[rgb]{ .929,  .788,  .788}\textbf{5.88s} \\
		\midrule
		\multirow{9}[4]{*}{Anime} & UniS & 8.77m & 8.77m & 8.74m & 8.73m & 8.75m & 8.79m & 8.77m & 8.78m & 8.73m & 8.79m & \cellcolor[rgb]{ .98,  .863,  .788}\underline{8.76m} \\
		& PopS & 14.96m & 14.70m & 14.69m & 14.59m & 15.02m & 14.81m & 14.80m & 15.04m & 14.98m & 14.93m & \cellcolor[rgb]{ .988,  .925,  .882}14.85m \\
		& 2stS & 49.55m & 48.37m & 48.98m & 48.95m & 47.92m & 49.60m & 49.80m & 49.27m & 49.84m & 48.94m & \cellcolor[rgb]{ .996,  .949,  .925}49.12m \\
		& HarS & 8.88m & 8.93m & 9.00m & 8.93m & 8.85m & 8.91m & 8.83m & 8.86m & 8.98m & 8.80m & \cellcolor[rgb]{ .98,  .863,  .788}8.90m \\
		& TransCF & 1.49h & 1.49h & 1.44h & 1.49h & 1.49h & 1.48h & 1.50h & 1.46h & 1.47h & 1.48h & \cellcolor[rgb]{ .996,  .973,  .961}1.48h \\
		& LRML & 12.03m & 12.04m & 12.04m & 11.97m & 11.97m & 11.97m & 11.93m & 11.96m & 11.99m & 12.00m & \cellcolor[rgb]{ .988,  .925,  .882}11.99m \\
		& CRML & 4.71h & 4.73h & 4.72h & 4.72h & 4.69h & 4.74h & 4.73h & 4.7h & 4.72h & 4.71h & 4.72h \\
		& NaiveCML & - & - & - & - & - & - & - & - & - & - & - \\
		\cmidrule{2-13}      & SFCML(ours) & 1.84m & 1.85m & 1.84m & 1.84m & 1.85m & 1.85m & 1.85m & 1.84m & 1.85m & 1.84m & \cellcolor[rgb]{ .973,  .796,  .678}\textbf{1.85m} \\
		\bottomrule
	\end{tabular}%
}
	\label{tab:eff_CiteULike_1m_Anime}%
\end{table}%

\begin{table}[htbp]
	\centering
	\caption{Comparisons against running time with respect to CML framework algorithms and SFCML(ours) on MovieLens-20m and Amazon-Book, where '-' means that we cannot complete the experiments due to the out-of-memory issue. Note that, here the 's', 'm' and 'h' represent the second, minute and hour, respectively. The best and second-best are highlighted in bold and underlined, respectively.}
	\scalebox{0.85}{
	\begin{tabular}{c|c|ccccccccccc}
		\toprule
		& Method & Epoch 1  & Epoch 2  & Epoch 3 & Epoch 4 & Epoch 5 & Epoch 6 & Epoch 7 & Epoch 8 & Epoch 9 & Epoch 10 & Average $\downarrow$ \\
		\midrule
		\multirow{9}[4]{*}{MovieLens-20m} & UniS & 17.34m & 16.98m & 17.21m & 17.10m & 16.86m & 17.00m & 16.93m & 17.05m & 17.00m & 17.18m & \cellcolor[rgb]{ .808,  .882,  .949}\underline{17.07m} \\
		& PopS & 41.52m & 41.15m & 41.44m & 41.60m & 41.18m & 41.26m & 41.40m & 41.45m & 41.43m & 41.50m & \cellcolor[rgb]{ .906,  .941,  .973}41.39m \\
		& 2stS & 2.22h & 2.20h & 2.17h & 2.15h & 2.16h & 2.17h & 2.06h & 2.05h & 2.19h & 2.07h & 2.14h \\
		& HarS & 17.66m & 17.66m & 17.80m & 17.70m & 17.73m & 17.33m & 17.43m & 17.50m & 17.69m & 17.45m & \cellcolor[rgb]{ .843,  .906,  .961}17.60m \\
		& TransCF & 4.49h & 4.43h & 4.47h & 4.51h & 4.50h & 4.48h & 4.44h & 4.47h & 4.46h & 4.47h & 4.47h \\
		& LRML & 23.35m & 23.46m & 23.56m & 23.35m & 23.37m & 23.72m & 23.60m & 23.56m & 23.35m & 23.46m & \cellcolor[rgb]{ .906,  .941,  .973}23.48m \\
		& CRML & 18.52h & 18.49h & 18.47h & 18.51h & 18.42h & 18.51h & 18.53h & 18.54h & 18.52h & 18.48h & 18.50h \\
		& NaiveCML & - & - & - & - & - & - & - & - & - & - & - \\
		\cmidrule{2-13}      & SFCML(ours) & 11.99m & 12.03m & 11.99m & 11.97m & 12.04m & 11.96m & 11.99m & 12.02m & 12.01m & 11.97m & \cellcolor[rgb]{ .741,  .843,  .933}\textbf{12.00m} \\
		\midrule
		\multirow{9}[8]{*}{Book} & UniS & 7.53m & 7.60m & 7.48m & 7.59m & 7.65m & 7.55m & 7.66m & 7.68m & 7.53m & 7.56m & \cellcolor[rgb]{ .945,  .835,  .835}\underline{7.58m} \\
		& PopS & 9.40m & 9.20m & 9.27m & 9.39m & 9.35m & 9.23m & 9.33m & 9.33m & 9.20m & 9.19m & \cellcolor[rgb]{ .961,  .882,  .882}9.29m \\
		& 2stS & 28.46m & 27.69m & 27.13m & 28.02m & 27.03m & 27.19m & 28.03m & 27.87m & 28.03m & 28.37m & 27.78m \\
		& HarS & 7.52m & 7.44m & 7.41m & 7.45m & 7.51m & 7.37m & 7.31m & 7.37m & 7.27m & 7.46m & \cellcolor[rgb]{ .929,  .788,  .788}\textbf{7.41m} \\
		& TransCF & 26.57m & 26.57m & 26.59m & 26.57m & 26.59m & 26.59m & 26.59m & 26.58m & 26.58m & 26.57m & 26.58m \\
		& LRML & 9.37m & 9.42m & 9.38m & 9.38m & 9.35m & 9.36m & 9.41m & 9.39m & 9.38m & 9.38m & \cellcolor[rgb]{ .961,  .882,  .882}9.38m \\
		& CRML & 18.93m & 18.80m & 18.71m & 18.94m & 18.81m & 18.62m & 18.84m & 18.77m & 18.82m & 18.80m & \cellcolor[rgb]{ .973,  .918,  .918}18.80m \\
		& NaiveCML & - & - & - & - & - & - & - & - & - & - & - \\
		\cmidrule{2-13}      & SFCML(ours) & 59.77m & 59.53m & 59.90m & 59.76m & 58.98m & 59.90m & 59.73m & 60.00m & 60.00m & 60.00m & 59.79m \\
		\bottomrule
	\end{tabular}%
}
	\label{tab:eff_20m_Book}%
\end{table}%

\end{document}